\documentclass[10pt,twocolumn,letterpaper]{article}

\usepackage[pagenumbers]{cvpr} %
\providecommand{\linenumbers}{}
\providecommand{\nolinenumbers}{}

\usepackage{tikz}
\usetikzlibrary{arrows.meta,calc}
\IfFileExists{fontawesome5.sty}{%
  \usepackage{fontawesome5}%
}{%
  \newcommand{\faLock}{\raisebox{0.05em}{\scriptsize$\blacksquare$}}%
}
\definecolor{pixblue}{RGB}{31,111,180}
\definecolor{latorange}{RGB}{226,121,24}
\definecolor{gradred}{RGB}{214,39,40}

\usepackage{placeins}

\usepackage{algorithm}
\usepackage{algpseudocode}
\AtBeginDocument{%
  \crefname{algorithm}{Algorithm}{Algorithms}%
  \Crefname{algorithm}{Algorithm}{Algorithms}%
}

\definecolor{cvprblue}{rgb}{0.21,0.49,0.74}
\usepackage[pagebackref,breaklinks,colorlinks,allcolors=cvprblue]{hyperref}
\hypersetup{
  pdftitle={L2P-DMD: Latent-Space Supervision for Distilling Few-Step Pixel Diffusion},
  pdfauthor={Zongze Wu, Nicholas Kolkin, Xianbang Wang, Ruiming Cao, Eli Shechtman}
}

\def\confName{CVPR}
\def\confYear{2027}

\title{L2P-DMD: Latent-Space Supervision for Distilling Few-Step Pixel Diffusion}

\author{
Zongze Wu$^{1}$ \quad Nicholas Kolkin$^{1}$ \quad Xianbang Wang$^{2}$ \quad Ruiming Cao$^{1}$ \quad Eli Shechtman$^{1}$
}

\begin{document}
\raggedbottom
\twocolumn[{%
\renewcommand\twocolumn[1][]{#1}%
\maketitle
\vspace{-2.7em}
{\centering\small $^{1}$Adobe \qquad $^{2}$MIT\par}
\vspace{0.8em}
\begin{center}
    \centering
    \noindent
    \begin{minipage}[b]{0.491\textwidth}%
        \includegraphics[width=\linewidth]{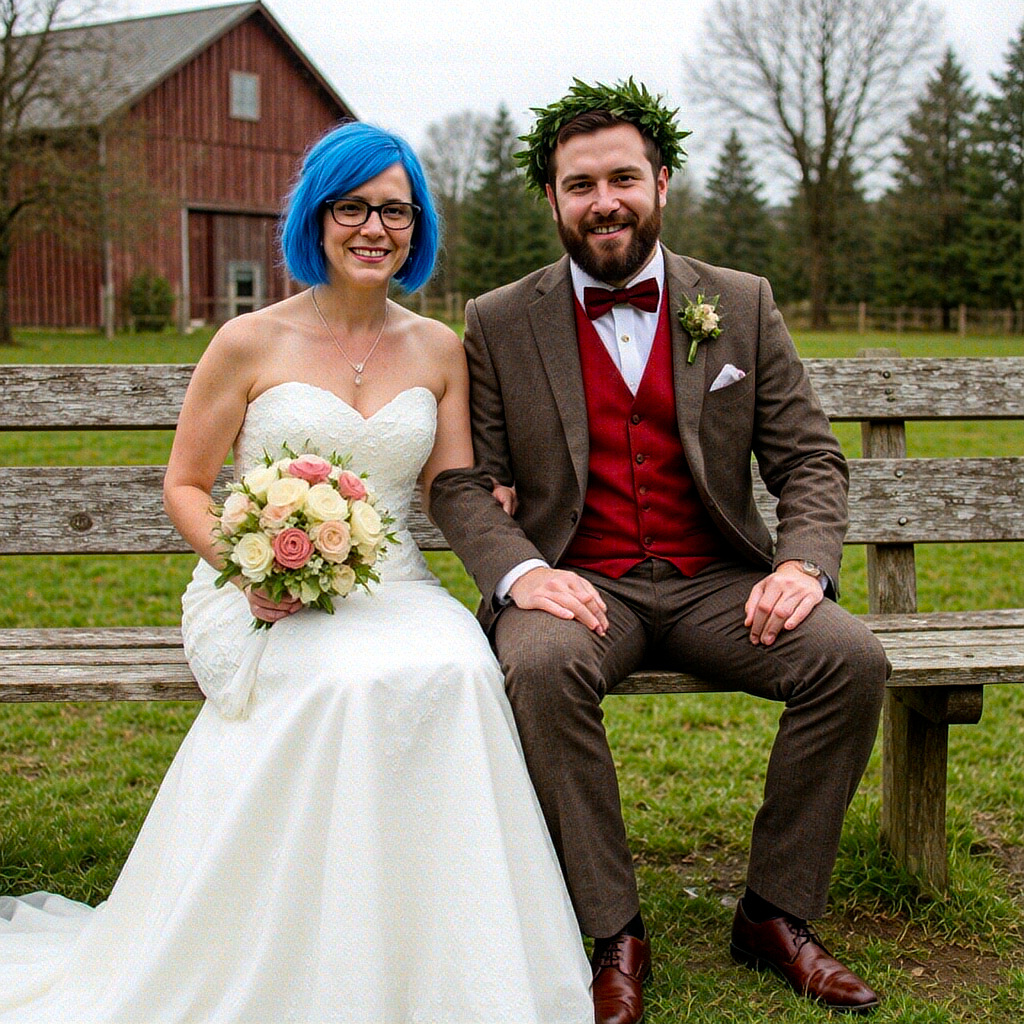}%
    \end{minipage}%
    \hspace{0.0085\textwidth}%
    \begin{minipage}[b]{0.241\textwidth}%
        \includegraphics[width=\linewidth]{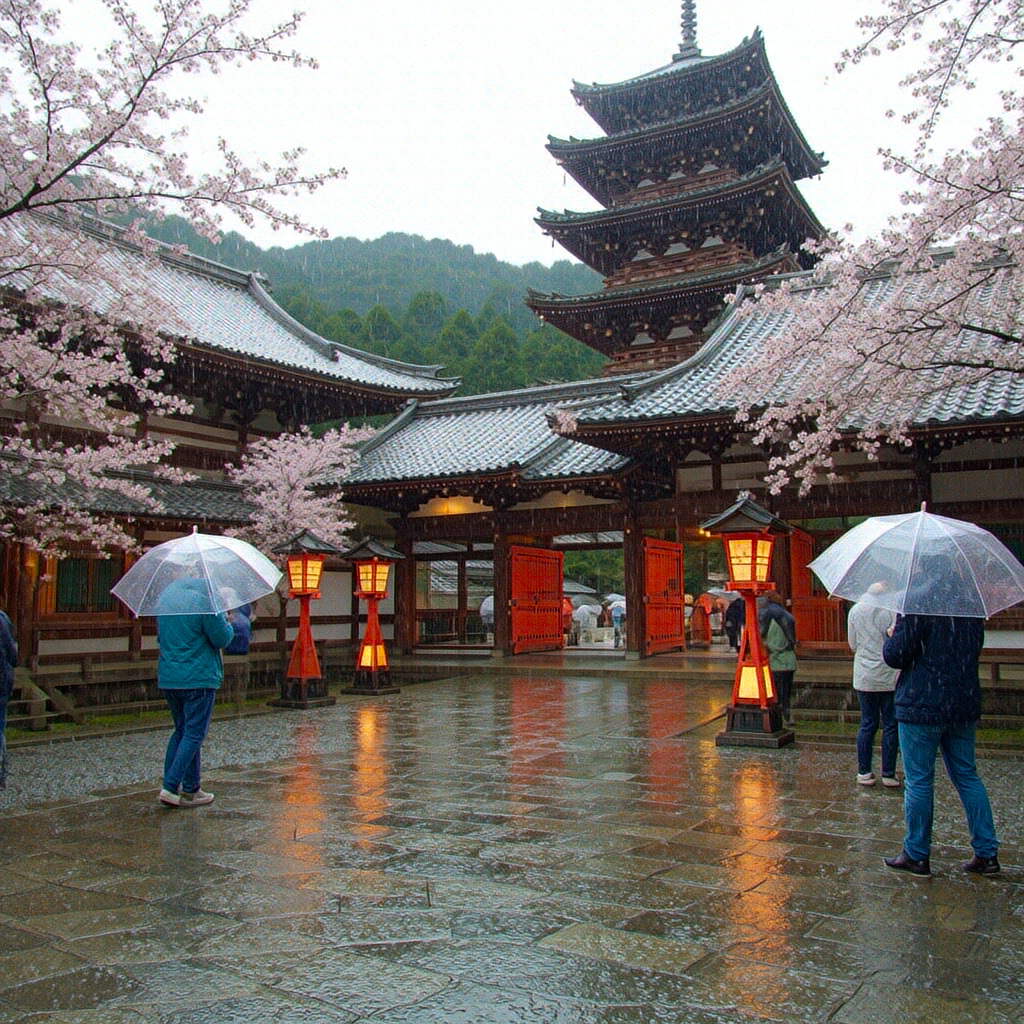}\par
        \vspace{0.0085\textwidth}%
        \includegraphics[width=\linewidth]{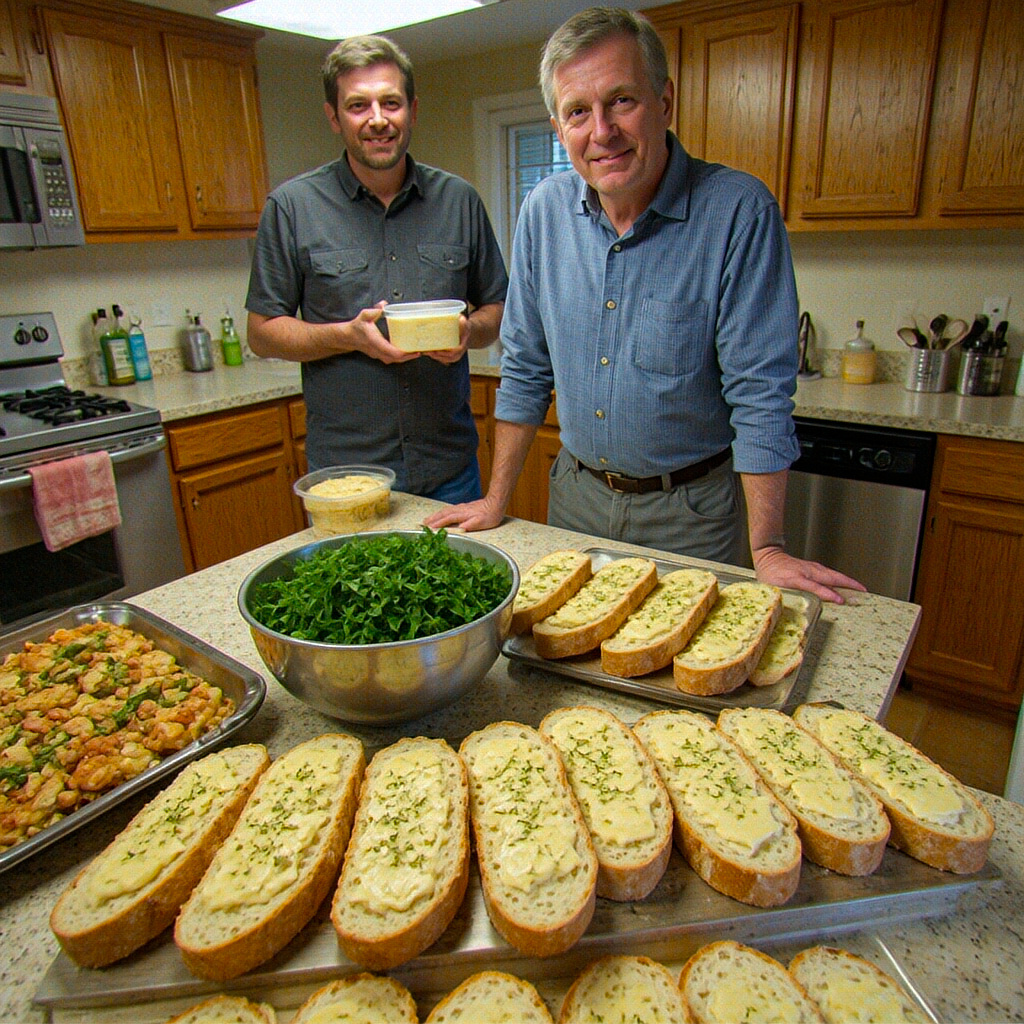}%
    \end{minipage}%
    \hspace{0.0085\textwidth}%
    \begin{minipage}[b]{0.241\textwidth}%
        \includegraphics[width=\linewidth]{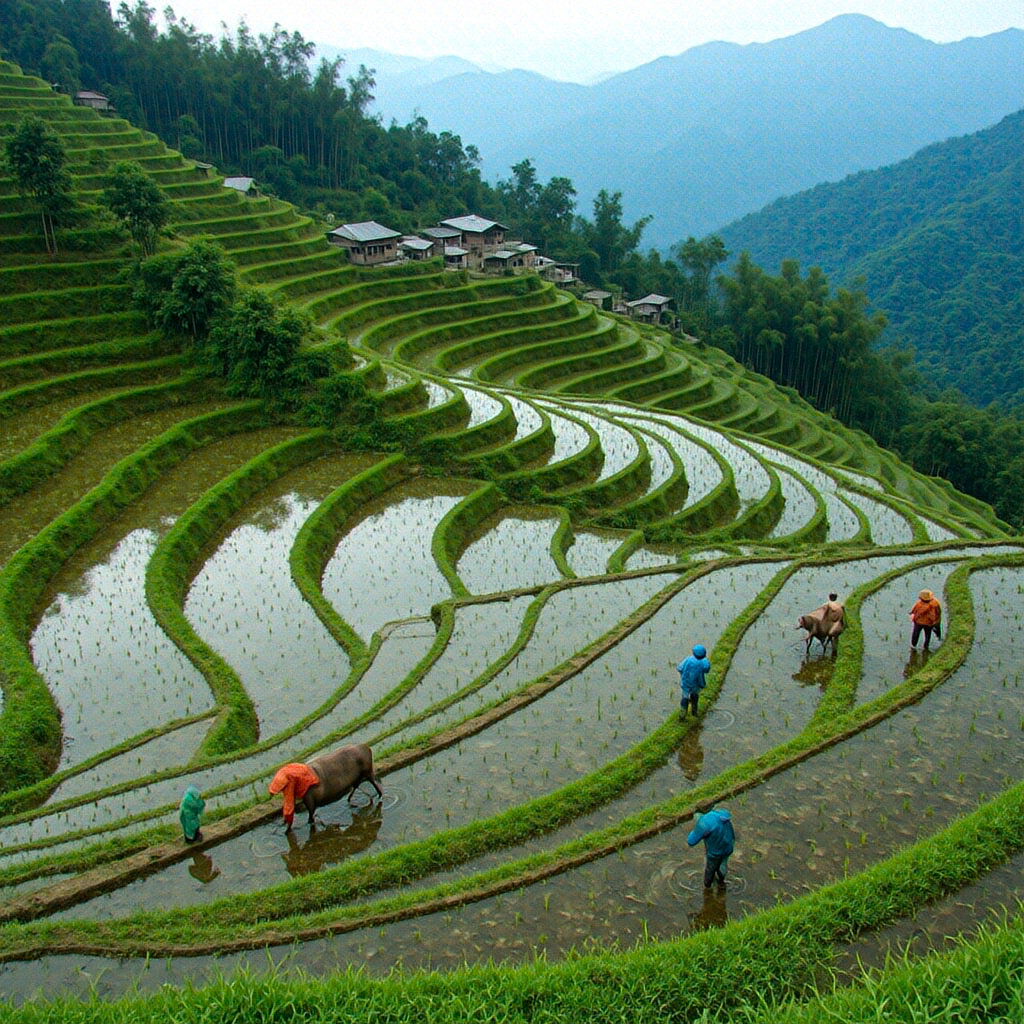}\par
        \vspace{0.0085\textwidth}%
        \includegraphics[width=0.482\linewidth]{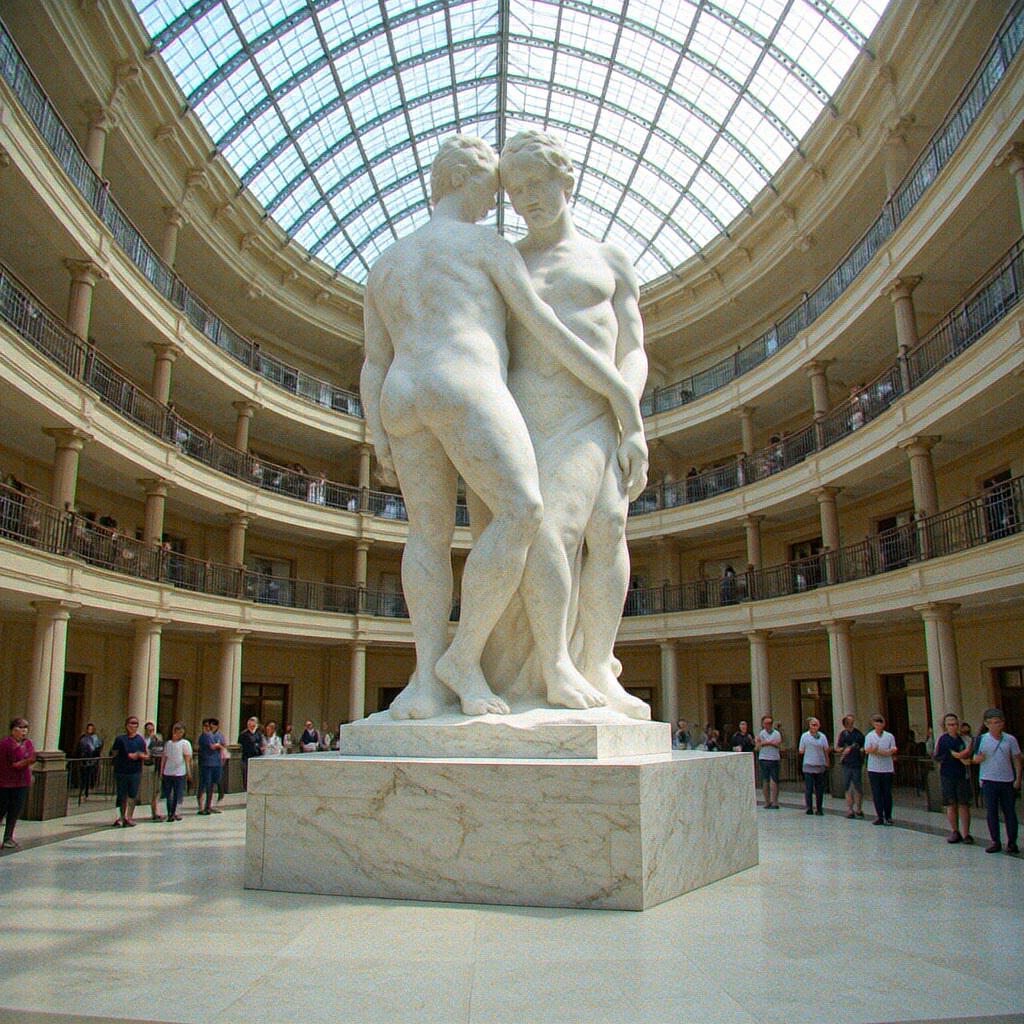}%
        \hfill
        \includegraphics[width=0.482\linewidth]{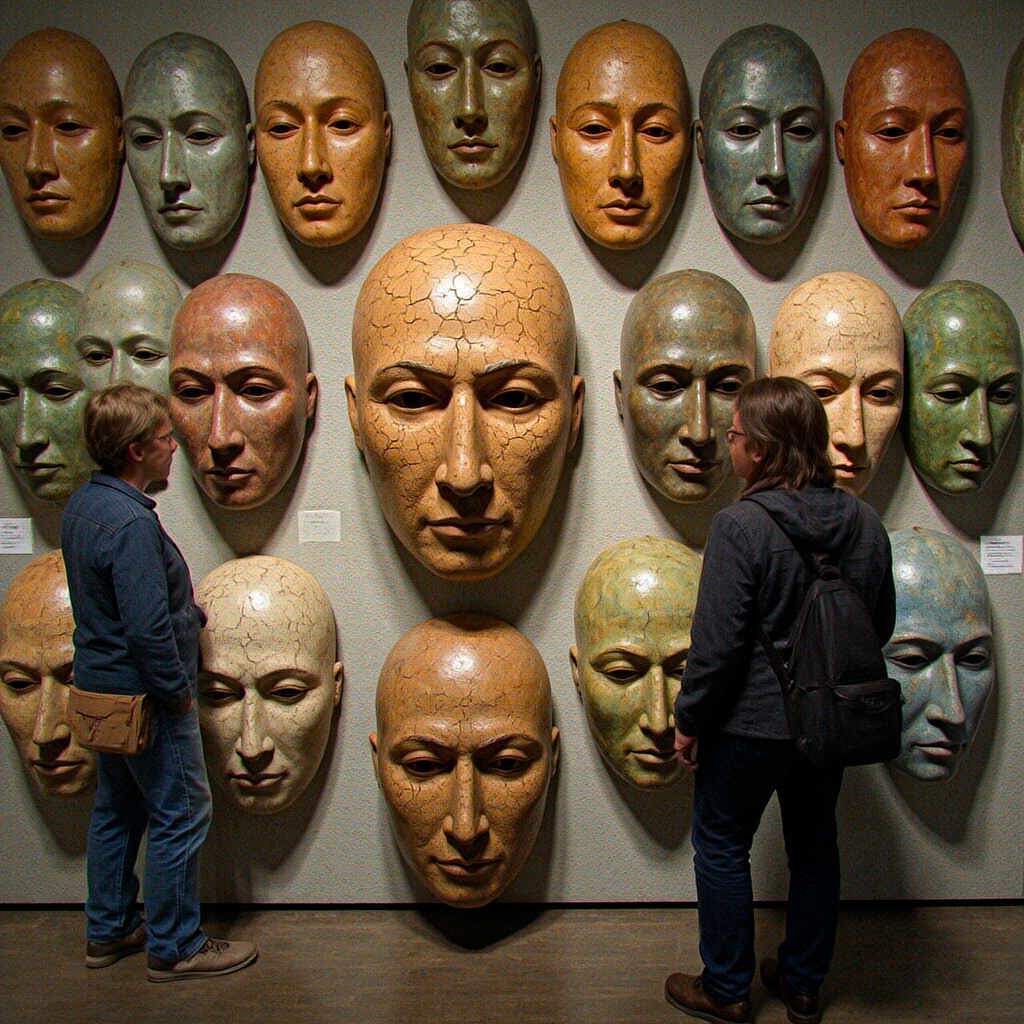}\par
        \vspace{0.035\linewidth}%
        \includegraphics[width=0.482\linewidth]{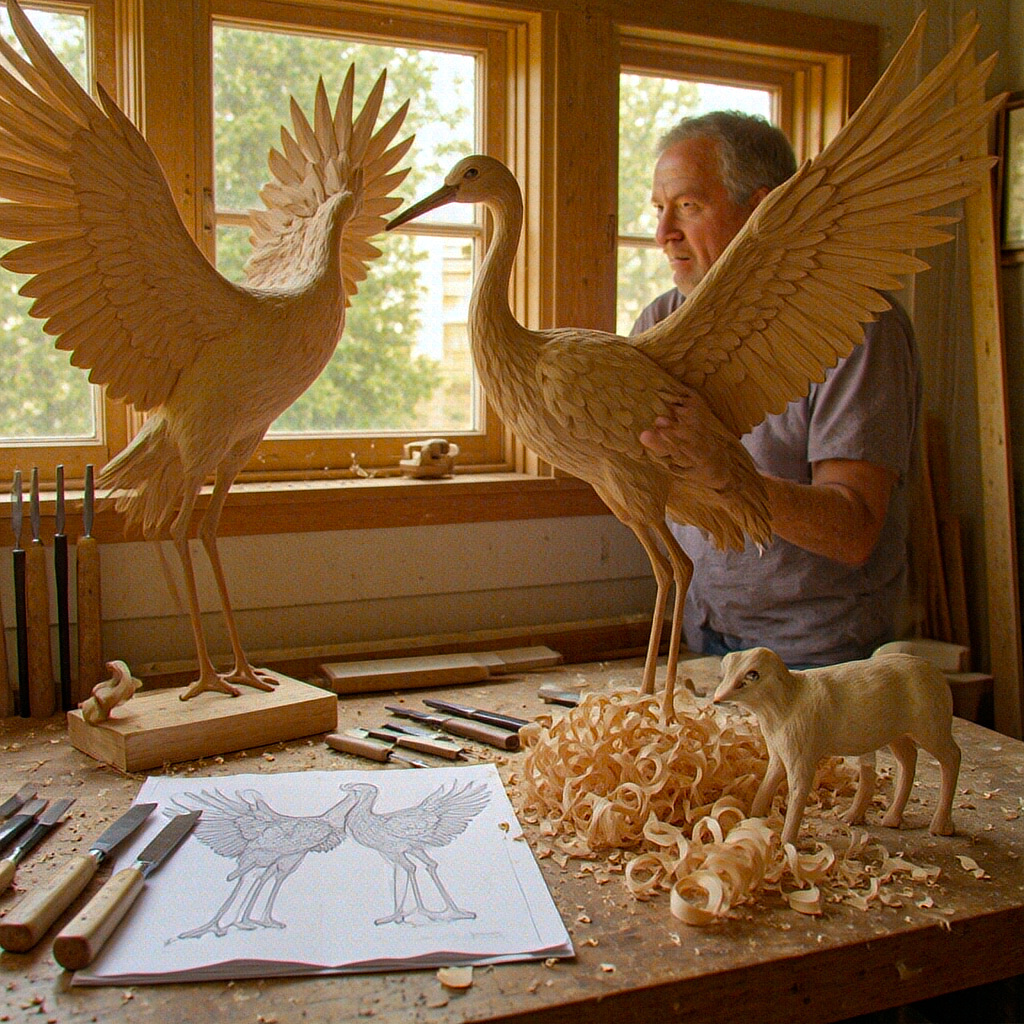}%
        \hfill
        \includegraphics[width=0.482\linewidth]{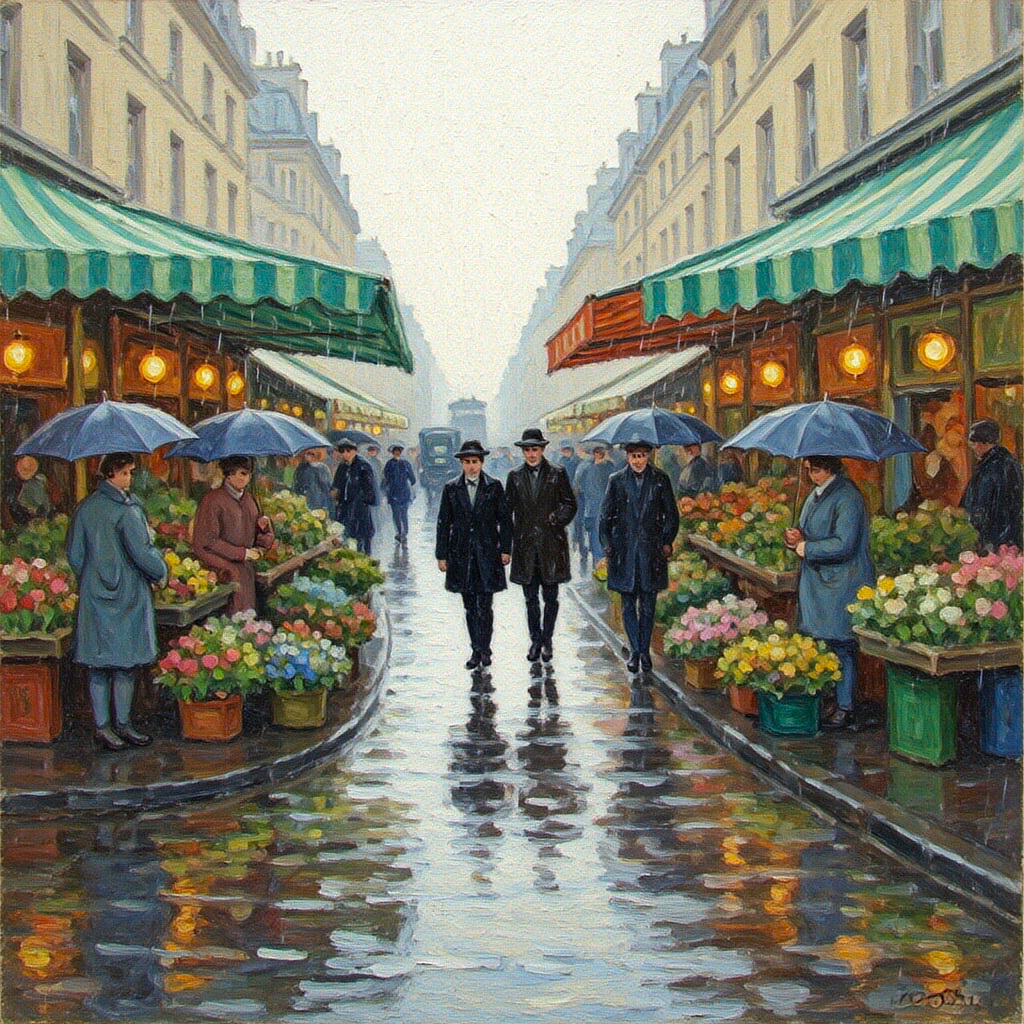}%
    \end{minipage}
    \captionsetup{hypcap=false}
    \captionof{figure}{\textbf{Four-step high-resolution pixel-space generation with L2P-DMD Training.}
    4-step $1024\!\times\!1024$ samples from the AsymFLUX.2 Klein 9B
    pixel-space student. Zoom into smaller tiles to see details.}
    \captionsetup{hypcap=true}
    \label{fig:l2p-teaser}
\end{center}

}]
\begin{abstract}
Few-step diffusion models enable efficient image generation, but their quality
depends on distribution matching. On ImageNet-256, we find that pixel-space
DMD supervision for few-step pixel generators converges more slowly and
yields a higher best FID than latent-space supervision. We therefore
introduce Latent-to-Pixel Distribution Matching Distillation (L2P-DMD), which
uses a latent teacher and online latent critic to supervise a pixel student
through a frozen encoder. L2P-DMD improves one-step FID by 24\% and converges
about $2\times$ faster than pixel matching. At
$1024\!\times\!1024$, FLUX.2 9B supervision improves a four-step AsymFLUX pixel
student over pixel matching on HPSv3 and DPG-Bench. Finally, L2P-DMD transfers
FLUX.3 supervision across VAEs and architectures to a 0.39B four-step
DreamLite pixel U-Net, reaching 14.77 HPSv3 and 85.07 DPG-Bench---absolute
gains of 4.51 and 0.47 over the official four-step latent DreamLite Mobile.
\end{abstract}

\section{Introduction}
\label{sec:intro}

Pixel-space diffusion models synthesize RGB images directly, avoiding the reconstruction bottleneck of an autoencoder and offering a promising route to fine-grained image generation \cite{hoogeboom2025sid2,chen2025pixelflow,li2026jit,chen2026asymmetric,vallaeys2026ujit,cai2026iris}. Combined with lightweight architectures and few-step sampling, they are attractive for efficient deployment. However, prior work reports that large-scale pixel-space pretraining converges substantially more slowly than latent-space pretraining \cite{jiang2026pixeltext}. In our distillation experiments, the gap is less extreme, but pixel-space supervision still requires more generator updates. The availability of capable pretrained latent diffusion models \cite{minimax2026h3image,qwen2026qwenimage21,bfl2026flux3imageklein} motivates a practical alternative: distilling their capabilities into lightweight, few-step pixel-space students.

Among step-distillation approaches, Distribution Matching Distillation
(DMD) has become a widely used distillation framework in both academia and
industry, supporting few-step image, video, and world-generation
systems~\cite{yin2024dmd,yin2024dmd2,
cai2025zimage,nvidia2026cosmos3,modeltc2025wanlightning,
tencent2025hyworldplay}. The framework trains a student using the difference
between two score estimates:
a frozen teacher provides the target distribution's score, and an online critic, trained by denoising student-generated samples, estimates the score of the student's current distribution.
Both scores are evaluated on the same noise-corrupted student output, assuming a shared representation space.
This creates a challenge when the strongest available teacher
provides scores in latent space but the desired student generates RGB pixels.
Resolving this mismatch requires choosing where to evaluate the
teacher--critic correction: in the student's pixel space or in the pretrained
teacher's latent space.
We call these alternatives pixel matching and latent matching, respectively.
In both cases, the student remains a pixel-space generator. Pixel matching
uses a frozen teacher and online critic
in pixel space, whereas latent matching uses a teacher's native
latent space. %

Our goal is to train a few-step pixel-space generator while matching in a
strong latent teacher's native space. We keep the student and its output in
pixel space, encode each generated RGB sample with the teacher's frozen VAE encoder,
and keep the online critic in the same latent space as the teacher. The
distribution-matching loss is computed entirely in latent space and back
propagated through the frozen encoder to update the pixel-space student.
Unlike decoding a latent target into pixel space, this update is stable in our
experiments and preserves direct RGB, VAE-free inference. This gradient path
itself is established: Magic3D backpropagates latent SDS through an LDM encoder
to optimize an RGB-producing renderer~\cite{lin2023magic3d}. We do not claim
this encoder-backpropagation mechanism as novel; our contribution is its
empirical evaluation within DMD against native pixel matching for few-step
pixel students.

We first compare pixel and latent matching on ImageNet-256 class-conditional
generation~\cite{deng2009imagenet}. Latent matching improves one-step FID~\cite{heusel2017fid} from
9.43 to 7.15 for a Pixel SiT student~\cite{ma2024sit,chen2026asymmetric} and
from 4.70 to 3.95 for an AsymJiT-H student~\cite{li2026jit,chen2026asymmetric}.
We then extend the comparison to $1024\!\times\!1024$ text-to-image generation
with AsymFLUX~\cite{chen2026asymmetric} and PixelDiT~\cite{yu2026pixeldit}, and
transfer FLUX~\cite{flux2} supervision to a 0.39B pixel-adapted
DreamLite~\cite{feng2026dreamlite} student across differences in architecture,
VAE, model scale, and conditioning stack.

Across the matched teacher DMD distillation comparisons reported
here~(\cref{tab:imagenet-main,tab:asymflux-main}), latent matching
outperforms pixel matching. This trend mirrors prior findings in flow-matching
training, where pixel-space models are
more sensitive to output parameterization~\cite{li2026jit} and converge more
slowly than latent-space models~\cite{jiang2026pixeltext}. We hypothesize that
computing the teacher--critic score difference in latent space produces a less
noisy training signal than computing it in pixel space, thereby improving
student performance.

In summary, our contributions are as follows:

\begin{itemize}
    \item We provide empirical evidence that, for the evaluated few-step pixel
    students and recipes, latent matching can outperform native pixel matching.
    \item We compare three cross-representation update rules and multiple
    teacher and training configurations. Under the tested recipes, encoder
    backpropagation trains stably, improves one-step FID in our primary
    ImageNet comparisons, and reaches the corresponding pixel-matching targets
    in fewer generator updates.
    \item We demonstrate high-quality few-step pixel-space text-to-image
    generation with latent matching across multiple student architectures and
    teacher latent spaces.
\end{itemize}

\section{Related Work}
\label{sec:related-work}

\noindent \textbf{Direct pixel-space generation.}
Early diffusion systems generated RGB images directly, including
classifier-guided ImageNet models and cascaded text-to-image or video systems
\cite{dhariwal2021adm,ho2022cascaded,nichol2022glide,saharia2022imagen,
balaji2022ediffi,ho2022video,ho2022imagenvideo,singer2023makeavideo}. Recent
work scales direct-pixel generation with recurrent,
multi-resolution, and hierarchical architectures, clean-image prediction, and
asymmetric parameterizations
\cite{jabri2023rin,hoogeboom2023simple,crowson2024hdit,gu2024matryoshka,
hoogeboom2025sid2,chen2025pixelflow,li2026jit,chen2026asymmetric,
chen2026dip,yu2026pixeldit,ma2026deco,nvidia2024edify,vallaeys2026ujit}.
Complementary work studies perceptual or representation guidance and semantic
waypoints~\cite{lei2026epg,ma2026pixelgen,shin2026pixelrepa,wang2026wit,
pan2026representation}. Although latent diffusion reduces denoising cost by
operating on compressed autoencoder representations
\cite{rombach2022ldm}, these results establish direct RGB generation as a
viable output formulation. We instead study how a pixel-space student can
inherit distribution-matching supervision from a latent teacher.

\noindent \textbf{Latent-to-pixel transfer.}
Rather than train in pixel space from scratch, L2P replaces a pretrained latent
model's interfaces while freezing its intermediate denoiser blocks, and
AsymFlow aligns a low-rank pixel subspace with the latent space
\cite{chen2026l2p,chen2026asymmetric}. Large-scale studies and Iris further
examine latent initialization for pixel-space post-training, while CrossFlow
maps latent inputs directly to pixels
\cite{jiang2026pixeltext,cai2026iris,wang2026crossflow}. These methods convert
the generator itself to produce pixels. Our approach instead keeps the latent
teacher and critic intact and back propagates their supervision through a
frozen encoder to a separate pixel-space student.

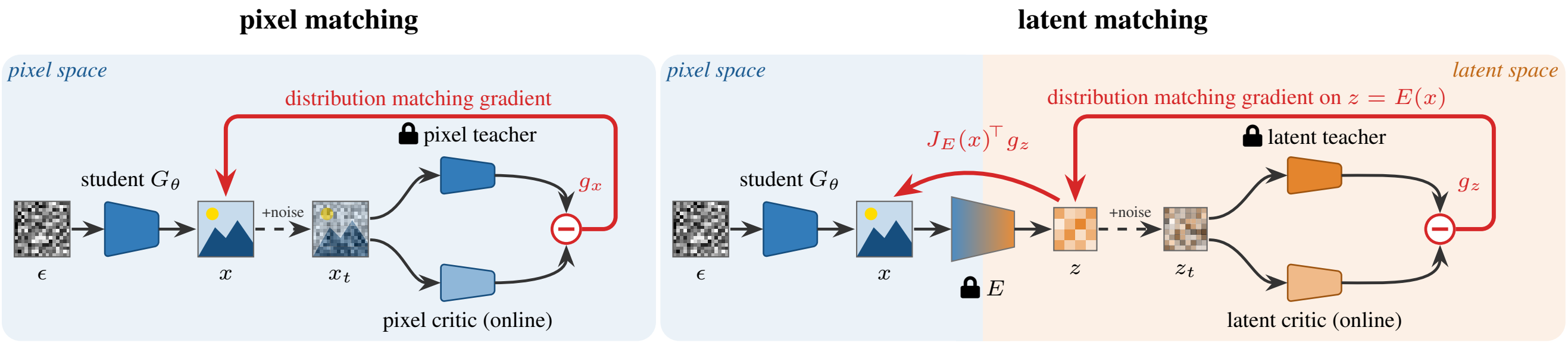
\begin{figure*}[!t]
    \centering
    \resizebox{\textwidth}{!}{\begin{tikzpicture}[
    font=\scriptsize,
    >=Stealth,
    flow/.style={->, line width=0.9pt, black!80},
    noisy/.style={->, line width=0.9pt, black!80, dashed},
    grad/.style={->, line width=1.3pt, gradred},
    fakeupd/.style={->, line width=0.9pt, black!55, densely dotted},
    lossbox/.style={draw, rounded corners=2pt, line width=0.7pt, align=center,
        inner sep=2pt, minimum height=0.42cm},
    region/.style={rounded corners=6pt, line width=0pt},
    spacelabel/.style={font=\scriptsize\itshape, anchor=north west, inner sep=2pt},
    title/.style={font=\small\bfseries, anchor=south},
]

\newcommand{\netbars}[5]{%
  \filldraw[fill=#3, fill opacity=#5, draw=#3!70!black, line width=0.5pt,
            rounded corners=1.2pt]
    ({#1-0.3},{#2-0.5*#4}) -- ({#1+0.3},{#2-0.3*#4})
    -- ({#1+0.3},{#2+0.3*#4}) -- ({#1-0.3},{#2+0.5*#4}) -- cycle;
}
\newcommand{\noisecells}[6]{%
  \pgfmathsetseed{#6}%
  \foreach \i in {1,...,#1} {%
    \foreach \j in {1,...,#1} {%
      \pgfmathtruncatemacro{\g}{max(0,min(100,50+38*(rnd+rnd+rnd-1.5)*1.4))}%
      \fill[black!\g!white, opacity=#5]
        ({#3+(\i-1)*#2},{#4+(\j-1)*#2}) rectangle ++(#2,#2);
    }%
  }%
}
\newcommand{\imgglyph}[3]{%
  \begin{scope}[shift={(#1,#2)}]
    \fill[white] (-0.31,-0.31) rectangle (0.31,0.31);
    \fill[pixblue!25] (-0.31,-0.31) rectangle (0.31,0.31);
    \fill[pixblue!70!black] (-0.31,-0.31) -- (-0.12,0.02) -- (0.0,-0.10)
        -- (0.13,0.10) -- (0.31,-0.16) -- (0.31,-0.31) -- cycle;
    \fill[yellow!80!orange] (-0.15,0.17) circle (0.07);
    \ifnum#3=1
      \noisecells{16}{0.03875}{-0.31}{-0.31}{0.45}{7}
    \fi
    \draw[black!60, line width=0.4pt] (-0.31,-0.31) rectangle (0.31,0.31);
  \end{scope}%
}
\newcommand{\latglyph}[3]{%
  \begin{scope}[shift={(#1,#2)}]
    \foreach \i/\j/\o in {0/0/70,1/0/35,2/0/55,3/0/20, 0/1/40,1/1/80,2/1/25,3/1/60,
                         0/2/15,1/2/50,2/2/90,3/2/45, 0/3/65,1/3/30,2/3/40,3/3/75} {%
      \fill[latorange!\o!white] (-0.24+0.12*\i,-0.24+0.12*\j) rectangle ++(0.12,0.12);
    }%
    \ifnum#3=1
      \noisecells{8}{0.06}{-0.24}{-0.24}{0.55}{11}
    \fi
    \draw[black!60, line width=0.4pt] (-0.24,-0.24) rectangle (0.24,0.24);
  \end{scope}%
}
\newcommand{\noiseglyph}[2]{%
  \begin{scope}[shift={(#1,#2)}]
    \noisecells{16}{0.03875}{-0.31}{-0.31}{1}{3}
    \draw[black!60, line width=0.4pt] (-0.31,-0.31) rectangle (0.31,0.31);
  \end{scope}%
}
\newcommand{\diffop}[2]{%
  \node[circle, draw=gradred, line width=0.9pt, fill=white, inner sep=0pt,
        minimum size=0.32cm] (#2) at (#1) {};
  \draw[gradred, line width=1.1pt] ($(#2)+(-0.09,0)$) -- ($(#2)+(0.09,0)$);
}

\begin{scope}[shift={(0,0)}]
  \fill[region, pixblue!9] (-0.45,-1.25) rectangle (6.85,1.95);
  \node[spacelabel, text=pixblue!80!black] at (-0.45,1.95) {pixel space};
  \node[title] at (3.2,2.05) {pixel matching};

  \noiseglyph{0}{0}
  \node[below] at (0,-0.33) {$\epsilon$};
  \netbars{1.0}{0}{pixblue}{0.62}{0.9}
  \node[above, align=center] at (1.0,0.32) {student $G_\theta$};
  \imgglyph{2.05}{0}{0}
  \node[below] at (2.05,-0.33) {$x$};
  \draw[flow] (0.33,0) -- (0.7,0);
  \draw[flow] (1.3,0) -- (1.72,0);

  \draw[noisy] (2.38,0) -- node[above=0pt, font=\tiny] {+noise} (3.0,0);
  \imgglyph{3.33}{0}{1}
  \node[below] at (3.33,-0.33) {$x_t$};

  \netbars{4.75}{0.6}{pixblue}{0.42}{0.9}
  \node[anchor=south] at (4.75,0.8) {\faLock\ pixel teacher};
  \netbars{4.75}{-0.6}{pixblue}{0.42}{0.45}
  \node[anchor=north] at (4.75,-0.8) {pixel critic (online)};
  \draw[flow] (3.66,0.12) to[out=0, in=180] (4.45,0.6);
  \draw[flow] (3.66,-0.12) to[out=0, in=180] (4.45,-0.6);

  \diffop{5.85,0}{dL}
  \draw[flow] (5.05,0.6) to[out=0, in=90] (dL.north);
  \draw[flow] (5.05,-0.6) to[out=0, in=-90] (dL.south);
  \draw[grad, rounded corners=4pt] (dL.east) -- (6.4,0) -- (6.4,1.26) -- (2.05,1.26) -- (2.05,0.36);
  \node[left, text=gradred] at (6.4,0.5) {$g_x$};
  \node[text=gradred, anchor=south] at (4.2,1.21) {distribution matching gradient};
\end{scope}

\begin{scope}[shift={(7.35,0)}]
  \fill[region, pixblue!9] (-0.45,-1.25) rectangle (3.15,1.95);
  \fill[region, latorange!11] (3.15,-1.25) rectangle (9.65,1.95);
  \fill[pixblue!9] (2.85,-1.25) rectangle (3.15,1.95);
  \fill[latorange!11] (3.15,-1.25) rectangle (3.45,1.95);
  \node[spacelabel, text=pixblue!80!black] at (-0.45,1.95) {pixel space};
  \node[spacelabel, text=latorange!85!black, anchor=north east] at (9.65,1.95) {latent space};
  \node[title] at (4.6,2.05) {latent matching};

  \noiseglyph{0}{0}
  \node[below] at (0,-0.33) {$\epsilon$};
  \netbars{1.0}{0}{pixblue}{0.62}{0.9}
  \node[above, align=center] at (1.0,0.32) {student $G_\theta$};
  \imgglyph{2.05}{0}{0}
  \node[below] at (2.05,-0.33) {$x$};
  \draw[flow] (0.33,0) -- (0.7,0);
  \draw[flow] (1.3,0) -- (1.72,0);

  \shade[left color=pixblue!80, right color=latorange!85, draw=black!55, line width=0.4pt]
      (2.8,-0.36) -- (3.5,-0.2) -- (3.5,0.2) -- (2.8,0.36) -- cycle;
  \node[anchor=north, align=center] at (3.15,-0.4) {\faLock\ $E$};
  \draw[flow] (2.38,0) -- (2.8,0);
  \draw[flow] (3.5,0) -- (3.92,0);
  \latglyph{4.18}{0}{0}
  \node[below] at (4.18,-0.27) {$z$};

  \draw[noisy] (4.44,0) -- node[above=0pt, font=\tiny] {+noise} (5.15,0);
  \latglyph{5.41}{0}{1}
  \node[below] at (5.41,-0.27) {$z_t$};

  \netbars{6.85}{0.6}{latorange}{0.42}{0.9}
  \node[anchor=south] at (6.85,0.8) {\faLock\ latent teacher};
  \netbars{6.85}{-0.6}{latorange}{0.42}{0.45}
  \node[anchor=north] at (6.85,-0.8) {latent critic (online)};
  \draw[flow] (5.67,0.12) to[out=0, in=180] (6.55,0.6);
  \draw[flow] (5.67,-0.12) to[out=0, in=180] (6.55,-0.6);

  \diffop{8.25,0}{dR}
  \draw[flow] (7.15,0.6) to[out=0, in=90] (dR.north);
  \draw[flow] (7.15,-0.6) to[out=0, in=-90] (dR.south);
  \draw[grad, rounded corners=4pt] (dR.east) -- (8.85,0) -- (8.85,1.26) -- (4.18,1.26) -- (4.18,0.3);
  \node[left, text=gradred] at (8.85,0.5) {$g_z$};
  \node[text=gradred, anchor=south] at (6.1,1.21) {distribution matching gradient on $z=E(x)$};
  \draw[grad] (4.0,0.32) to[out=150, in=30] (2.15,0.36);
  \node[text=gradred, anchor=south] at (3.1,0.74) {$J_E(x)^{\!\top} g_z$};
\end{scope}
\end{tikzpicture}}
    \caption{\textbf{Pixel matching and latent matching.}
    \emph{Left:} pixel matching noises the pixel-space student's output and
    compares a frozen pixel teacher with an online pixel critic.
    \emph{Right:} latent matching keeps the student's output in pixel space but encodes it with
    the frozen encoder $E$ of a latent teacher. The latent teacher and the
    latent critic produce the score difference $g_z$ on $z=E(x)$; back
    propagation through $E$ delivers it to the student as $J_E(x)^{\top} g_z$.}
    \label{fig:dmd-vs-l2p}
\end{figure*}

\noindent \textbf{Few-step diffusion distillation.}
Step distillation compresses multi-step teachers into few-step students for
fast inference through trajectory-based
methods~\cite{luhman2021knowledge,salimans2022progressive,meng2023guided,
berthelot2023tract,liu2024instaflow}, consistency
methods~\cite{song2023consistency,luo2023lcm,luo2023lcmlora,kim2024ctm},
adversarial methods~\cite{sauer2024add,sauer2024ladd,lin2024sdxllightning,
kang2024diffusion2gan}, and score- or distribution-matching
methods~\cite{nguyen2024swiftbrush,zhou2024sid,yin2024dmd,yin2024dmd2,
zhou2026editstream}.
These methods typically evaluate the teacher--student correction in a common
representation. We study latent teacher--critic systems that supervise
pixel-output students, including transfers across architectures.

\noindent \textbf{Few-step distribution matching.}
DMD and DMD2 match a student distribution to a pretrained teacher using an
online critic~\cite{yin2024dmd,yin2024dmd2}. Related distribution-level methods
develop score-implicit and general-divergence objectives, moment matching,
adversarial variants, and scalable or phased few-step training
\cite{luo2024sim,xu2025fdistill,salimans2024moment,lu2025dmdx,
ge2026senseflow,fan2026phaseddmd}. Pixel MeanFlow trains a one-step latent-free
generator directly, while DMA$^2$ adapts DMD to native pixel-space teachers
with RGB-specific objectives~\cite{lu2026pmf,lin2026dma2}.
Cross-Space Distillation handles mismatched VAEs and latent resolutions by
learning a compact latent mapper~\cite{nguyen2026crossspace}. Magic3D back
propagates latent SDS through an LDM encoder to an RGB-producing
renderer~\cite{lin2023magic3d}; more generally, score-based supervision can
propagate through differentiable generators or
maps~\cite{luo2023diffinstruct,wang2023prolificdreamer,
nguyen2024swiftbrush,wang2023sjc}. Representation Distribution Matching instead
matches generated and reference feature distributions under frozen
encoders~\cite{feng2026rdm}, without the teacher--critic score difference used
here. PixSDS identifies encoder-null pixel directions as a related cross-space
risk~\cite{skorokhodov2026pixsds}. Our experiments apply this established
encoder-backpropagation interface to few-step pixel-output DMD students, with
both the teacher and online critic remaining in latent space; no learned mapper
or intermediate pixel teacher is required.

\section{Evaluating Latent Matching for Pixel-Space Students}
\label{sec:method}
\label{sec:experiments}
\label{sec:imagenet-comparison}

Prior work reports that pixel-space flow matching can require more training than
latent-space flow matching~\cite{jiang2026pixeltext}.  We first compare native
pixel and latent DMD systems using a common evaluation
protocol, then
empirically compare encoder-mediated latent matching with native pixel matching
for the same initialized few-step pixel-space students.

\subsection{Slow DMD convergence of pixel models}
\label{sec:native-dmd-comparison}

We first compare two ImageNet-256 systems.  \emph{Native latent DMD}
uses a latent SiT-XL/2 student, teacher, and online critic; \emph{native pixel
DMD} uses the corresponding Pixel SiT-XL student, teacher, and critic.  We obtain Pixel SiT-XL by fine-tuning the latent
SiT-XL/2 checkpoint into pixel space following
AsymFlow~\cite{chen2026asymmetric}; the supplementary appendix provides the
conversion and training details. We tune training-time teacher guidance
for the evaluated configurations and use the same evaluation protocol for
both native systems. The supplementary material lists the recorded CFG
searches and actual training budgets. For each system, we report the lowest
one-step EMA FID-50K observed across its evaluated guidance settings and
training checkpoints.

\begin{table}[!t]
    \centering
    \small
    \setlength{\tabcolsep}{4.0pt}
    \caption{\textbf{Native latent DMD converges faster and achieves
    better sample quality than native pixel DMD on ImageNet-256.}}
    \label{tab:native-dmd-comparison}
    \begin{tabular}{@{}llc@{}}
        \toprule
        Native system / student & Guidance & Best FID $\downarrow$ \\
        \midrule
        Latent / Latent SiT-XL/2 & CFG 1.6 & \textbf{3.98 @ 70k} \\
        Pixel / Pixel SiT-XL & CFG 1.4 & 9.43 @ 200k \\
        \bottomrule
    \end{tabular}
\end{table}

The latent system attains substantially better quality
(\cref{tab:native-dmd-comparison}).  Its convergence is also faster in our training runs: the CFG-1.6 latent run reaches 3.98 at 70k updates,
whereas the best pixel run reaches 9.43 at 200k.  

However, native latent DMD does not solve our deployment goal: its student
still emits latents and requires a VAE decoder at inference.  This raises a
natural question: can we perform DMD matching in latent space to achieve faster
convergence and better sample quality, while keeping the few-step student's
output in pixel space?

\subsection{Latent Matching for a Pixel-Space Student}
\label{sec:method-direct}
\label{sec:method-latent}
\label{sec:imagenet-effectiveness}

We apply standard DMD to the encoded outputs of a pixel-space student. The
student generates pixels $x=G_\theta(\epsilon,c)$, which we map to $z=E(x)$
using the latent teacher's frozen VAE encoder $E$. The teacher and online critic
operate in this latent space, and their DMD correction backpropagates through
$E$ to update the pixel student. This gradient path is not new: Magic3D already
backpropagates latent SDS through a VAE encoder to optimize an RGB-producing
renderer~\cite{lin2023magic3d}. We adopt this established chain-rule mechanism
rather than claim it as a new gradient estimator. Our contribution is the
empirical study of this interface for amortized few-step direct-pixel students,
including its comparison with native pixel matching. The encoder, teacher, and
critic are used only during training; inference retains only the few-step pixel
student and requires no VAE (\cref{fig:dmd-vs-l2p}).

\begin{table}[!t]
    \centering
    \footnotesize
    \setlength{\tabcolsep}{3.0pt}
    \caption{\textbf{Latent matching yields faster convergence and a lower
    best FID for pixel-space students.}  The same advantage holds
    for both Pixel SiT and AsymJiT-H students.}
    \label{tab:imagenet-main}
    \begin{tabular}{@{}llcc@{}}
        \toprule
        Student & Matching & Best FID & \shortstack{Pixel target\\(first hit)} \\
        \midrule
        Pixel SiT-XL & Pixel matching & 9.43 @ 200k & 200k \\
                     & Latent matching & \textbf{7.15 @ 200k} & \textbf{80k (9.09)} \\
        \midrule
        AsymJiT-H/16 & Pixel matching & 4.70 @ 170k & 170k \\
                     & Latent matching & \textbf{3.95 @ 90k} & \textbf{40k (4.57)} \\
        \bottomrule
    \end{tabular}
\end{table}

\begin{figure}[!t]
    \centering
    \includegraphics[width=0.9\linewidth]{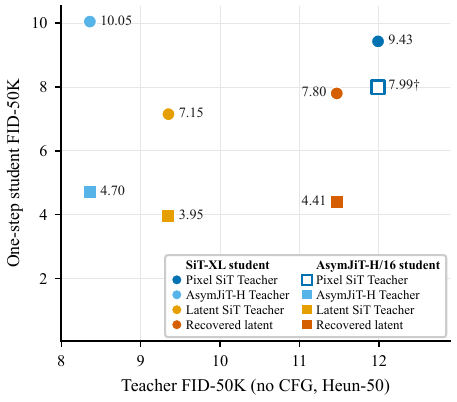}
    \caption{\textbf{Latent matching remains stronger in the paired
    native-versus-latent comparisons when teacher quality and model size favor
    pixel matching.}  Across the completed paired comparisons, latent
    teacher--critic systems produce lower-FID students than their pixel
    counterparts.  In particular, the smaller recovered-latent teacher
    produces better students than the larger, higher-quality AsymJiT-H pixel
    teacher.  Colors denote teacher--critic pairs and shapes denote student
    architectures. The open marker is an auxiliary cross-teacher control at
    its available 150k checkpoint and is not used in the paired-comparison
    claim.}
    \label{fig:teacher-student-fid}
\end{figure}

\begin{figure*}[!t]
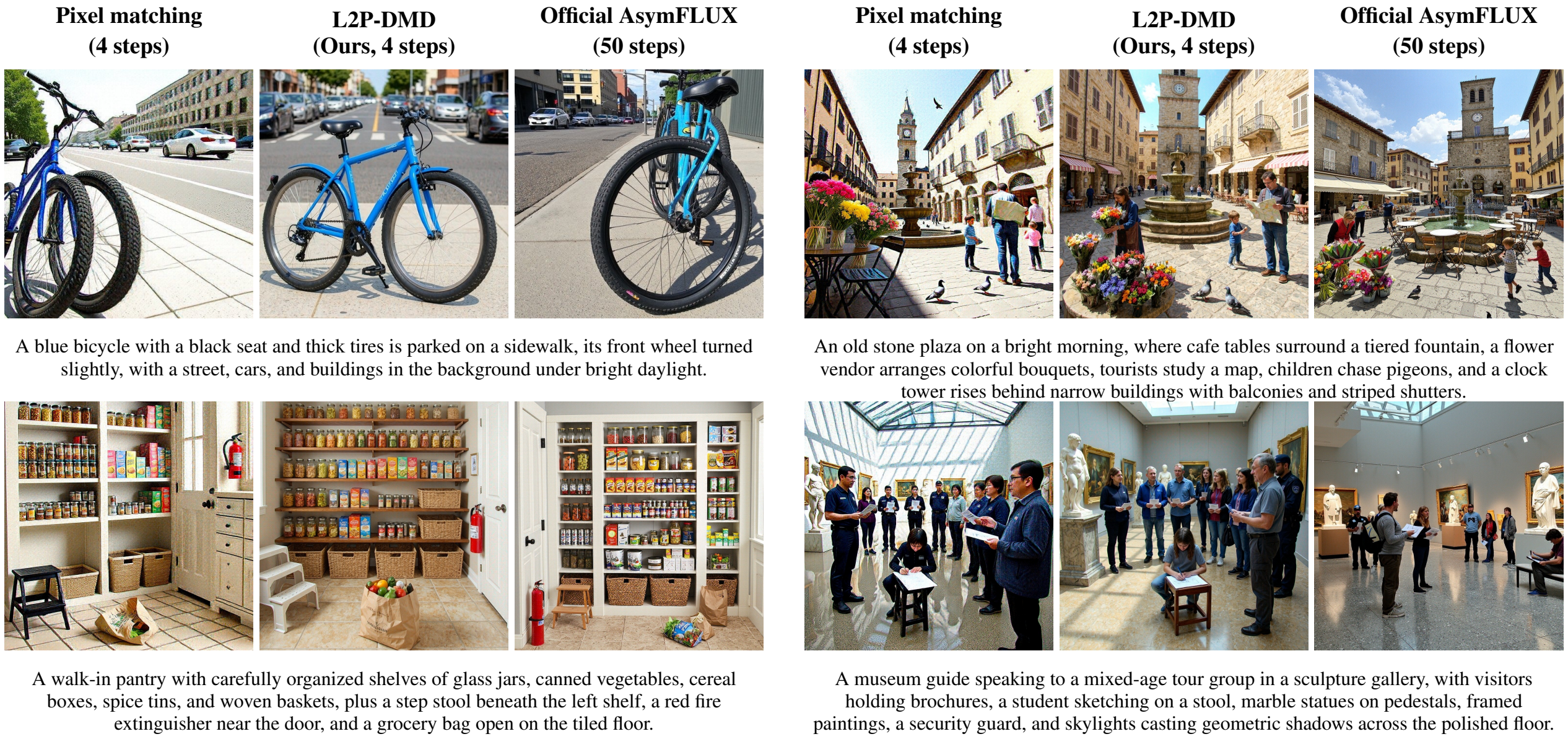

    \centering
    \begingroup
    \setlength{\tabcolsep}{0pt}
    \newcommand{\asymfluxlegacyimage}[2]{%
        \includegraphics[width=0.152\textwidth]{figure/asymflux/selected13_assets/legacy#1_#2.jpg}}
    \newcommand{\asymfluxselectedimage}[2]{%
        \includegraphics[width=0.152\textwidth]{figure/asymflux/selected13_assets/seed#1_#2.jpg}}
    \newcommand{\asymfluxprompt}[1]{%
        \parbox[t]{0.455\textwidth}{\centering\fontsize{6.0}{6.8}\selectfont
        #1}}
    \begin{tabular}{@{}c@{\hspace{0.2em}}c@{\hspace{0.2em}}c@{\hspace{1.25em}}c@{\hspace{0.2em}}c@{\hspace{0.2em}}c@{}}
        \scriptsize\bfseries\shortstack{Pixel matching\\(4 steps)} &
        \scriptsize\bfseries\shortstack{L2P-DMD\\(Ours, 4 steps)} &
        \scriptsize\bfseries\shortstack{Official AsymFLUX\\(50 steps)} &
        \scriptsize\bfseries\shortstack{Pixel matching\\(4 steps)} &
        \scriptsize\bfseries\shortstack{L2P-DMD\\(Ours, 4 steps)} &
        \scriptsize\bfseries\shortstack{Official AsymFLUX\\(50 steps)} \\[0.2em]
        \asymfluxlegacyimage{00032}{qwen_p2p} &
        \asymfluxlegacyimage{00032}{qwen_l2p} &
        \asymfluxlegacyimage{00032}{asymflux50} &
        \asymfluxselectedimage{00001}{qwen_p2p} &
        \asymfluxselectedimage{00001}{qwen_l2p} &
        \asymfluxselectedimage{00001}{asymflux50} \\[-0.15em]
        \multicolumn{3}{@{}c@{\hspace{1.25em}}}{\asymfluxprompt{A blue bicycle with a black seat and thick tires is parked on a sidewalk, its front wheel turned slightly, with a street, cars, and buildings in the background under bright daylight.}} &
        \multicolumn{3}{c@{}}{\asymfluxprompt{An old stone plaza on a bright morning, where cafe tables surround a tiered fountain, a flower vendor arranges colorful bouquets, tourists study a map, children chase pigeons, and a clock tower rises behind narrow buildings with balconies and striped shutters.}} \\[0.45em]
        \asymfluxselectedimage{00038}{qwen_p2p} &
        \asymfluxselectedimage{00038}{qwen_l2p} &
        \asymfluxselectedimage{00038}{asymflux50} &
        \asymfluxselectedimage{00022}{qwen_p2p} &
        \asymfluxselectedimage{00022}{qwen_l2p} &
        \asymfluxselectedimage{00022}{asymflux50} \\[-0.15em]
        \multicolumn{3}{@{}c@{\hspace{1.25em}}}{\asymfluxprompt{A walk-in pantry with carefully organized shelves of glass jars, canned vegetables, cereal boxes, spice tins, and woven baskets, plus a step stool beneath the left shelf, a red fire extinguisher near the door, and a grocery bag open on the tiled floor.}} &
        \multicolumn{3}{c@{}}{\asymfluxprompt{A museum guide speaking to a mixed-age tour group in a sculpture gallery, with visitors holding brochures, a student sketching on a stool, marble statues on pedestals, framed paintings, a security guard, and skylights casting geometric shadows across the polished floor.}}
    \end{tabular}
    \endgroup
    \captionsetup{skip=10pt}
    \caption{\textbf{AsymFLUX qualitative comparison under matched numeric
    seeds.} Pixel matching exhibits structural
    inconsistencies, oversaturated colors, and visible high-frequency noise,
    while latent matching produces more coherent structures and cleaner images.}
    \label{fig:asymflux-qualitative}
\end{figure*}

For Pixel SiT, latent matching improves the standard-recipe result from 9.43
to 7.15 and first beats the pixel-matching target at 80k rather than 200k
updates.  For AsymJiT-H~\cite{li2026jit,chen2026asymmetric}, it improves 4.70
to 3.95 and beats the native target at 40k rather than 170k
(\cref{tab:imagenet-main}).  The student architecture and initialization are
fixed within each pair, and all rows use the same evaluation protocol. These
paired comparisons establish
our central empirical finding: for the two direct-pixel students evaluated
here, latent matching achieves lower one-step FID and reaches the corresponding
native pixel-matching targets in fewer generator updates when each route is
evaluated at its reported tuned operating point.

\subsection{Effect of VAE Round Trips}
\label{sec:backprop-controls}

Mapping through a VAE is not lossless: round trips do not recover the original input exactly.
A DMD update computed in one
representation may therefore be distorted when transferred to the other.  We also study 
how much this mapping error can degrade DMD.

We construct a test in which the student, teacher, and online
critic all remain in latent space.  The baseline applies latent DMD directly
to the student output $z$.  In the stress test, we first decode $z$ to pixels,
re-encode the result as $\widetilde z=E(D(z))$, and then compute the same
latent-space DMD loss. 
Direct latent
matching reaches 3.98 FID at 70k updates, whereas the decoder--encoder route
reaches 4.37 at 90k, a roughly 0.4-FID degradation.  This experiment indicates that the VAE round trip
introduces a real but small penalty.  Our pixel-student route
backpropagates only through the encoder, so its penalty should be even
smaller.

\subsection{Comparison of Latent-to-Pixel Update Rules}
\label{sec:backprop-rules}

We compare three ways of delivering the latent DMD correction to the pixel
student. Let $z_0=E(x_0)$ denote the encoded student output and $g_z$ the
latent DMD score difference. The standard chain-rule update is
$J_E(x_0)^\top g_z$. It uses the same encoder-Jacobian path as latent SDS in
Magic3D~\cite{lin2023magic3d}, with the DMD teacher--critic correction providing
$g_z$. We compare this established encoder-backpropagation route with two
decoder-based alternatives. \emph{Decoded endpoints}
uses $D(z_{\mathrm{critic}})-D(z_{\mathrm{real}})$: it decodes the critic and
teacher predictions separately and then subtracts them.  \emph{Decoded step}
instead maps the score difference around the current student latent using
$D(z_0)-D(z_0-g_z)$.  Because this signal does not constrain pixel-space
directions discarded by the encoder, these directions can drift during
training.  We therefore add the weak reconstruction restore
$0.1[x_0-D(z_0)]$, which pulls the student output toward its VAE
reconstruction.

\begin{table}[!t]
    \centering
    \small
    \setlength{\tabcolsep}{4.0pt}
    \caption{\textbf{Only encoder backpropagation transfers the latent DMD
    signal stably to the pixel student.}}
    \label{tab:backprop-summary}
    \begin{tabular}{@{}lcc@{}}
        \toprule
        Backpropagation & Best FID $\downarrow$ & Outcome \\
        \midrule
        Decoded endpoints & 382.72 @ 20k & collapsed \\
        Decoded step + 0.1 restore & 47.36 @ 30k & 236.97 @ 200k \\
        Encoder VJP & \textbf{7.15 @ 200k} & stable \\
        \bottomrule
    \end{tabular}
\end{table}

Among the three transfer rules, only backpropagation through the frozen
encoder yields stable optimization, reaching 7.15 FID.  Decoded endpoints
collapse, while the decoded step with reconstruction initially improves but
eventually degrades (\cref{tab:backprop-summary}). The supplement gives
implementation-level definitions, measured current-recipe checkpoint
trajectories, and matched VAE controls.

\subsection{Controls for Teacher Quality and Size}
\label{sec:analysis}

Latent SiT is a
stronger sampler than Pixel SiT (no-CFG FID: 9.35 versus 11.99).  The student
improvement could therefore reflect better teacher quality rather than the
latent matching route.  We therefore test whether the gain persists after
controlling for teacher quality and model size.

First, we perturb the latent SiT weights and fine-tune for 1.6k updates to
construct a \emph{recovered latent} control whose no-CFG FID of 11.47 closely
matches the Pixel SiT teacher's 11.99.  With the same Pixel SiT student,
recovered latent matching reaches 7.80 FID, compared with 9.43 for native
Pixel SiT matching at their tuned operating points under the common evaluation
protocol. The supplementary material documents its construction and the
distinction between the checkpoint-selection reference and offline teacher
evaluation. Approximately matching the
teachers' unguided sampling FID therefore does not remove the downstream gap:
teacher FID alone cannot explain the advantage of the latent route.

Second, AsymJiT-H is larger (approximately
0.95B parameters versus 0.68B) with better unguided FID versus latent SiT (8.36 versus 9.35).  Nevertheless, the Pixel SiT student reaches
10.05 with the AsymJiT-H teacher and 7.15 with latent SiT.  For an
AsymJiT-H student initialized from that pixel teacher, native matching reaches
4.70 whereas latent matching reaches 3.95.  A larger pixel teacher with
better unguided FID isn't necessarily the better DMD teacher.

Finally, the recovered latent teacher has worse no-CFG FID than AsymJiT-H
(11.47 versus 8.36), yet it produces better downstream students: 7.80 versus
10.05 for the Pixel SiT student and 4.41 versus 4.70 for the AsymJiT-H
student.  Across experiments, the latent-matching advantage persists even
when pixel teachers are larger with better sampling : a
smaller latent teacher with worse no-CFG FID still produces better students
than pixel matching.  This provides further evidence that latent matching is
more effective than pixel matching in the evaluated configurations
(\cref{fig:teacher-student-fid}).

In summary, these experiments assess an established encoder-backpropagation
interface for DMD training of direct-pixel students. The central finding is
empirical: latent matching improves one-step FID over native pixel matching in
the paired ImageNet configurations evaluated here and reaches the corresponding
pixel-matching targets in fewer generator updates. Encoder backpropagation is
also the only stable update rule among the three tested under the reported
recipes. The supplement provides further diagnostics of the observed route
gap.

\FloatBarrier

\section{Text-to-Image Experiments}
\label{sec:scaling-transfer}

Our ImageNet experiments indicate that latent matching produces higher-quality students than pixel matching (while converging faster), and that
a teacher's sampling quality alone does not predict the quality of its
distilled student. In this section we establish that these observations extend to
$1024\!\times\!1024$ text-to-image generation in three increasingly
challenging settings.

\begin{enumerate}[leftmargin=*,itemsep=2pt,topsep=3pt]
    \item \textbf{Paired latent and pixel teachers.}
    Chen et al. ~\cite{chen2026asymmetric} propose a finetuning strategy that converts  pretrained latent FLUX.2 into pixel-space AsymFLUX. We use this pair to test whether an intermediate pixel space teacher is even desirable when training a pixel space student. We find that although 50-step AsymFLUX scores higher on sample quality metrics, its few-step student is worse than the student trained
    with the latent FLUX.2 as the teacher (\cref{tab:asymflux-main}), analogous to our findings on ImageNet.
    \item \textbf{A pixel teacher trained directly in RGB.}
    We additionally evaluate PixelDiT~\cite{yu2026pixeldit}, a multi-step pixel
    model trained from scratch directly in RGB space, to test whether the result
    extends beyond the paired FLUX.2/AsymFLUX setting. We find that strong
    latent FLUX teacher--critic systems surpass the evaluated native PixelDiT
    teacher--critic system, and the PixelDiT student improves monotonically
    across the FLUX.2 4B, 9B, and 32B teachers
    (\cref{tab:pixeldit-results}).
    \item \textbf{Cross-family large-to-small transfer.}
    Finally, we confirm L2P-DMD remains effective across a
    substantially larger source--target gap: from large latent FLUX
    Transformers to a compact convolutional U-Net
    (DreamLite~\cite{feng2026dreamlite}) pixel student with a different VAE
    latent space and conditioning stack. The resulting 4-step student
    (\cref{tab:dreamlite-results}) substantially improves HPSv3 over the
    official four-step latent DreamLite Mobile checkpoint while attaining
    comparable DPG-Bench performance.
\end{enumerate}

\noindent \textbf{Shared training setup.}
Across AsymFLUX, PixelDiT, and DreamLite, we ODE-initialize each
four-step pixel student from its corresponding multi-step pixel model before
DMD training. DMD samples text conditions from 100K synthetic prompts,
generated from scene graphs and independent templates with an equal mixture of
DPG-style compositional prompts and GenEval-style object-centric prompts.
We also apply a text-conditioned GAN loss in pixel space.
Full dataset construction, ODE
initialization, and model-specific hyper-parameters are provided in the
supplementary material.

\noindent \textbf{Evaluation.}
HPSv3 (Human Preference Score v3) uses a VLM trained on human preferences to
assess prompt fidelity and overall perceptual quality~\cite{ma2025hpsv3}.
DPG-Bench (Dense Prompt Graph Benchmark) measures text--image
alignment on dense prompts using dependency-aware VQA
questions~\cite{hu2024ella}. Higher is better for both metrics.

\subsection{AsymFLUX: Paired Latent and Pixel Teachers}
\label{sec:asymflux-results}

AsymFLUX~\cite{chen2026asymmetric} converts and fine-tunes a pretrained latent
FLUX.2 model into a pixel generator rather than training a pixel model from
scratch. For the paired comparison, both runs start from the same
ODE-initialized four-step AsymFLUX student, use the same training data,
optimization schedule, update budget, and evaluation protocol. Pixel matching
uses the AsymFLUX teacher--critic system, whereas latent matching uses the
corresponding FLUX.2 teacher--critic system.

\begin{table}[!t]
    \centering
    \small
    \setlength{\tabcolsep}{4.0pt}
    \caption{\textbf{With the same ODE-initialized AsymFLUX pixel student, the
    latent teacher--critic setting outperforms the pixel teacher--critic setting
    across both HPSv3 and DPG.}}
    \label{tab:asymflux-main}
    \resizebox{\columnwidth}{!}{%
    \begin{tabular}{@{}lccc@{}}
        \toprule
        Model / matching setup & Steps & HPSv3 $\uparrow$ & DPG $\uparrow$ \\
        \midrule
        Pixel matching & 4 & 14.64 & 85.66 \\
        Latent matching & 4 & \textbf{14.84} & \textbf{86.33} \\
        \midrule
        AsymFLUX.2 Klein 9B (pixel) & 50 & 14.80 & 86.61 \\
        FLUX.2 Klein 9B Base (latent) & 50 & 10.58 & 85.28 \\
        \bottomrule
    \end{tabular}
    }
\end{table}

Despite the better scores of AsymFLUX.2 Klein 9B (14.80 HPSv3 and 86.61 DPG, compared with 10.58 and 85.28 for latent FLUX.2 Klein 9B Base), it is a worse teacher.
The L2P-DMD student reaches 14.84 HPSv3 and 86.33 DPG, compared
with 14.64 and 85.66 for pixel matching (\cref{tab:asymflux-main}).
This inversion mirrors the ImageNet Pixel SiT experiment
(\cref{fig:teacher-student-fid}), where teacher sampling quality is also
insufficient to rank downstream students.

\Cref{fig:asymflux-qualitative} shows four representative comparisons for the
students trained with pixel and latent matching alongside the official
multi-step AsymFLUX.2 reference. 
Pixel matching duplicates the bicycle's front wheel, places the tourist's map
behind his back instead of in front, and collapses the pantry's grocery bag
into an implausible shape. Its images also appear oversaturated and contain
visible high-frequency speckle, most clearly in the museum example shown last.
Qualitatively, latent matching better follows the specified object counts and
spatial relations while producing fewer artifacts and more realistic textures.
Additional qualitative comparisons are provided
in the supplementary material.

\subsection{PixelDiT as a Native Pixel-Space Teacher}
\label{sec:pixeldit-results}

Next, we evaluate PixelDiT~\cite{yu2026pixeldit}, a pixel-space model trained
from scratch, to test whether the result extends to another pixel-teacher
family. Holding the PixelDiT student architecture and initialization fixed, we
compare its native pixel teacher--critic system with latent FLUX
teacher--critic systems under a common evaluation protocol.

\begin{table}[!t]
    \centering
    \footnotesize
    \setlength{\tabcolsep}{3.0pt}
    \caption{\textbf{Scaling the FLUX.2 latent teacher--critic size
    consistently improves the PixelDiT student across both HPSv3 and DPG.}}
    \label{tab:pixeldit-results}
    \resizebox{\columnwidth}{!}{%
    \begin{tabular}{@{}llccc@{}}
        \toprule
        Matching setup / system & Teacher / model & Steps & HPSv3 $\uparrow$ & DPG $\uparrow$ \\
        \midrule
        Pixel matching & PixelDiT-1.3B & 4 & 10.69 & 81.09 \\
        Latent matching & FLUX.2 Klein-base-4B & 4 & 9.50 & 78.21 \\
        Latent matching & FLUX.2 Klein-base-9B & 4 & 10.89 & 83.50 \\
        Latent matching & FLUX.2-dev 32B & 4 & \textbf{11.97} & 84.50 \\
        Latent matching & FLUX.3 Klein Base & 4 & 11.21 & \textbf{84.71} \\
        \midrule
        Official PixelDiT & - & 50 & 10.63 & 84.73 \\
        Official FLUX.2 Klein-base-4B & - & 50 & 10.51 & 82.88 \\
        Official FLUX.2 Klein-base-9B & - & 50 & 10.58 & 85.28 \\
        Official FLUX.2-dev & - & 50 & 12.14 & 86.79 \\
        Official FLUX.3 Klein Base & - & 50 & 11.91 & 86.17 \\
        \bottomrule
    \end{tabular}
    }
\end{table}

Several recently released large-scale text-to-image models operate in learned
latent spaces
\cite{minimax2026h3image,qwen2026qwenimage21,bfl2026flux3imageklein}.
This concentration of model capacity in latent space makes large latent models
a particularly valuable source of supervision for pixel generators. L2P-DMD
can use these models directly as teacher--critic systems, without first
requiring a comparably capable pixel-space teacher, while the deployed student
retains few-step, direct-RGB, VAE-free inference.

To test whether this design provides a useful scaling axis, we build on the
finding of Wu et al.~\cite{wu2026bettercritic} that jointly scaling the teacher
and critic benefits a fixed student. We keep the PixelDiT student architecture
and initialization fixed and evaluate latent FLUX.2~\cite{flux2}
teacher--critic systems at 4B, 9B, and 32B under the same evaluation protocol.
As shown in \Cref{tab:pixeldit-results}, both HPSv3 and DPG increase across
these three evaluated systems. This empirical trend motivates a practical
route toward stronger few-step pixel
generators: distill the best available large latent diffusion models rather
than constrain the teacher to the student's output representation. Among the
evaluated latent systems, FLUX.2-dev 32B achieves the highest HPSv3 (11.97),
while the more recent FLUX.3~\cite{bfl2026flux3imageklein} achieves the highest
DPG (84.71). The latter reaches 11.21 HPSv3 and essentially matches the official
50-step PixelDiT reference on DPG ($84.71$ versus $84.73$) in only four steps.

\begin{figure}[!t]
    \centering
    \begin{minipage}[c][\dimexpr\textheight-\textfloatsep-1pt\relax][c]{\columnwidth}
    \centering
    \begingroup
    \setlength{\tabcolsep}{0.8pt}
    \newcommand{\pixeldittablewidth}{%
        \dimexpr0.93\columnwidth+4\tabcolsep\relax}
    \newsavebox{\pixelditcaptionbox}
    \newcommand{\pixelditcaptionline}[1]{%
        \sbox{\pixelditcaptionbox}{\fontsize{5.2}{5.6}\selectfont #1}%
        \ifdim\wd\pixelditcaptionbox>\pixeldittablewidth
            \resizebox{\pixeldittablewidth}{!}{\usebox{\pixelditcaptionbox}}%
        \else
            \usebox{\pixelditcaptionbox}%
        \fi
    }
    \newcommand{\pixelditrow}[4]{%
        \includegraphics[width=0.31\columnwidth]{#1} &
        \includegraphics[width=0.31\columnwidth]{#2} &
        \includegraphics[width=0.31\columnwidth]{#3} \\[-0.6ex]
        \multicolumn{3}{@{}c@{}}{\makebox[\pixeldittablewidth][c]{%
            \pixelditcaptionline{#4}}} \\[0.4ex]
    }
    \newcommand{\pixelditheader}[2]{%
        \shortstack{{\scriptsize\bfseries #1}\\[-0.2ex]
        {\fontsize{4.8}{5.2}\selectfont #2}}}
    \begin{tabular}{@{}ccc@{}}
        \pixelditheader{Pixel matching}{PixelDiT \textrightarrow{} PixelDiT} &
        \pixelditheader{L2P-DMD (Ours)}{Latent FLUX.3 \textrightarrow{} PixelDiT} &
        \pixelditheader{Official PixelDiT}{(50-step)} \\[-0.2ex]
        \pixelditrow
        {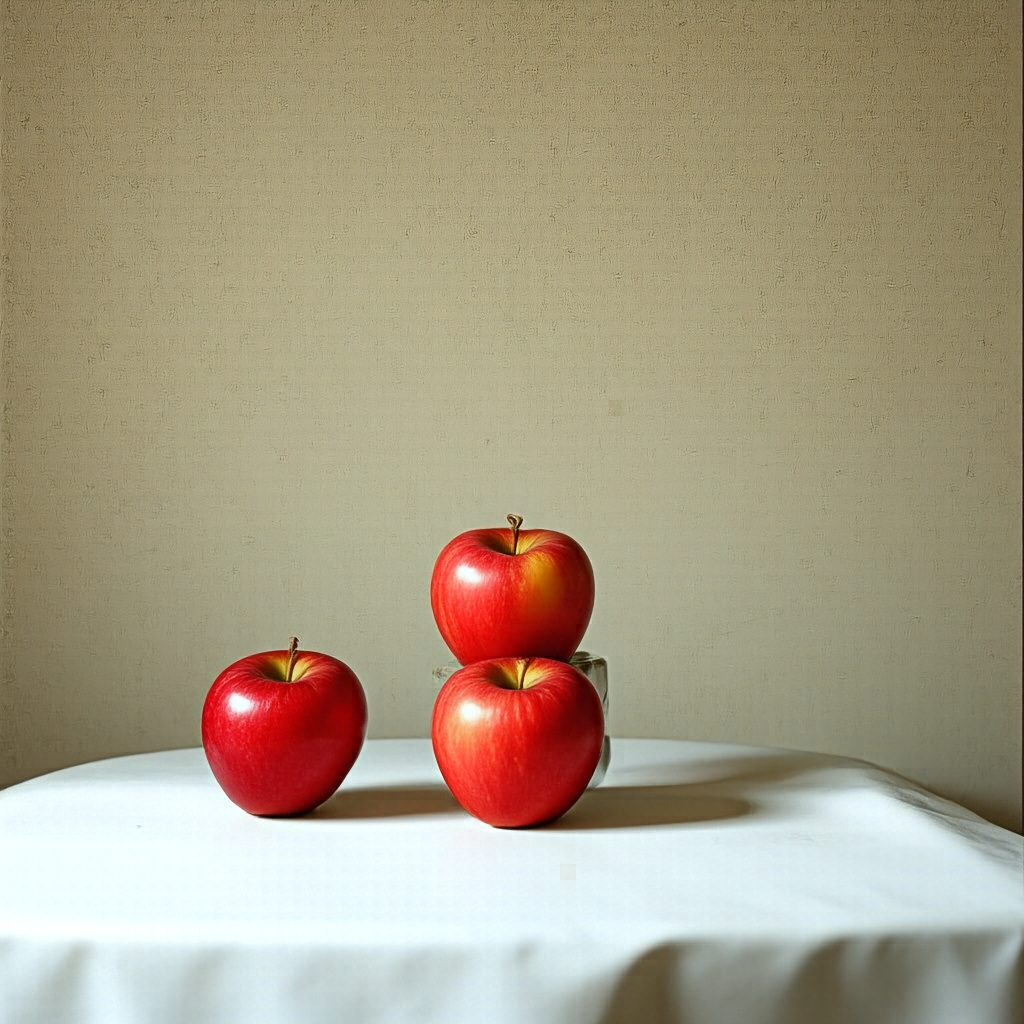}
        {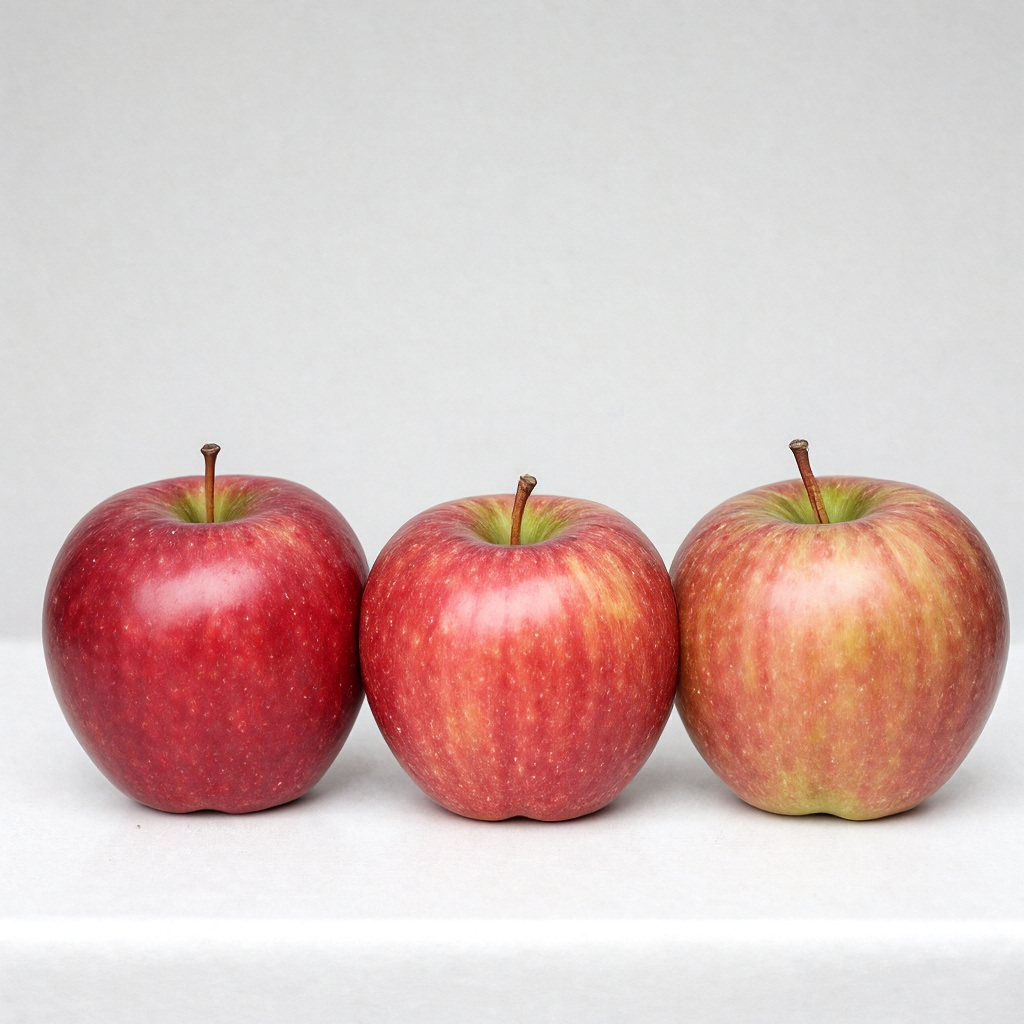}
        {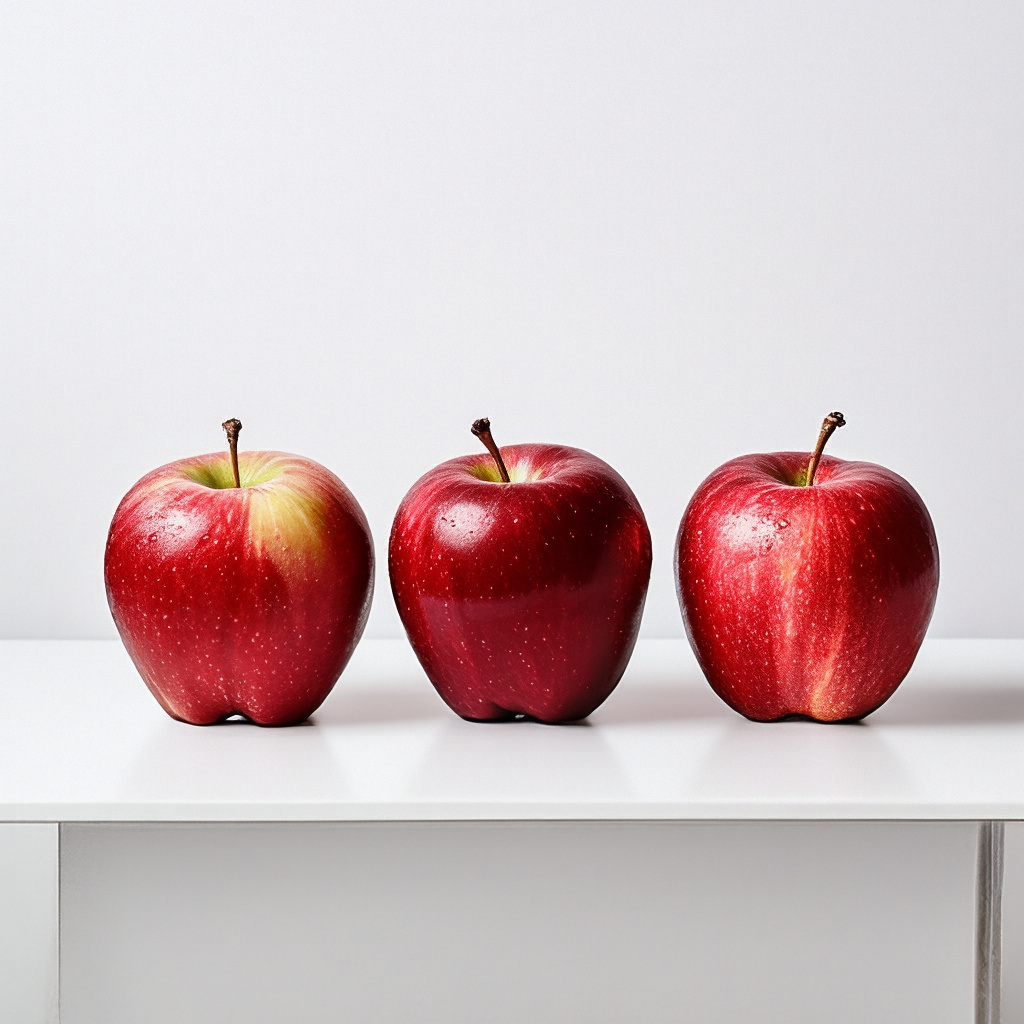}
        {Exactly three red apples arranged in a row on a white table, no other objects, studio photograph.}
        \pixelditrow
        {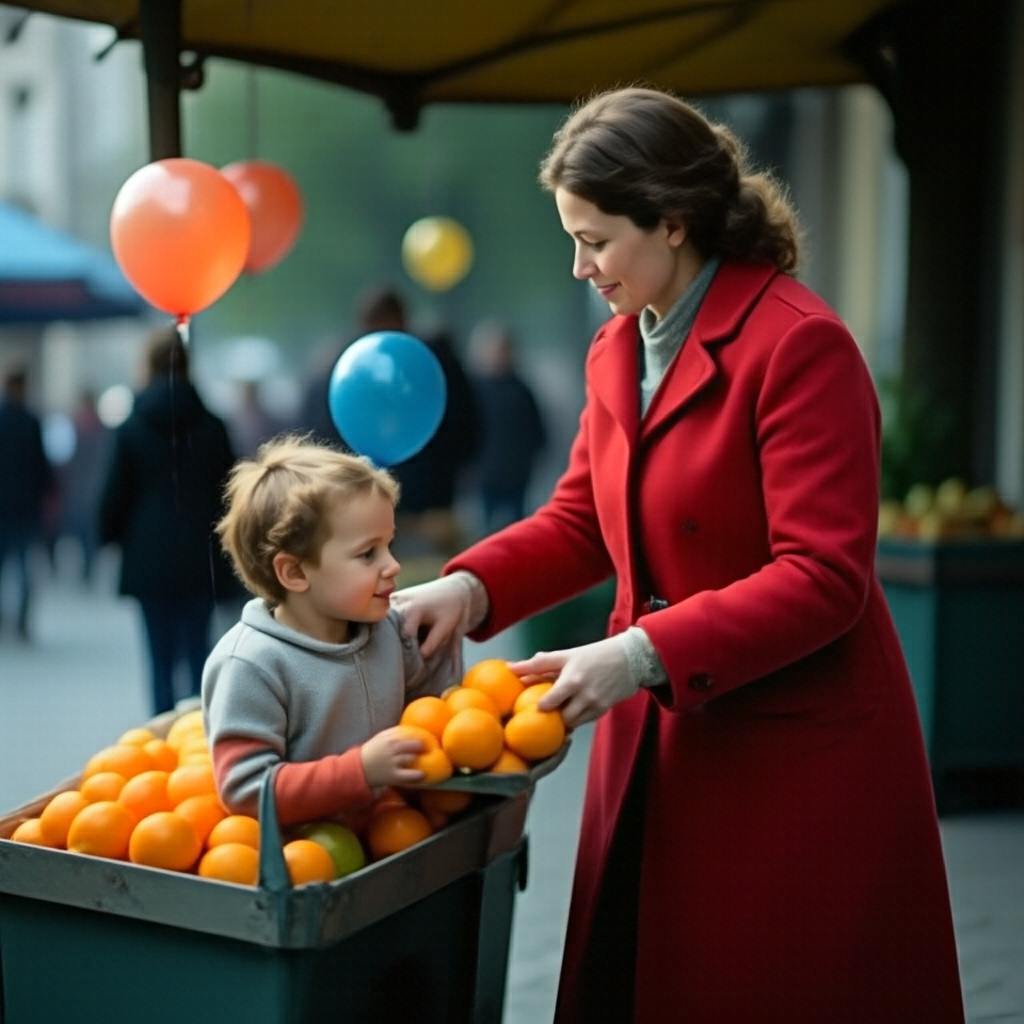}
        {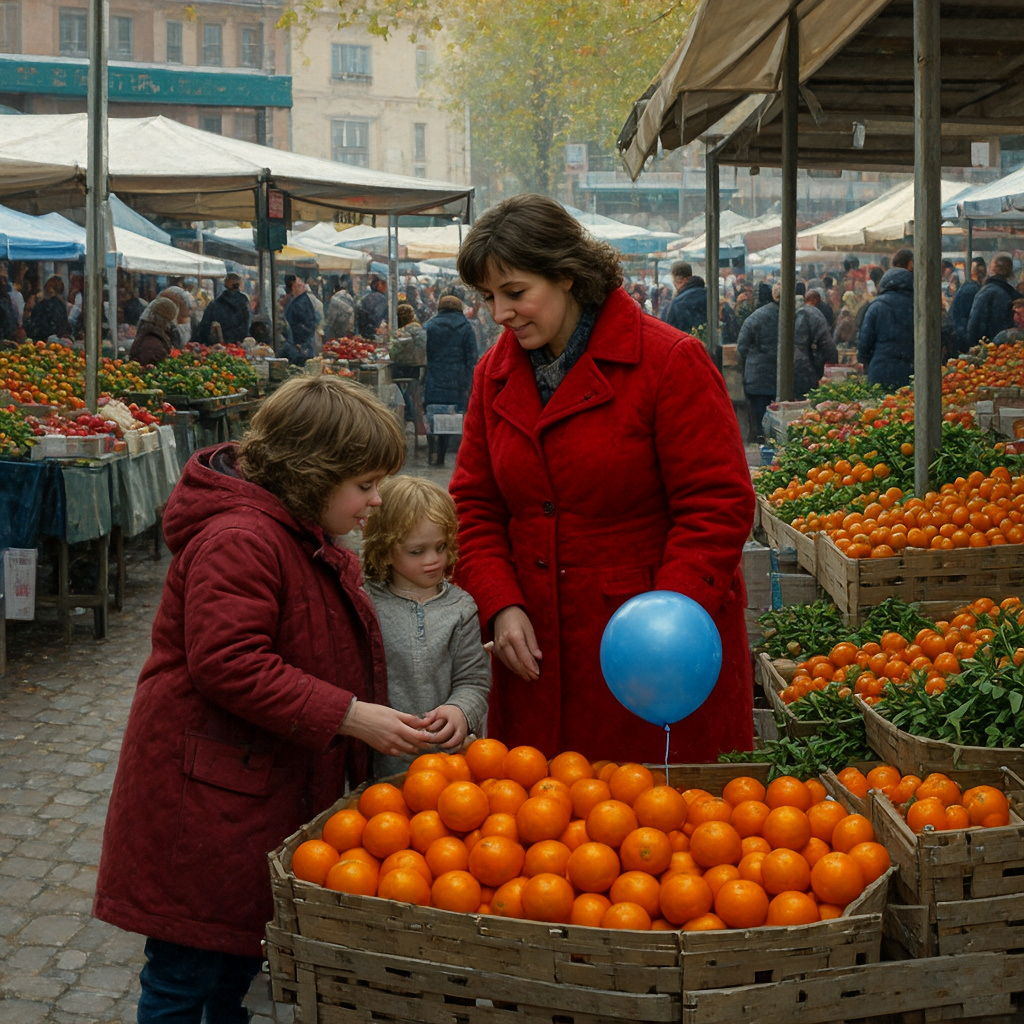}
        {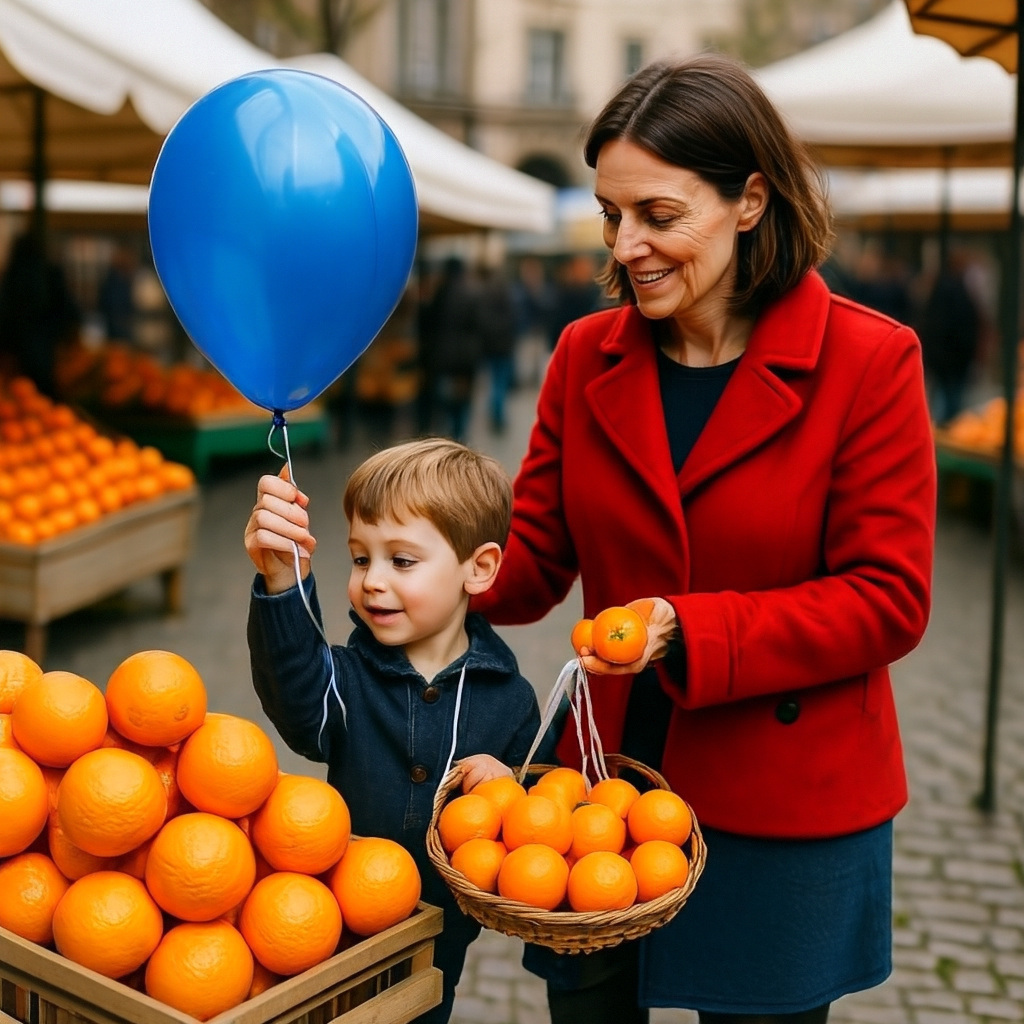}
        {An outdoor market with a woman in a red coat buying oranges and a child holding a blue balloon.}
        \pixelditrow
        {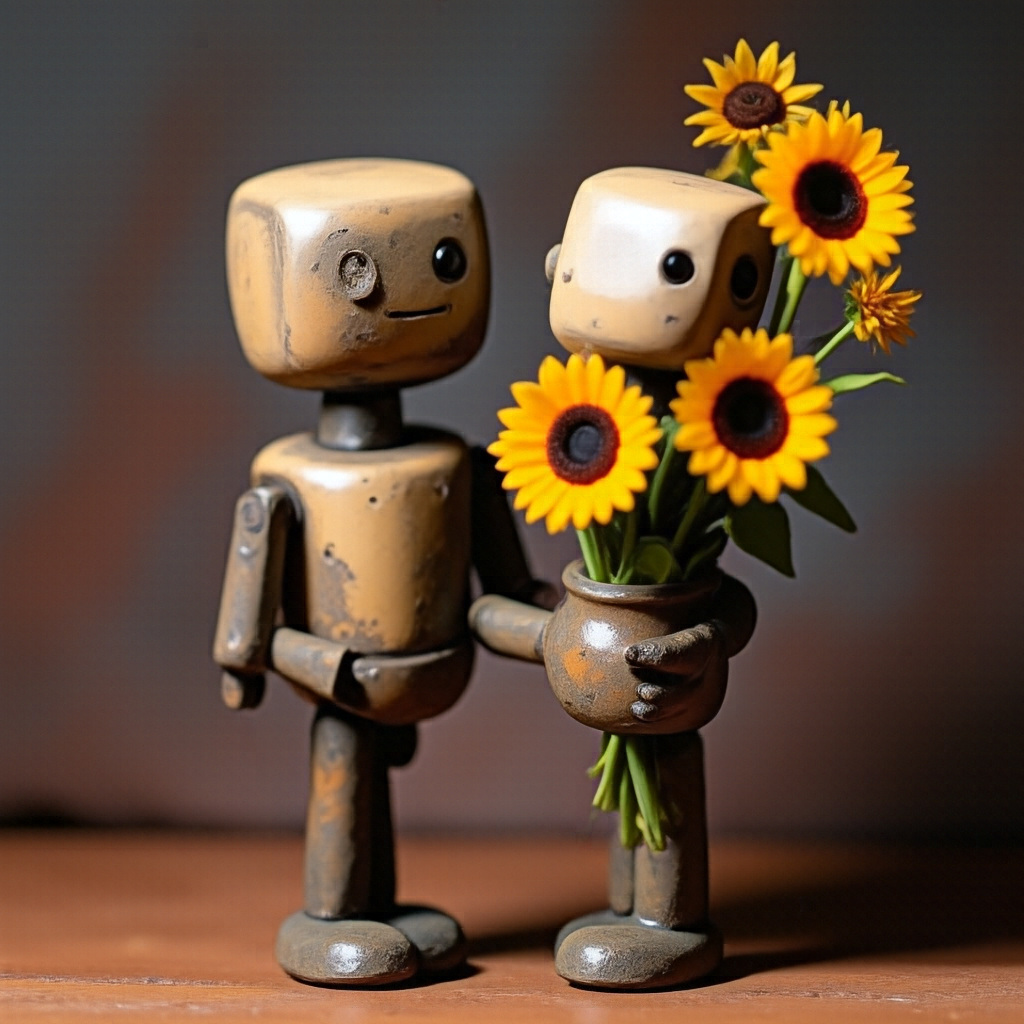}
        {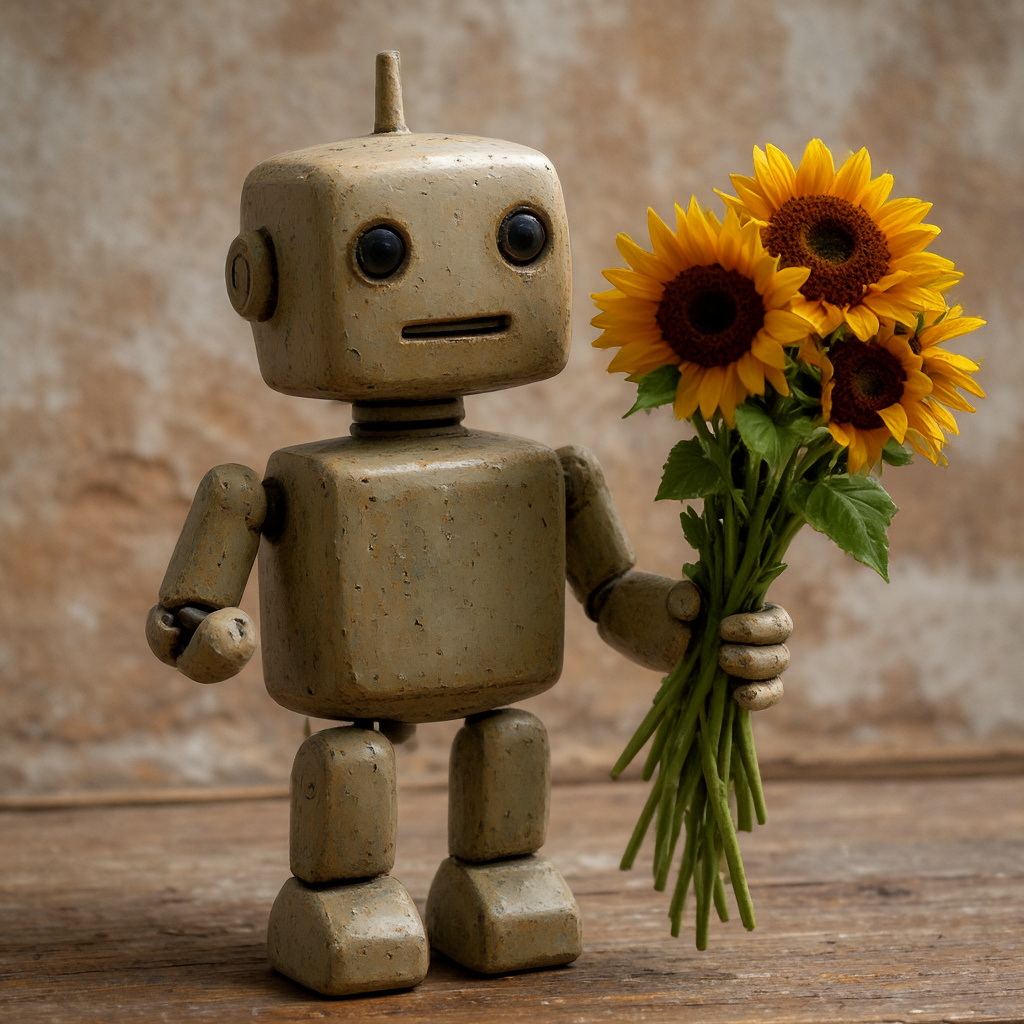}
        {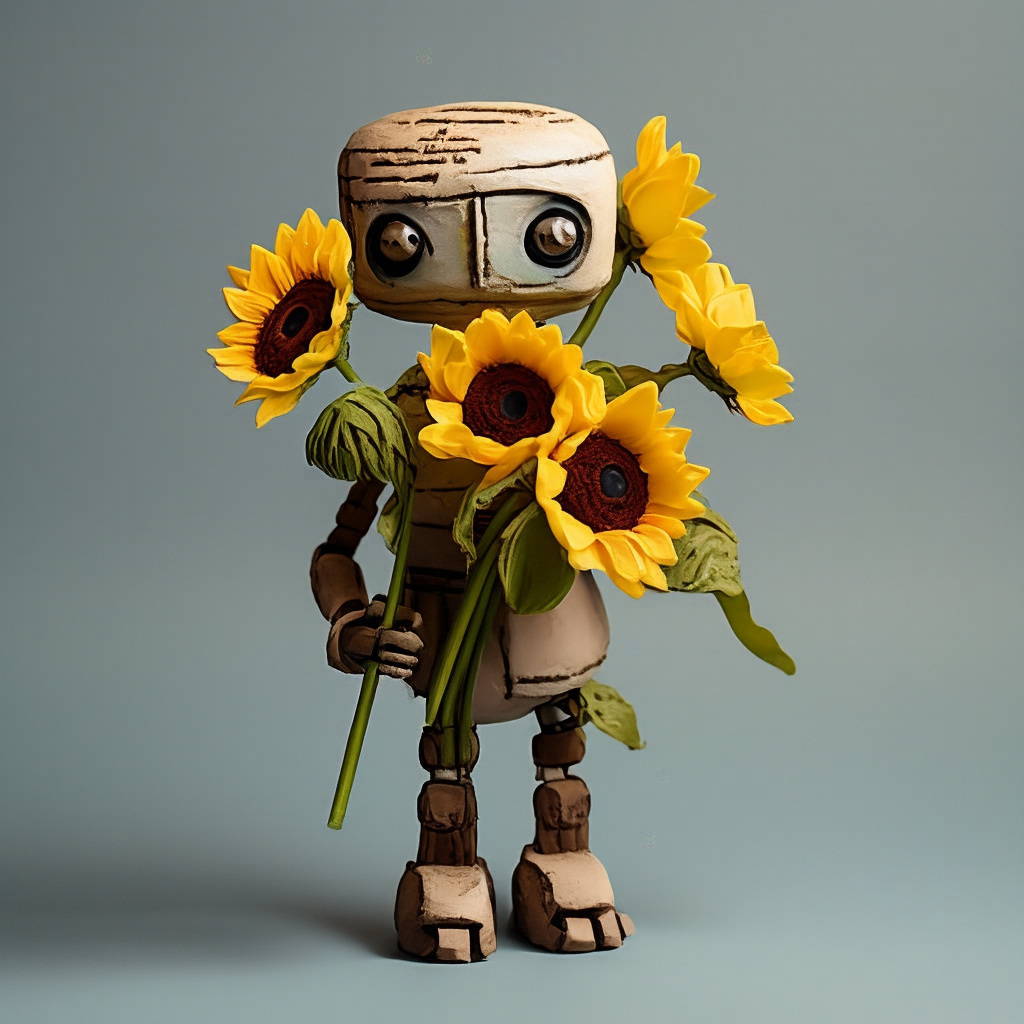}
        {A ceramic robot holding a bouquet of sunflowers, playful stop-motion style with visible clay texture.}
        \pixelditrow
        {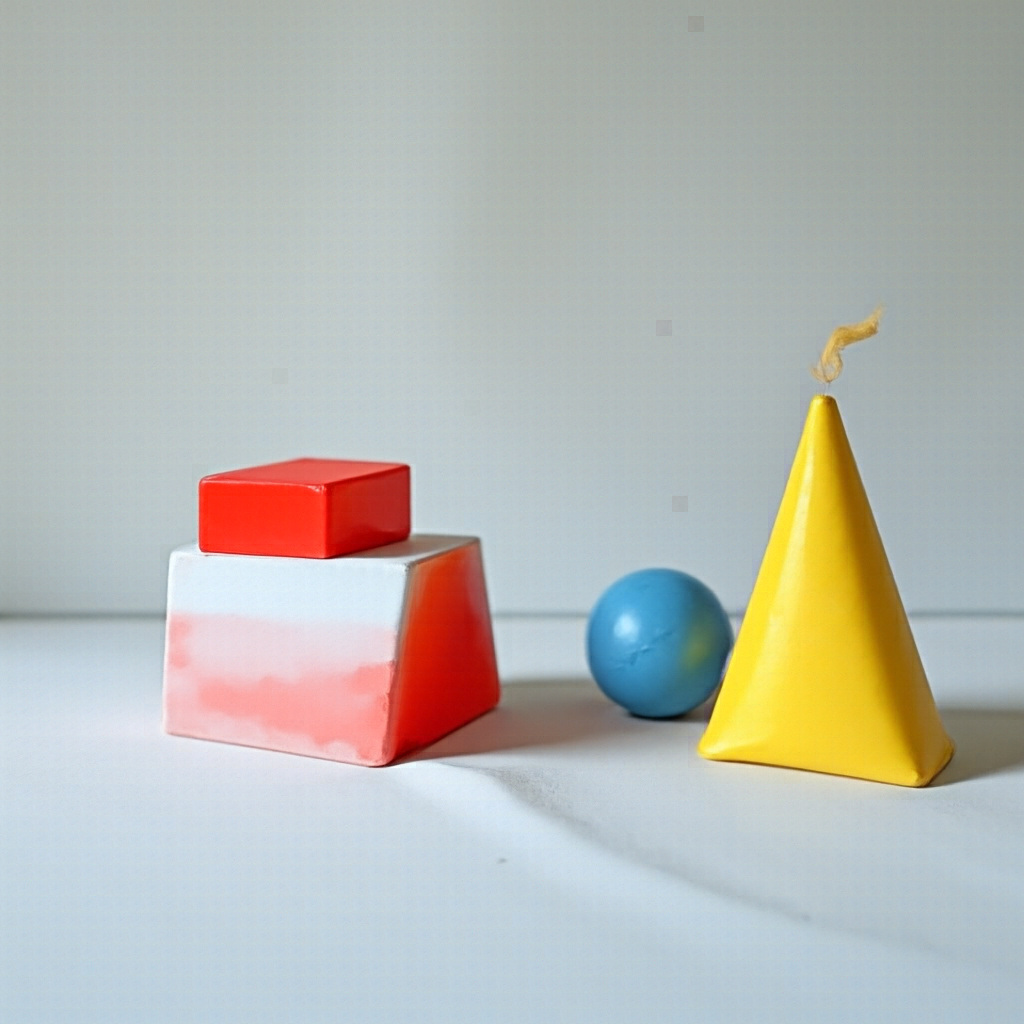}
        {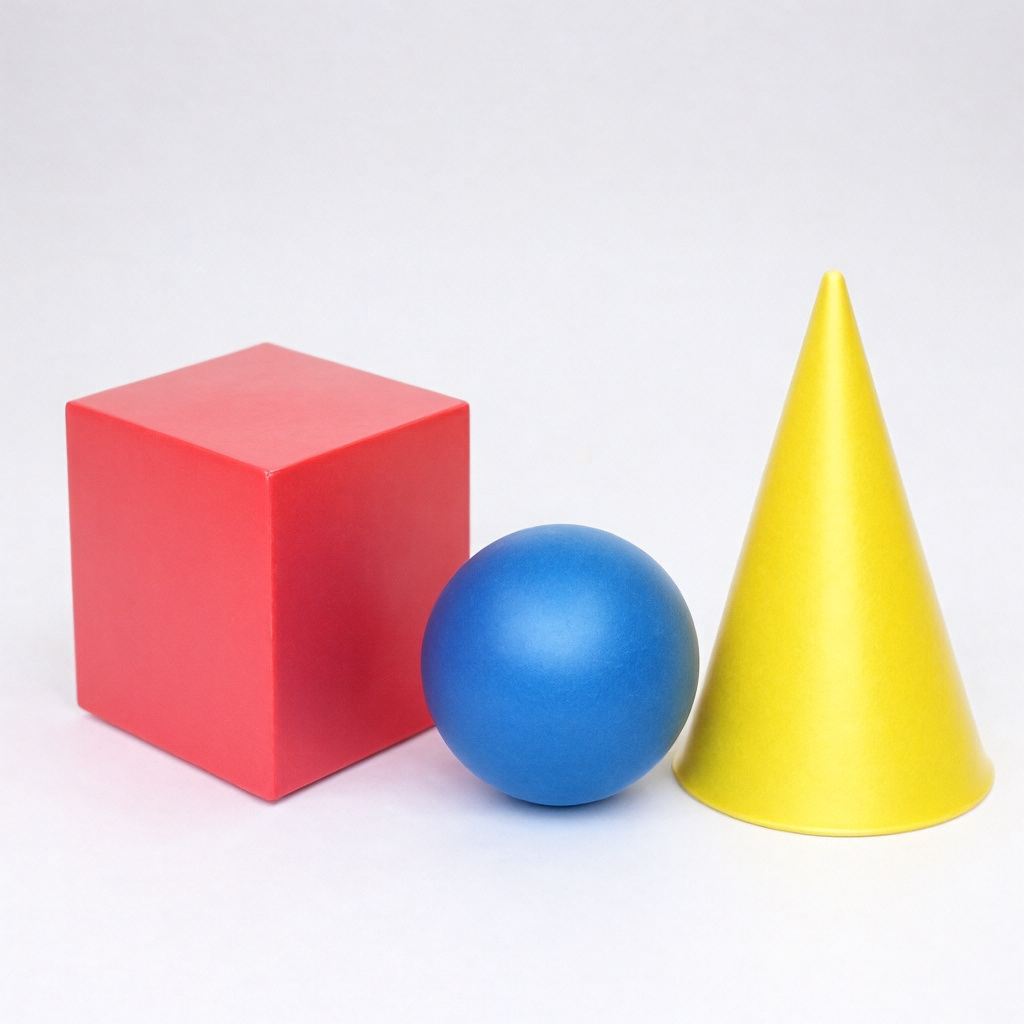}
        {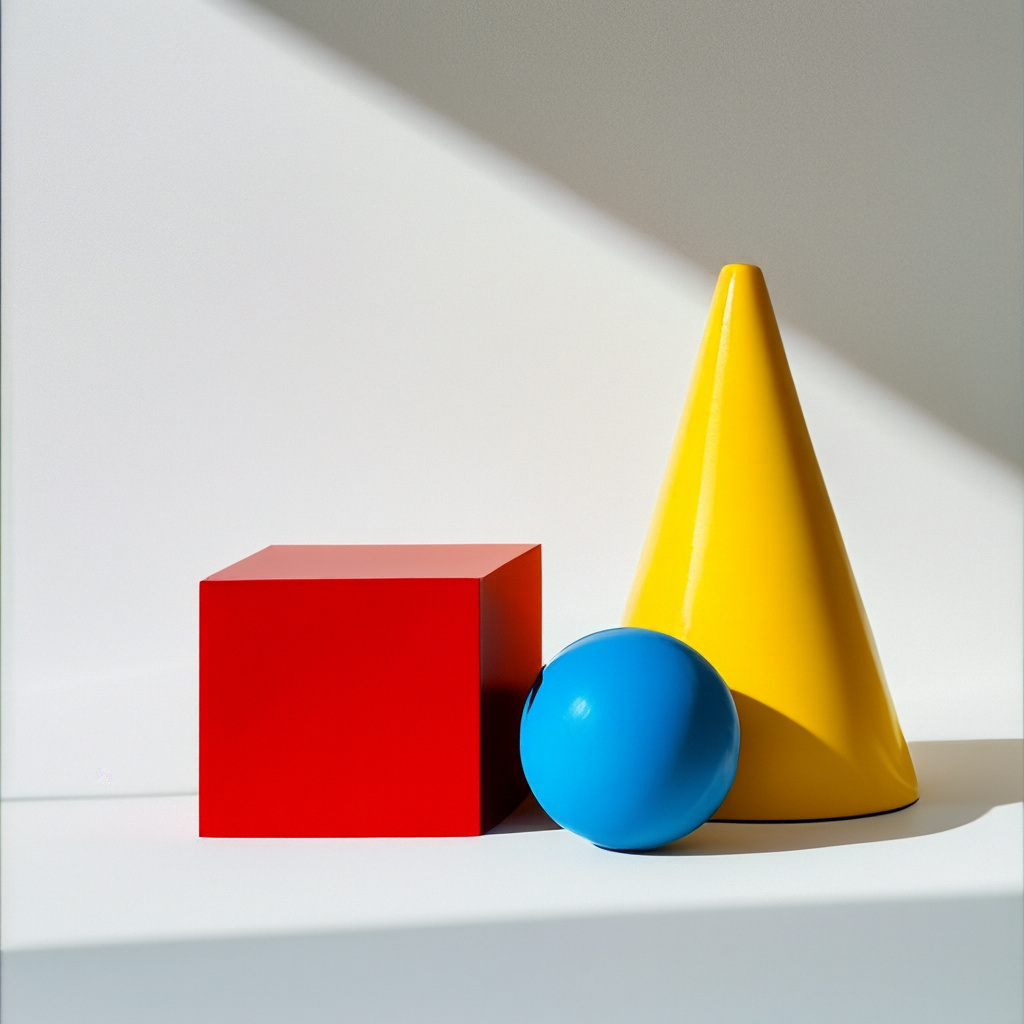}
        {A red cube, a blue sphere and a yellow cone side by side on a white surface, studio lighting.}
        \pixelditrow
        {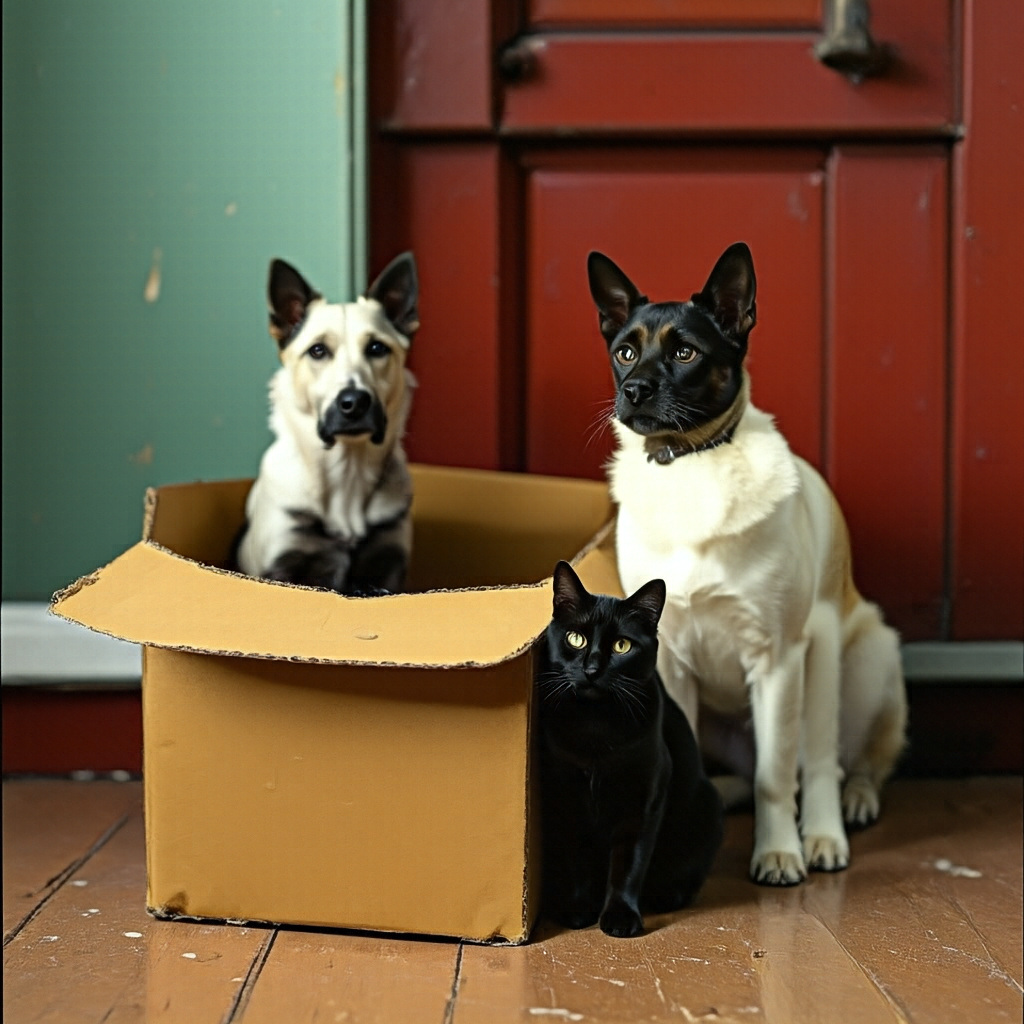}
        {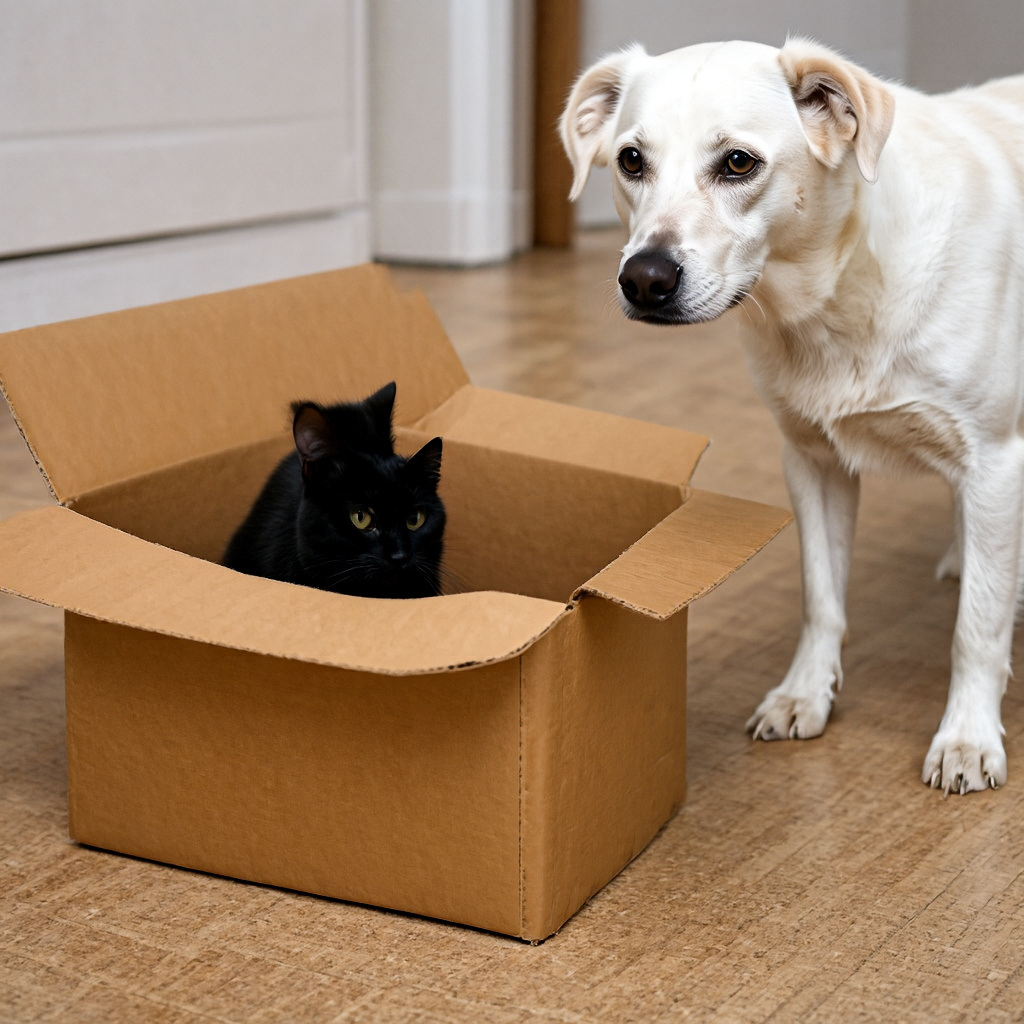}
        {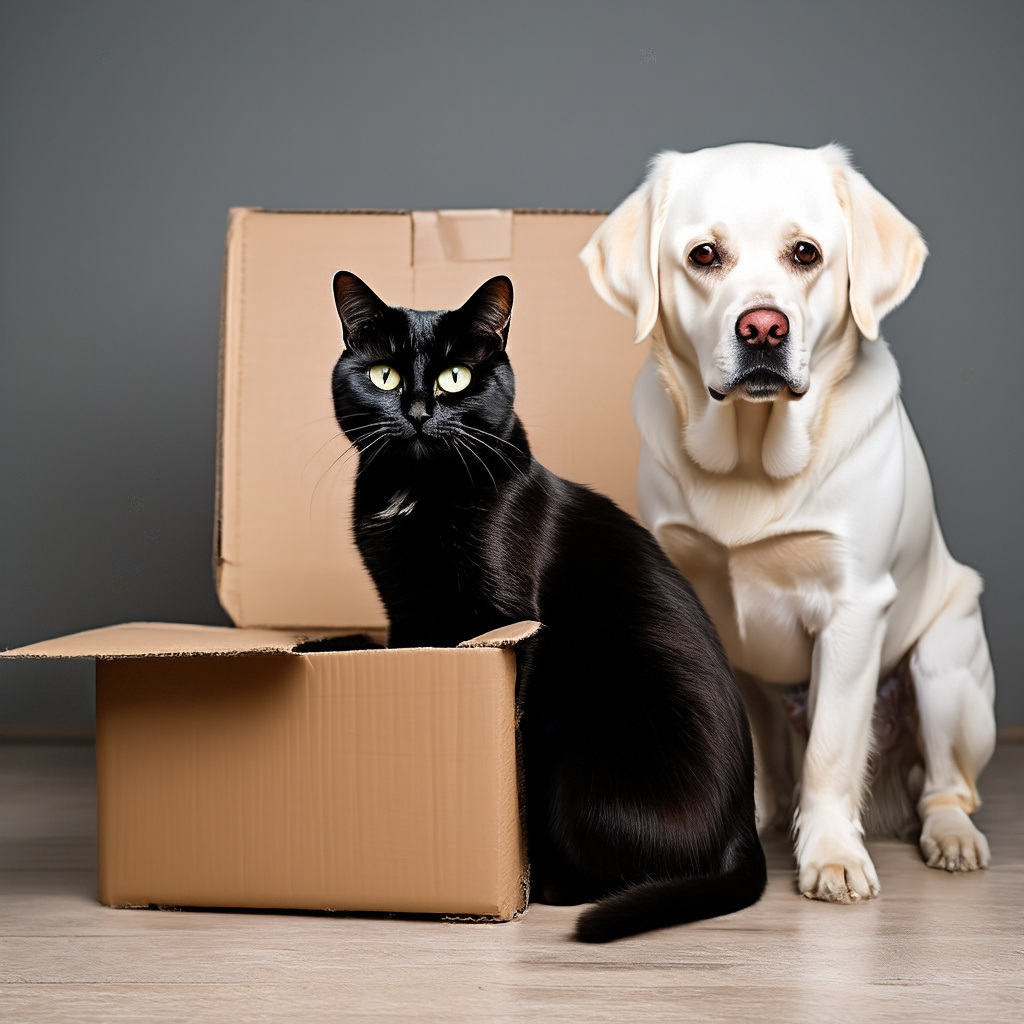}
        {A black cat sitting inside a cardboard box while a white dog stands beside the box.}
        \pixelditrow
        {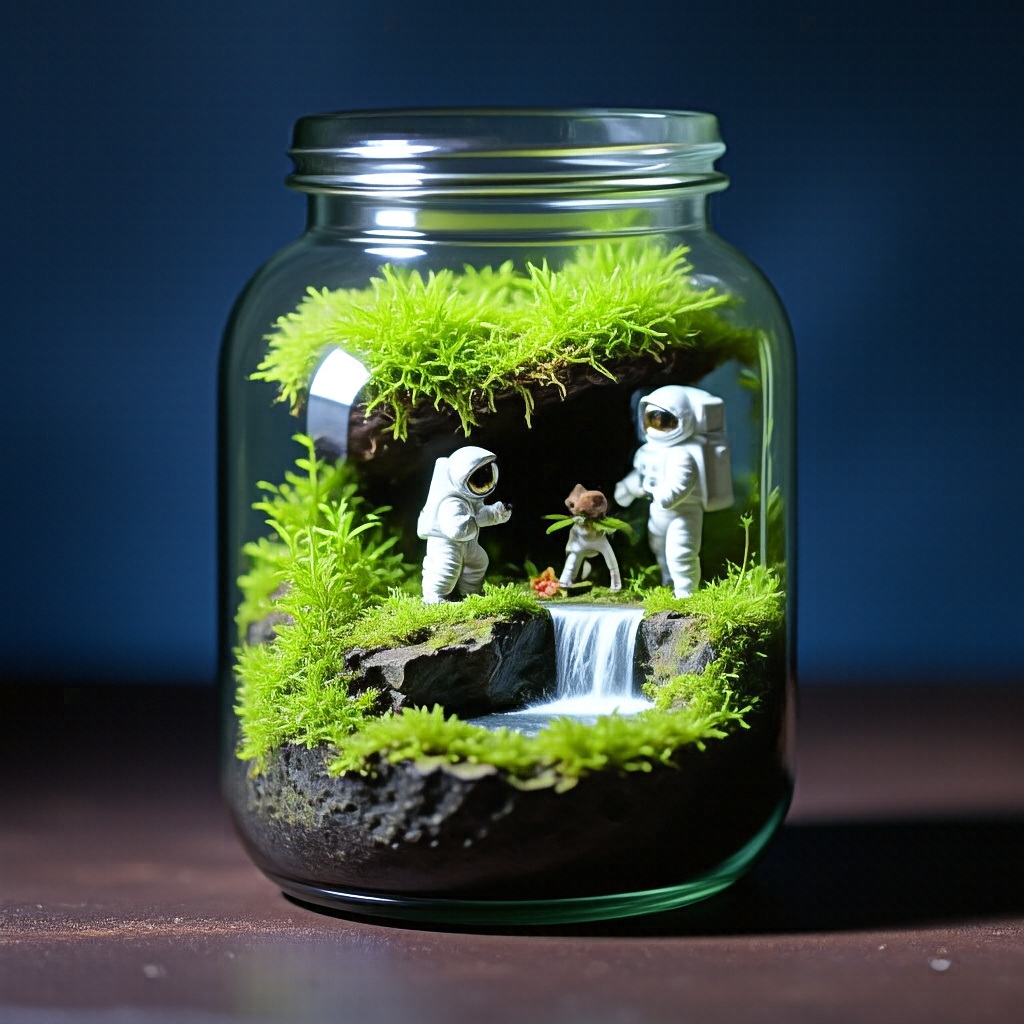}
        {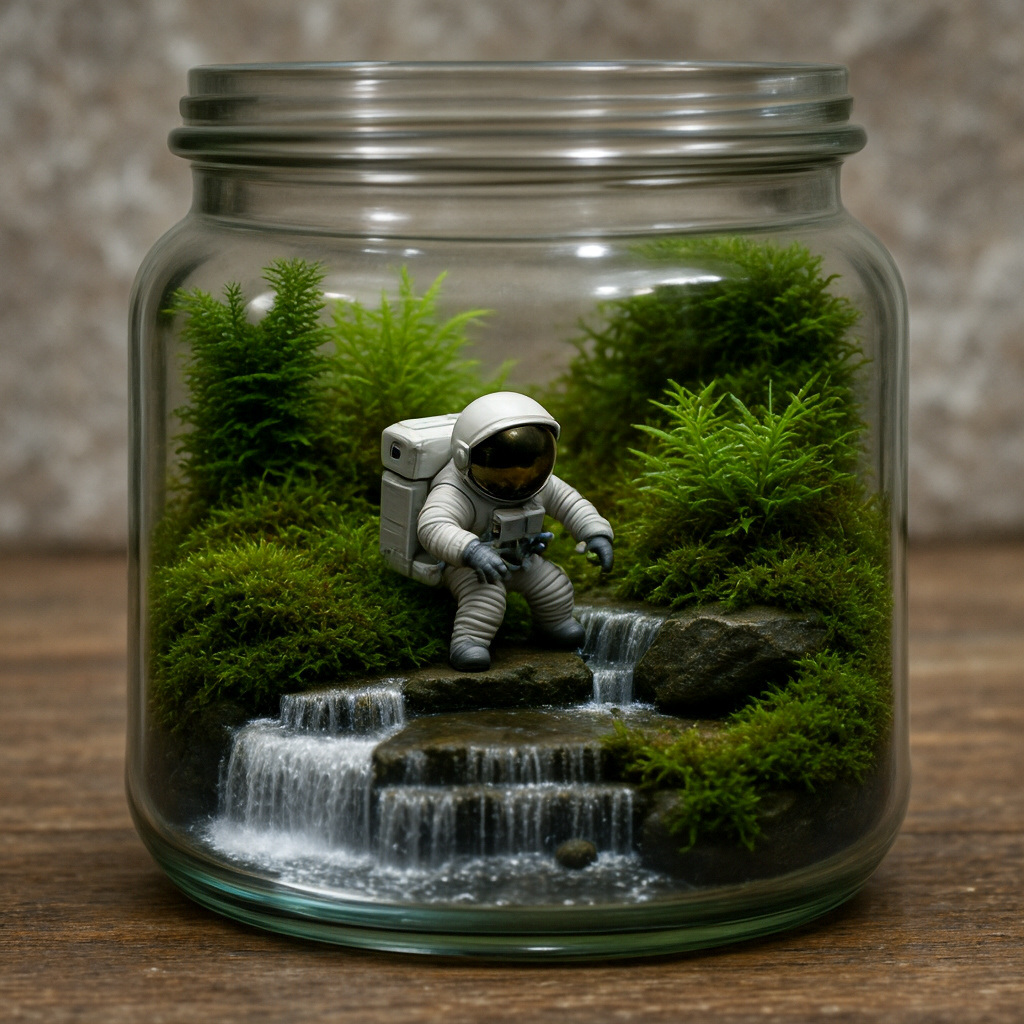}
        {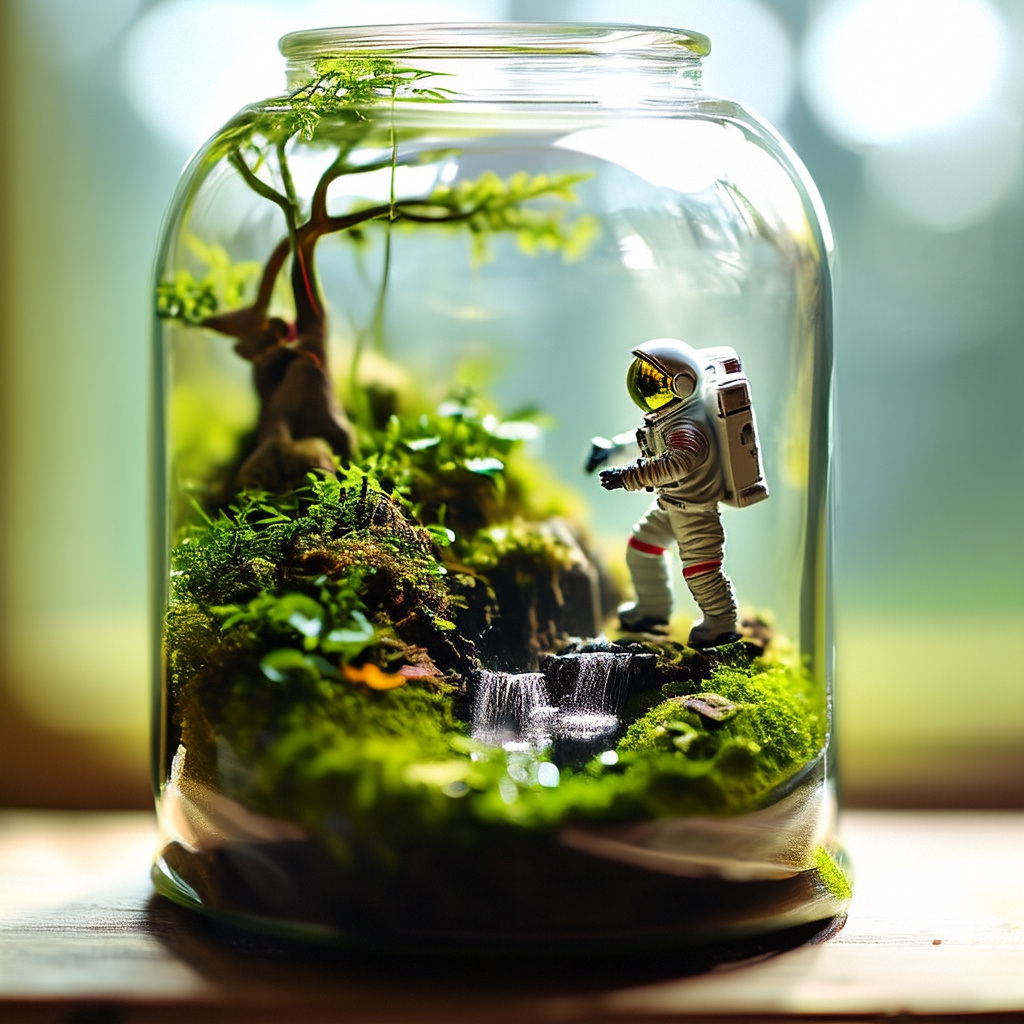}
        {A tiny astronaut exploring a terrarium inside a glass jar, miniature waterfalls and moss.}
    \end{tabular}
    \endgroup
    \captionsetup{skip=6pt}
    \caption{\textbf{PixelDiT qualitative comparison under matched numeric seeds.}
    Pixel matching exhibits implausible layouts, object-count errors, and
    weaker prompt following. Latent matching produces more coherent layouts
    and better preserves the requested objects and spatial relations.}
    \label{fig:pixeldit-qualitative}
    \end{minipage}
\end{figure}

\Cref{fig:pixeldit-qualitative} compares the same three systems on selected
gallery examples with matched prompts. Pixel matching stacks one apple on
another despite the requested row, places the child's lower body inside the
orange bin, generates an extra robot fused with the sunflowers, renders the red
cube as a two-tier block. These errors span physical layout, prompt
following, object identity, and counting. In the same rows, FLUX.3 latent
matching places the apples in a row, separates the child from the produce bins,
depicts one robot holding a distinct bouquet, renders the requested geometric
objects. Across these selected
examples, latent matching follows object counts and spatial relations more
faithfully than pixel matching.

\subsection{DreamLite: Large-to-Small Transfer Across Architectures and VAEs}
\label{sec:dreamlite-results}

Next we show L2P-DMD remains effective across a substantially
larger student--teacher gap: from large latent FLUX Transformers to a compact
convolutional U-Net pixel student adapted from
DreamLite~\cite{feng2026dreamlite}, whose original latent model uses a different
VAE and conditioning stack.

Its 0.39B backbone follows a pruned mobile U-Net design and is paired
with a lightweight VAE. The released model is a strong baseline, post-trained with high-quality supervised fine-tuning and reinforcement learning with HPSv3 as the generation reward,
and its Mobile variant is a well-tuned baseline distilled to four sampling steps.

We leverage the latent FLUX.2 models~\cite{flux2} as teachers, spanning 4B to 32B
parameters, roughly $10\times$--$82\times$ the size of DreamLite. We first use
AsymFlow to fine-tune DreamLite-base into a pixel generator and then perform
latent matching with the FLUX teacher--critic pair. The resulting student produces
pixels directly in four steps.

\begin{table}[!t]
    \centering
    \footnotesize
    \setlength{\tabcolsep}{3.0pt}
    \caption{\textbf{FLUX.3 latent matching improves the DreamLite student
    over Official Mobile across both HPSv3 and DPG despite differences in
    source-model VAE latent spaces and architectures (Transformer vs.
    convolutional U-Net).}}
    \label{tab:dreamlite-results}
    \resizebox{\columnwidth}{!}{%
    \begin{tabular}{@{}llccc@{}}
        \toprule
        Model & Teacher/critic & Steps & HPSv3 $\uparrow$ & DPG $\uparrow$ \\
        \midrule
        Ours & Klein-base-4B & 4 & 13.81 & 83.49 \\
        Ours & Klein-base-9B & 4 & 13.33 & 83.64 \\
        Ours & FLUX.2-dev 32B & 4 & 14.59 & 84.70 \\
        Ours & FLUX.3 Klein Base (5.6B) & 4 & \textbf{14.77} & \textbf{85.07} \\
        \midrule
        Official Mobile & native latent model & 4 & 10.26 & 84.60 \\
        Official Base & - & 28 & 10.64 & 85.53 \\
        \bottomrule
    \end{tabular}
    }
\end{table}

Across FLUX.2 parameter counts, DPG rises from 83.49 to 83.64 and 84.70 as the teacher--critic
system grows, whereas HPSv3 is non-monotonic at 13.81, 13.33, and 14.59
(\cref{tab:dreamlite-results}). These results indicate correlation between teacher parameter count 
and prompt alignment, but not a uniform improvement across both metrics. Using
 FLUX.3 as the teacher gives the strongest result, reaching 14.77 HPSv3 and 85.07 DPG. Relative to Official Mobile, it improves HPSv3 by 4.51 and DPG by 0.47. Relative to Official 28-step Base,
our 4-step model improves HPSv3 by 4.13 and trails DPG by 0.46.

\begin{figure}[!t]
    \centering
    \begin{minipage}[c][\dimexpr\textheight-\textfloatsep-1pt\relax][c]{\columnwidth}
    \centering
    \begingroup
    \setlength{\tabcolsep}{0.8pt}
    \newcommand{\dreamlitetablewidth}{%
        \dimexpr0.90\columnwidth+4\tabcolsep\relax}
    \newsavebox{\dreamlitecaptionbox}
    \newcommand{\dreamlitecaptionline}[1]{%
        \sbox{\dreamlitecaptionbox}{\fontsize{5.2}{5.6}\selectfont #1}%
        \ifdim\wd\dreamlitecaptionbox>\dreamlitetablewidth
            \resizebox{\dreamlitetablewidth}{!}{\usebox{\dreamlitecaptionbox}}%
        \else
            \usebox{\dreamlitecaptionbox}%
        \fi
    }
    \newcommand{\dreamliterow}[4]{%
        \includegraphics[width=0.30\columnwidth]{#1} &
        \includegraphics[width=0.30\columnwidth]{#2} &
        \includegraphics[width=0.30\columnwidth]{#3} \\[-0.6ex]
        \multicolumn{3}{@{}c@{}}{\makebox[\dreamlitetablewidth][c]{%
            \dreamlitecaptionline{#4}}} \\[0.4ex]
    }
    \begin{tabular}{@{}ccc@{}}
        {\scriptsize\bfseries\shortstack{Official DreamLite\\(4-step)}} &
        {\scriptsize\bfseries\shortstack{L2P-DMD DreamLite\\(Ours, 4-step)}} &
        {\scriptsize\bfseries\shortstack{Official DreamLite\\(28-step)}} \\[-0.2ex]
        \dreamliterow
        {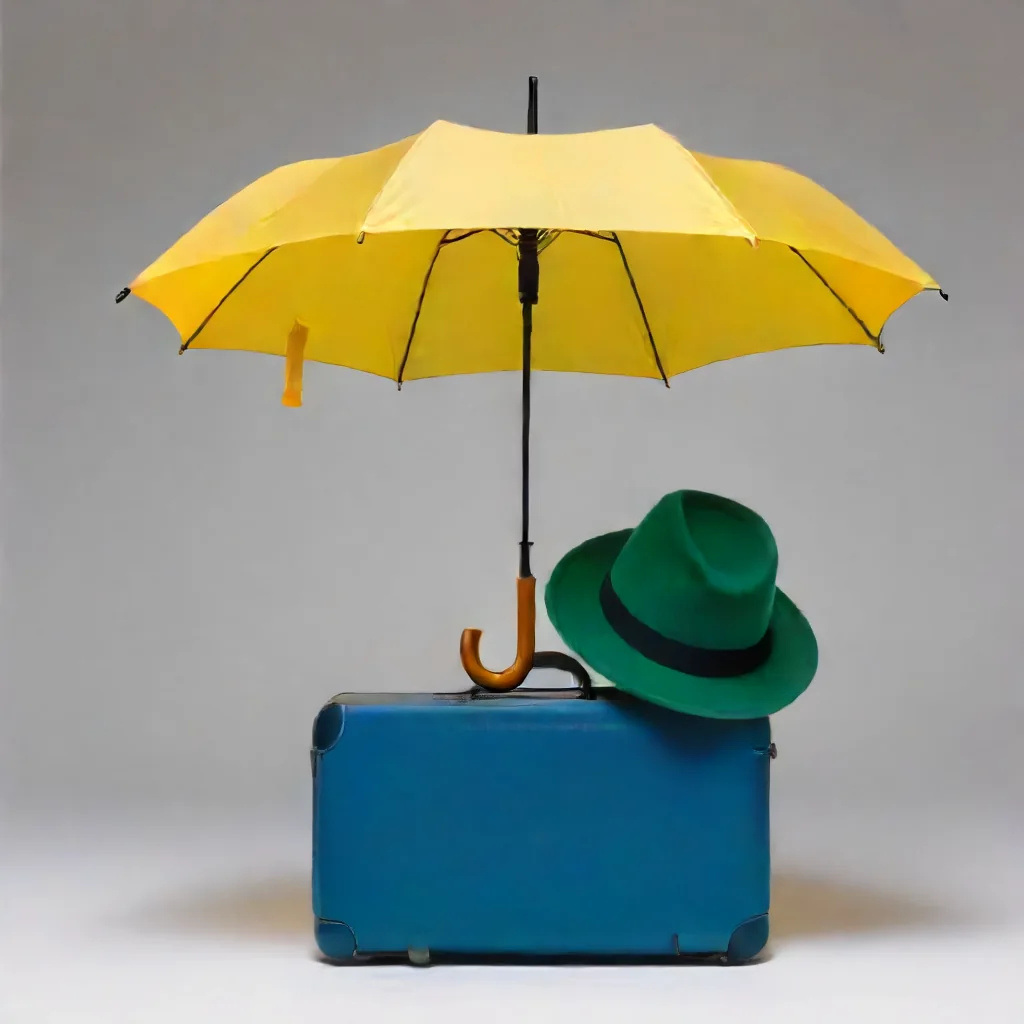}
        {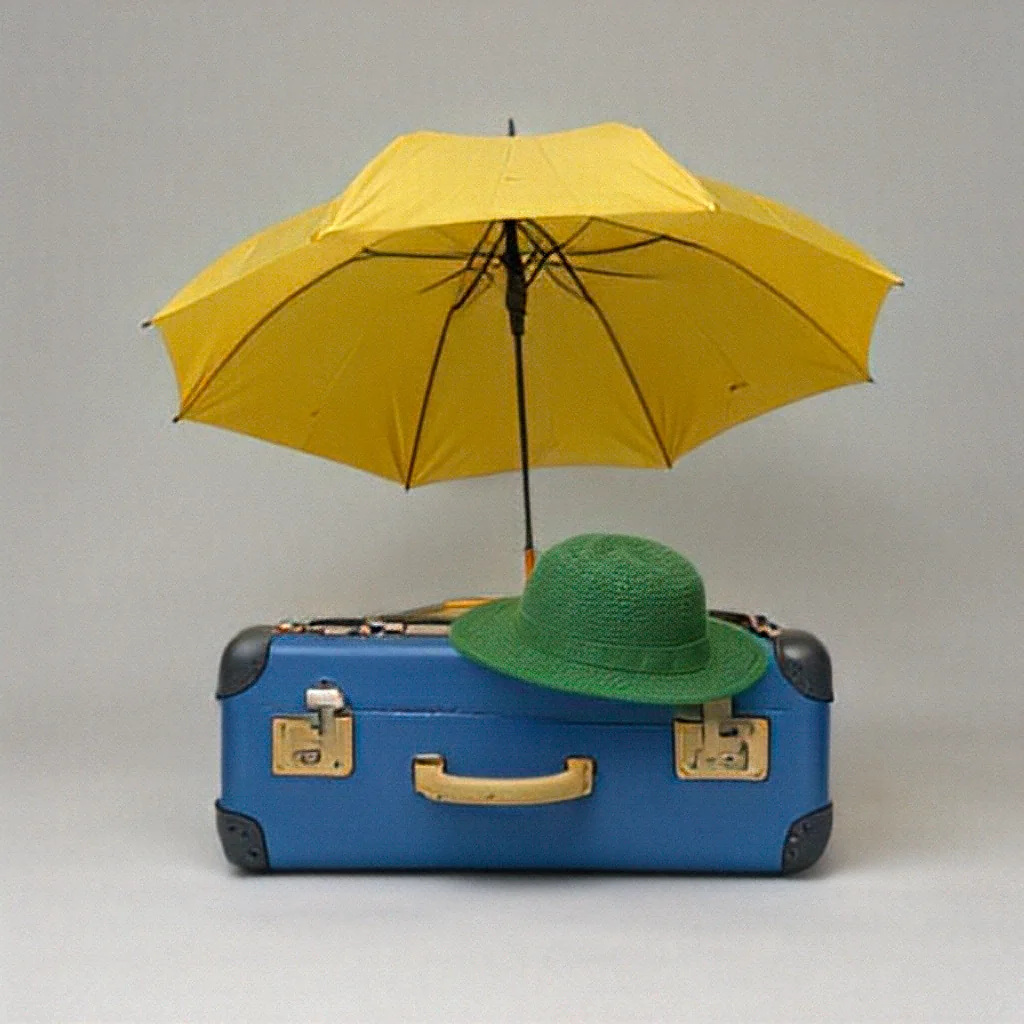}
        {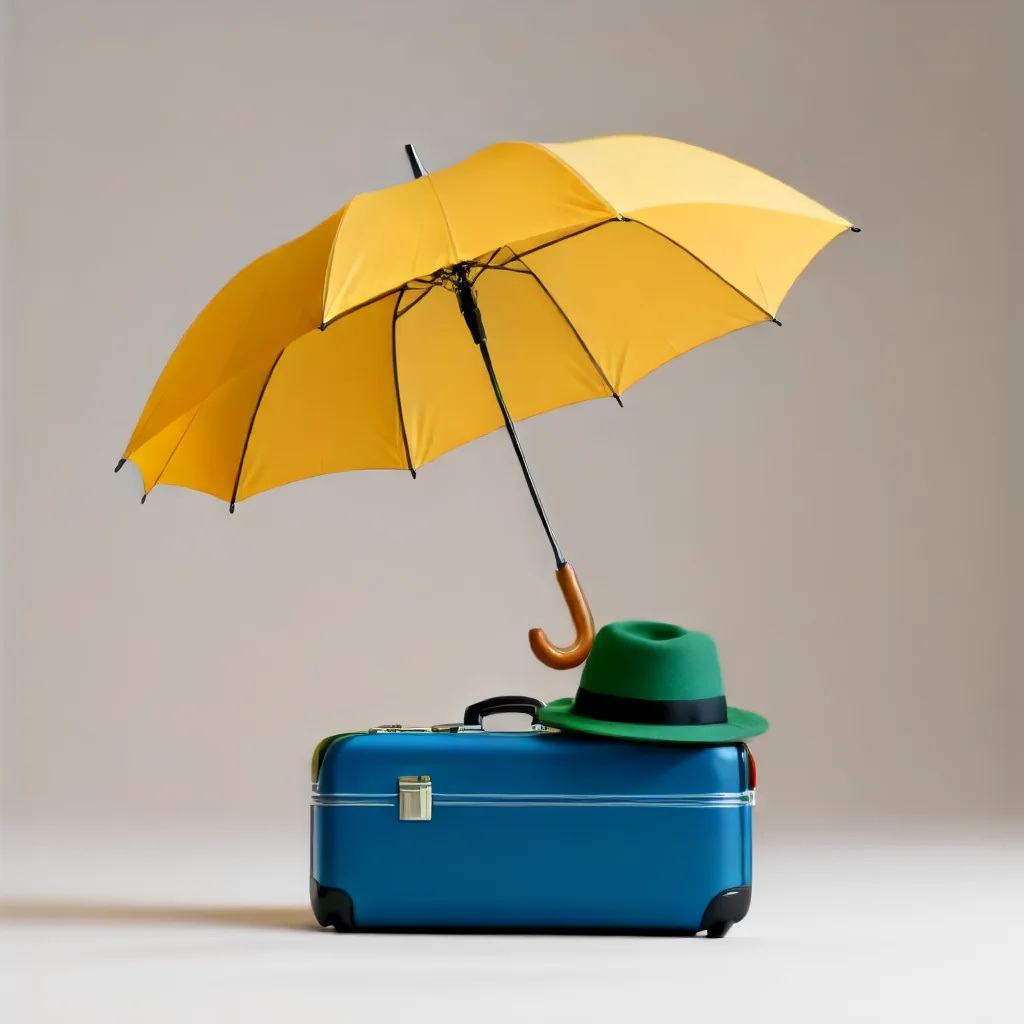}
        {A yellow umbrella above a blue suitcase, with a green hat placed between them, plain studio background.}
        \dreamliterow
        {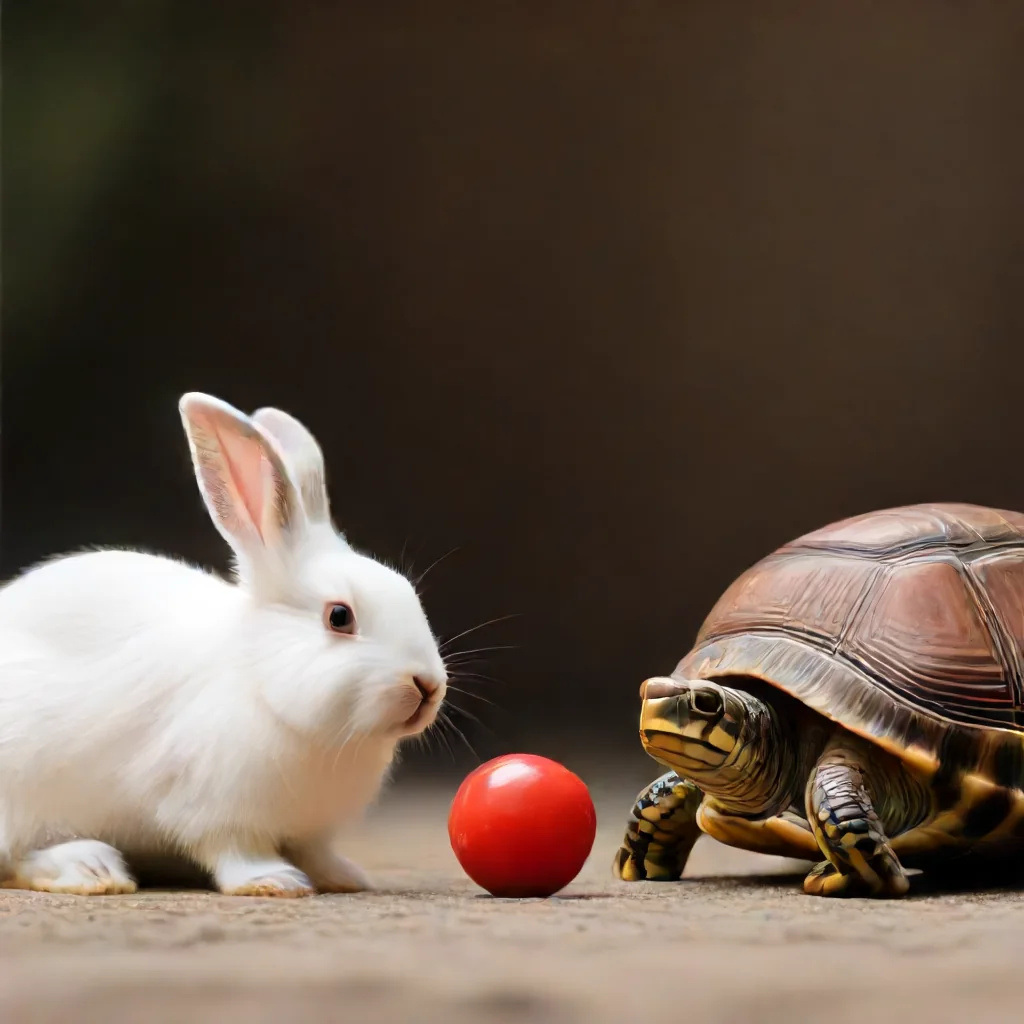}
        {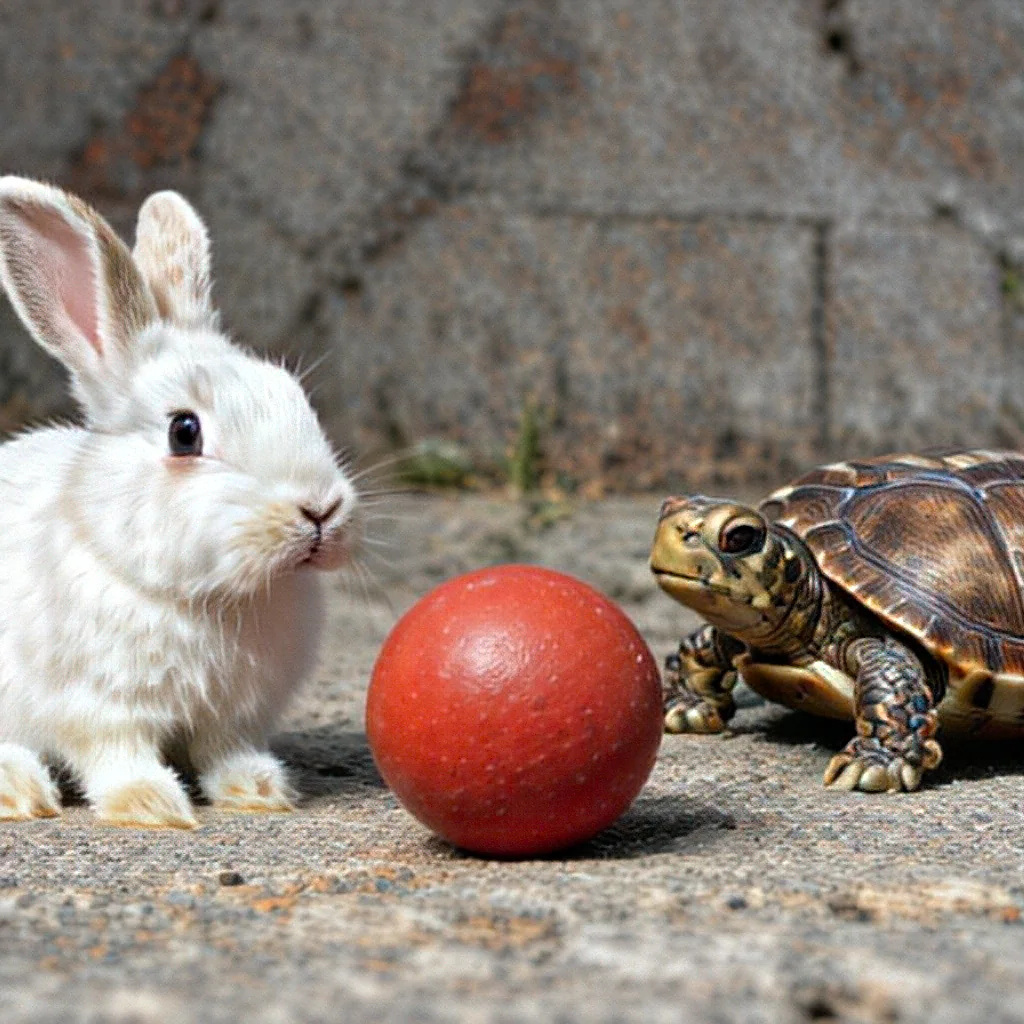}
        {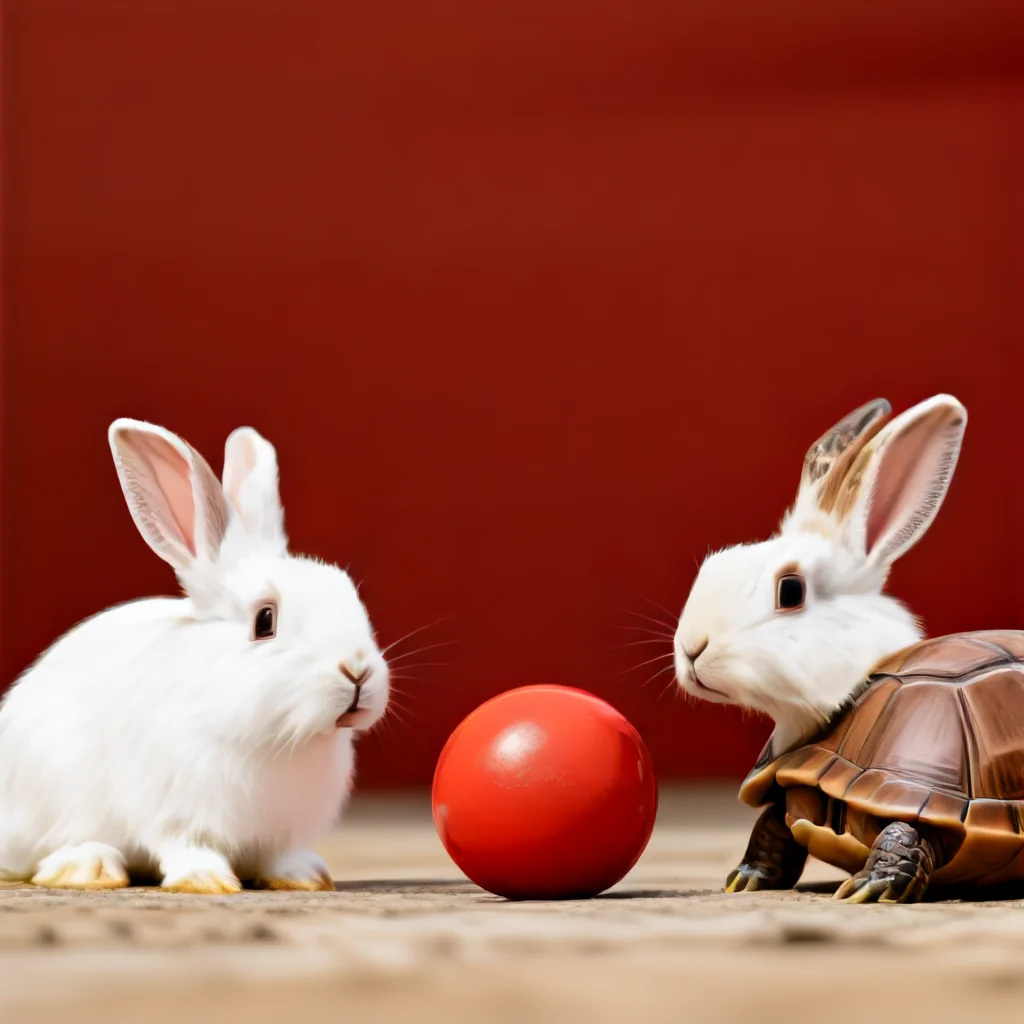}
        {A white rabbit to the left of a brown turtle, both facing a red ball in the center.}
        \dreamliterow
        {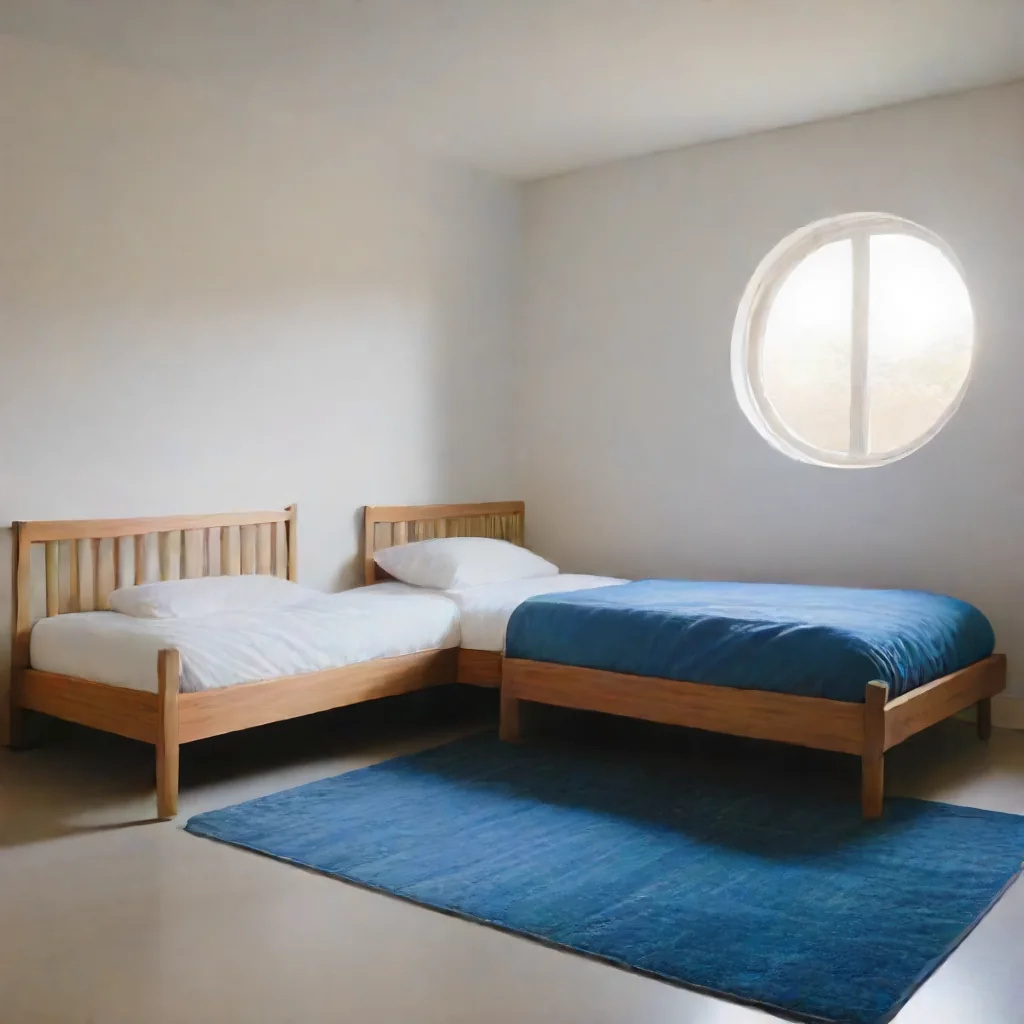}
        {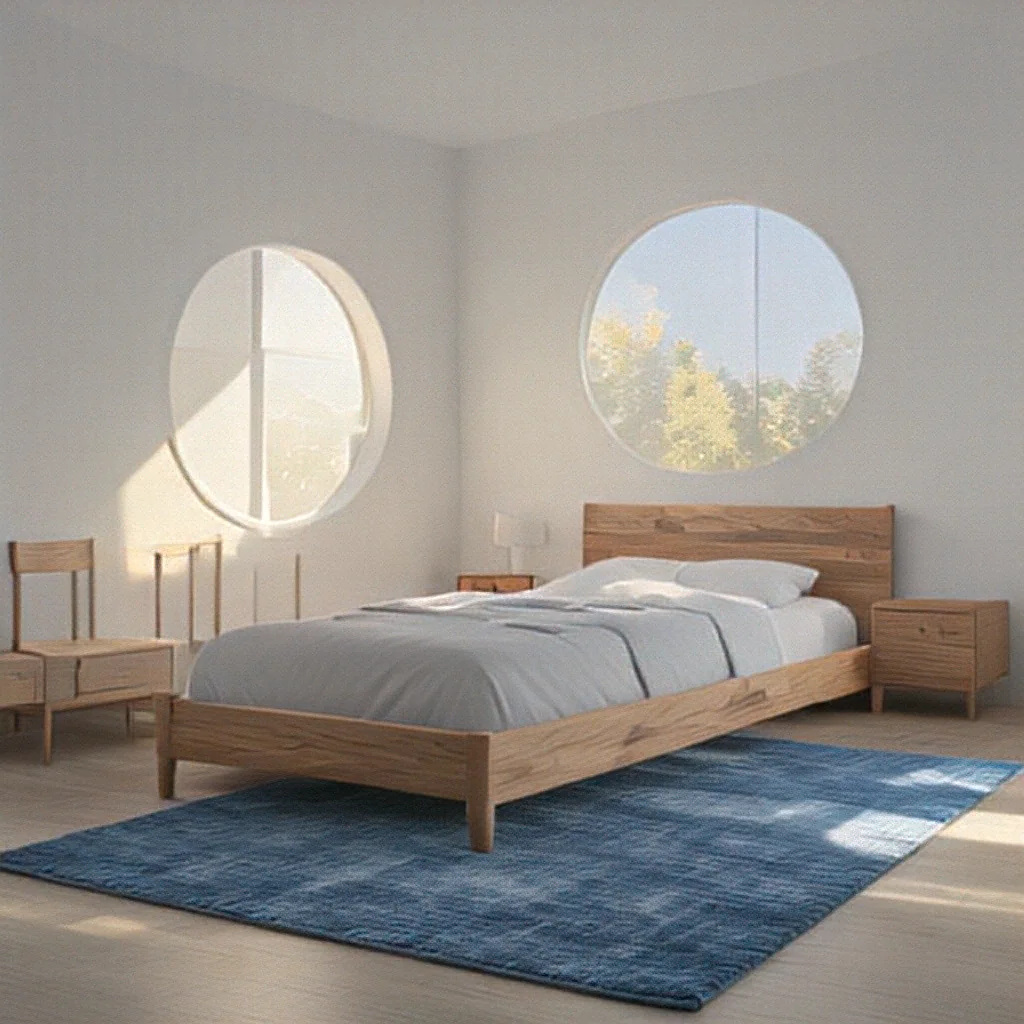}
        {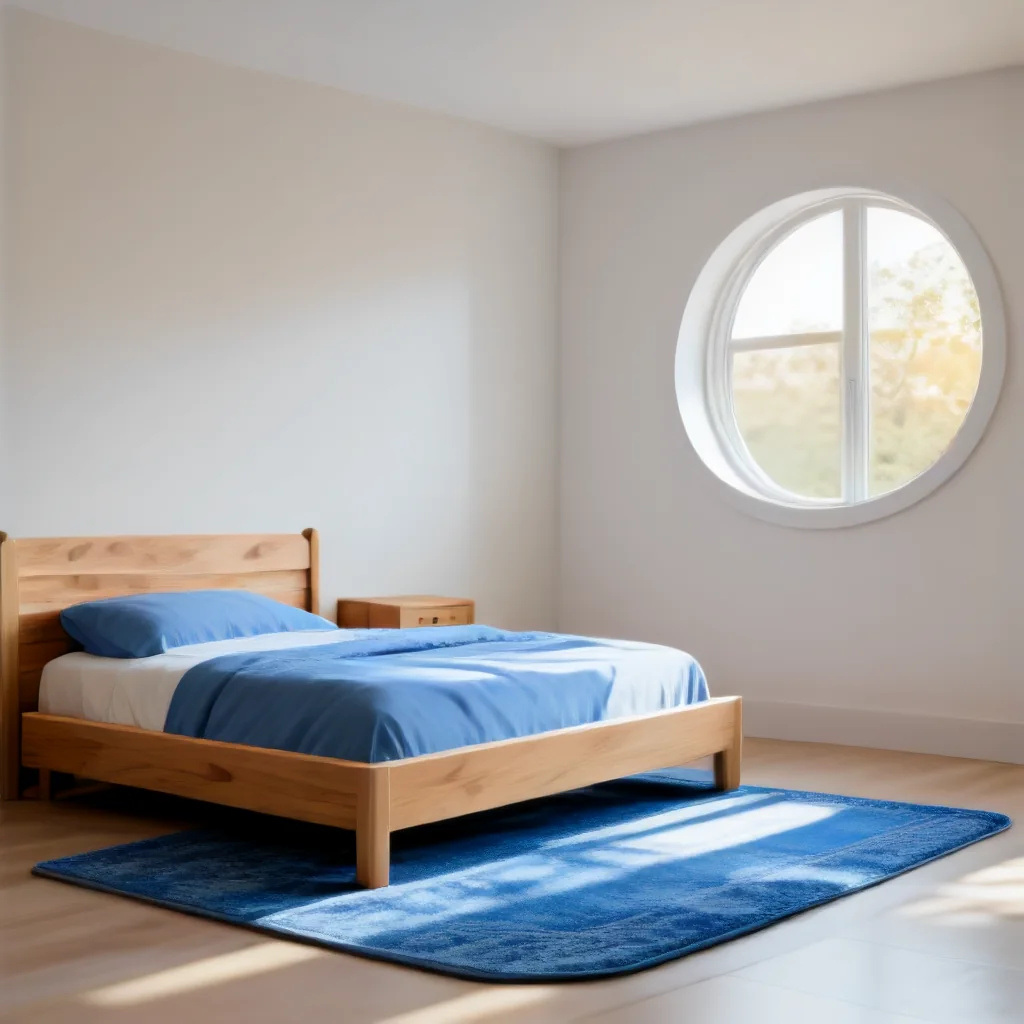}
        {A minimalist bedroom with a round window, a low oak bed and a blue rug, soft morning sunlight.}
        \dreamliterow
        {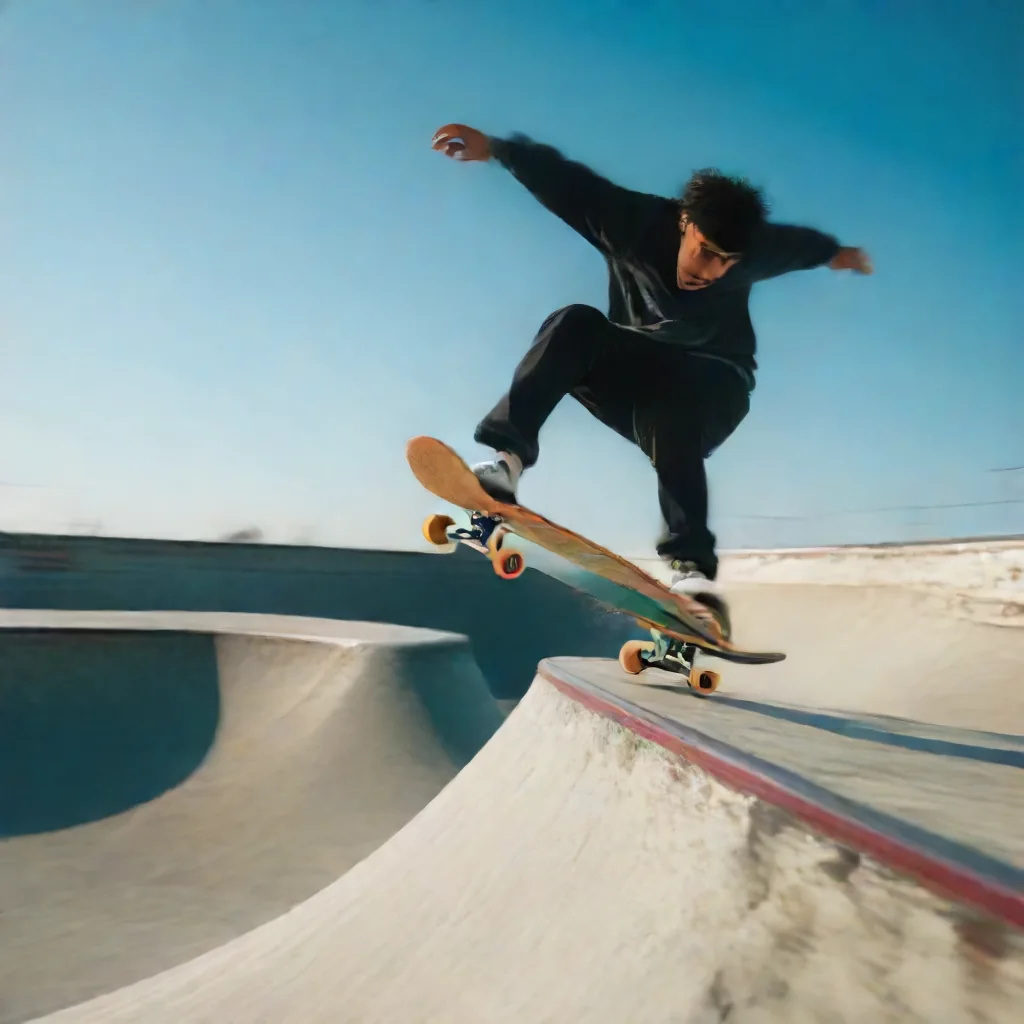}
        {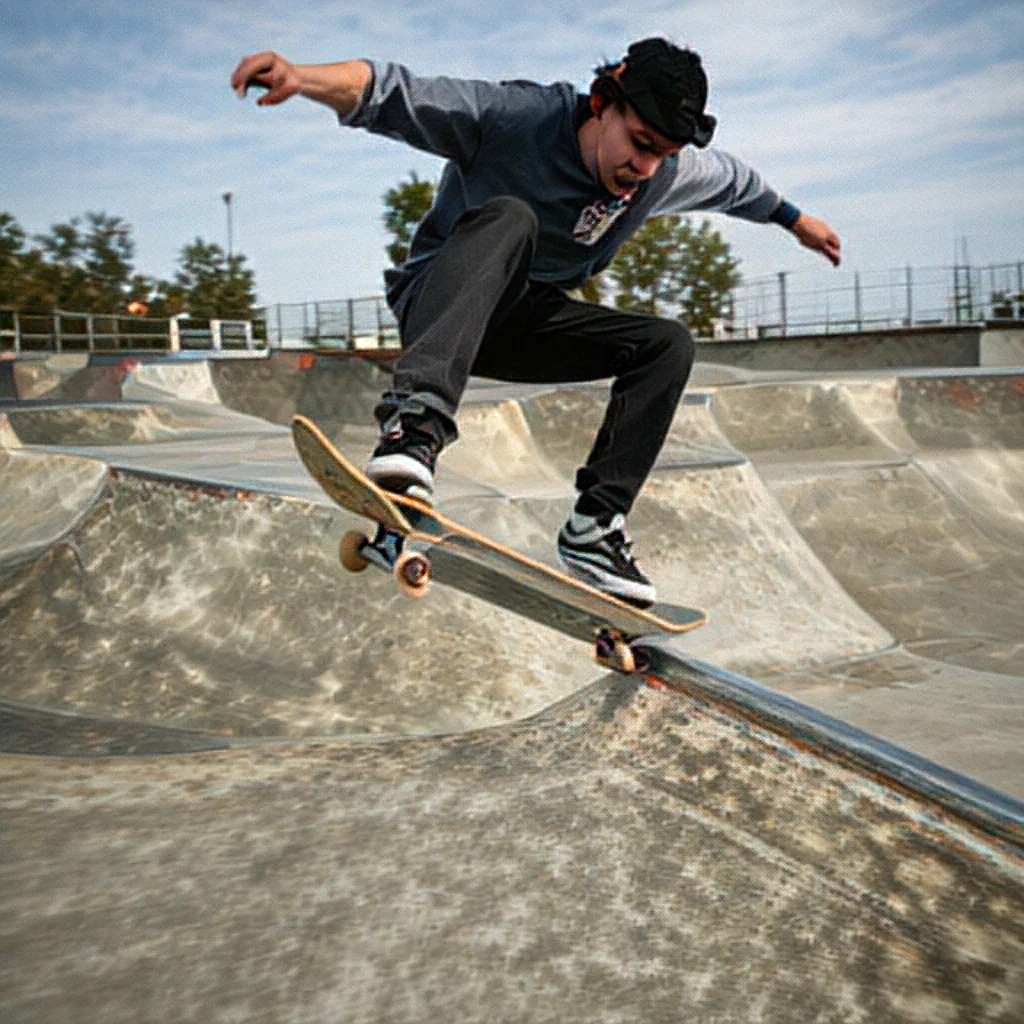}
        {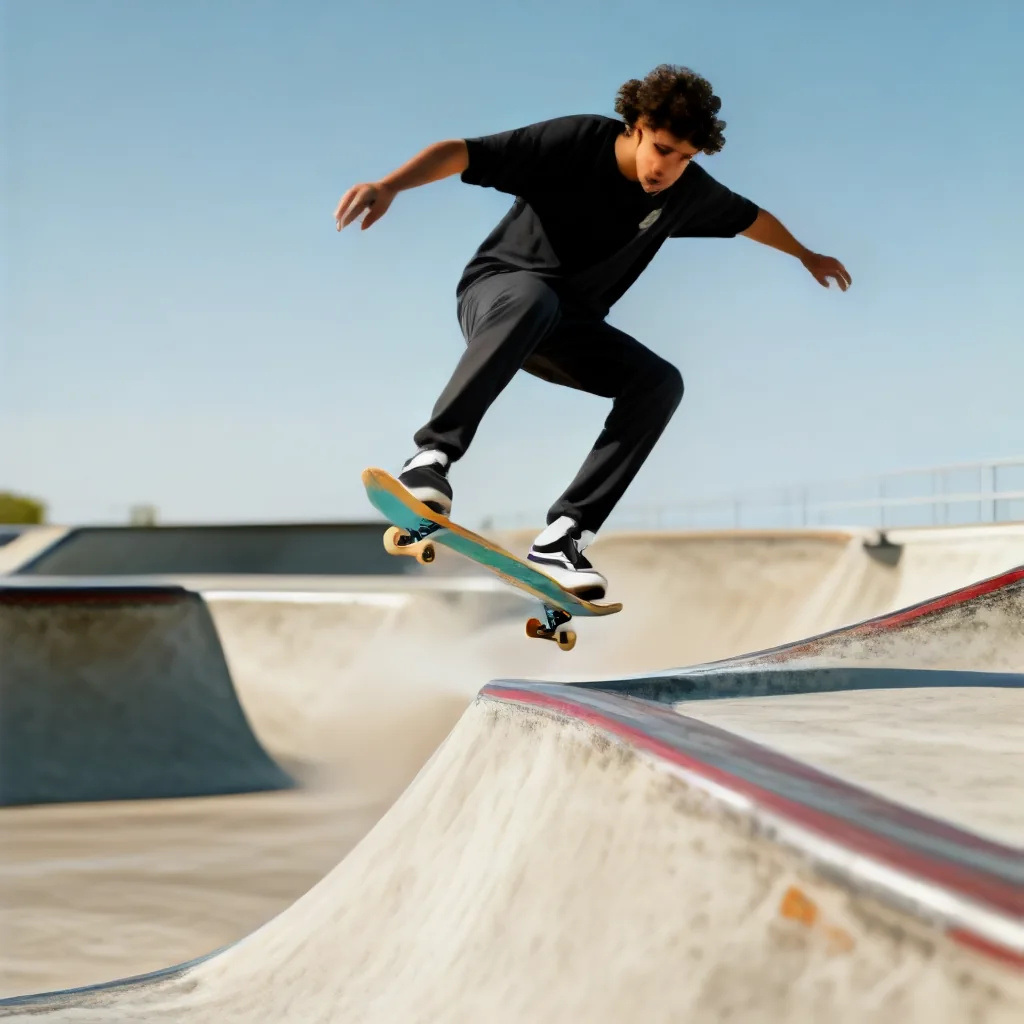}
        {A skateboarder performing a kickflip above a concrete ramp, sharp subject with realistic motion blur.}
        \dreamliterow
        {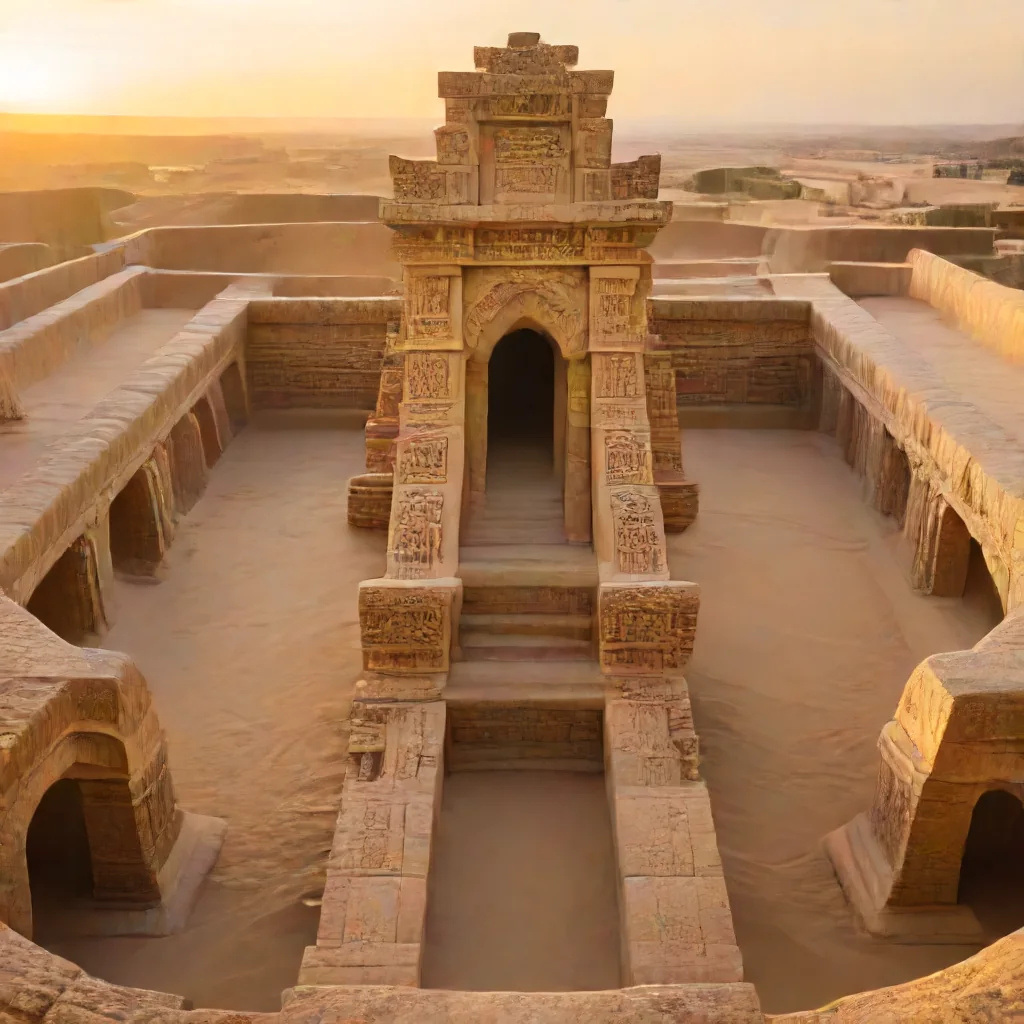}
        {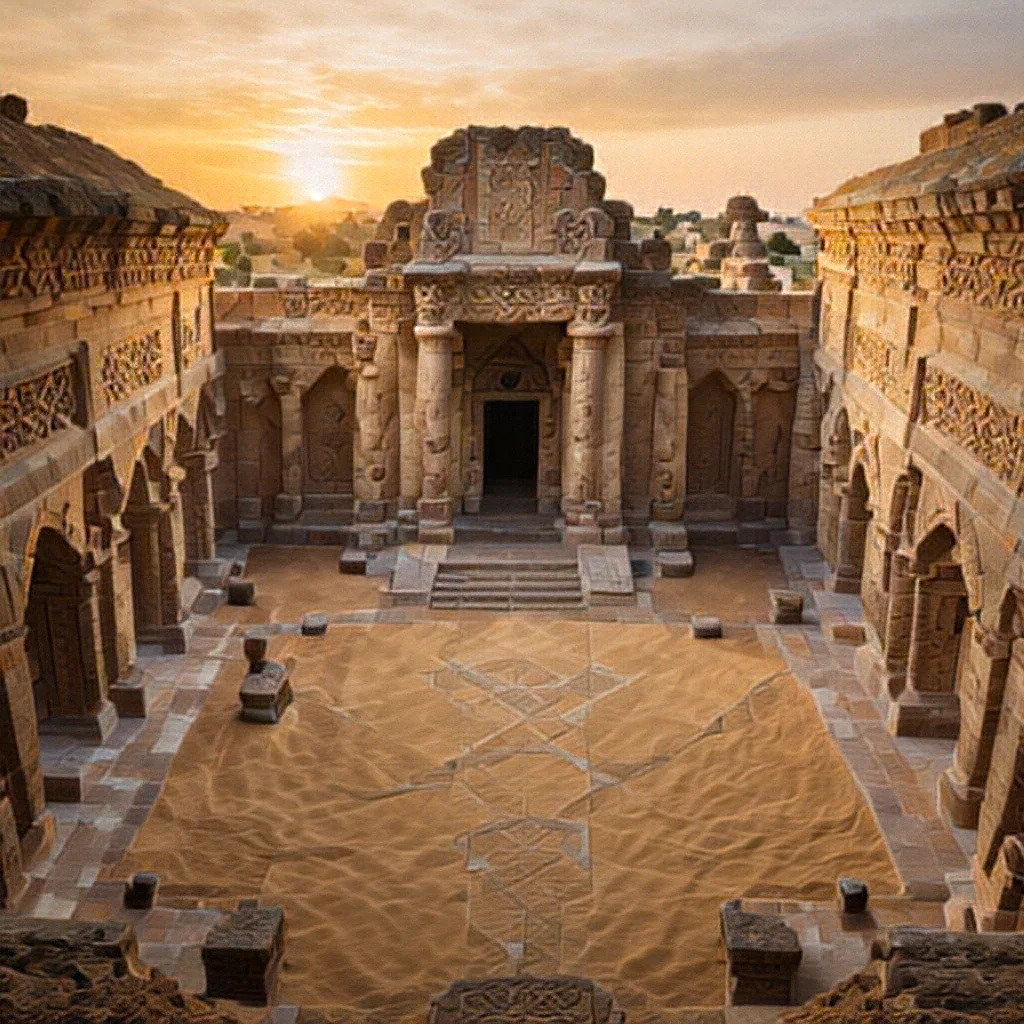}
        {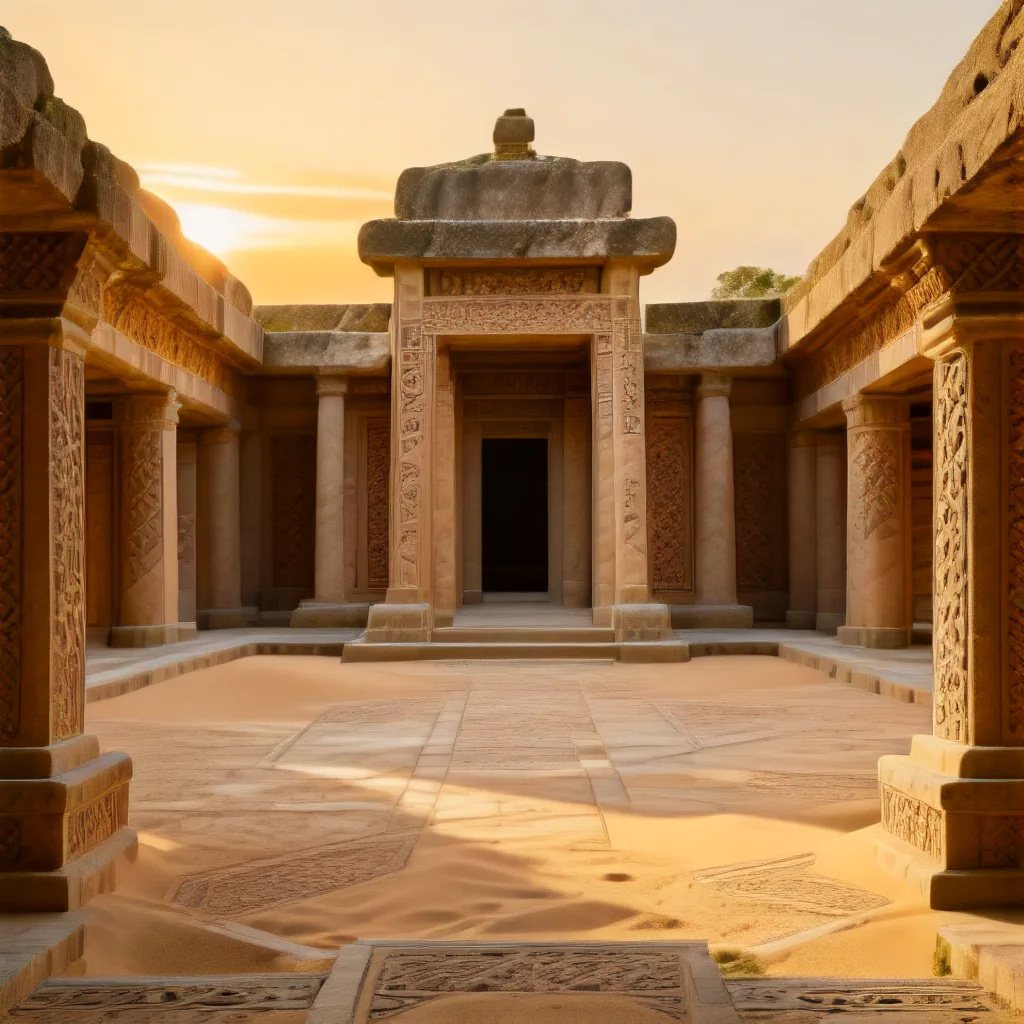}
        {A symmetrical photograph of an ancient sandstone temple courtyard at golden hour, intricate carvings.}
        \dreamliterow
        {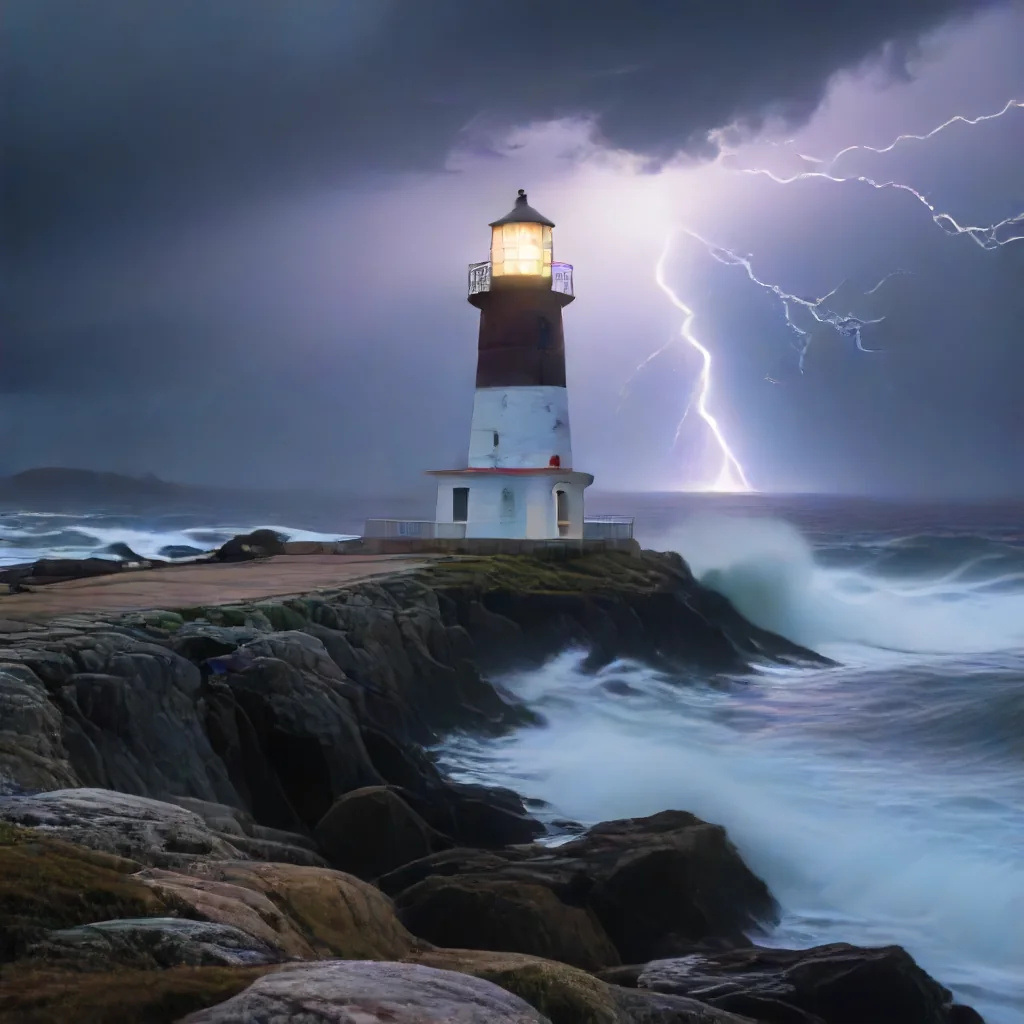}
        {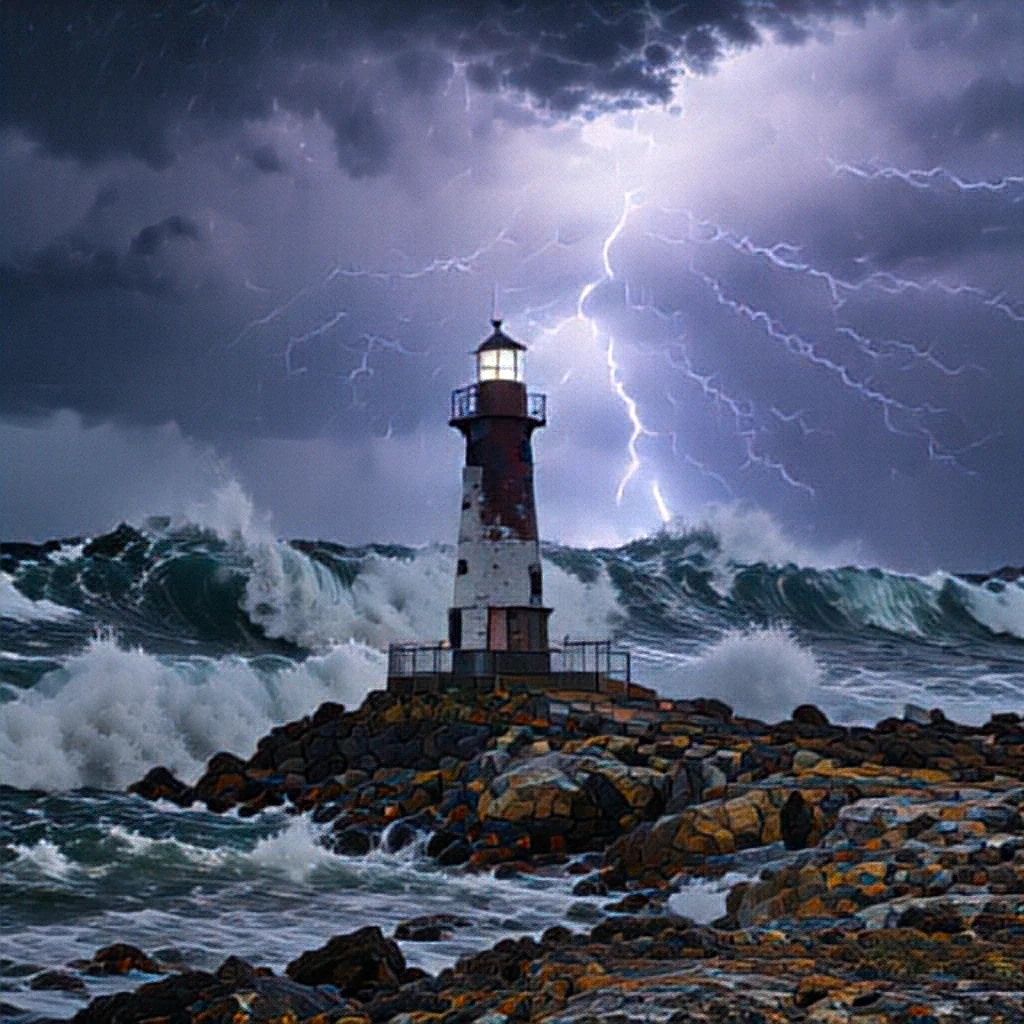}
        {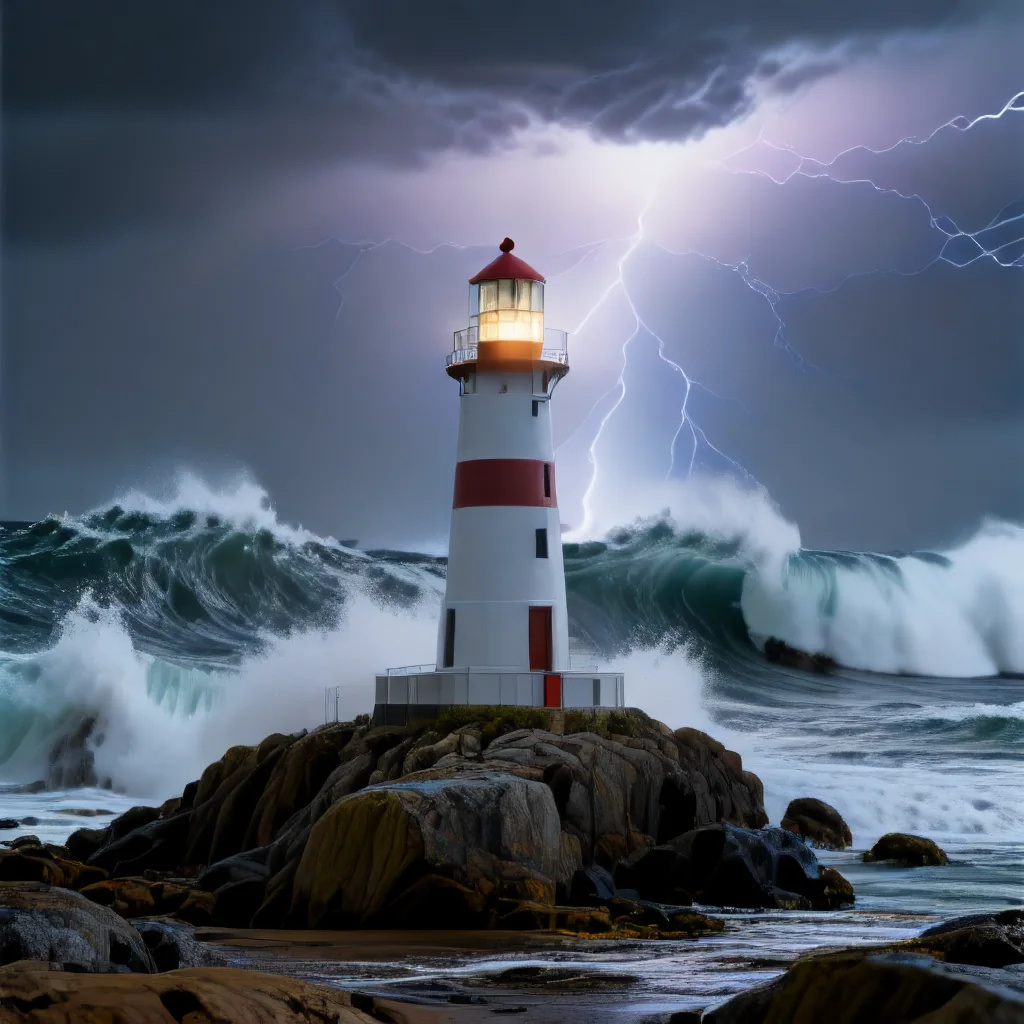}
        {A lighthouse on a rocky coast during a thunderstorm, a lightning bolt illuminating enormous waves.}
    \end{tabular}
    \endgroup
    \captionsetup{skip=6pt}
    \caption{\textbf{DreamLite qualitative comparison under numeric seeds.}
    Across these selected examples, Latent FLUX.3 $\rightarrow$ Pixel DreamLite
    adds richer object, material, and scene detail than the official DreamLite
    baselines, whose backgrounds are often smoother and less structured.}
    \label{fig:dreamlite-qualitative}
    \end{minipage}
\end{figure}

\Cref{fig:dreamlite-qualitative} compares our latent-FLUX.3-to-pixel-DreamLite
student with the Official Mobile and Base models on selected examples. The
three images in each triplet use the same prompt and numeric seed.
Across
these examples, the latent-matched pixel student renders richer
materials and backgrounds than the official latent DreamLite models.

Our DreamLite experiment separates training-time capacity from deployment-time architecture.
A large latent Transformer supplies the teacher--critic supervision, while the
deployed student remains a 0.39B mobile U-Net with four-step, direct-pixel
inference. Direct pixel output removes the VAE decoder and its associated
inference cost.

\section{Discussion and Conclusion}
\label{sec:conclusion}

We introduce L2P-DMD, which supervises few-step pixel generators in a latent
teacher's native space through a frozen encoder while preserving direct-RGB,
VAE-free inference. Across ImageNet-256 and $1024\!\times\!1024$ text-to-image
experiments, latent matching outperforms pixel matching in the evaluated paired
comparisons and transfers across architectures, latent spaces, and model scales,
decoupling the student's output and DMD matching spaces. Future work should
improve transfer efficiency and test broader teacher--student families.

\FloatBarrier
\clearpage
\section*{LLM Usage Statement}
AI chatbots (ChatGPT and Claude) were used to polish the manuscript, and GitHub Copilot (GPT-6.1 Sol) was used to assist with implementing the experiments. All research ideas, methodological contributions, experimental designs, results, and scientific claims were developed and verified by the authors, who take full responsibility for the content of this paper.

{
    \small
    \bibliographystyle{ieeenat_fullname}
    \bibliography{main}

\begin{thebibliography}{79}
\providecommand{\natexlab}[1]{#1}
\providecommand{\url}[1]{\texttt{#1}}
\expandafter\ifx\csname urlstyle\endcsname\relax
  \providecommand{\doi}[1]{doi: #1}\else
  \providecommand{\doi}{doi: \begingroup \urlstyle{rm}\Url}\fi

\bibitem[Balaji et~al.(2022)Balaji, Nah, Huang, Vahdat, Song, Zhang, Kreis,
  Aittala, Aila, Laine, Catanzaro, Karras, and Liu]{balaji2022ediffi}
Yogesh Balaji, Seungjun Nah, Xun Huang, Arash Vahdat, Jiaming Song, Qinsheng
  Zhang, Karsten Kreis, Miika Aittala, Timo Aila, Samuli Laine, Bryan
  Catanzaro, Tero Karras, and Ming-Yu Liu.
\newblock ediff-i: Text-to-image diffusion models with an ensemble of expert
  denoisers.
\newblock \emph{arXiv preprint arXiv:2211.01324}, 2022.

\bibitem[Berthelot et~al.(2023)Berthelot, Autef, Lin, Yap, Zhai, Hu, Zheng,
  Talbott, and Gu]{berthelot2023tract}
David Berthelot, Arnaud Autef, Jierui Lin, Dian~Ang Yap, Shuangfei Zhai, Siyuan
  Hu, Daniel Zheng, Walter Talbott, and Eric Gu.
\newblock {TRACT}: Denoising diffusion models with transitive closure
  time-distillation.
\newblock \emph{arXiv preprint arXiv:2303.04248}, 2023.

\bibitem[{Black Forest Labs}(2025)]{flux2}
{Black Forest Labs}.
\newblock {FLUX.2}: Frontier visual intelligence.
\newblock \url{https://bfl.ai/blog/flux-2}, 2025.

\bibitem[{Black Forest Labs}(2026)]{bfl2026flux3imageklein}
{Black Forest Labs}.
\newblock {FLUX} 3 image klein.
\newblock \url{https://bfl.ai/blog/flux-3-image-klein}, 2026.

\bibitem[Chen et~al.(2026{\natexlab{a}})Chen, Ackermann, Kim, Wetzstein, and
  Guibas]{chen2026asymmetric}
Hansheng Chen, Jan Ackermann, Minseo Kim, Gordon Wetzstein, and Leonidas
  Guibas.
\newblock Asymmetric flow models.
\newblock In \emph{Advances in Neural Information Processing Systems},
  2026{\natexlab{a}}.

\bibitem[Chen et~al.(2025)Chen, Ge, Zhang, Sun, and Luo]{chen2025pixelflow}
Shoufa Chen, Chongjian Ge, Shilong Zhang, Peize Sun, and Ping Luo.
\newblock {PixelFlow}: Pixel-space generative models with flow.
\newblock \emph{arXiv preprint arXiv:2504.07963}, 2025.

\bibitem[Chen et~al.(2026{\natexlab{b}})Chen, Zhu, Chen, Zhang, Chen, Zeng,
  Zhang, Wang, Yang, and Tai]{chen2026l2p}
Zhennan Chen, Junwei Zhu, Xu Chen, Jiangning Zhang, Jiawei Chen, Zhuoqi Zeng,
  Wei Zhang, Chengjie Wang, Jian Yang, and Ying Tai.
\newblock {L2P}: Unlocking latent potential for pixel generation.
\newblock In \emph{Advances in Neural Information Processing Systems},
  2026{\natexlab{b}}.

\bibitem[Chen et~al.(2026{\natexlab{c}})Chen, Zhu, Chen, Zhang, Hu, Zhao, Wang,
  Yang, and Tai]{chen2026dip}
Zhennan Chen, Junwei Zhu, Xu Chen, Jiangning Zhang, Xiaobin Hu, Hanzhen Zhao,
  Chengjie Wang, Jian Yang, and Ying Tai.
\newblock {DiP}: Taming diffusion models in pixel space.
\newblock In \emph{Proceedings of the IEEE/CVF Conference on Computer Vision
  and Pattern Recognition}, pages 36136--36146, 2026{\natexlab{c}}.

\bibitem[Crowson et~al.(2024)Crowson, Baumann, Birch, Abraham, Kaplan, and
  Shippole]{crowson2024hdit}
Katherine Crowson, Stefan~Andreas Baumann, Alex Birch, Tanishq~Mathew Abraham,
  Daniel~Z. Kaplan, and Enrico Shippole.
\newblock Scalable high-resolution pixel-space image synthesis with hourglass
  diffusion transformers.
\newblock In \emph{Proceedings of the 41st International Conference on Machine
  Learning}, pages 9550--9575, 2024.

\bibitem[Deng et~al.(2009)Deng, Dong, Socher, Li, Li, and
  Fei-Fei]{deng2009imagenet}
Jia Deng, Wei Dong, Richard Socher, Li-Jia Li, Kai Li, and Li Fei-Fei.
\newblock {ImageNet}: A large-scale hierarchical image database.
\newblock In \emph{2009 IEEE Conference on Computer Vision and Pattern
  Recognition}, pages 248--255, 2009.

\bibitem[Dhariwal and Nichol(2021)]{dhariwal2021adm}
Prafulla Dhariwal and Alexander Nichol.
\newblock Diffusion models beat {GANs} on image synthesis.
\newblock In \emph{Advances in Neural Information Processing Systems}, pages
  8780--8794, 2021.

\bibitem[Fan et~al.(2026)Fan, Qiu, Wu, Wang, Lin, Ren, Lin, Gong, and
  Yang]{fan2026phaseddmd}
Xiangyu Fan, Zesong Qiu, Zhuguanyu Wu, Fanzhou Wang, Zhiqian Lin, Tianxiang
  Ren, Dahua Lin, Ruihao Gong, and Lei Yang.
\newblock Phased {DMD}: Few-step distribution matching distillation via score
  matching within subintervals.
\newblock In \emph{Proceedings of the IEEE/CVF Conference on Computer Vision
  and Pattern Recognition}, pages 41667--41676, 2026.

\bibitem[Feng et~al.(2026{\natexlab{a}})Feng, Wei, Chen, Pan, Liu, Yan, Gao,
  and Zuo]{feng2026dreamlite}
Kailai Feng, Yuxiang Wei, Bo Chen, Yang Pan, Songwei Liu, Chenqian Yan, Yuan
  Gao, and Wangmeng Zuo.
\newblock {DreamLite}: A lightweight on-device unified model for image
  generation and editing.
\newblock In \emph{Computer Vision -- ECCV 2026}, pages 214--232. Springer,
  2026{\natexlab{a}}.

\bibitem[Feng et~al.(2026{\natexlab{b}})Feng, Li, Zablocki, Cord, and
  Alahi]{feng2026rdm}
Lan Feng, Wuyang Li, {\'E}loi Zablocki, Matthieu Cord, and Alexandre Alahi.
\newblock Representation distribution matching for one-step visual generation.
\newblock \emph{arXiv preprint arXiv:2607.02375}, 2026{\natexlab{b}}.

\bibitem[Ge et~al.(2026)Ge, Zhang, Xu, Zhang, Zhang, Wang, and
  Zhang]{ge2026senseflow}
Xingtong Ge, Xin Zhang, Tongda Xu, Yi Zhang, Xinjie Zhang, Yan Wang, and Jun
  Zhang.
\newblock {SenseFlow}: Scaling distribution matching for flow-based
  text-to-image distillation.
\newblock In \emph{International Conference on Learning Representations}, 2026.

\bibitem[Gu et~al.(2024)Gu, Zhai, Zhang, Susskind, and
  Jaitly]{gu2024matryoshka}
Jiatao Gu, Shuangfei Zhai, Yizhe Zhang, Joshua Susskind, and Navdeep Jaitly.
\newblock Matryoshka diffusion models.
\newblock In \emph{International Conference on Learning Representations}, 2024.

\bibitem[Heusel et~al.(2017)Heusel, Ramsauer, Unterthiner, Nessler, and
  Hochreiter]{heusel2017fid}
Martin Heusel, Hubert Ramsauer, Thomas Unterthiner, Bernhard Nessler, and Sepp
  Hochreiter.
\newblock {GANs} trained by a two time-scale update rule converge to a local
  {Nash} equilibrium.
\newblock In \emph{Advances in Neural Information Processing Systems}, 2017.

\bibitem[Ho et~al.(2022{\natexlab{a}})Ho, Chan, Saharia, Whang, Gao, Gritsenko,
  Kingma, Poole, Norouzi, Fleet, and Salimans]{ho2022imagenvideo}
Jonathan Ho, William Chan, Chitwan Saharia, Jay Whang, Ruiqi Gao, Alexey
  Gritsenko, Diederik~P. Kingma, Ben Poole, Mohammad Norouzi, David~J. Fleet,
  and Tim Salimans.
\newblock Imagen video: High definition video generation with diffusion models.
\newblock \emph{arXiv preprint arXiv:2210.02303}, 2022{\natexlab{a}}.

\bibitem[Ho et~al.(2022{\natexlab{b}})Ho, Saharia, Chan, Fleet, Norouzi, and
  Salimans]{ho2022cascaded}
Jonathan Ho, Chitwan Saharia, William Chan, David~J. Fleet, Mohammad Norouzi,
  and Tim Salimans.
\newblock Cascaded diffusion models for high fidelity image generation.
\newblock \emph{Journal of Machine Learning Research}, 23\penalty0
  (47):\penalty0 1--33, 2022{\natexlab{b}}.

\bibitem[Ho et~al.(2022{\natexlab{c}})Ho, Salimans, Gritsenko, Chan, Norouzi,
  and Fleet]{ho2022video}
Jonathan Ho, Tim Salimans, Alexey Gritsenko, William Chan, Mohammad Norouzi,
  and David~J. Fleet.
\newblock Video diffusion models.
\newblock In \emph{Advances in Neural Information Processing Systems}, pages
  8633--8646, 2022{\natexlab{c}}.

\bibitem[Hoogeboom et~al.(2023)Hoogeboom, Heek, and
  Salimans]{hoogeboom2023simple}
Emiel Hoogeboom, Jonathan Heek, and Tim Salimans.
\newblock Simple diffusion: End-to-end diffusion for high resolution images.
\newblock In \emph{Proceedings of the 40th International Conference on Machine
  Learning}, pages 13213--13232, 2023.

\bibitem[Hoogeboom et~al.(2025)Hoogeboom, Mensink, Heek, Lamerigts, Gao, and
  Salimans]{hoogeboom2025sid2}
Emiel Hoogeboom, Thomas Mensink, Jonathan Heek, Kay Lamerigts, Ruiqi Gao, and
  Tim Salimans.
\newblock Simpler diffusion: 1.5 {FID} on {ImageNet512} with pixel-space
  diffusion.
\newblock In \emph{Proceedings of the IEEE/CVF Conference on Computer Vision
  and Pattern Recognition}, pages 18062--18071, 2025.

\bibitem[Hu et~al.(2024)Hu, Wang, Fang, Fu, Cheng, and Yu]{hu2024ella}
Xiwei Hu, Rui Wang, Yixiao Fang, Bin Fu, Pei Cheng, and Gang Yu.
\newblock {ELLA}: Equip diffusion models with {LLM} for enhanced semantic
  alignment.
\newblock \emph{arXiv preprint arXiv:2403.05135}, 2024.

\bibitem[Jabri et~al.(2023)Jabri, Fleet, and Chen]{jabri2023rin}
Allan Jabri, David~J. Fleet, and Ting Chen.
\newblock Scalable adaptive computation for iterative generation.
\newblock In \emph{Proceedings of the 40th International Conference on Machine
  Learning}, pages 14569--14589, 2023.

\bibitem[Jiang et~al.(2026)Jiang, Du, Chen, Liu, Wang, Zheng, Yang, Cai, Hao,
  Jiang, Gao, Yang, and Hoi]{jiang2026pixeltext}
Dengyang Jiang, Ruoyi Du, Zhennan Chen, Dongyang Liu, Zanyi Wang, Mingzhe
  Zheng, Xiangpeng Yang, Huanqia Cai, Aiming Hao, Yuming Jiang, Peng Gao, Harry
  Yang, and Steven Hoi.
\newblock An empirical study of training pixel-space text-to-image diffusion
  models.
\newblock \emph{arXiv preprint arXiv:2608.16887}, 2026.

\bibitem[Kang et~al.(2024)Kang, Zhang, Barnes, Paris, Kwak, Park, Shechtman,
  Zhu, and Park]{kang2024diffusion2gan}
Minguk Kang, Richard Zhang, Connelly Barnes, Sylvain Paris, Suha Kwak, Jaesik
  Park, Eli Shechtman, Jun-Yan Zhu, and Taesung Park.
\newblock Distilling diffusion models into conditional {GANs}.
\newblock In \emph{European Conference on Computer Vision}, pages 428--447,
  2024.

\bibitem[Kim et~al.(2024)Kim, Lai, Liao, Murata, Takida, Uesaka, He, Mitsufuji,
  and Ermon]{kim2024ctm}
Dongjun Kim, Chieh-Hsin Lai, Wei-Hsiang Liao, Naoki Murata, Yuhta Takida,
  Toshimitsu Uesaka, Yutong He, Yuki Mitsufuji, and Stefano Ermon.
\newblock Consistency trajectory models: Learning probability flow {ODE}
  trajectory of diffusion.
\newblock In \emph{International Conference on Learning Representations}, 2024.

\bibitem[Lei et~al.(2026)Lei, Liu, Berner, Yu, Zheng, Wu, and Chu]{lei2026epg}
Jiachen Lei, Keli Liu, Julius Berner, Haiming Yu, Hongkai Zheng, Jiahong Wu,
  and Xiangxiang Chu.
\newblock There is no {VAE}: End-to-end pixel-space generative modeling via
  self-supervised pre-training.
\newblock In \emph{International Conference on Learning Representations}, 2026.

\bibitem[Li and He(2026)]{li2026jit}
Tianhong Li and Kaiming He.
\newblock Back to basics: Let denoising generative models denoise.
\newblock In \emph{Proceedings of the IEEE/CVF Conference on Computer Vision
  and Pattern Recognition}, pages 36115--36125, 2026.

\bibitem[Li~Cai and Garabito(2026)]{cai2026iris}
Hanqiu Li~Cai and Chema Garabito.
\newblock {Iris-3B}: Going beyond the latent with pixel-space diffusion
  training, conversion and fine-tuning.
\newblock \emph{arXiv preprint arXiv:2610.09450}, 2026.

\bibitem[Lin et~al.(2023)Lin, Gao, Tang, Takikawa, Zeng, Huang, Kreis, Fidler,
  Liu, and Lin]{lin2023magic3d}
Chen-Hsuan Lin, Jun Gao, Luming Tang, Towaki Takikawa, Xiaohui Zeng, Xun Huang,
  Karsten Kreis, Sanja Fidler, Ming-Yu Liu, and Tsung-Yi Lin.
\newblock {Magic3D}: High-resolution text-to-3d content creation.
\newblock In \emph{Proceedings of the IEEE/CVF Conference on Computer Vision
  and Pattern Recognition}, pages 300--309, 2023.

\bibitem[Lin et~al.(2024)Lin, Wang, and Yang]{lin2024sdxllightning}
Shanchuan Lin, Anran Wang, and Xiao Yang.
\newblock {SDXL-Lightning}: Progressive adversarial diffusion distillation.
\newblock \emph{arXiv preprint arXiv:2402.13929}, 2024.

\bibitem[Lin et~al.(2026)Lin, Zhang, Zhou, Zheng, Liu, Yang, Lin, Yang, and
  Nguyen]{lin2026dma2}
Xin Lin, Zhifei Zhang, Yuqian Zhou, Haitian Zheng, Shaoteng Liu, Lehan Yang,
  Zhe Lin, Ming-Hsuan Yang, and Truong Nguyen.
\newblock {DMA}$^2$: Pixel-space distribution matching with adversarial and
  anchor losses.
\newblock \emph{arXiv preprint arXiv:2609.38156}, 2026.

\bibitem[Liu et~al.(2024)Liu, Zhang, Ma, Peng, and Liu]{liu2024instaflow}
Xingchao Liu, Xiwen Zhang, Jianzhu Ma, Jian Peng, and Qiang Liu.
\newblock {InstaFlow}: One step is enough for high-quality diffusion-based
  text-to-image generation.
\newblock In \emph{International Conference on Learning Representations}, 2024.

\bibitem[Lu et~al.(2025)Lu, Ren, Xia, Lin, Wang, Xiao, Ma, Xie, and
  Lai]{lu2025dmdx}
Yanzuo Lu, Yuxi Ren, Xin Xia, Shanchuan Lin, Xing Wang, Xuefeng Xiao, Andy~J.
  Ma, Xiaohua Xie, and Jian-Huang Lai.
\newblock Adversarial distribution matching for diffusion distillation towards
  efficient image and video synthesis.
\newblock In \emph{Proceedings of the IEEE/CVF International Conference on
  Computer Vision}, pages 16818--16829, 2025.

\bibitem[Lu et~al.(2026)Lu, Lu, Sun, Zhao, Jiang, Wang, Li, Geng, and
  He]{lu2026pmf}
Yiyang Lu, Susie Lu, Qiao Sun, Hanhong Zhao, Zhicheng Jiang, Xianbang Wang,
  Tianhong Li, Zhengyang Geng, and Kaiming He.
\newblock One-step latent-free image generation with pixel mean flows.
\newblock \emph{arXiv preprint arXiv:2601.22158}, 2026.

\bibitem[Luhman and Luhman(2021)]{luhman2021knowledge}
Eric Luhman and Troy Luhman.
\newblock Knowledge distillation in iterative generative models for improved
  sampling speed.
\newblock \emph{arXiv preprint arXiv:2101.02388}, 2021.

\bibitem[Luo et~al.(2023{\natexlab{a}})Luo, Tan, Huang, Li, and
  Zhao]{luo2023lcm}
Simian Luo, Yiqin Tan, Longbo Huang, Jian Li, and Hang Zhao.
\newblock Latent consistency models: Synthesizing high-resolution images with
  few-step inference.
\newblock \emph{arXiv preprint arXiv:2310.04378}, 2023{\natexlab{a}}.

\bibitem[Luo et~al.(2023{\natexlab{b}})Luo, Tan, Patil, Gu, von Platen, Passos,
  Huang, Li, and Zhao]{luo2023lcmlora}
Simian Luo, Yiqin Tan, Suraj Patil, Daniel Gu, Patrick von Platen,
  Apolin{\'a}rio Passos, Longbo Huang, Jian Li, and Hang Zhao.
\newblock {LCM-LoRA}: A universal stable-diffusion acceleration module.
\newblock \emph{arXiv preprint arXiv:2311.05556}, 2023{\natexlab{b}}.

\bibitem[Luo et~al.(2023{\natexlab{c}})Luo, Hu, Zhang, Sun, Li, and
  Zhang]{luo2023diffinstruct}
Weijian Luo, Tianyang Hu, Shifeng Zhang, Jiacheng Sun, Zhenguo Li, and Zhihua
  Zhang.
\newblock Diff-instruct: A universal approach for transferring knowledge from
  pre-trained diffusion models.
\newblock In \emph{Advances in Neural Information Processing Systems},
  2023{\natexlab{c}}.

\bibitem[Luo et~al.(2024)Luo, Huang, Geng, Kolter, and Qi]{luo2024sim}
Weijian Luo, Zemin Huang, Zhengyang Geng, J.~Zico Kolter, and Guo-Jun Qi.
\newblock One-step diffusion distillation through score implicit matching.
\newblock In \emph{Advances in Neural Information Processing Systems}, 2024.

\bibitem[Ma et~al.(2024)Ma, Goldstein, Albergo, Boffi, Vanden-Eijnden, and
  Xie]{ma2024sit}
Nanye Ma, Mark Goldstein, Michael~S. Albergo, Nicholas~M. Boffi, Eric
  Vanden-Eijnden, and Saining Xie.
\newblock {SiT}: Exploring flow and diffusion-based generative models with
  scalable interpolant transformers.
\newblock In \emph{European Conference on Computer Vision}, 2024.

\bibitem[Ma et~al.(2025)Ma, Wu, Sun, and Li]{ma2025hpsv3}
Yuhang Ma, Xiaoshi Wu, Keqiang Sun, and Hongsheng Li.
\newblock {HPSv3}: Towards wide-spectrum human preference score.
\newblock In \emph{Proceedings of the IEEE/CVF International Conference on
  Computer Vision}, pages 15086--15095, 2025.

\bibitem[Ma et~al.(2026{\natexlab{a}})Ma, Wei, Wang, Zhang, and
  Tian]{ma2026deco}
Zehong Ma, Longhui Wei, Shuai Wang, Shiliang Zhang, and Qi Tian.
\newblock {DeCo}: Frequency-decoupled pixel diffusion for end-to-end image
  generation.
\newblock In \emph{Proceedings of the IEEE/CVF Conference on Computer Vision
  and Pattern Recognition}, pages 43600--43610, 2026{\natexlab{a}}.

\bibitem[Ma et~al.(2026{\natexlab{b}})Ma, Xu, and Zhang]{ma2026pixelgen}
Zehong Ma, Ruihan Xu, and Shiliang Zhang.
\newblock {PixelGen}: Improving pixel diffusion with perceptual supervision.
\newblock \emph{arXiv preprint arXiv:2602.02493}, 2026{\natexlab{b}}.

\bibitem[Meng et~al.(2023)Meng, Rombach, Gao, Kingma, Ermon, Ho, and
  Salimans]{meng2023guided}
Chenlin Meng, Robin Rombach, Ruiqi Gao, Diederik Kingma, Stefano Ermon,
  Jonathan Ho, and Tim Salimans.
\newblock On distillation of guided diffusion models.
\newblock In \emph{Proceedings of the IEEE/CVF Conference on Computer Vision
  and Pattern Recognition}, pages 14297--14306, 2023.

\bibitem[{MiniMax}(2026)]{minimax2026h3image}
{MiniMax}.
\newblock {MiniMax-H3-Image}.
\newblock \url{https://github.com/MiniMax-AI/MiniMax-H3-Image}, 2026.

\bibitem[{ModelTC}(2025)]{modeltc2025wanlightning}
{ModelTC}.
\newblock {Wan2.2-Lightning}.
\newblock GitHub repository,
  \url{https://github.com/ModelTC/LightX2V-Wan2.2-Lightning}, 2025.
\newblock Accessed 2026-11-24.

\bibitem[Nguyen et~al.(2026)Nguyen, Nguyen, Vu, Dao, Nguyen, Dao, Nguyen, Tran,
  Nguyen, Nguyen, Pham, Metaxas, Patel, and Tran]{nguyen2026crossspace}
Anh Nguyen, Ngan Nguyen, Duc Vu, Trung Dao, Viet Nguyen, Quan Dao, Kien Nguyen,
  Chi Tran, Phong Nguyen, Khoi Nguyen, Cuong Pham, Dimitris Metaxas, Vishal~M.
  Patel, and Anh Tran.
\newblock Cross-space distillation: Teaching one-step students with modern
  diffusion teachers.
\newblock In \emph{Computer Vision -- ECCV 2026}, pages 275--292. Springer,
  2026.

\bibitem[Nguyen and Tran(2024)]{nguyen2024swiftbrush}
Thuan~Hoang Nguyen and Anh Tran.
\newblock {SwiftBrush}: One-step text-to-image diffusion model with variational
  score distillation.
\newblock In \emph{Proceedings of the IEEE/CVF Conference on Computer Vision
  and Pattern Recognition}, pages 7807--7816, 2024.

\bibitem[Nichol et~al.(2022)Nichol, Dhariwal, Ramesh, Shyam, Mishkin, McGrew,
  Sutskever, and Chen]{nichol2022glide}
Alexander~Quinn Nichol, Prafulla Dhariwal, Aditya Ramesh, Pranav Shyam, Pamela
  Mishkin, Bob McGrew, Ilya Sutskever, and Mark Chen.
\newblock {GLIDE}: Towards photorealistic image generation and editing with
  text-guided diffusion models.
\newblock In \emph{Proceedings of the 39th International Conference on Machine
  Learning}, pages 16784--16804, 2022.

\bibitem[{NVIDIA}(2026)]{nvidia2026cosmos3}
{NVIDIA}.
\newblock Models -- {Cosmos-Framework}.
\newblock NVIDIA documentation,
  \url{https://nvidia.github.io/cosmos-framework/models/}, 2026.
\newblock Section ``Super Text2Image and Image2Video''; accessed 2026-11-24.

\bibitem[{NVIDIA} et~al.(2024){NVIDIA}, Atzmon, Bala, Balaji, Cai, Cui, Fan,
  Ge, Gururani, Huffman, Isaac, Jannaty, Karras, Lam, Lewis, Licata, Lin, Liu,
  Ma, Mallya, Martino-Tarr, Mendez, Nah, Pruett, Reda, Song, Wang, Wei, Zeng,
  Zeng, and Zhang]{nvidia2024edify}
{NVIDIA}, Yuval Atzmon, Maciej Bala, Yogesh Balaji, Tiffany Cai, Yin Cui,
  Jiaojiao Fan, Yunhao Ge, Siddharth Gururani, Jacob Huffman, Ronald Isaac,
  Pooya Jannaty, Tero Karras, Grace Lam, J.~P. Lewis, Aaron Licata, Yen-Chen
  Lin, Ming-Yu Liu, Qianli Ma, Arun Mallya, Ashlee Martino-Tarr, Doug Mendez,
  Seungjun Nah, Chris Pruett, Fitsum Reda, Jiaming Song, Ting-Chun Wang,
  Fangyin Wei, Xiaohui Zeng, Yu Zeng, and Qinsheng Zhang.
\newblock Edify image: High-quality image generation with pixel space laplacian
  diffusion models.
\newblock \emph{arXiv preprint arXiv:2411.07126}, 2024.

\bibitem[Pan et~al.(2026)Pan, Chen, Shi, Hu, Yue, Wang, Wang, Li, and
  Yang]{pan2026representation}
Zihao Pan, Mengting Chen, Cheng Shi, Taihang Hu, Zhengrong Yue, Zikang Wang,
  Yuan Wang, Wei Li, and Sibei Yang.
\newblock Learning representations to guide pixel diffusion transformers.
\newblock \emph{arXiv preprint arXiv:2610.12978}, 2026.

\bibitem[{Qwen Team}(2026)]{qwen2026qwenimage21}
{Qwen Team}.
\newblock {Qwen-Image-2.1}.
\newblock GitHub repository, \url{https://github.com/QwenLM/Qwen-Image-2.1},
  2026.
\newblock Accessed 2026-11-24.

\bibitem[Rombach et~al.(2022)Rombach, Blattmann, Lorenz, Esser, and
  Ommer]{rombach2022ldm}
Robin Rombach, Andreas Blattmann, Dominik Lorenz, Patrick Esser, and Bj{\"o}rn
  Ommer.
\newblock High-resolution image synthesis with latent diffusion models.
\newblock In \emph{Proceedings of the IEEE/CVF Conference on Computer Vision
  and Pattern Recognition}, 2022.

\bibitem[Saharia et~al.(2022)Saharia, Chan, Saxena, Li, Whang, Denton,
  Ghasemipour, Lopes, Ayan, Salimans, Ho, Fleet, and
  Norouzi]{saharia2022imagen}
Chitwan Saharia, William Chan, Saurabh Saxena, Lala Li, Jay Whang, Emily~L.
  Denton, Kamyar Ghasemipour, Raphael~Gontijo Lopes, Burcu~Karagol Ayan, Tim
  Salimans, Jonathan Ho, David~J. Fleet, and Mohammad Norouzi.
\newblock Photorealistic text-to-image diffusion models with deep language
  understanding.
\newblock In \emph{Advances in Neural Information Processing Systems}, pages
  36479--36494, 2022.

\bibitem[Salimans and Ho(2022)]{salimans2022progressive}
Tim Salimans and Jonathan Ho.
\newblock Progressive distillation for fast sampling of diffusion models.
\newblock In \emph{International Conference on Learning Representations}, 2022.

\bibitem[Salimans et~al.(2024)Salimans, Mensink, Heek, and
  Hoogeboom]{salimans2024moment}
Tim Salimans, Thomas Mensink, Jonathan Heek, and Emiel Hoogeboom.
\newblock Multistep distillation of diffusion models via moment matching.
\newblock In \emph{Advances in Neural Information Processing Systems}, 2024.

\bibitem[Sauer et~al.(2024{\natexlab{a}})Sauer, Boesel, Dockhorn, Blattmann,
  Esser, and Rombach]{sauer2024ladd}
Axel Sauer, Frederic Boesel, Tim Dockhorn, Andreas Blattmann, Patrick Esser,
  and Robin Rombach.
\newblock Fast high-resolution image synthesis with latent adversarial
  diffusion distillation.
\newblock In \emph{{SIGGRAPH} Asia 2024 Conference Papers}, pages
  106:1--106:11, 2024{\natexlab{a}}.

\bibitem[Sauer et~al.(2024{\natexlab{b}})Sauer, Lorenz, Blattmann, and
  Rombach]{sauer2024add}
Axel Sauer, Dominik Lorenz, Andreas Blattmann, and Robin Rombach.
\newblock Adversarial diffusion distillation.
\newblock In \emph{European Conference on Computer Vision}, pages 87--103,
  2024{\natexlab{b}}.

\bibitem[Shin et~al.(2026)Shin, Kim, and Shim]{shin2026pixelrepa}
Jaeyo Shin, Jiwook Kim, and Hyunjung Shim.
\newblock Representation alignment for just image transformers is not easier
  than you think.
\newblock \emph{arXiv preprint arXiv:2603.14366}, 2026.

\bibitem[Singer et~al.(2023)Singer, Polyak, Hayes, Yin, An, Zhang, Hu, Yang,
  Ashual, Gafni, Parikh, Gupta, and Taigman]{singer2023makeavideo}
Uriel Singer, Adam Polyak, Thomas Hayes, Xi Yin, Jie An, Songyang Zhang, Qiyuan
  Hu, Harry Yang, Oron Ashual, Oran Gafni, Devi Parikh, Sonal Gupta, and Yaniv
  Taigman.
\newblock Make-a-video: Text-to-video generation without text-video data.
\newblock In \emph{International Conference on Learning Representations}, 2023.

\bibitem[Skorokhodov(2026)]{skorokhodov2026pixsds}
Vsevolod Skorokhodov.
\newblock {PixSDS}: Why latent {SDS} makes noisy pixels.
\newblock \emph{arXiv preprint arXiv:2608.12997}, 2026.

\bibitem[Song et~al.(2023)Song, Dhariwal, Chen, and
  Sutskever]{song2023consistency}
Yang Song, Prafulla Dhariwal, Mark Chen, and Ilya Sutskever.
\newblock Consistency models.
\newblock In \emph{Proceedings of the 40th International Conference on Machine
  Learning}, pages 32211--32252, 2023.

\bibitem[Sun et~al.(2025)Sun, Zhang, Wang, Wu, Wang, Wang, Wang, Zhang, Wang,
  and Guo]{tencent2025hyworldplay}
Wenqiang Sun, Haiyu Zhang, Haoyuan Wang, Junta Wu, Zehan Wang, Zhenwei Wang,
  Yunhong Wang, Jun Zhang, Tengfei Wang, and Chunchao Guo.
\newblock {WorldPlay}: Towards long-term geometric consistency for real-time
  interactive world modeling.
\newblock \emph{arXiv preprint arXiv:2512.14614}, 2025.

\bibitem[Vallaeys et~al.(2026)Vallaeys, Verbeek, and Cord]{vallaeys2026ujit}
Th{\'e}ophane Vallaeys, Jakob Verbeek, and Matthieu Cord.
\newblock Pixel-space video generation with {U-JiT}.
\newblock \emph{arXiv preprint arXiv:2611.04764}, 2026.

\bibitem[Wang et~al.(2023{\natexlab{a}})Wang, Du, Li, Yeh, and
  Shakhnarovich]{wang2023sjc}
Haochen Wang, Xiaodan Du, Jiahao Li, Raymond~A. Yeh, and Greg Shakhnarovich.
\newblock Score jacobian chaining: Lifting pretrained 2d diffusion models for
  3d generation.
\newblock In \emph{Proceedings of the IEEE/CVF Conference on Computer Vision
  and Pattern Recognition}, pages 12619--12629, 2023{\natexlab{a}}.

\bibitem[Wang et~al.(2026{\natexlab{a}})Wang, Li, and Guo]{wang2026wit}
Hainuo Wang, Mingjia Li, and Xiaojie Guo.
\newblock {WiT}: Waypoint diffusion transformers for alleviating trajectory
  conflict in pixel-space image generation.
\newblock \emph{arXiv preprint arXiv:2603.15132}, 2026{\natexlab{a}}.

\bibitem[Wang et~al.(2026{\natexlab{b}})Wang, Zhang, Li, Jiang, Zhong, Bo, and
  Zhang]{wang2026crossflow}
Xiyuan Wang, Xiao Zhang, Yang Li, Ruoxi Jiang, Zhao Zhong, Liefeng Bo, and
  Muhan Zhang.
\newblock {CrossFlow}: One-step generation across latent and pixel spaces.
\newblock \emph{arXiv preprint arXiv:2606.19970}, 2026{\natexlab{b}}.

\bibitem[Wang et~al.(2023{\natexlab{b}})Wang, Lu, Wang, Bao, Li, Su, and
  Zhu]{wang2023prolificdreamer}
Zhengyi Wang, Cheng Lu, Yikai Wang, Fan Bao, Chongxuan Li, Hang Su, and Jun
  Zhu.
\newblock {ProlificDreamer}: High-fidelity and diverse text-to-3d generation
  with variational score distillation.
\newblock In \emph{Advances in Neural Information Processing Systems},
  2023{\natexlab{b}}.

\bibitem[Wu et~al.(2026)Wu, Wang, Geng, Cao, Zhang, Zhang, Shechtman, and
  Cao]{wu2026bettercritic}
Zongze Wu, Xianbang Wang, Zhengyang Geng, Mingdeng Cao, Zhoutong Zhang, Richard
  Zhang, Eli Shechtman, and Ruiming Cao.
\newblock A better critic matters for distribution matching distillation.
\newblock \emph{arXiv preprint arXiv:2610.42135}, 2026.

\bibitem[Xu et~al.(2025)Xu, Nie, and Vahdat]{xu2025fdistill}
Yilun Xu, Weili Nie, and Arash Vahdat.
\newblock One-step diffusion models with {$f$}-divergence distribution
  matching.
\newblock \emph{arXiv preprint arXiv:2502.15681}, 2025.

\bibitem[Yin et~al.(2024{\natexlab{a}})Yin, Gharbi, Park, Zhang, Shechtman,
  Durand, and Freeman]{yin2024dmd2}
Tianwei Yin, Micha{\"e}l Gharbi, Taesung Park, Richard Zhang, Eli Shechtman,
  Fr{\'e}do Durand, and William~T. Freeman.
\newblock Improved distribution matching distillation for fast image synthesis.
\newblock In \emph{Advances in Neural Information Processing Systems},
  2024{\natexlab{a}}.

\bibitem[Yin et~al.(2024{\natexlab{b}})Yin, Gharbi, Zhang, Shechtman, Durand,
  Freeman, and Park]{yin2024dmd}
Tianwei Yin, Micha{\"e}l Gharbi, Richard Zhang, Eli Shechtman, Fr{\'e}do
  Durand, William~T. Freeman, and Taesung Park.
\newblock One-step diffusion with distribution matching distillation.
\newblock In \emph{Proceedings of the IEEE/CVF Conference on Computer Vision
  and Pattern Recognition}, pages 6613--6623, 2024{\natexlab{b}}.

\bibitem[Yu et~al.(2026)Yu, Xiong, Nie, Sheng, Liu, and Luo]{yu2026pixeldit}
Yongsheng Yu, Wei Xiong, Weili Nie, Yichen Sheng, Shiqiu Liu, and Jiebo Luo.
\newblock {PixelDiT}: Pixel diffusion transformers for image generation.
\newblock In \emph{Proceedings of the IEEE/CVF Conference on Computer Vision
  and Pattern Recognition}, pages 14273--14282, 2026.

\bibitem[{Z-Image Team} et~al.(2025){Z-Image Team}, Cai, Cao, Du, Gao, Hao,
  Hoi, Hou, Huang, Jiang, Jiang, Jin, Li, Li, Li, Liu, Liu, Wu, Yu, Zhan,
  Zhang, Zhang, Zhou, and Zhou]{cai2025zimage}
{Z-Image Team}, Huanqia Cai, Sihan Cao, Ruoyi Du, Peng Gao, Aiming Hao, Steven
  Hoi, Zhaohui Hou, Shijie Huang, Dengyang Jiang, Yuming Jiang, Xin Jin,
  Liangchen Li, Zhen Li, Zhong-Yu Li, David Liu, Dongyang Liu, Qilong Wu, Feng
  Yu, Zechao Zhan, Chi Zhang, Shifeng Zhang, Ruikai Zhou, and Shilin Zhou.
\newblock {Z-Image}: An efficient image generation foundation model with
  single-stream diffusion transformer.
\newblock \emph{arXiv preprint arXiv:2511.22699}, 2025.

\bibitem[Zhou et~al.(2024)Zhou, Zheng, Wang, Yin, and Huang]{zhou2024sid}
Mingyuan Zhou, Huangjie Zheng, Zhendong Wang, Mingzhang Yin, and Hai Huang.
\newblock Score identity distillation: Exponentially fast distillation of
  pretrained diffusion models for one-step generation.
\newblock In \emph{Proceedings of the 41st International Conference on Machine
  Learning}, pages 62307--62331, 2024.

\bibitem[Zhou et~al.(2026)Zhou, Zhou, Wu, Smith, Zhang, Luo, Shechtman, and
  Lin]{zhou2026editstream}
Yuqian Zhou, Zhenghong Zhou, Zongze Wu, Cameron Smith, Richard Zhang, Jiebo
  Luo, Eli Shechtman, and Zhe Lin.
\newblock {EditStream}: A unified autoregressive framework for interactive
  video generation and editing.
\newblock \emph{arXiv preprint arXiv:2608.21424}, 2026.

\end{thebibliography}
}
\clearpage
\onecolumn
\appendix
\setcounter{figure}{0}
\renewcommand{\thefigure}{S\arabic{figure}}
\renewcommand{\theHfigure}{S\arabic{figure}}
\setcounter{table}{0}
\renewcommand{\thetable}{S\arabic{table}}
\renewcommand{\theHtable}{S\arabic{table}}
\setcounter{equation}{0}
\renewcommand{\theequation}{S\arabic{equation}}
\renewcommand{\theHequation}{S\arabic{equation}}
\setcounter{algorithm}{0}
\renewcommand{\thealgorithm}{S\arabic{algorithm}}
\providecommand{\theHalgorithm}{S\arabic{algorithm}}

\begin{center}
    {\Large\bfseries \thetitle\par}
    \vspace{0.6em}
    {\large Supplementary Material}
\end{center}

\linenumbers
\section{Additional Qualitative Results}
\label{sec:supp-qualitative}

\subsection{Additional AsymFLUX Qualitative Comparisons}
\label{sec:supp-asymflux-gallery}

\begin{center}
    \captionsetup{hypcap=false}
    \captionof{figure}{\textbf{Additional AsymFLUX qualitative comparisons.}
    The pixel- and latent-matching Qwen-GAN students at DMD training
    iteration 3,000 (600 student updates) use four-step
    sampling without inference-time CFG; the AsymFLUX.2 Klein 9B pixel
    reference uses 50-step UniPC sampling with CFG 4.0 and APG.}
    \label{fig:supp-asymflux-qualitative}
\end{center}

\nolinenumbers
\begingroup
\setlength{\parindent}{0pt}
\newcommand{\asymfluxsuppsample}[5]{%
  \begin{minipage}[t]{\textwidth}%
    \centering
    \begin{tabular}{@{}c@{\hspace{0.018\textwidth}}c@{\hspace{0.018\textwidth}}c@{}}
      \scriptsize\bfseries Pixel matching &
      \scriptsize\bfseries Latent matching (Ours) &
      \scriptsize\bfseries Multi-step \\[0.15em]
      \includegraphics[width=#1]{#2} &
      \includegraphics[width=#1]{#3} &
      \includegraphics[width=#1]{#4}
    \end{tabular}\par
    \vspace{0.25em}%
    \parbox{0.91\textwidth}{\footnotesize\raggedright #5\par}%
  \end{minipage}%
  \par\vspace{0.35em}%
}

\asymfluxsuppsample
{0.24\textwidth}
{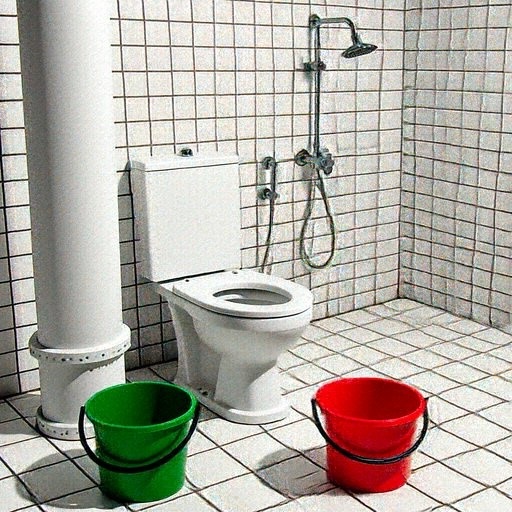}
{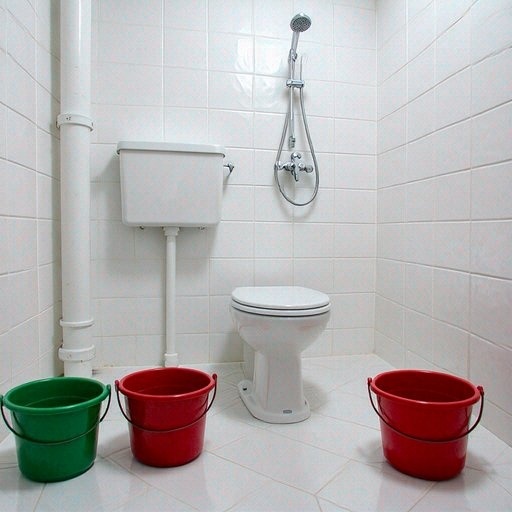}
{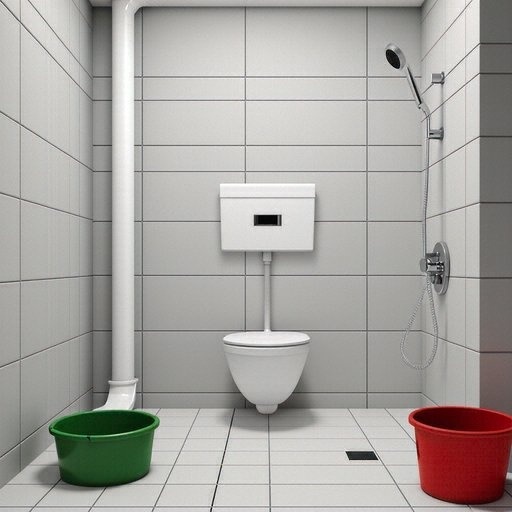}
{A white tiled bathroom contains a large white pipe, a wall-mounted toilet tank, a floor-level squat toilet, a shower head, and two buckets, one green and one red, on the floor.}
\asymfluxsuppsample
{0.24\textwidth}
{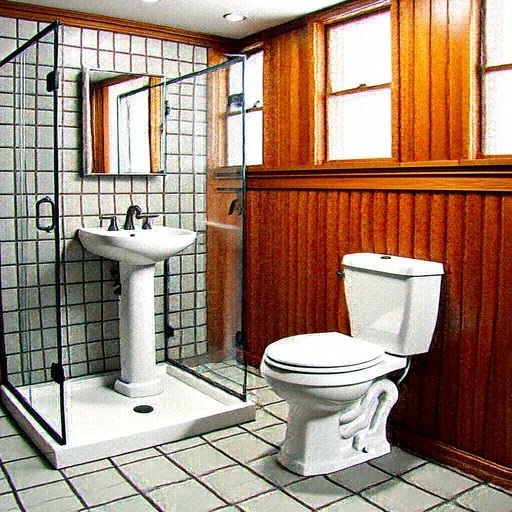}
{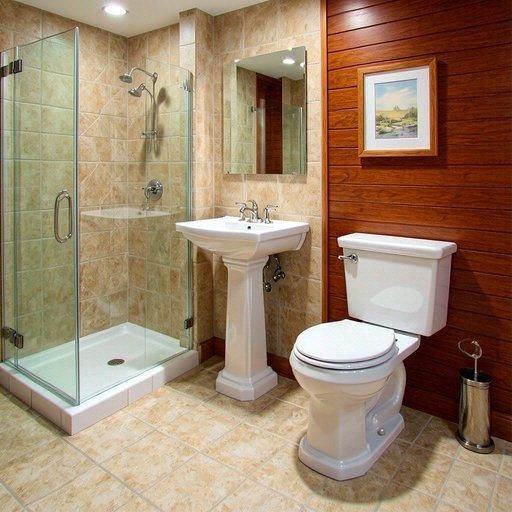}
{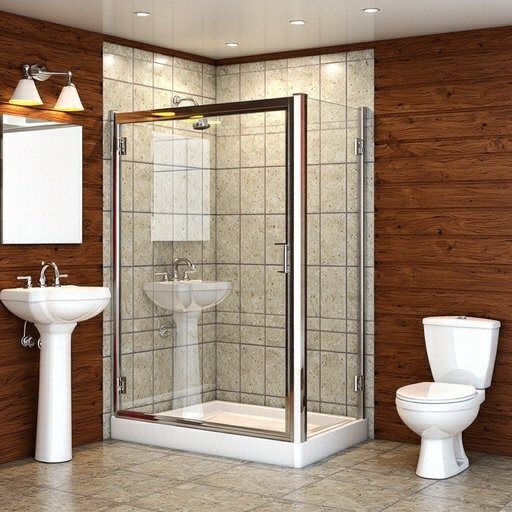}
{A bathroom features a glass-enclosed shower with tiled walls, a white pedestal sink, and a white toilet, arranged with a mirror above the sink, wooden paneling in the background, and tiled flooring under bright overhead lighting.}
\asymfluxsuppsample
{0.24\textwidth}
{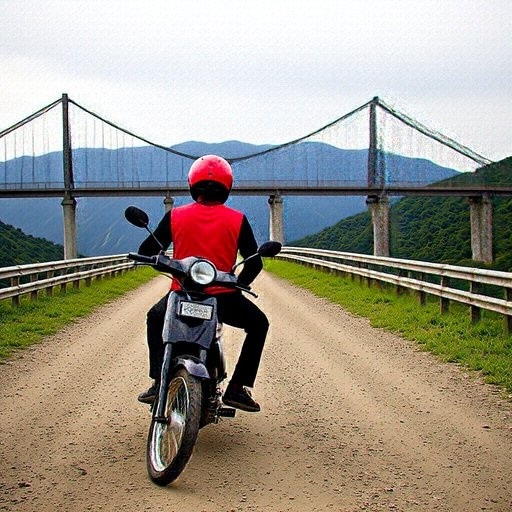}
{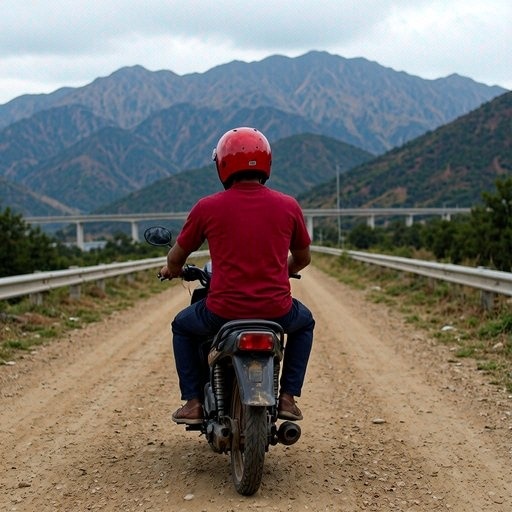}
{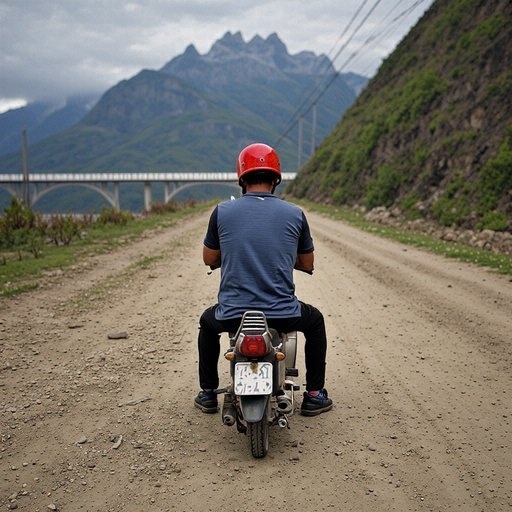}
{A man in a red helmet and shirt sits on a small moped on a dirt road, facing a mountain range with a bridge in the background under overcast light.}
\asymfluxsuppsample
{0.27\textwidth}
{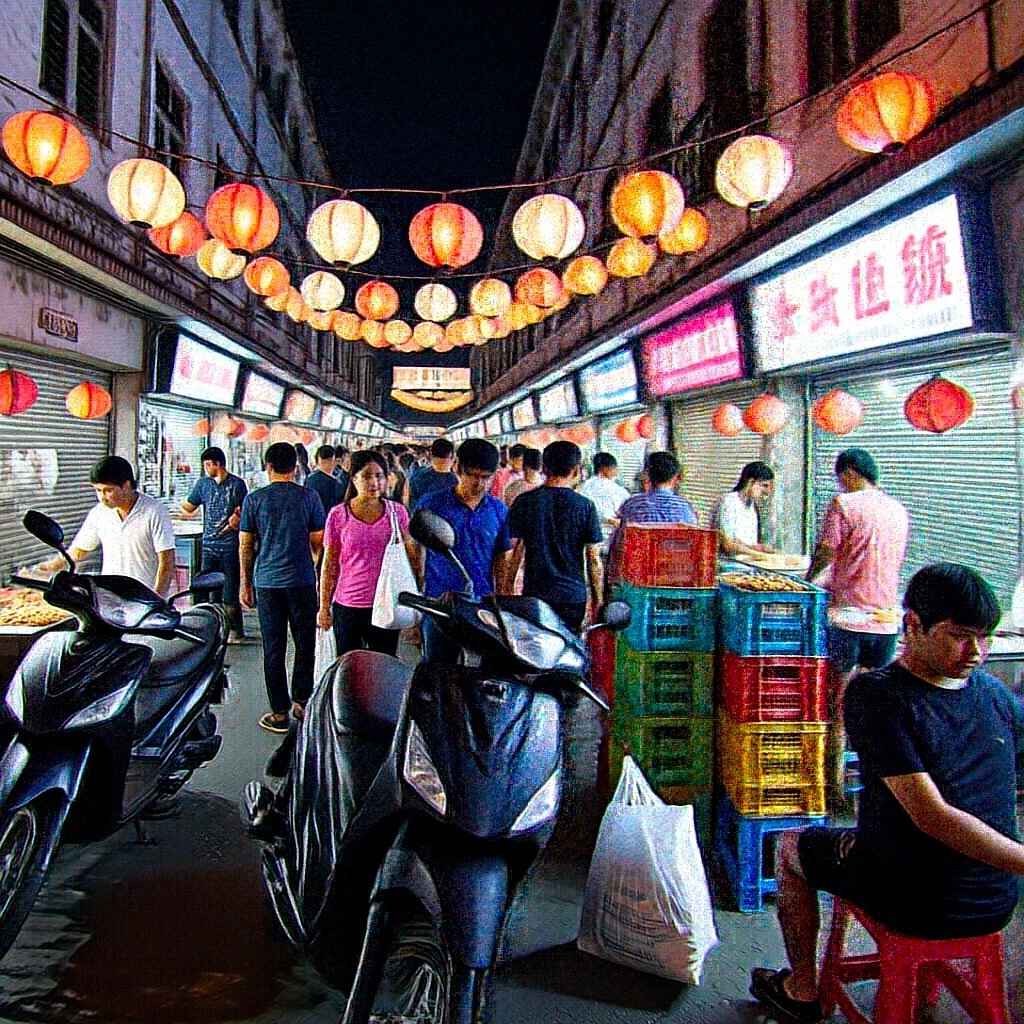}
{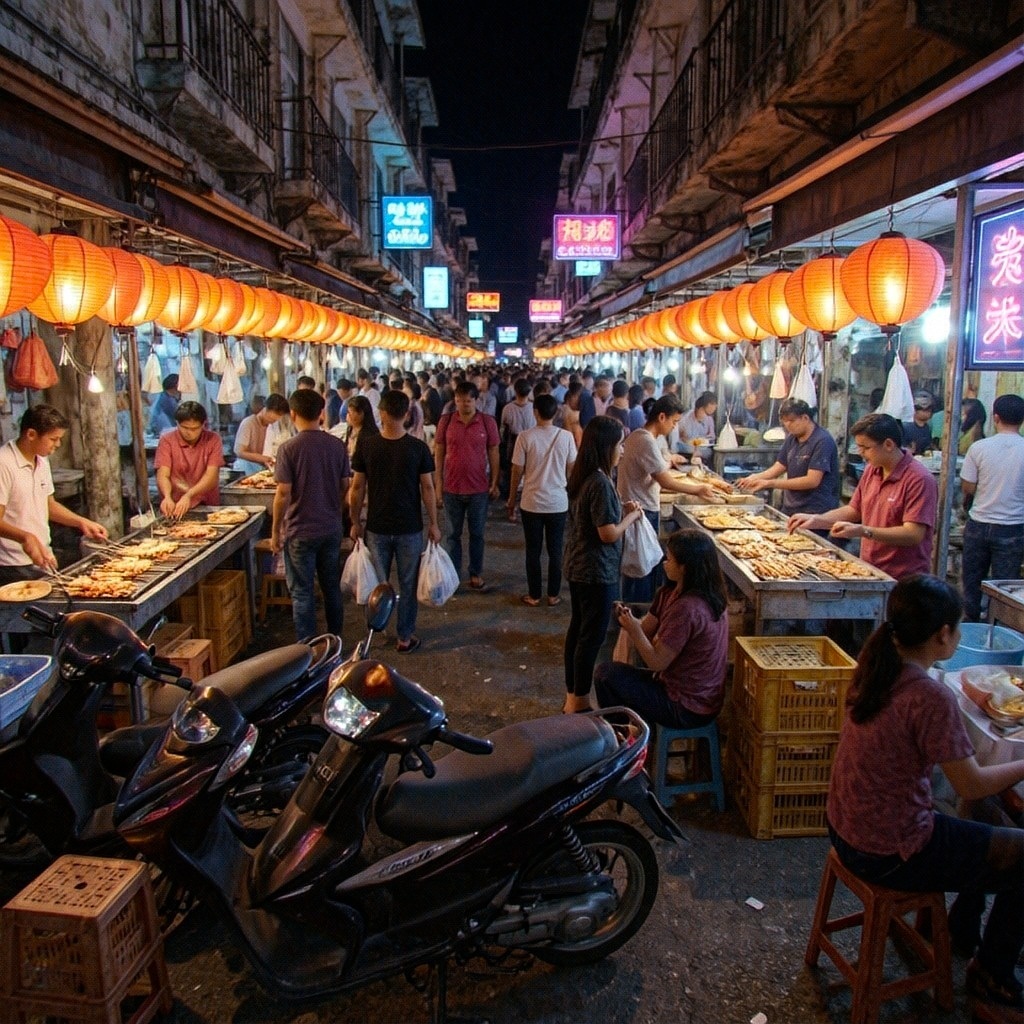}
{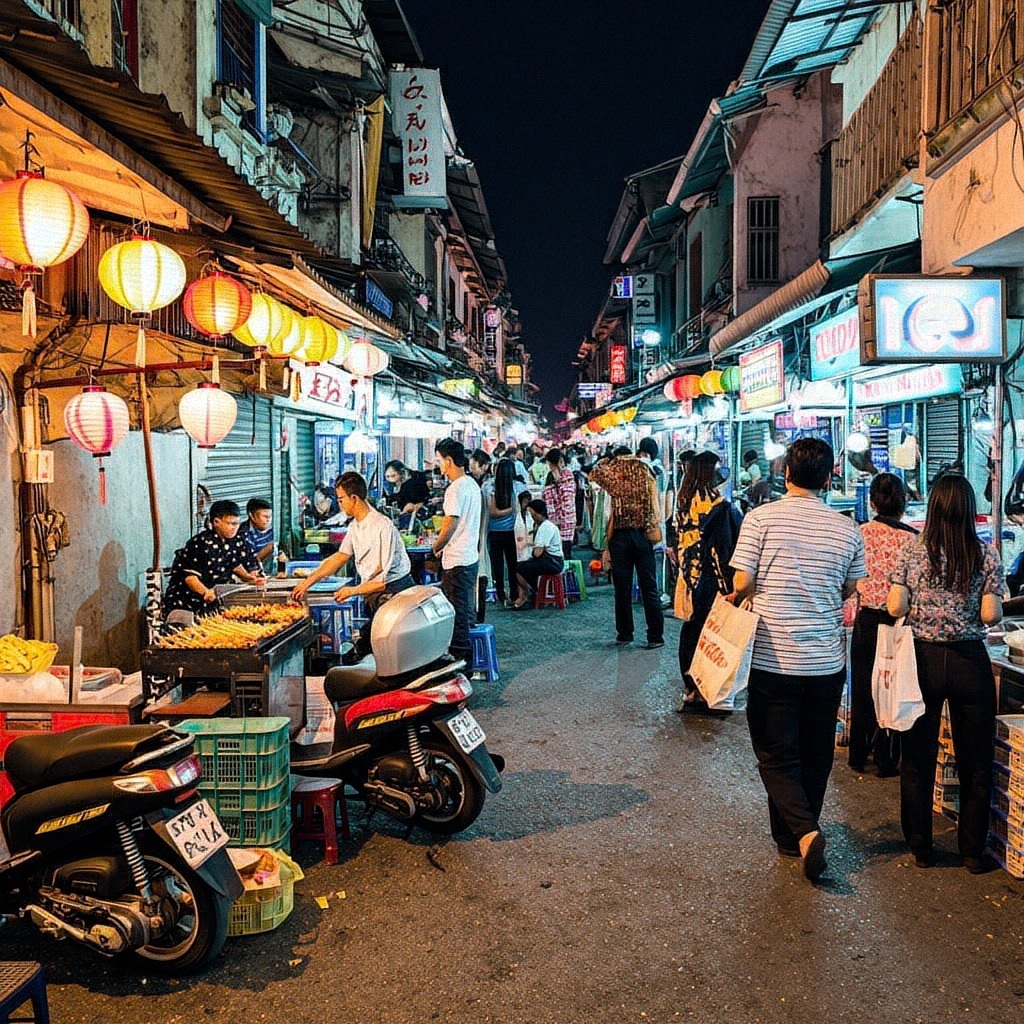}
{A crowded night-market alley lined with food stalls, where vendors grill skewers beneath paper lanterns, families carry shopping bags, two scooters are parked near stacked crates, diners sit on small stools, and neon signs glow against weathered balconies overhead.}
\asymfluxsuppsample
{0.27\textwidth}
{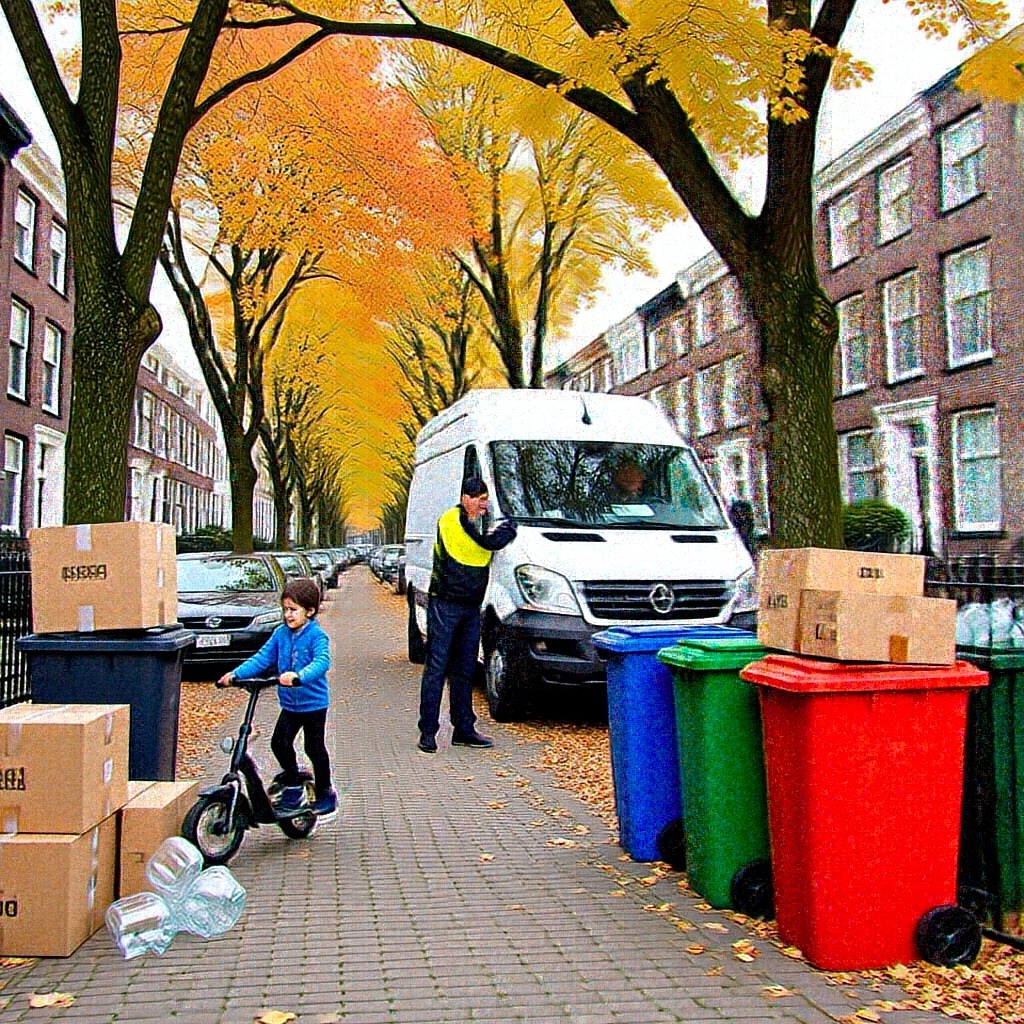}
{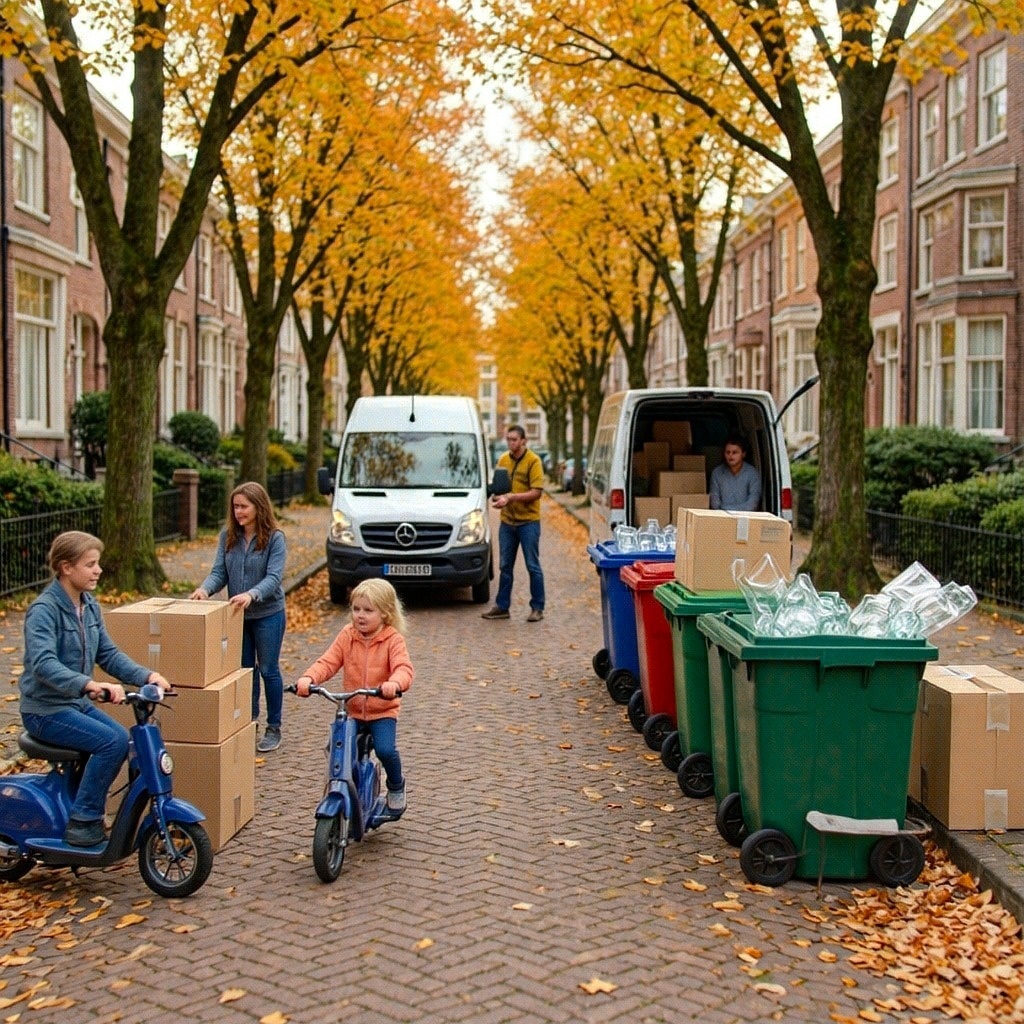}
{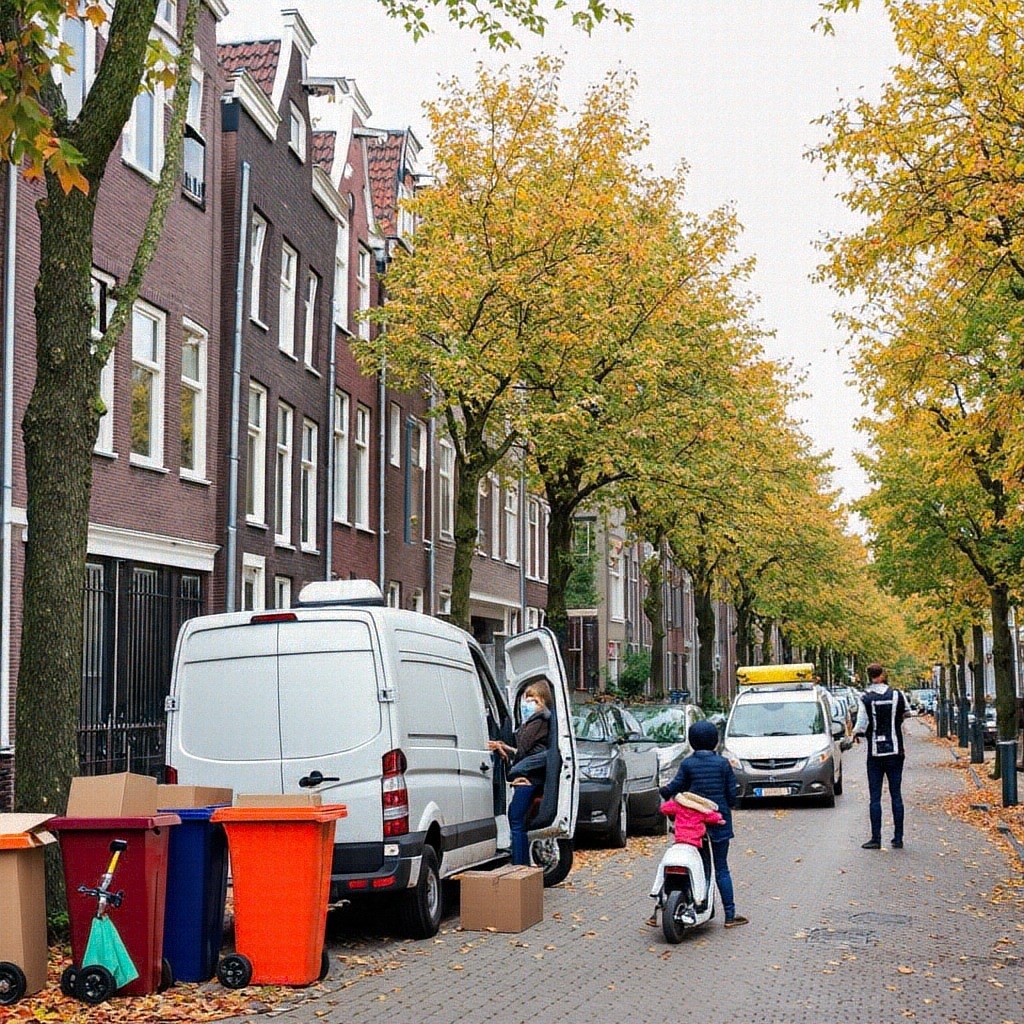}
{A residential street on recycling day, where neighbors sort cardboard and glass into colored bins, a courier unloads packages from a white van, a child rides a scooter, and rows of narrow houses stand beneath mature maple trees in soft autumn light.}
\asymfluxsuppsample
{0.27\textwidth}
{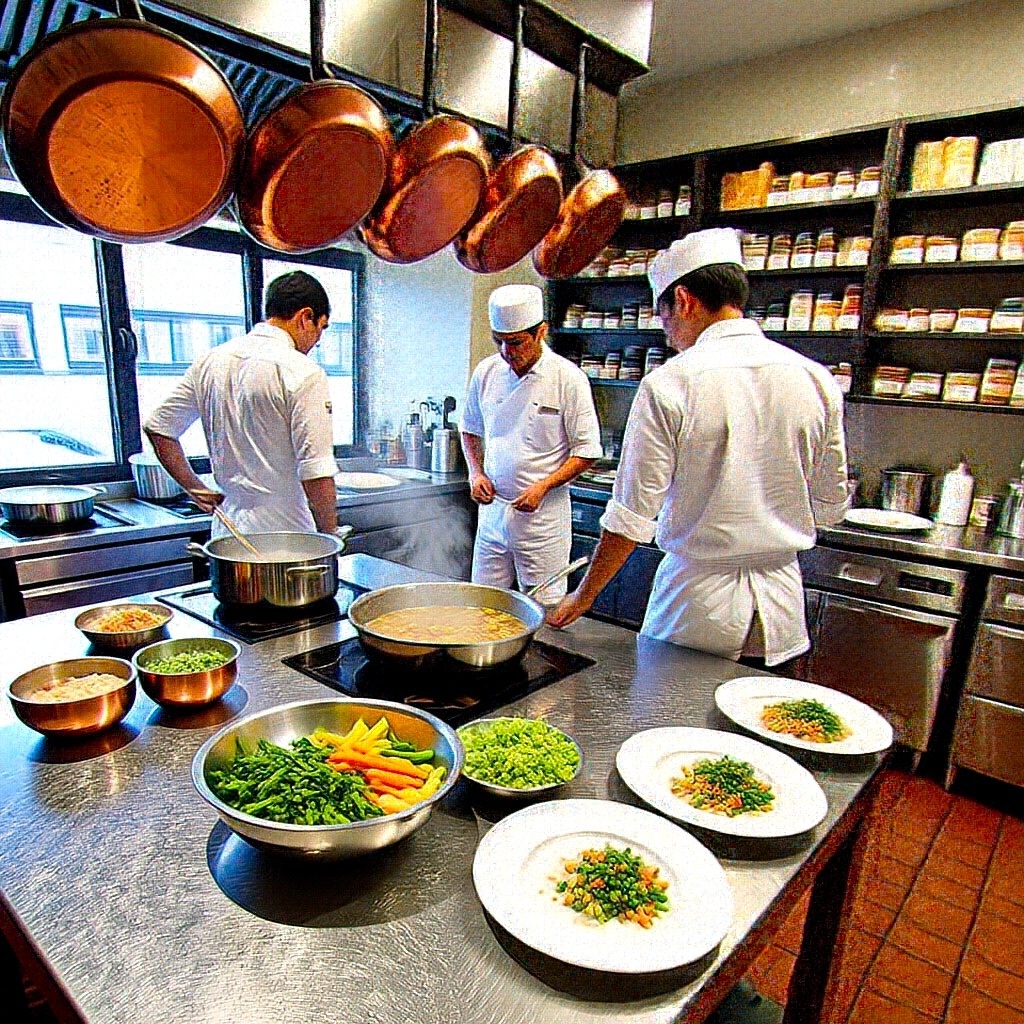}
{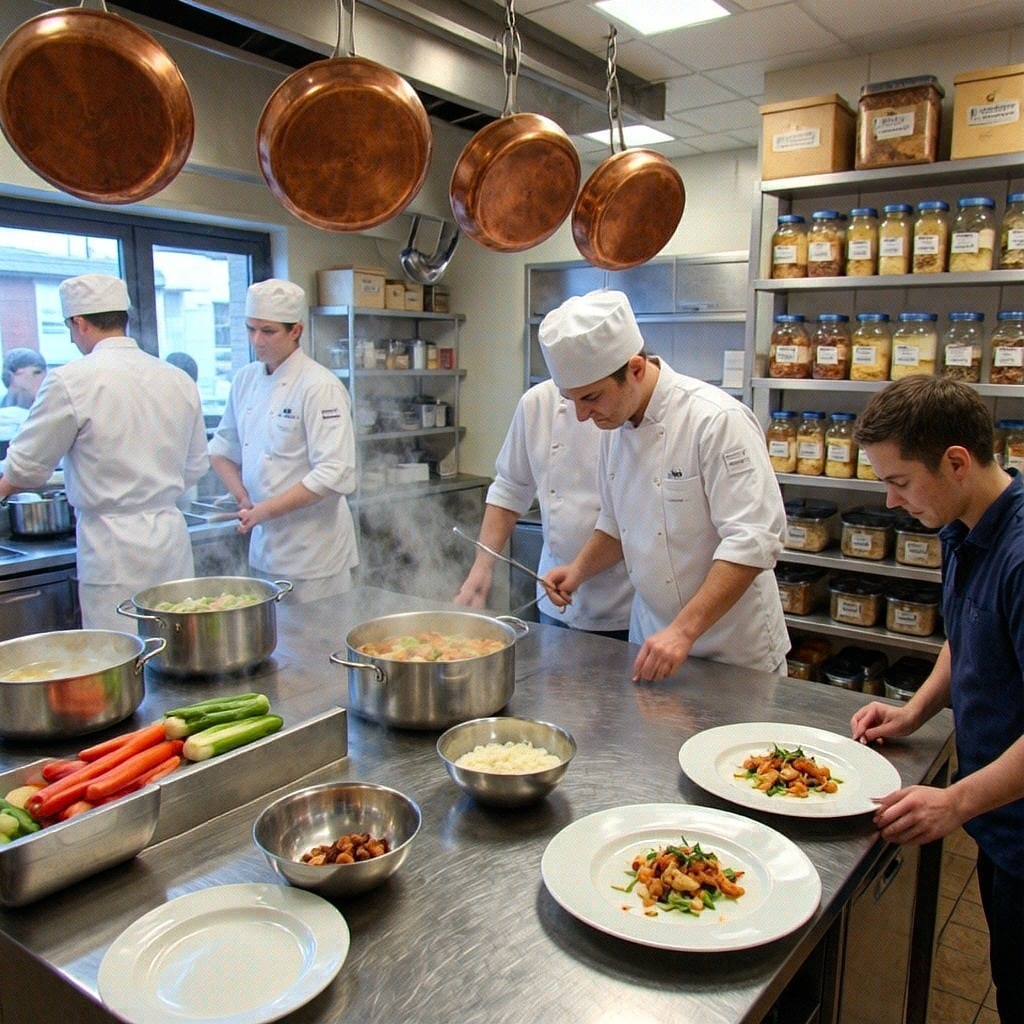}
{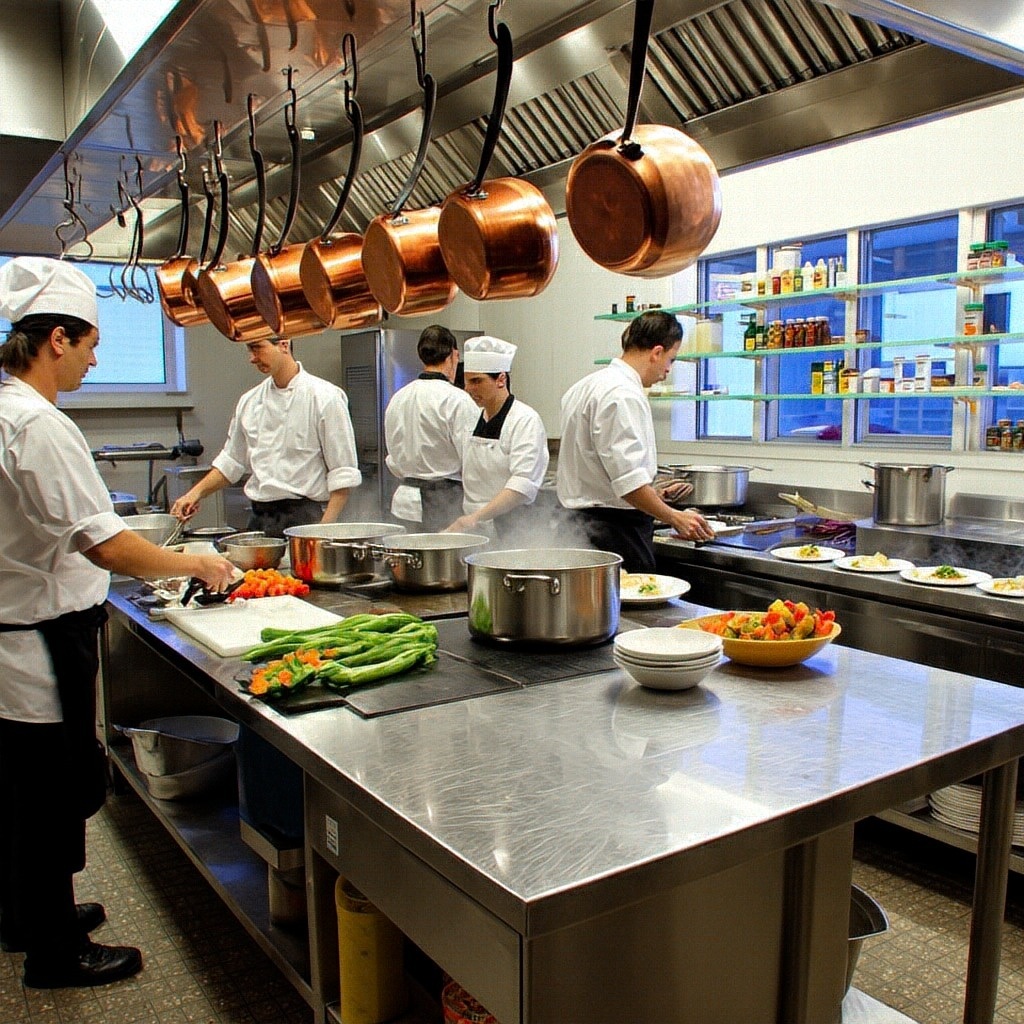}
{A bustling restaurant kitchen during dinner service, with four chefs working around a stainless-steel island, copper pans hanging overhead, vegetables and mixing bowls on the counter, steam rising from stockpots, plated meals waiting by a service window, and shelves of labeled ingredients.}
\asymfluxsuppsample
{0.27\textwidth}
{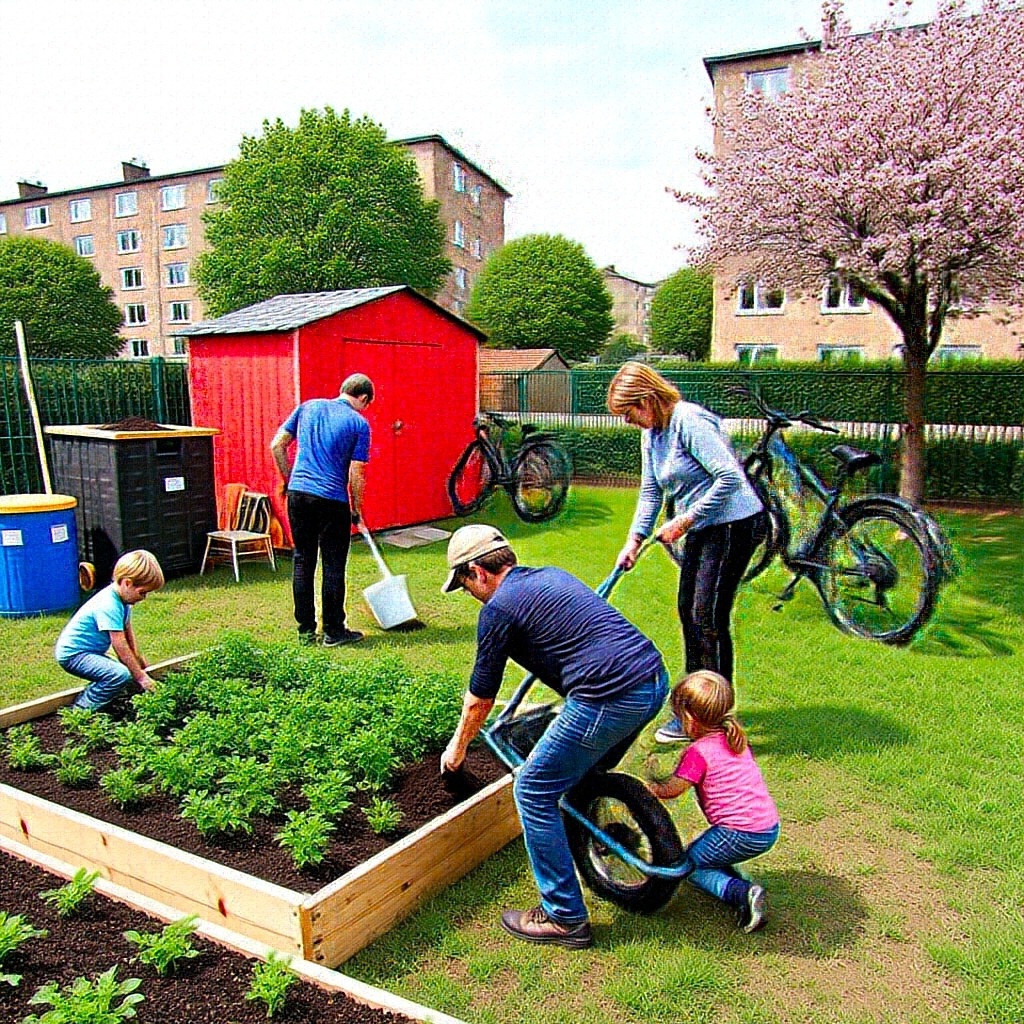}
{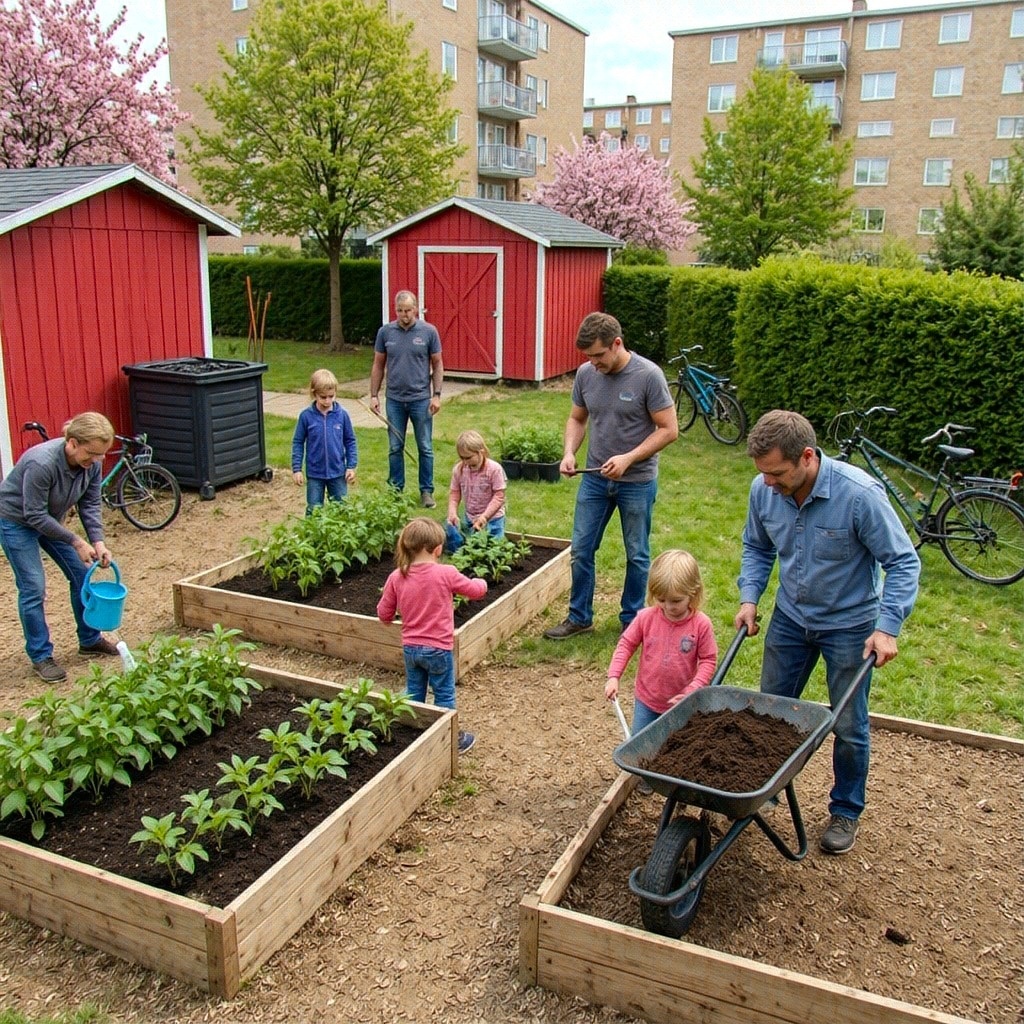}
{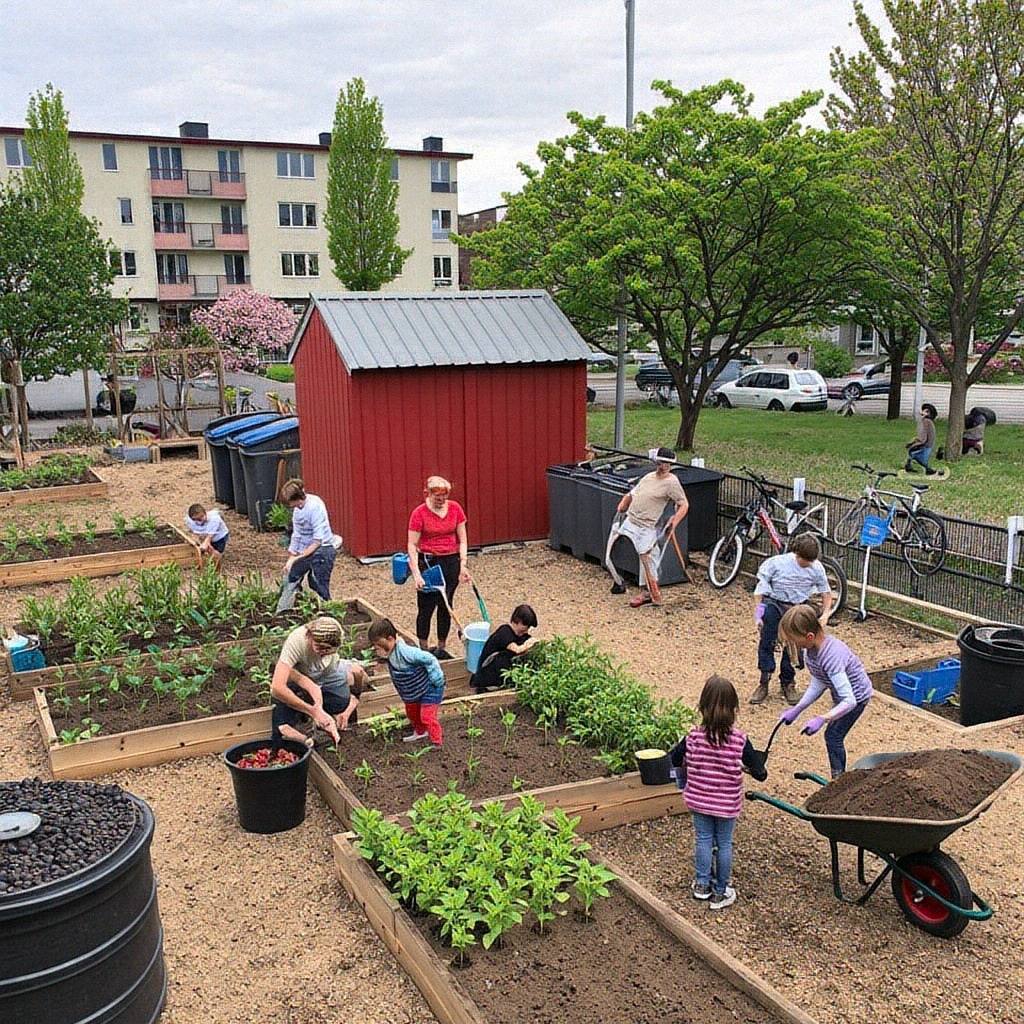}
{A community garden workday with adults and children planting seedlings in raised beds, one person watering herbs, another pushing a wheelbarrow of soil, and a red shed, compost bins, tools, bicycles, apartment buildings, and flowering trees surrounding the group.}
\asymfluxsuppsample
{0.27\textwidth}
{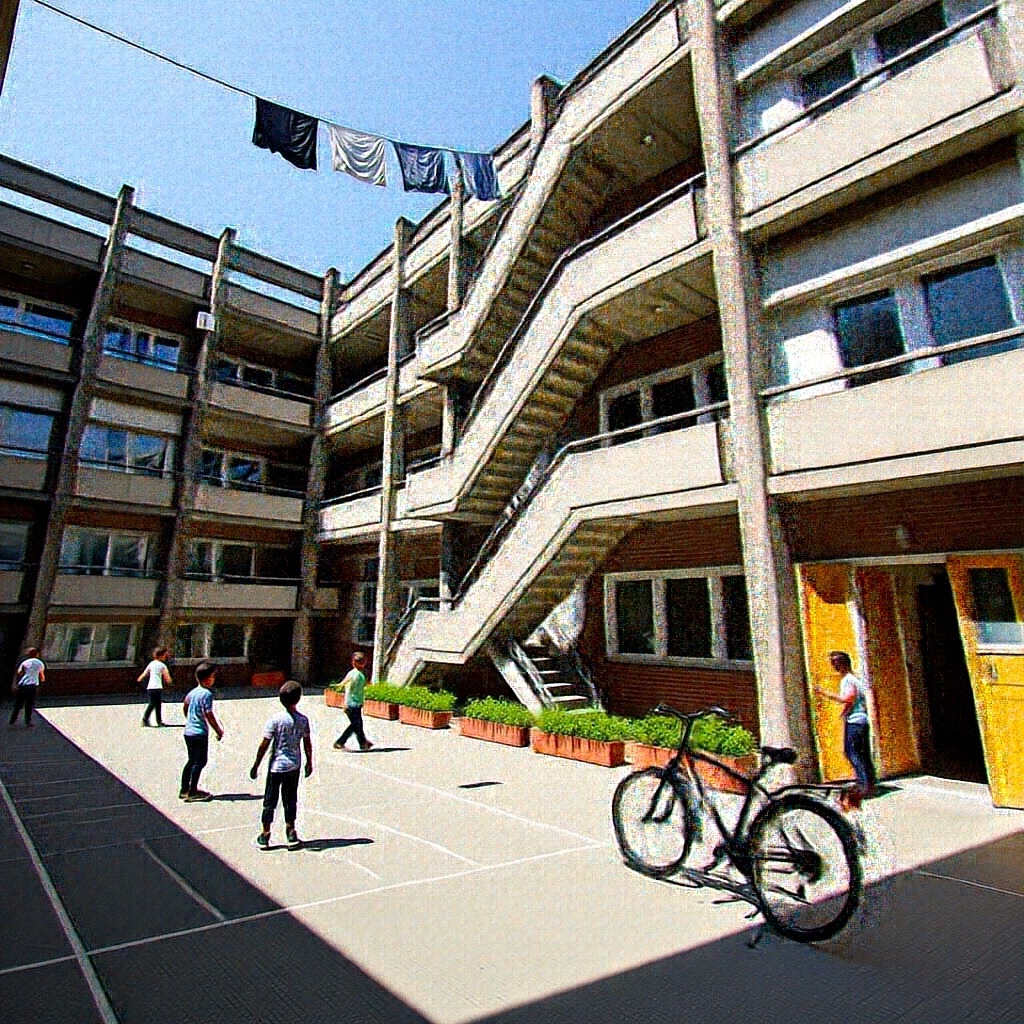}
{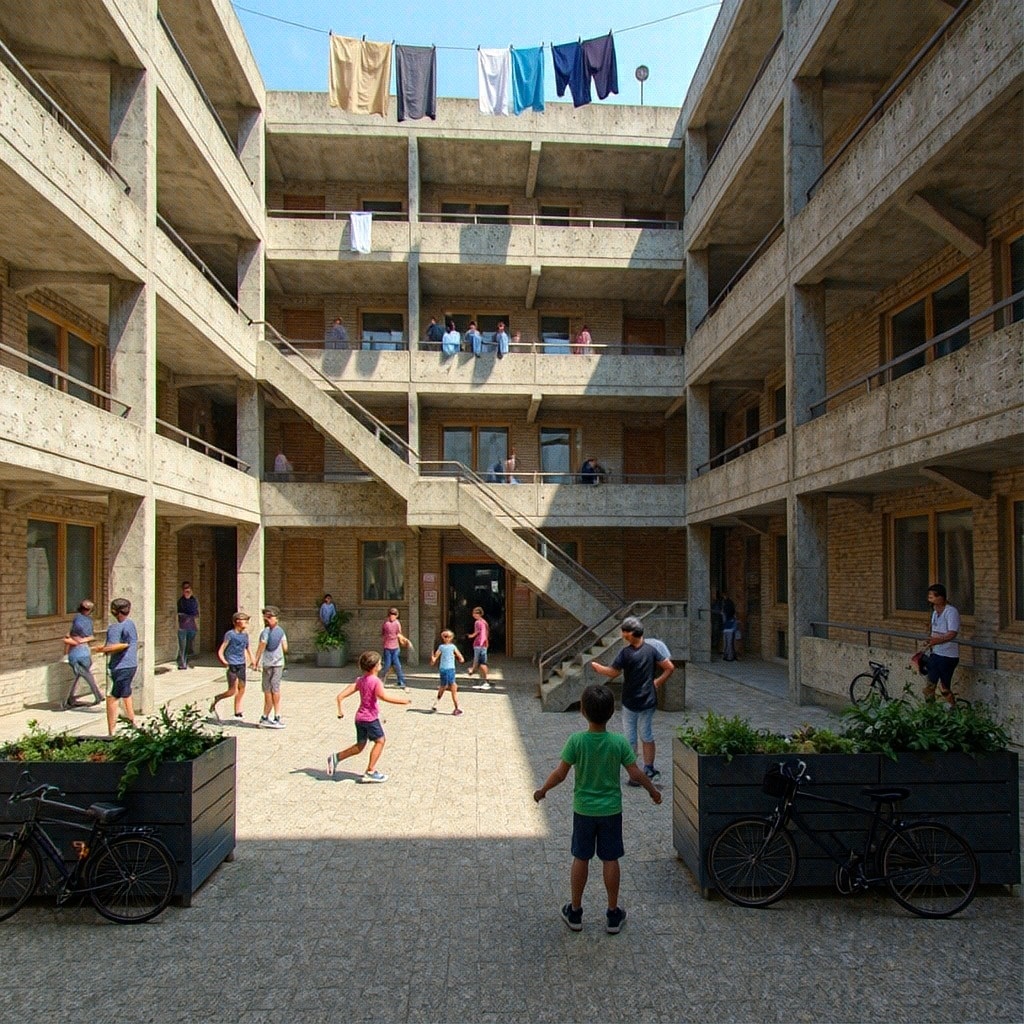}
{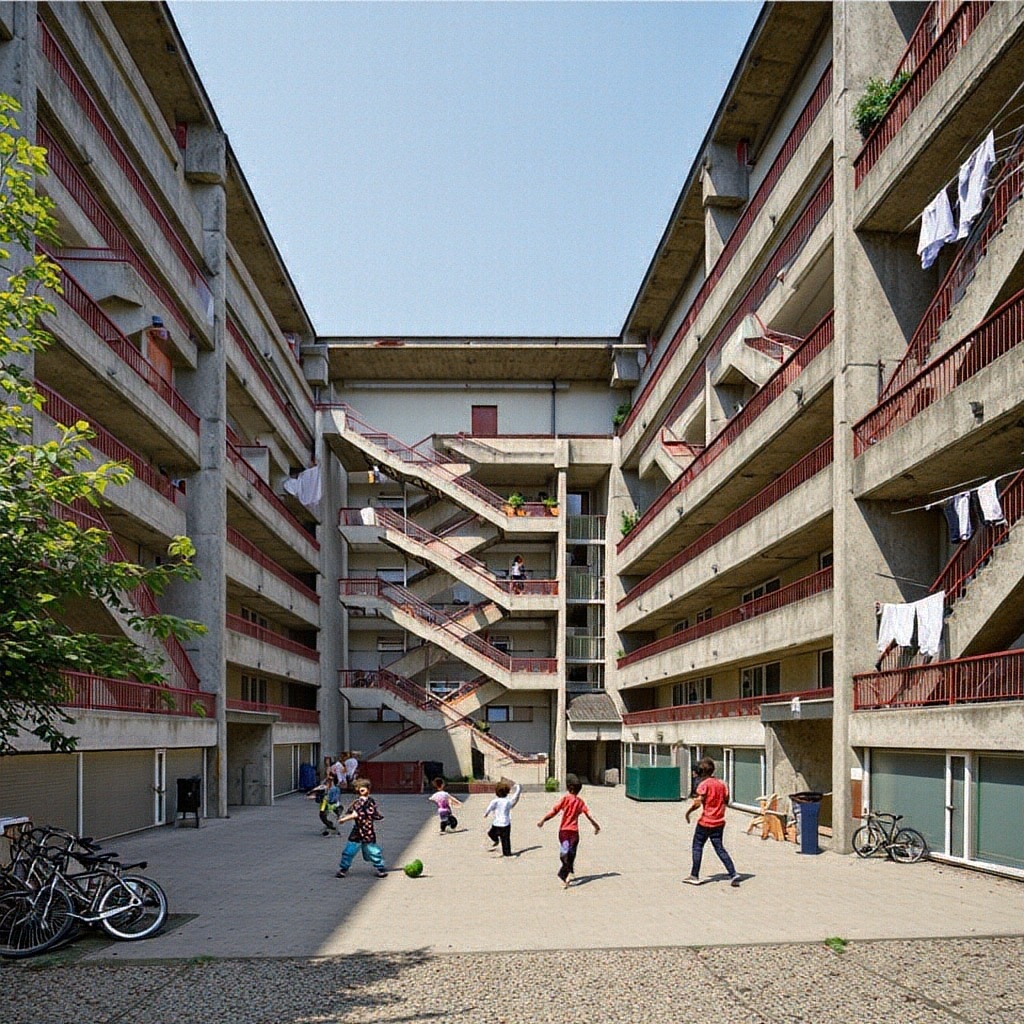}
{A large brutalist housing complex enclosing a lively courtyard, with concrete balconies, exterior staircases, children playing basketball, residents tending planters, bicycles beside an entrance, laundry hanging overhead, and bold afternoon shadows emphasizing the architecture.}
\asymfluxsuppsample
{0.27\textwidth}
{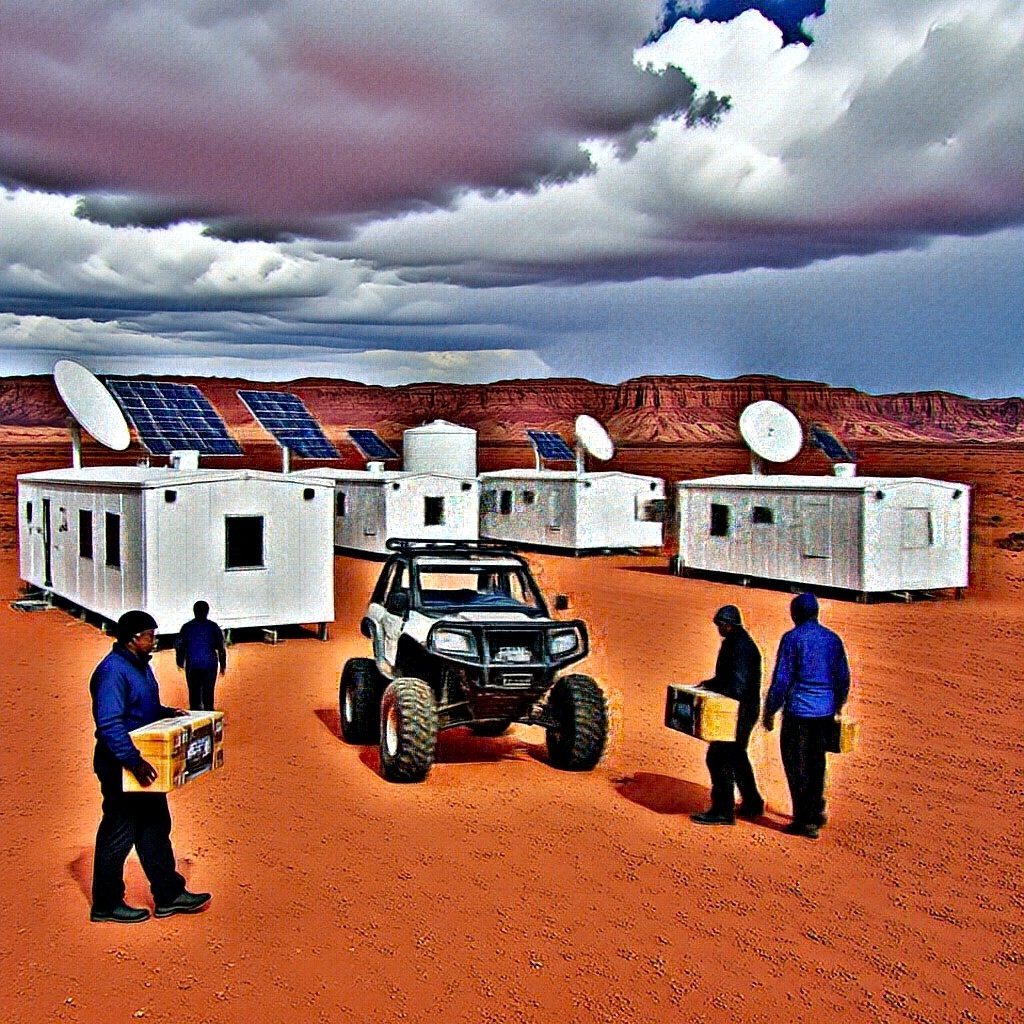}
{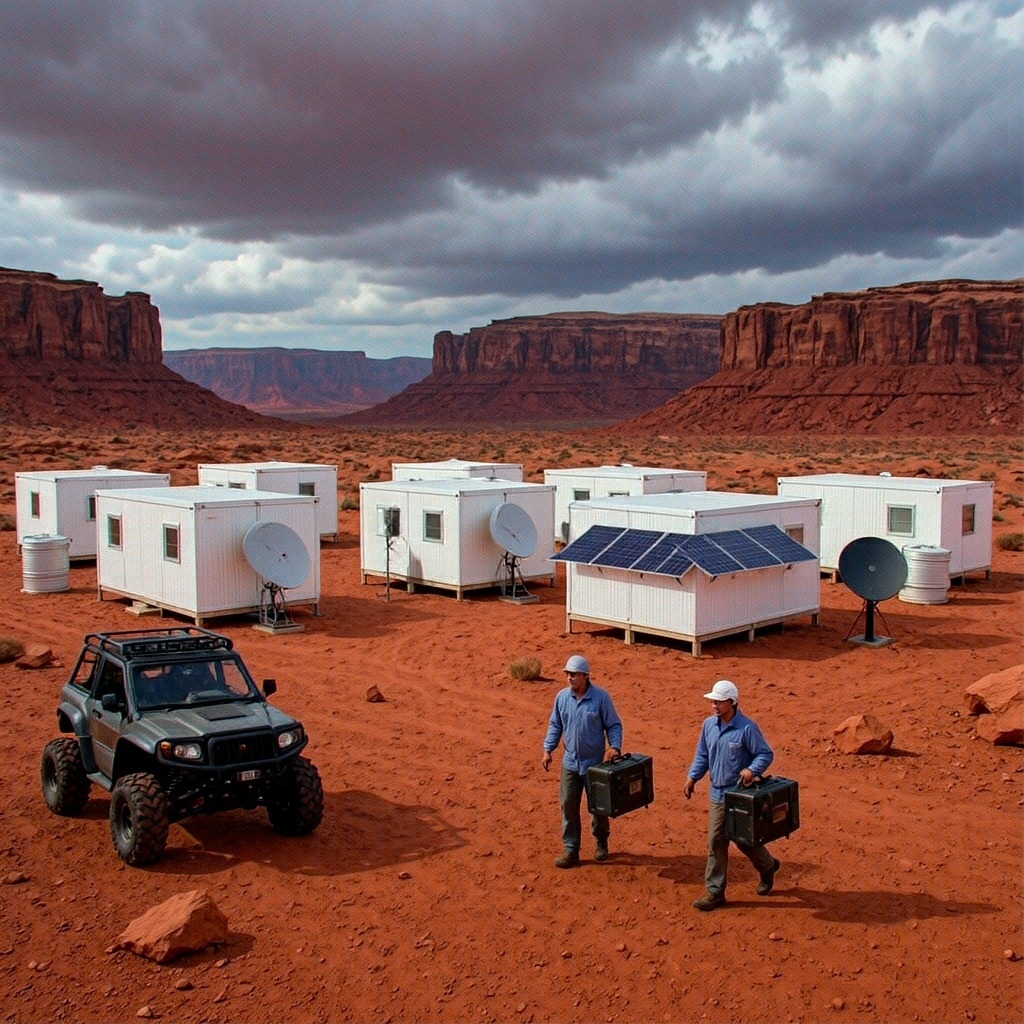}
{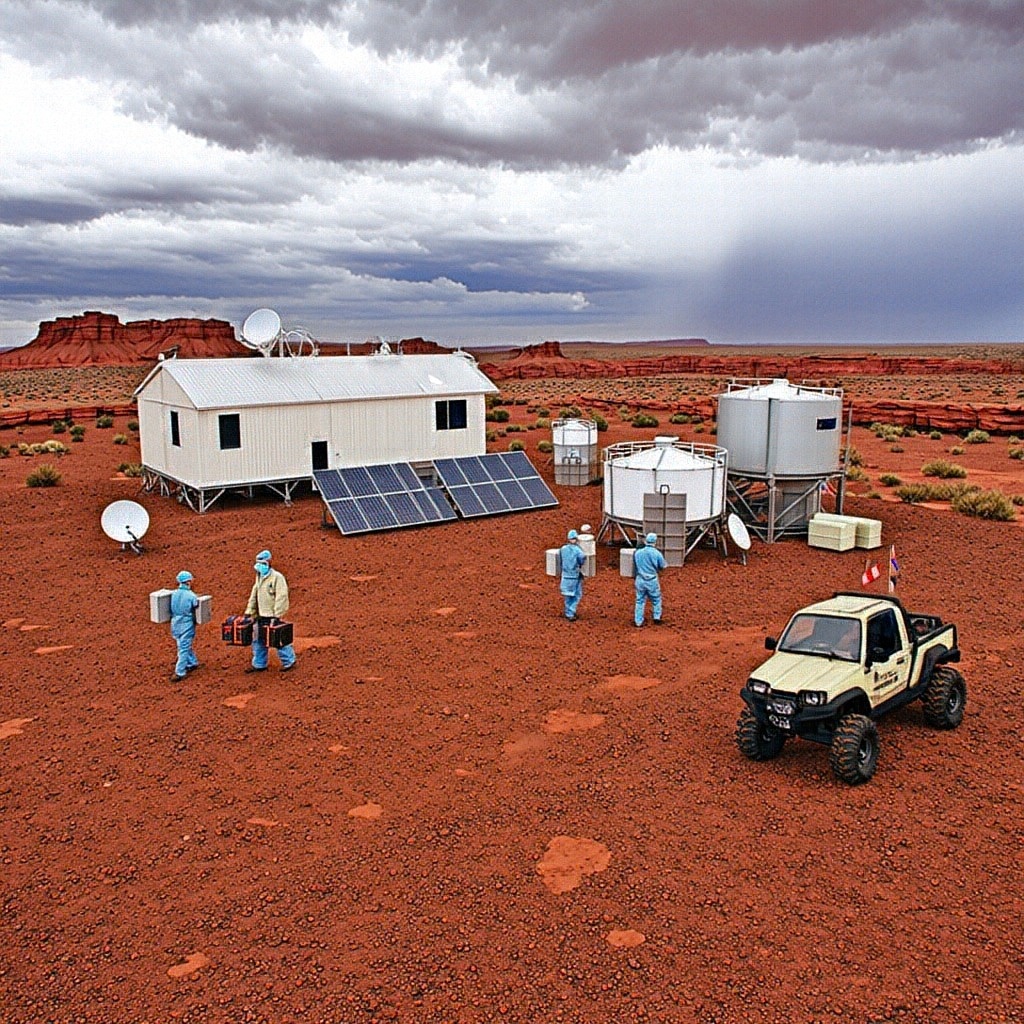}
{A remote desert research outpost composed of white modular buildings, solar panels, satellite dishes, water tanks, an all-terrain vehicle, scientists carrying equipment cases, and red rock formations stretching beneath a dramatic stormy sky.}
\asymfluxsuppsample
{0.27\textwidth}
{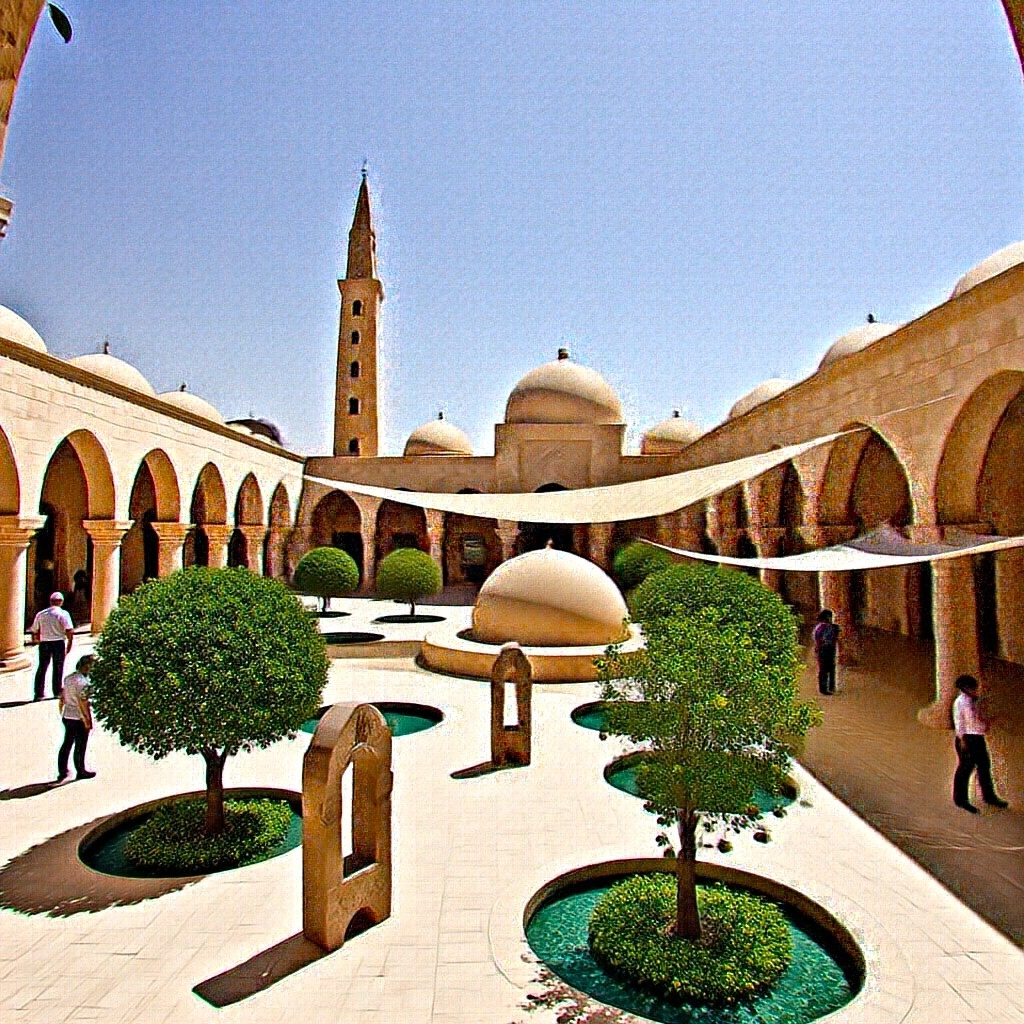}
{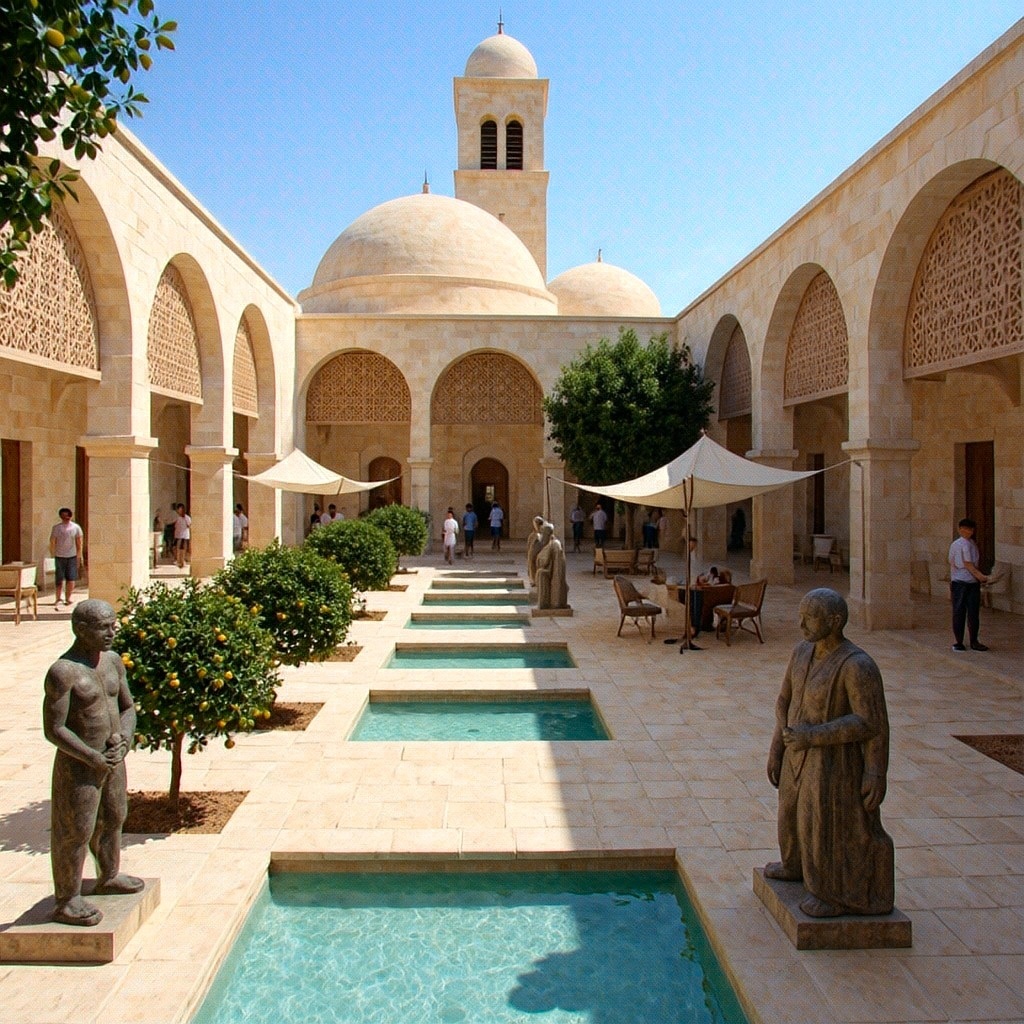}
{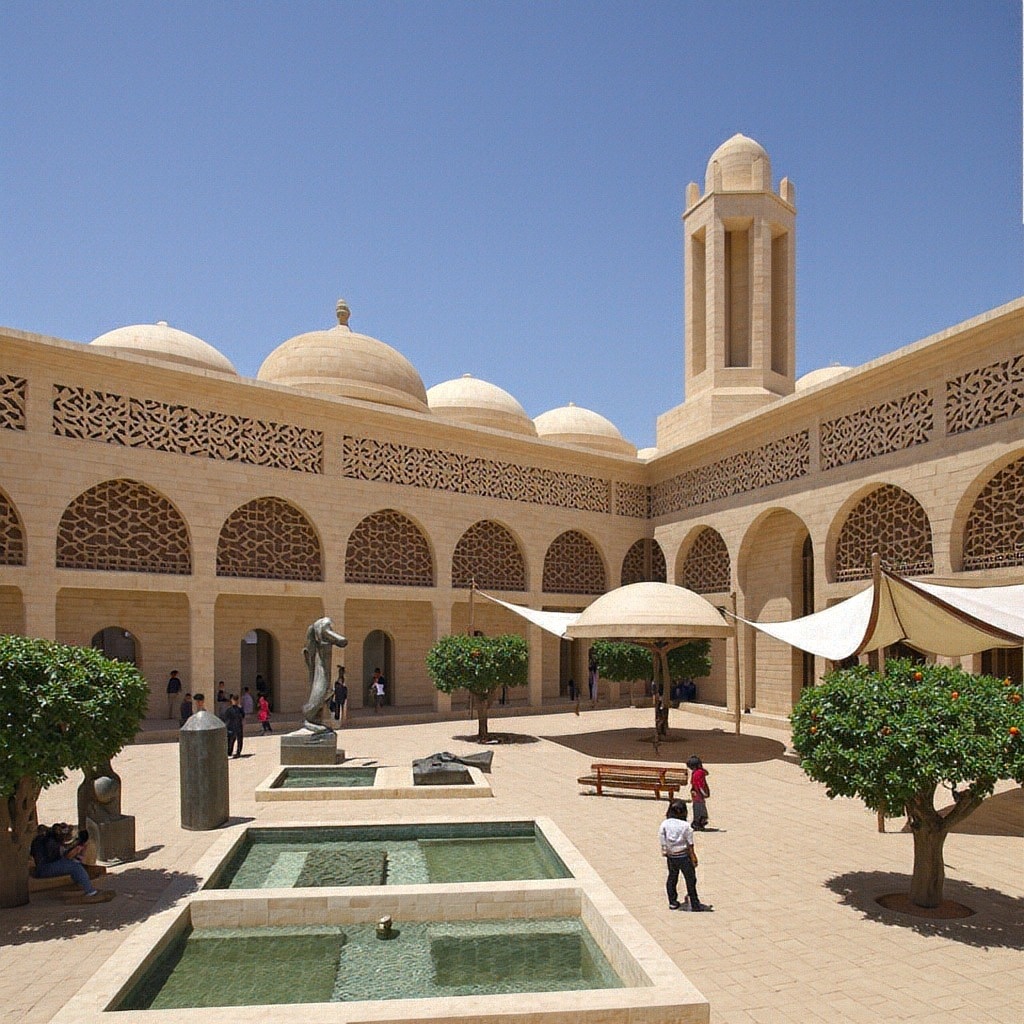}
{A cultural center built around a shaded courtyard, combining pale stone arcades, patterned screens, shallow pools, domed roofs, and a slender tower, while visitors gather near sculptures, citrus trees, and fabric canopies in bright desert light.}
\asymfluxsuppsample
{0.27\textwidth}
{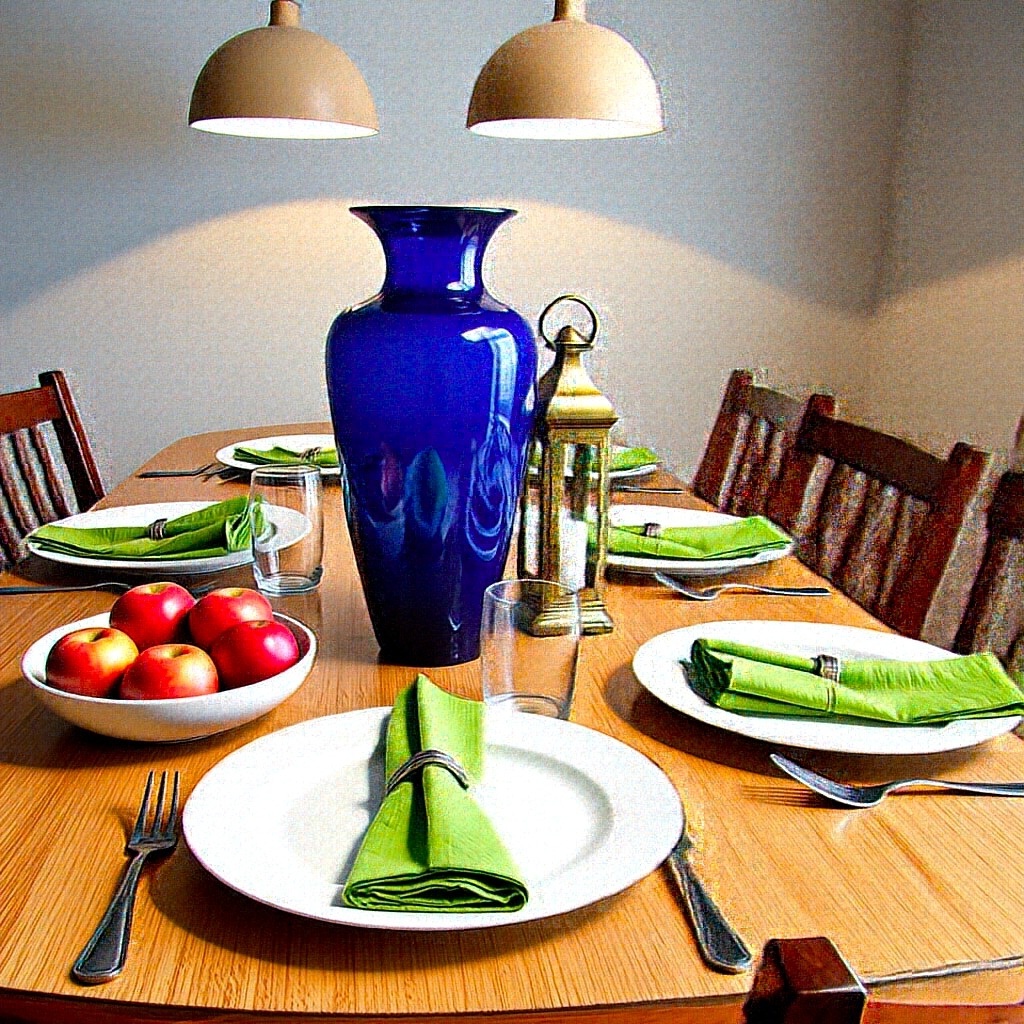}
{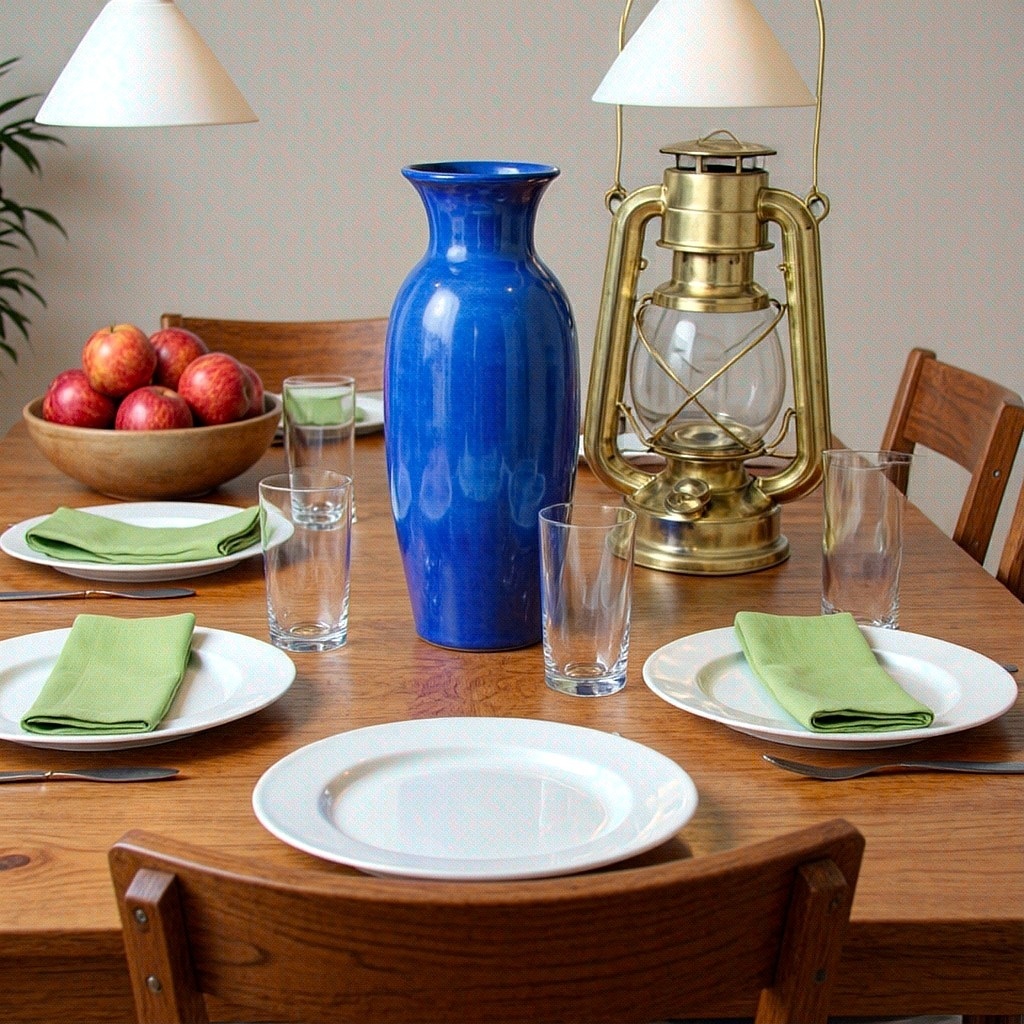}
{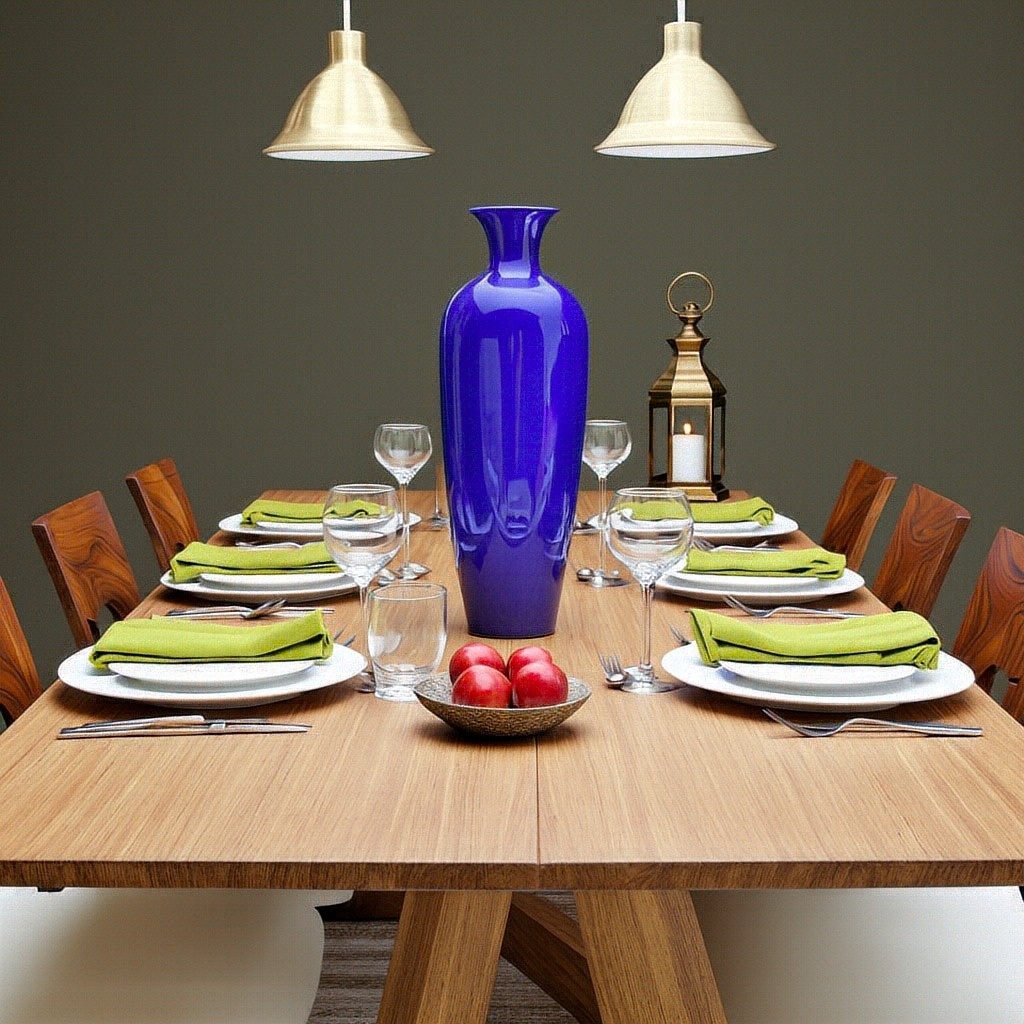}
{A long dining table set for six people, with a tall blue vase exactly in the center, three red apples in a bowl to its left, a brass lantern to its right, white plates, folded green napkins, clear glasses, and wooden chairs aligned beneath two pendant lamps.}
\asymfluxsuppsample
{0.27\textwidth}
{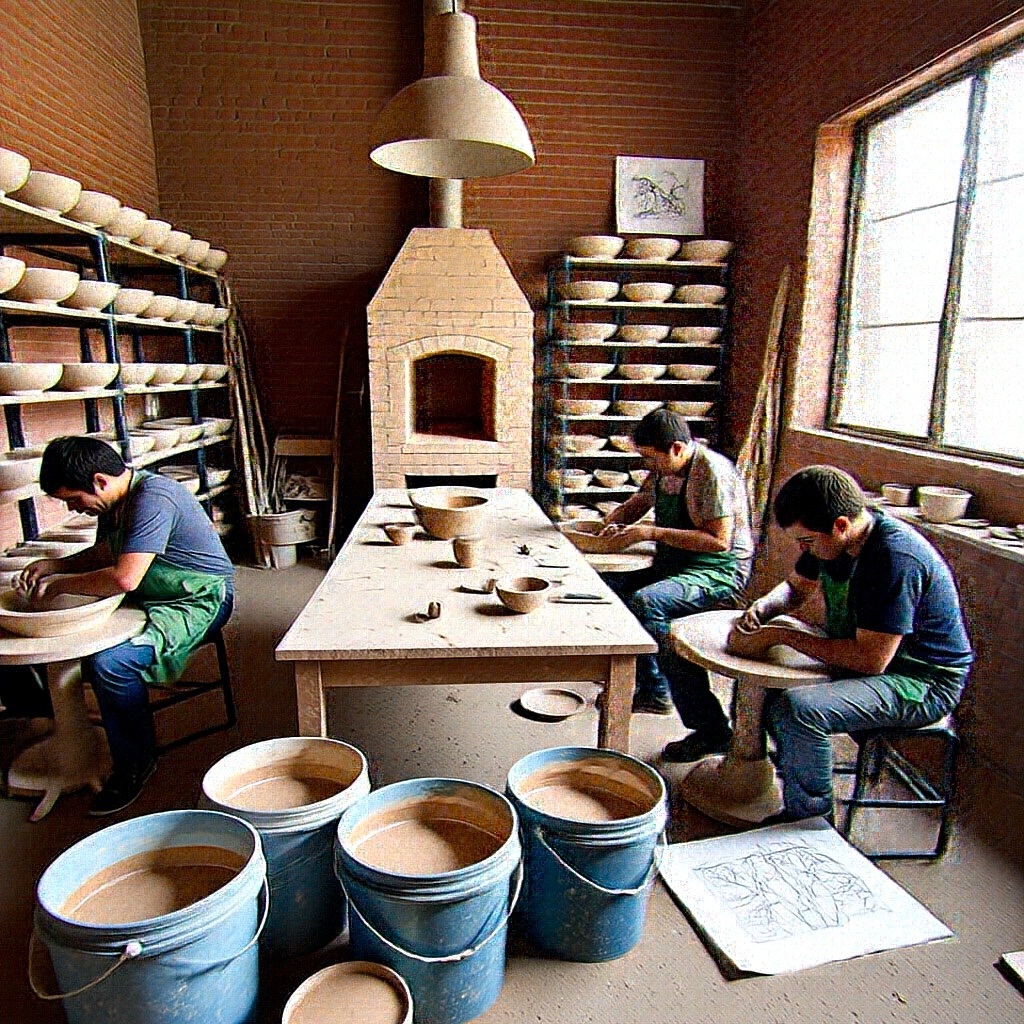}
{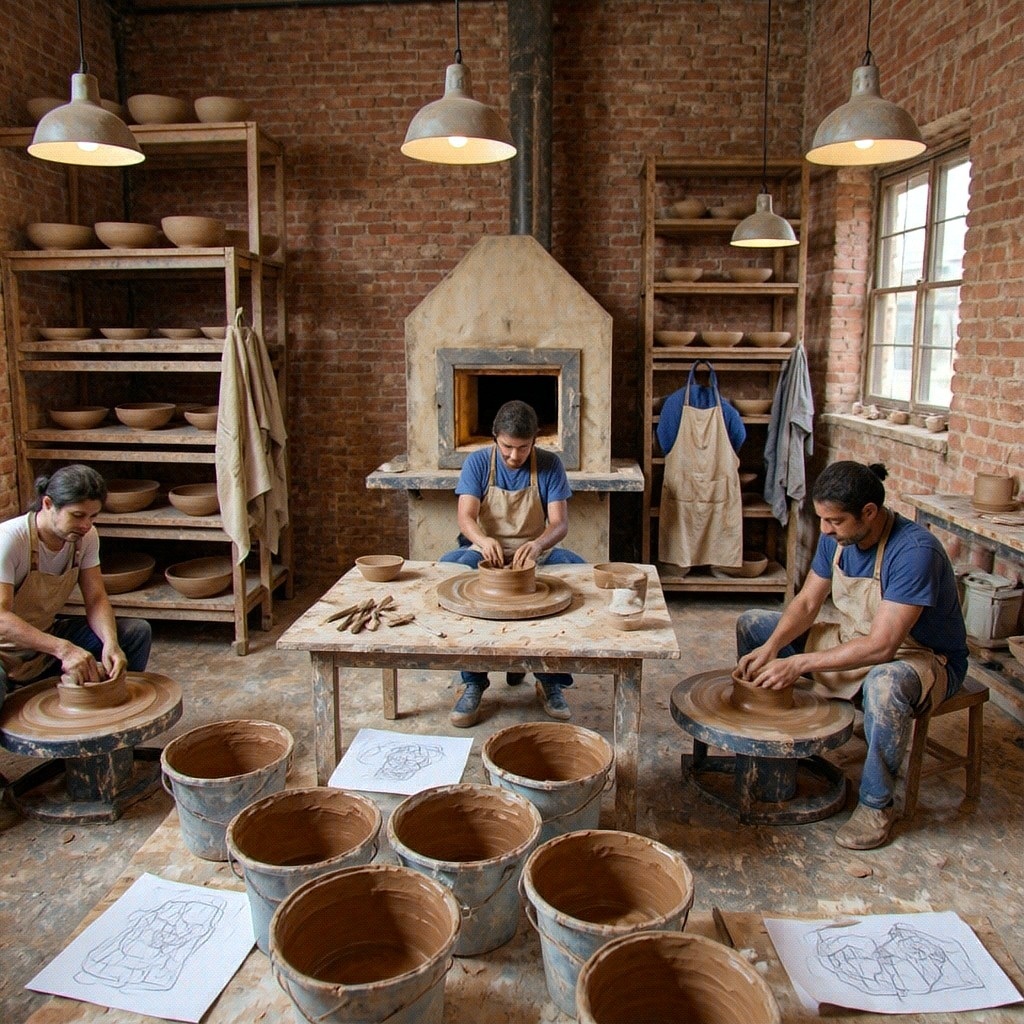}
{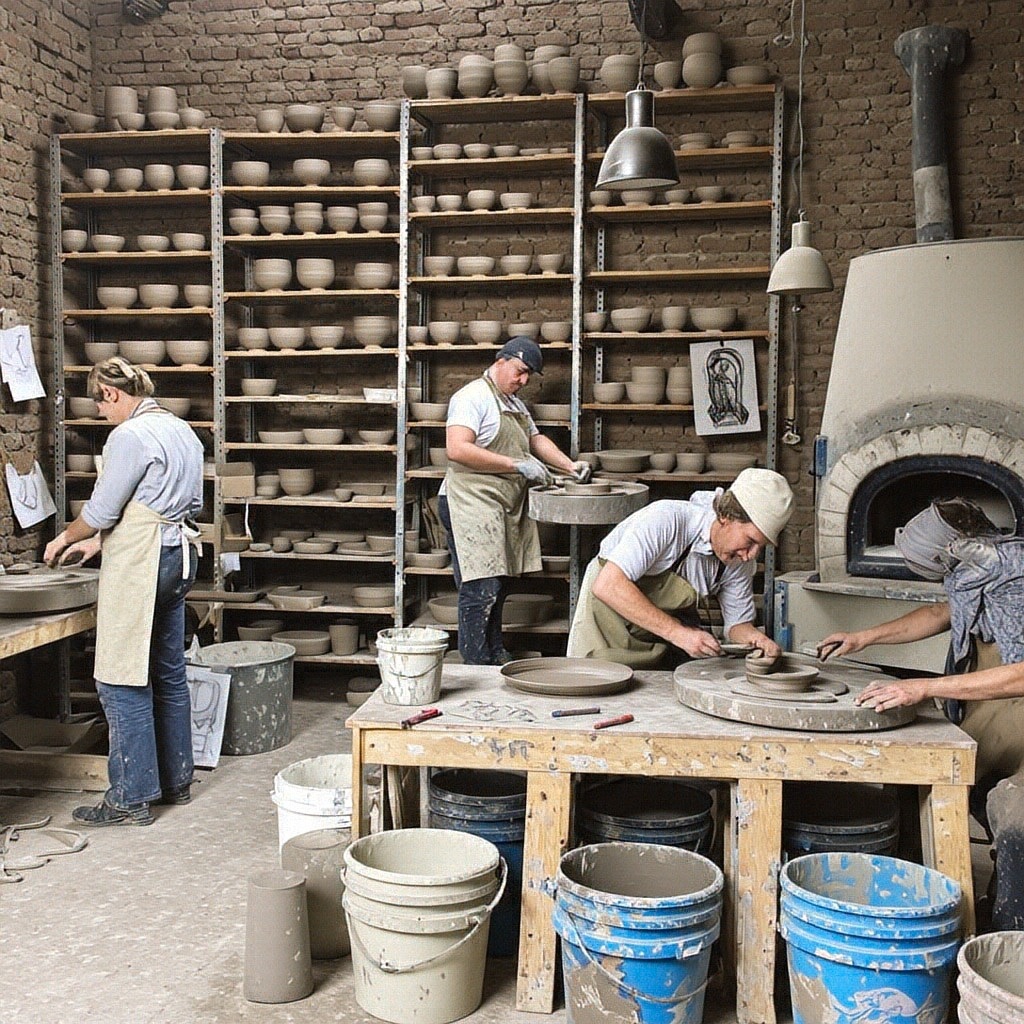}
{A pottery workshop where three artisans shape clay at wheels, finished bowls dry on tall shelves, a large kiln stands open, buckets of glaze surround a central table, and aprons, tools, sketches, and hanging lamps fill the dusty brick room.}
\asymfluxsuppsample
{0.27\textwidth}
{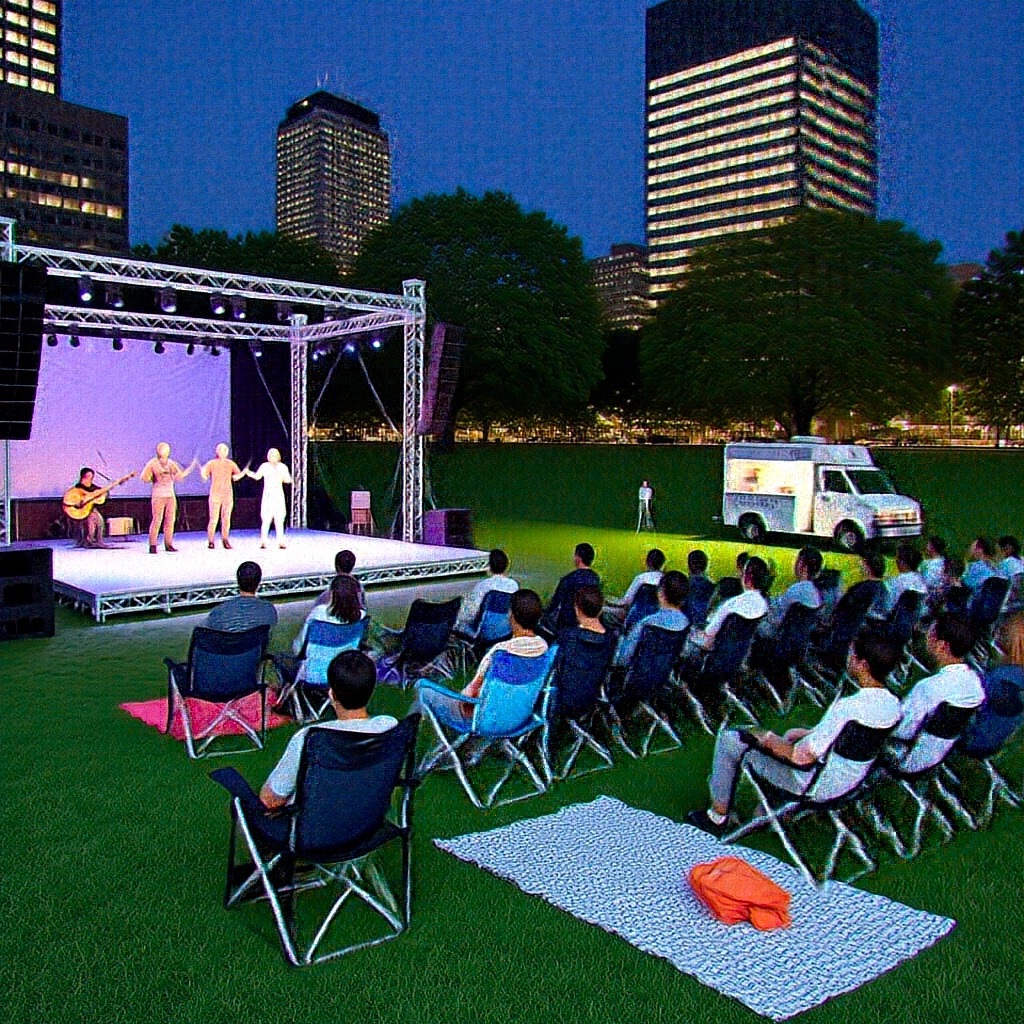}
{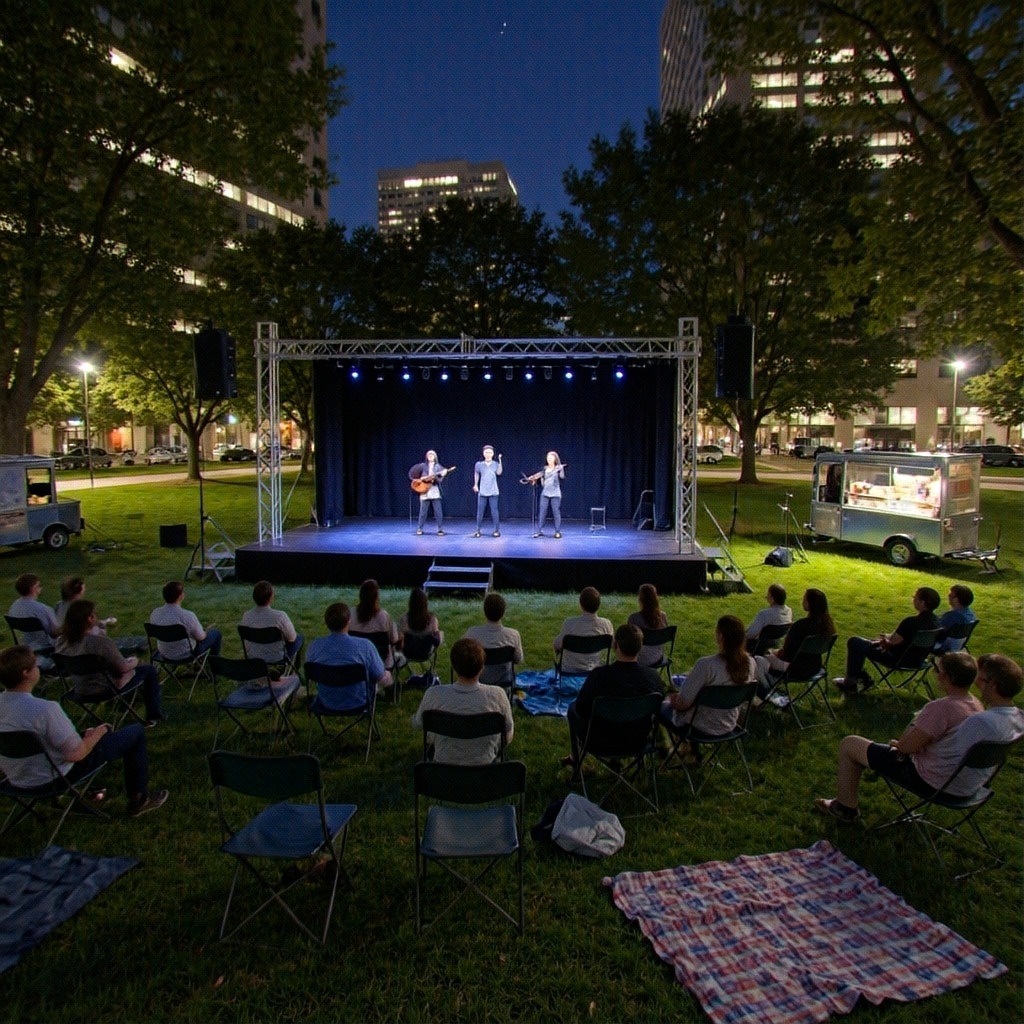}
{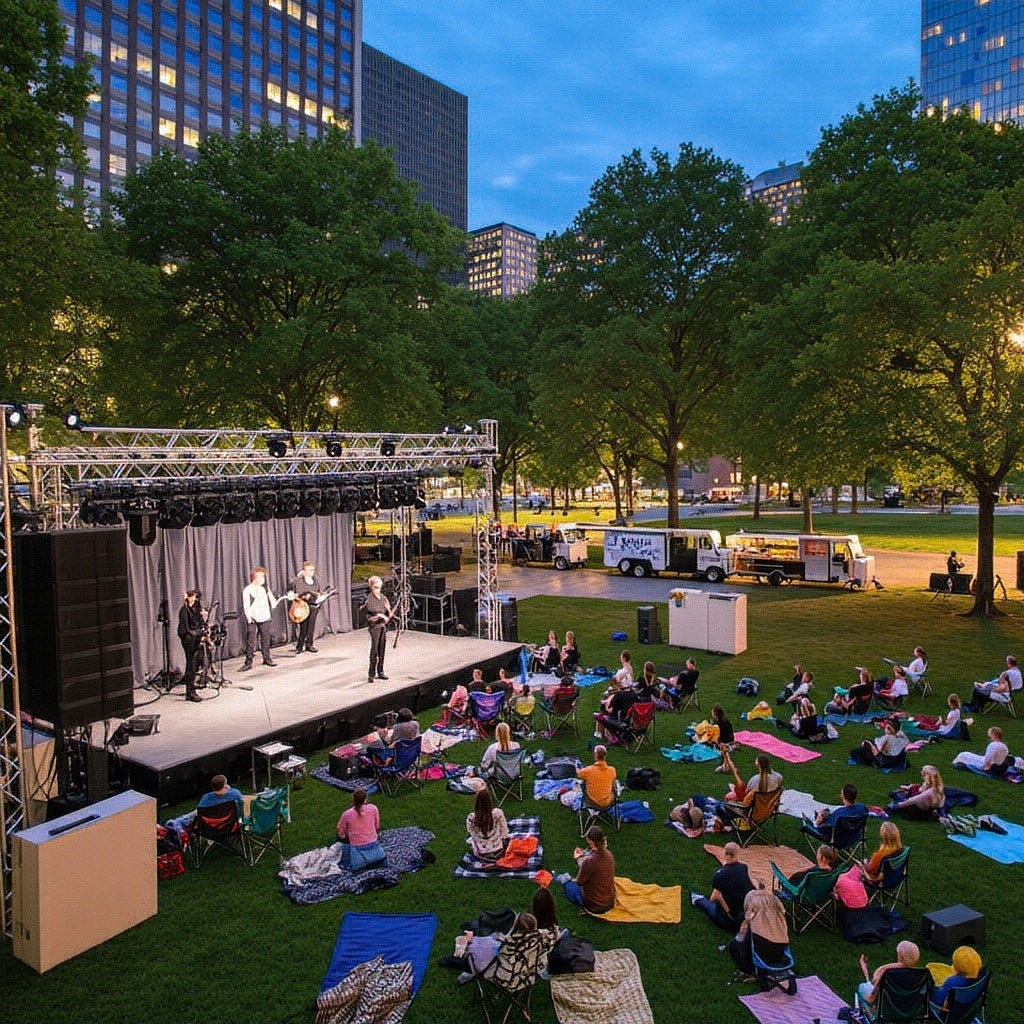}
{An outdoor theater performance in a city park, with actors on a temporary stage, musicians seated to one side, an audience on folding chairs and blankets, lighting rigs, food carts, tall trees, and office buildings glowing beyond the lawn at night.}

\endgroup
\linenumbers

\subsection{Additional Samples from the Pixel-Space Student Trained with Latent Matching}
\label{sec:supp-l2p-gallery}

Additional samples that follow the same settings as the main paper's teaser shows eight samples selected from a 38-image review gallery.

\nolinenumbers
\begingroup
\setlength{\parindent}{0pt}
\newcommand{\ltwpsample}[2]{%
  \begin{minipage}[t]{0.22\textwidth}%
    \centering
    \includegraphics[width=\linewidth]{#1}\par%
    \vspace{0.25em}%
    {\tiny\raggedright #2\par}%
  \end{minipage}%
}

\ltwpsample
{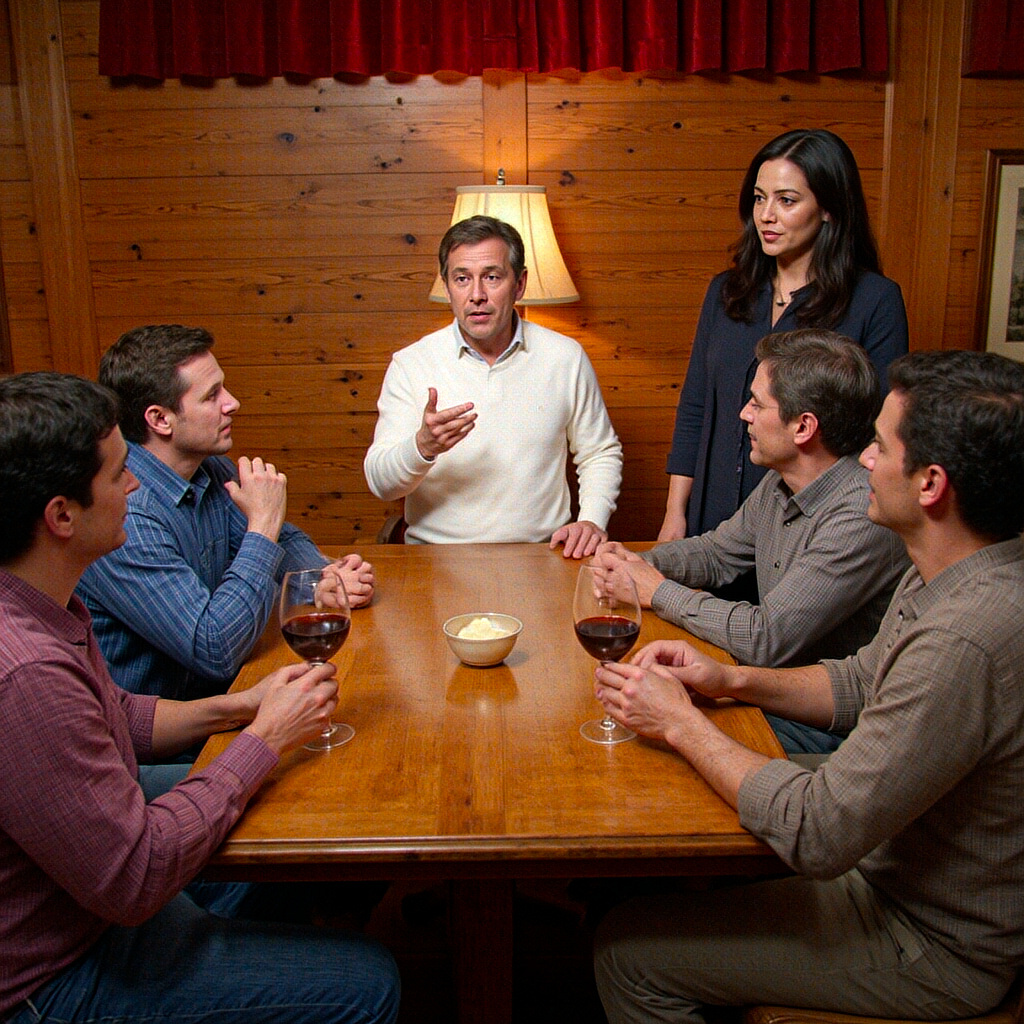}
{A group of people sit around a wooden table in a wood-paneled room, a man in a white sweater gestures while speaking, others hold wine glasses, a woman stands at the table, a red curtain hangs above, a lamp illuminates the scene.}
\hfill
\ltwpsample
{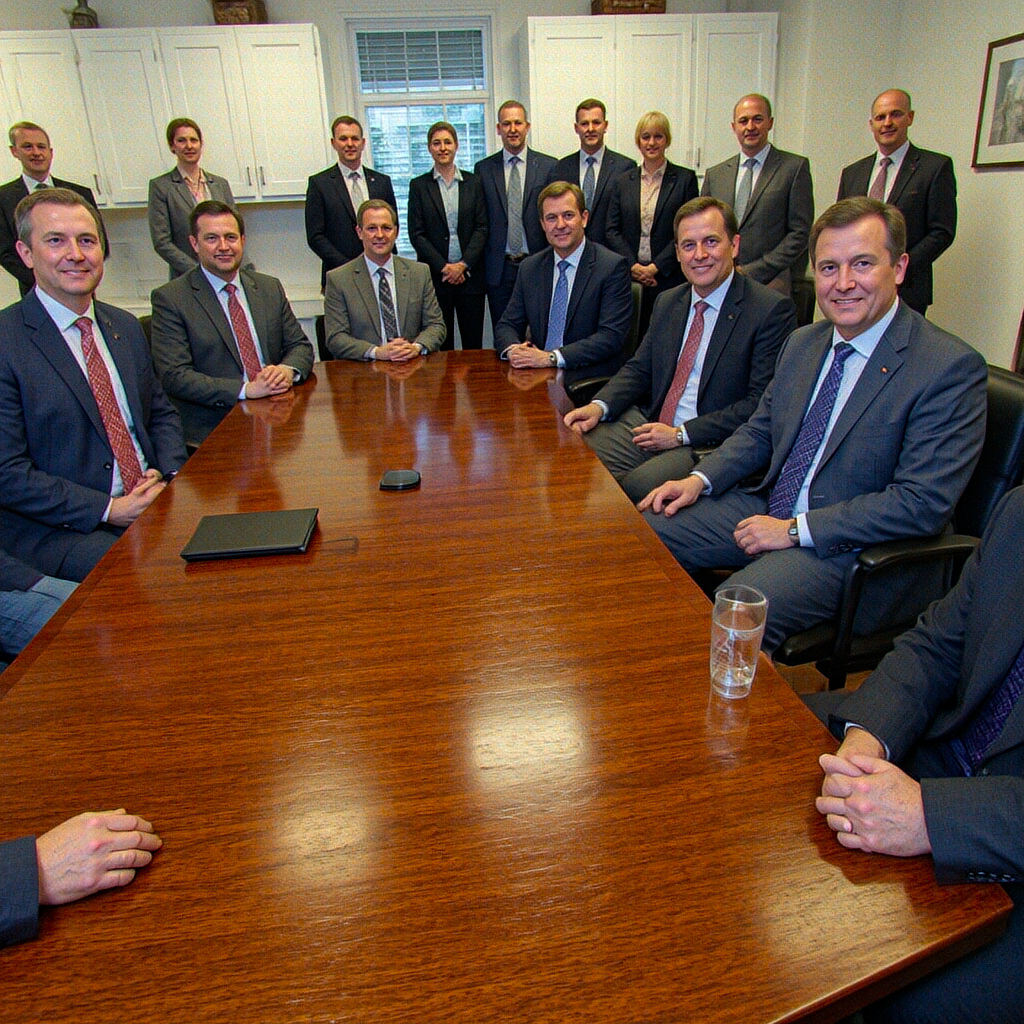}
{A group of people in dress clothes sits around a large wooden conference table, with others standing behind them, in a room with white cabinets and a window in the background, under even indoor lighting.}
\hfill
\ltwpsample
{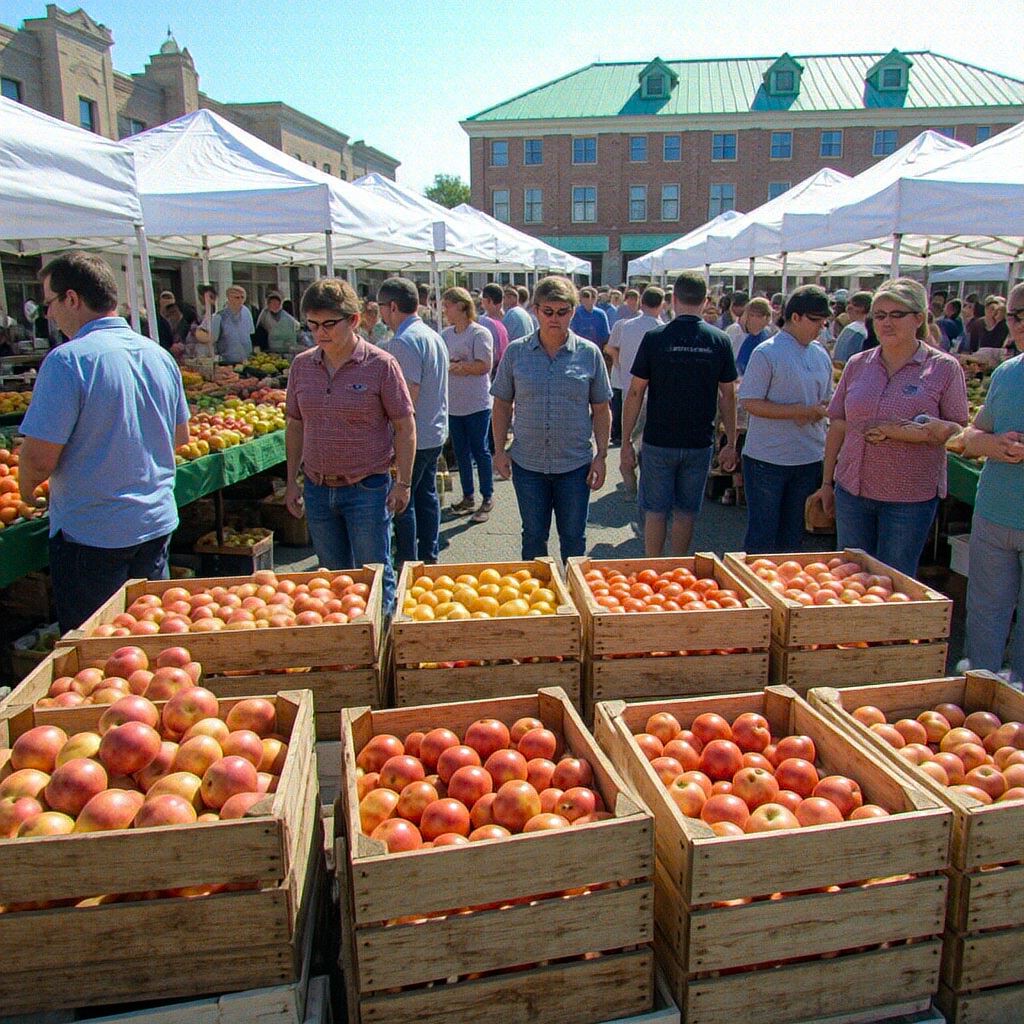}
{A crowd of people shops at a farmers market with wooden crates of apples, tomatoes, and other produce in the foreground, under white canopies, with buildings and a green-roofed structure in the background under bright sunlight.}
\hfill
\ltwpsample
{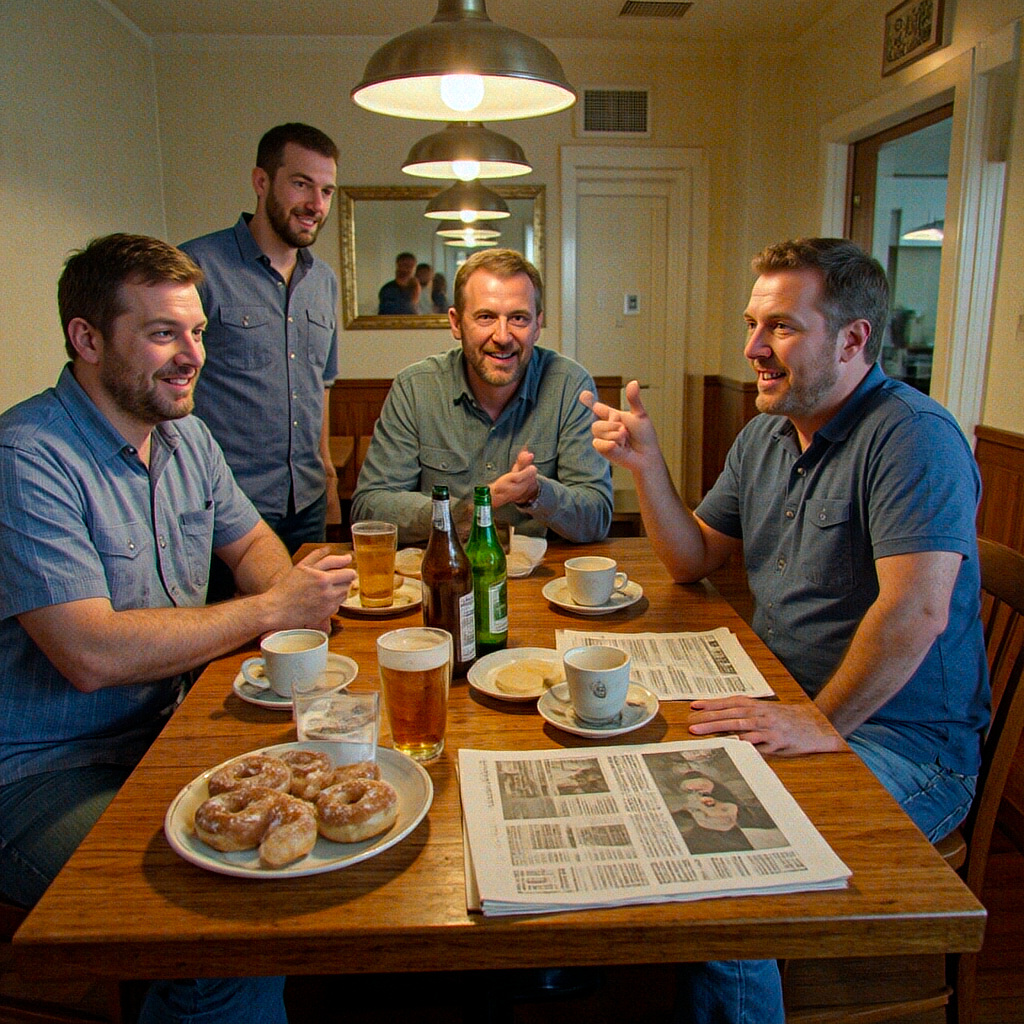}
{Three men sit at a wooden table in a restaurant, one standing, one leaning, one gesturing, with donuts, beer, coffee, newspapers, and bottles on the table, under hanging lights, with a mirror and doorway in the background.}
\par\vspace{0.35em}

\ltwpsample
{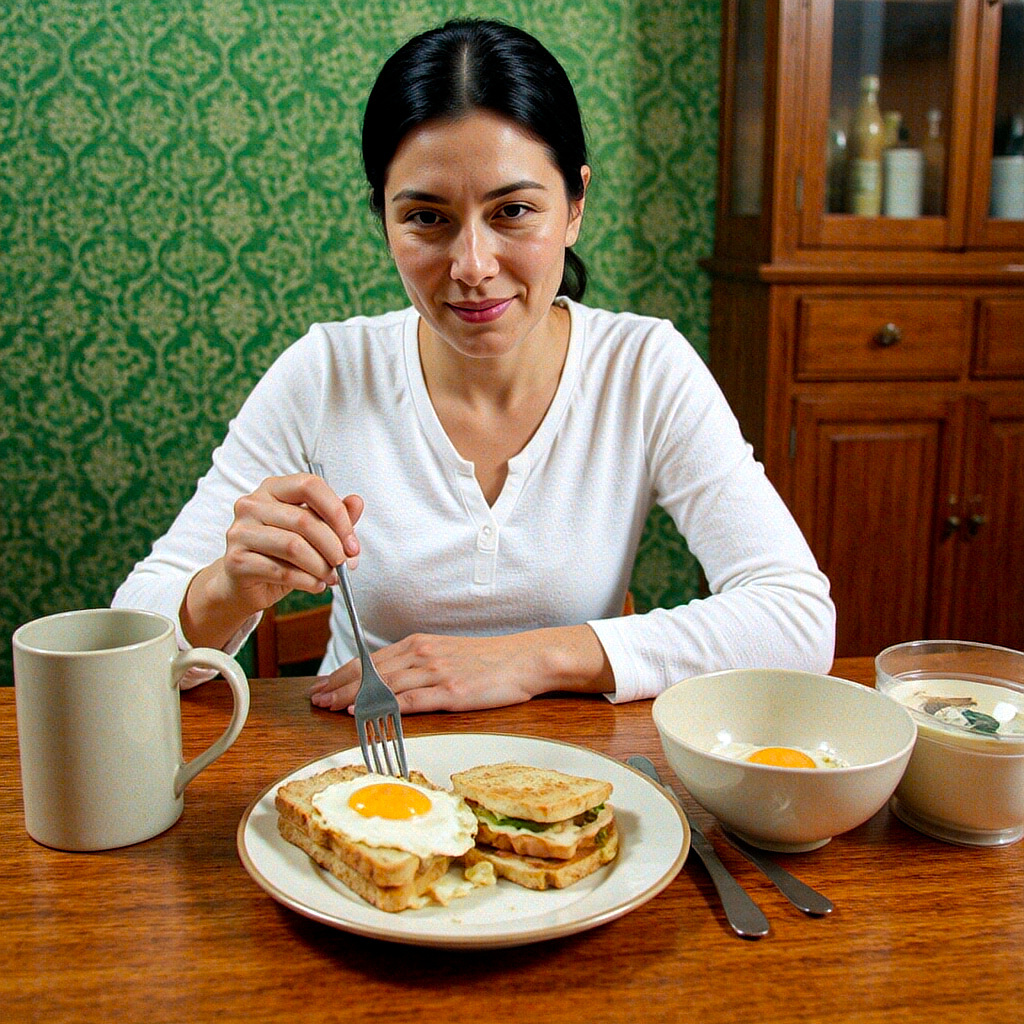}
{A woman with dark hair tied back sits at a wooden table, wearing a white long-sleeved shirt, holding a fork over a plate with a sandwich and eggs, with a mug, bowl, and container nearby, against a background of green patterned wallpaper and a wooden cabinet.}
\hfill
\ltwpsample
{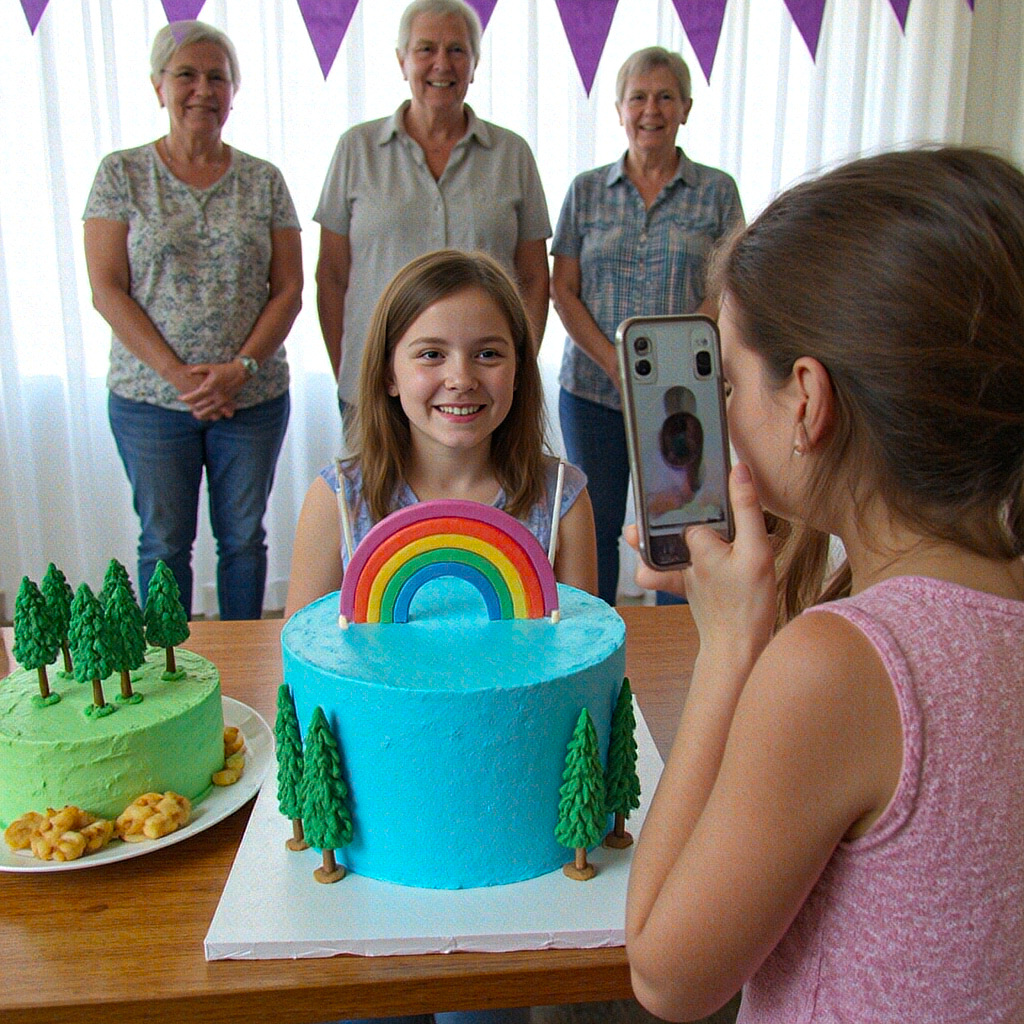}
{A girl holds a phone to photograph a blue cake with a rainbow topper, a green cake with trees, and a plate of food on a table, with three people standing behind her under a white curtain and purple bunting.}
\hfill
\ltwpsample
{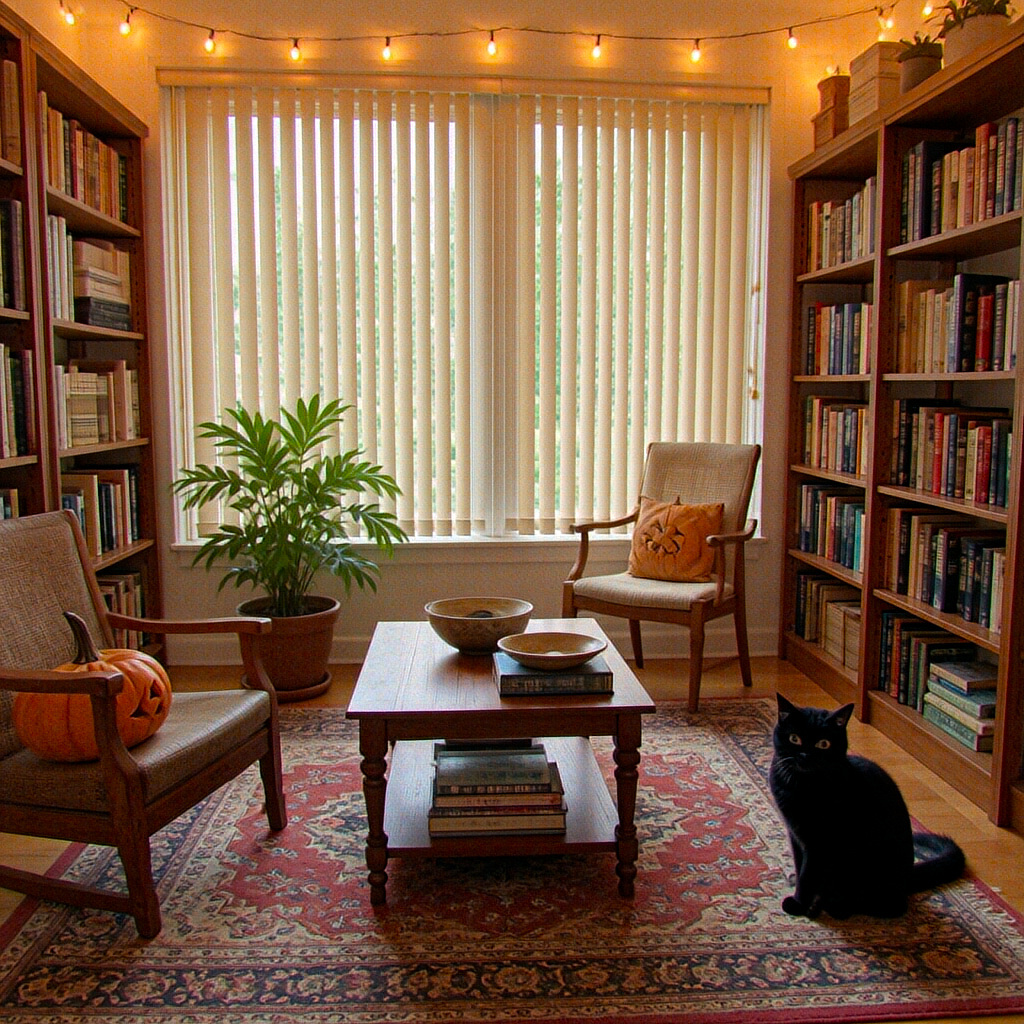}
{A living room features a large window with closed vertical blinds, a bookshelf filled with books, a potted plant, a wooden chair with a carved pumpkin, a coffee table with books and a bowl, a patterned rug, and a black cat, all under warm string lights.}
\hfill
\ltwpsample
{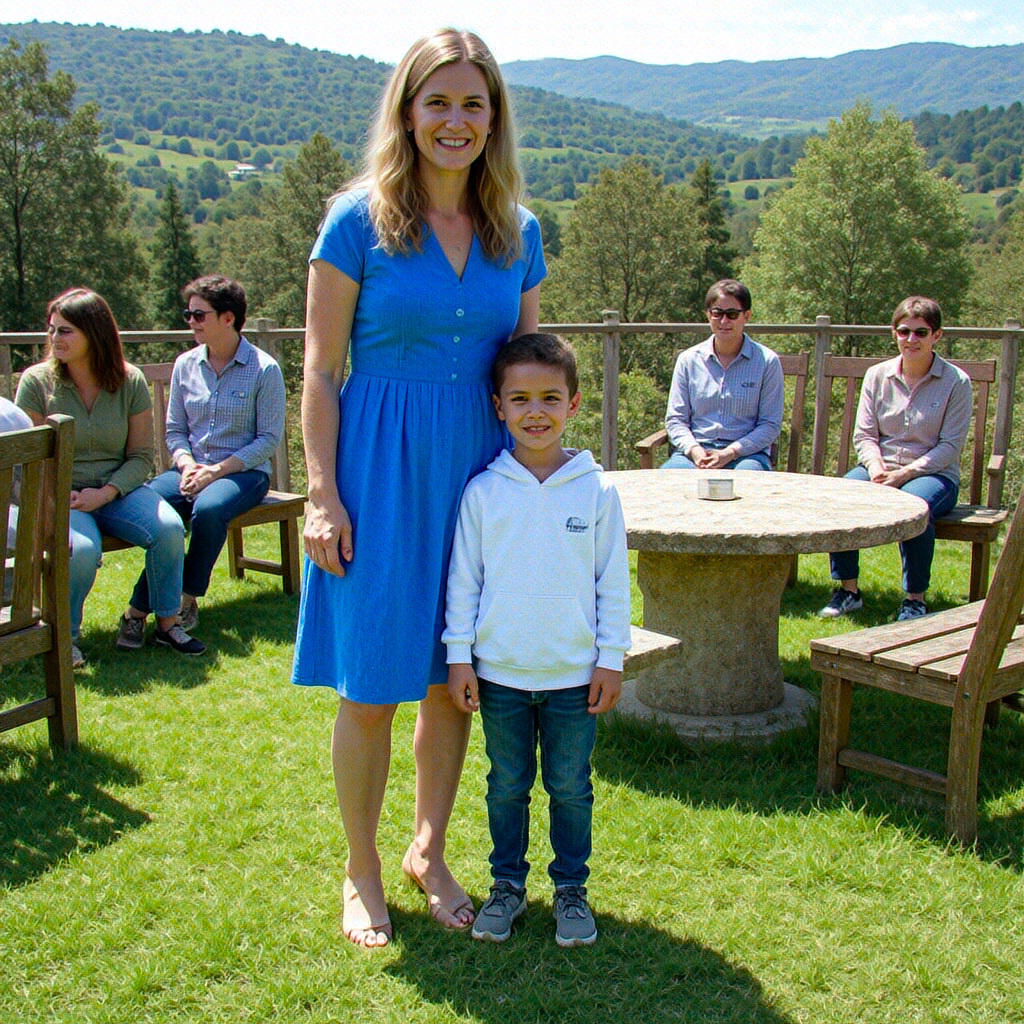}
{A woman in a blue dress stands beside a young child in a white hoodie, while others sit on wooden benches and a stone table in a grassy outdoor setting, with hills and trees in the background under bright daylight.}
\par\vspace{0.35em}

\ltwpsample
{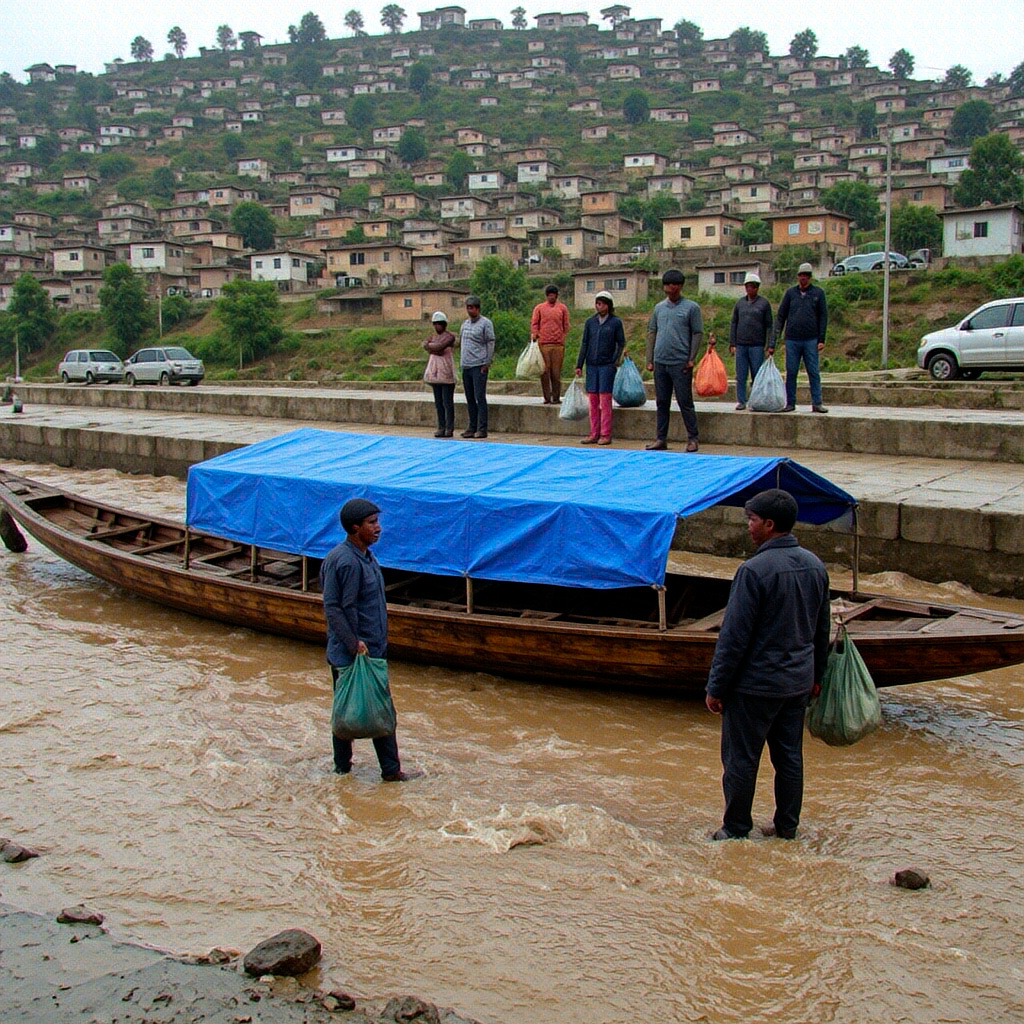}
{A long wooden boat with a blue tarp is docked at a concrete pier, where several people stand with bags and bundles, while a muddy river flows in the foreground and a hillside village with houses and vehicles rises in the background under overcast light.}
\hfill
\ltwpsample
{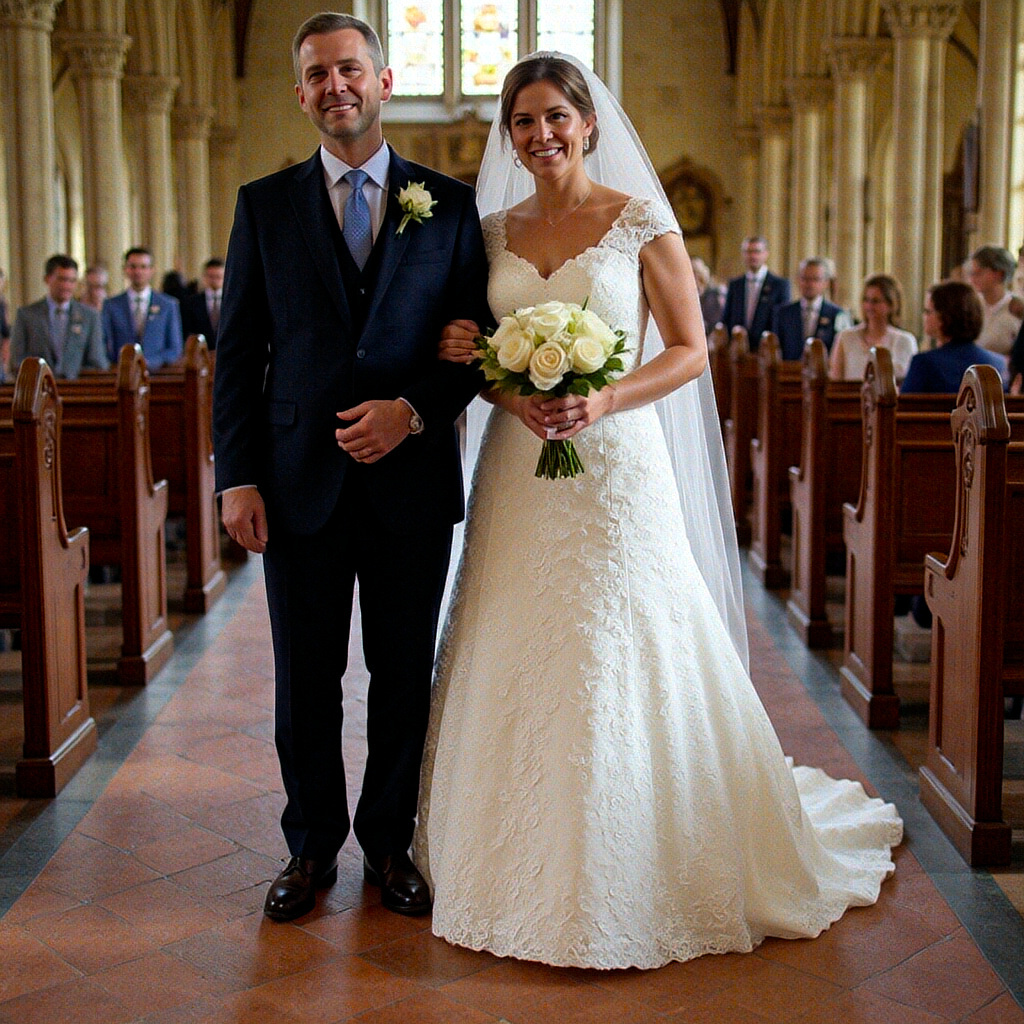}
{A man in a dark suit and tie stands beside a woman in a white wedding dress and veil, holding a bouquet, in a church aisle with pews and people in the background.}
\hfill
\ltwpsample
{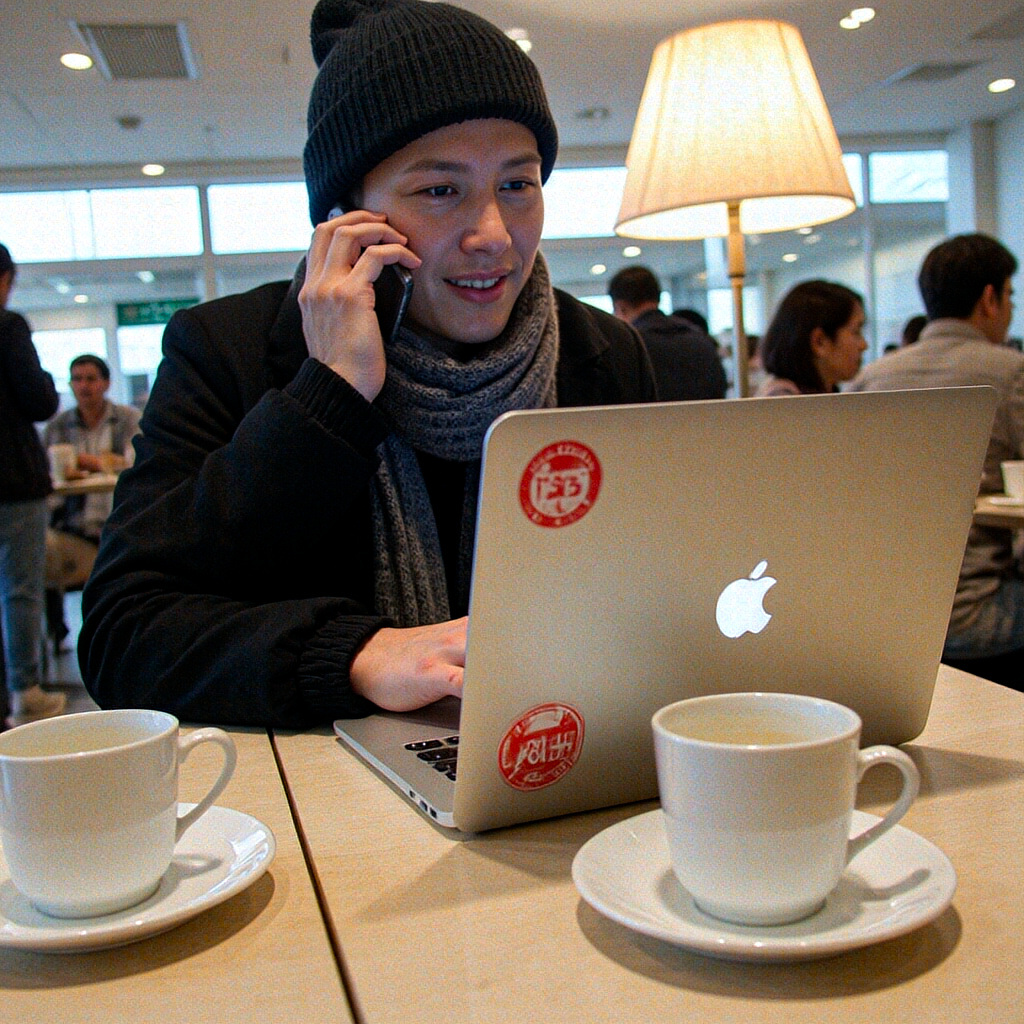}
{A person wearing a beanie, scarf, and black jacket sits at a table, holding a phone to their ear while using a silver laptop with a red and white sticker, a white cup and saucer on the table, in a brightly lit indoor space with other people and a lamp in the background.}
\hfill
\ltwpsample
{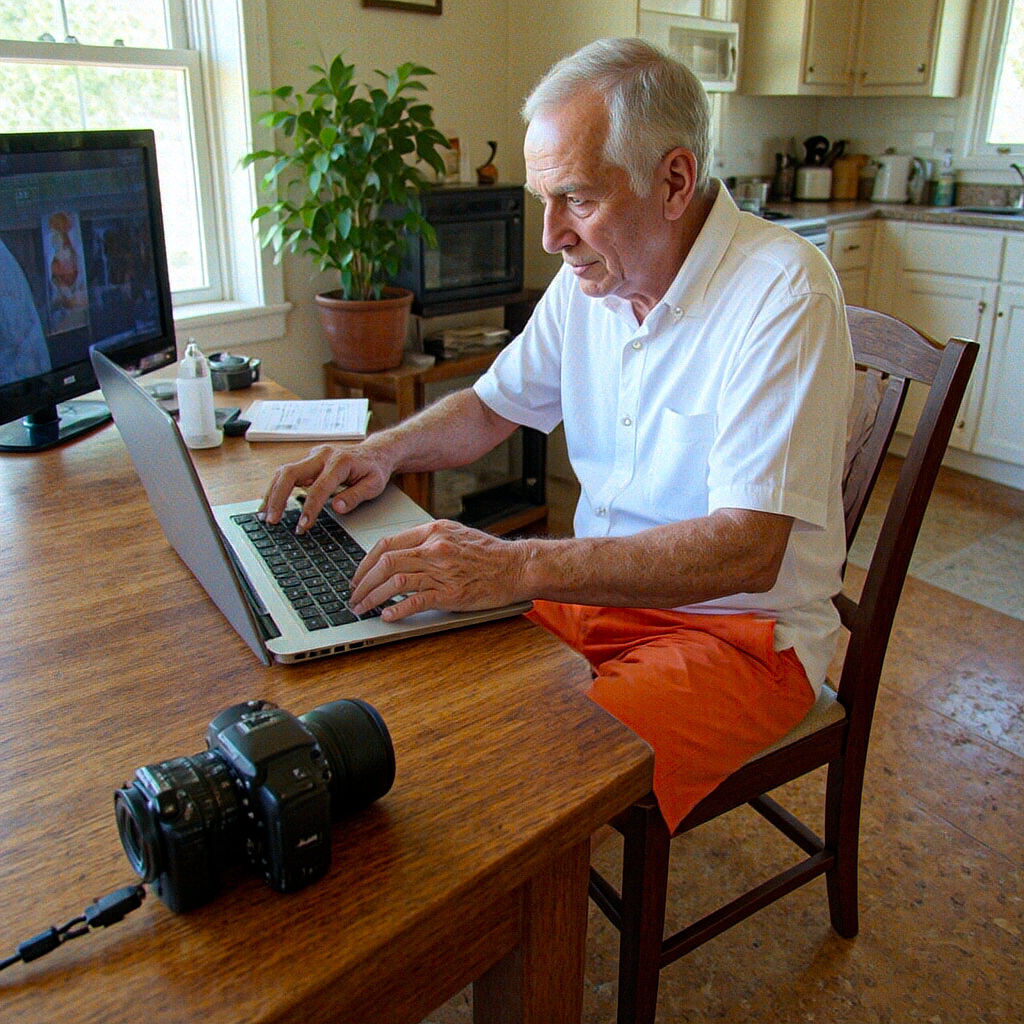}
{An older man in a white shirt and orange shorts sits at a wooden desk, looking at a laptop, with a camera beside it, in a room with a window, a television, and a kitchen in the background, under natural light.}
\par\vspace{0.35em}

\ltwpsample
{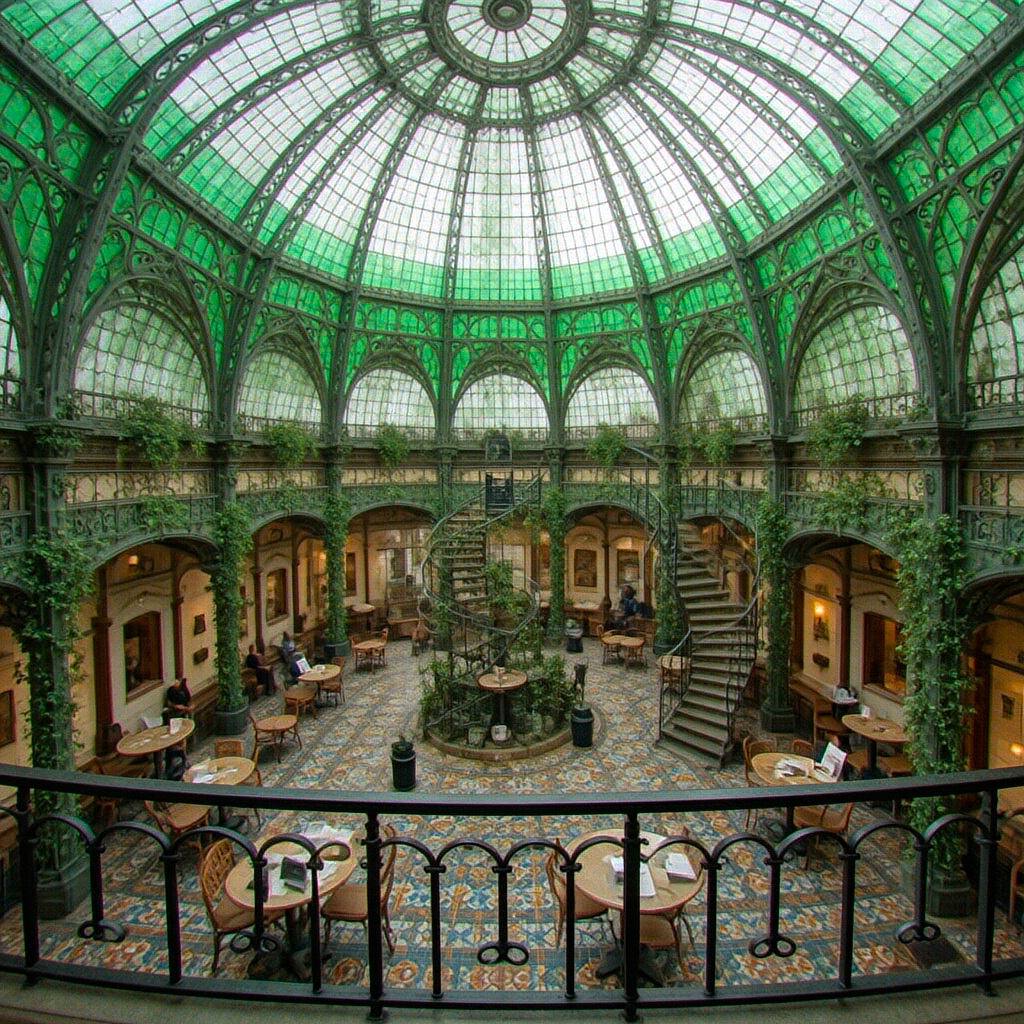}
{An ornate Art Nouveau winter garden with curved iron ribs, green stained glass, patterned tile floors, spiral staircases, hanging vines, café tables, and people reading beneath a tall central dome, photographed from an upper balcony in soft overcast daylight.}
\hfill
\ltwpsample
{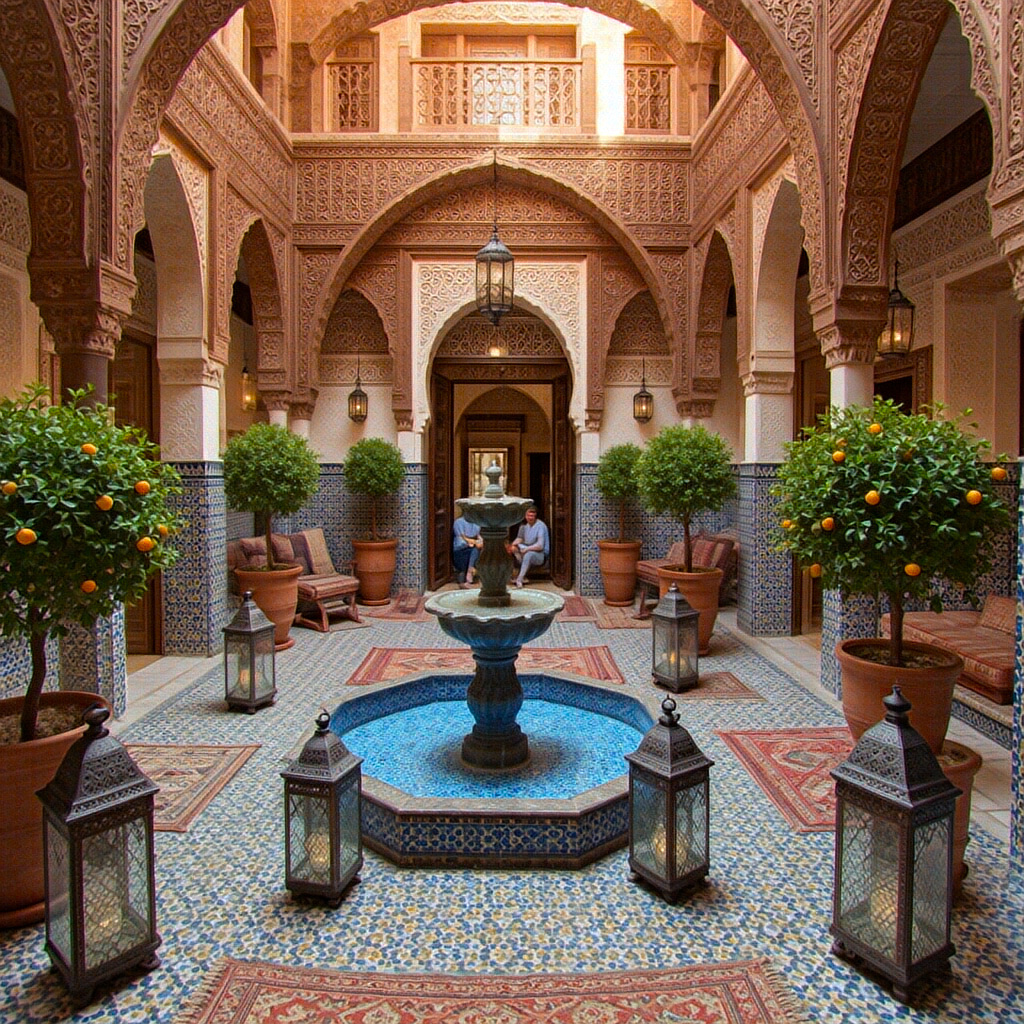}
{A traditional Moroccan riad courtyard centered on a blue mosaic fountain, surrounded by horseshoe arches, carved cedar balconies, patterned zellige walls, potted orange trees, woven rugs, and lanterns; two guests sit near an open doorway while warm afternoon light enters from above.}
\hfill
\ltwpsample
{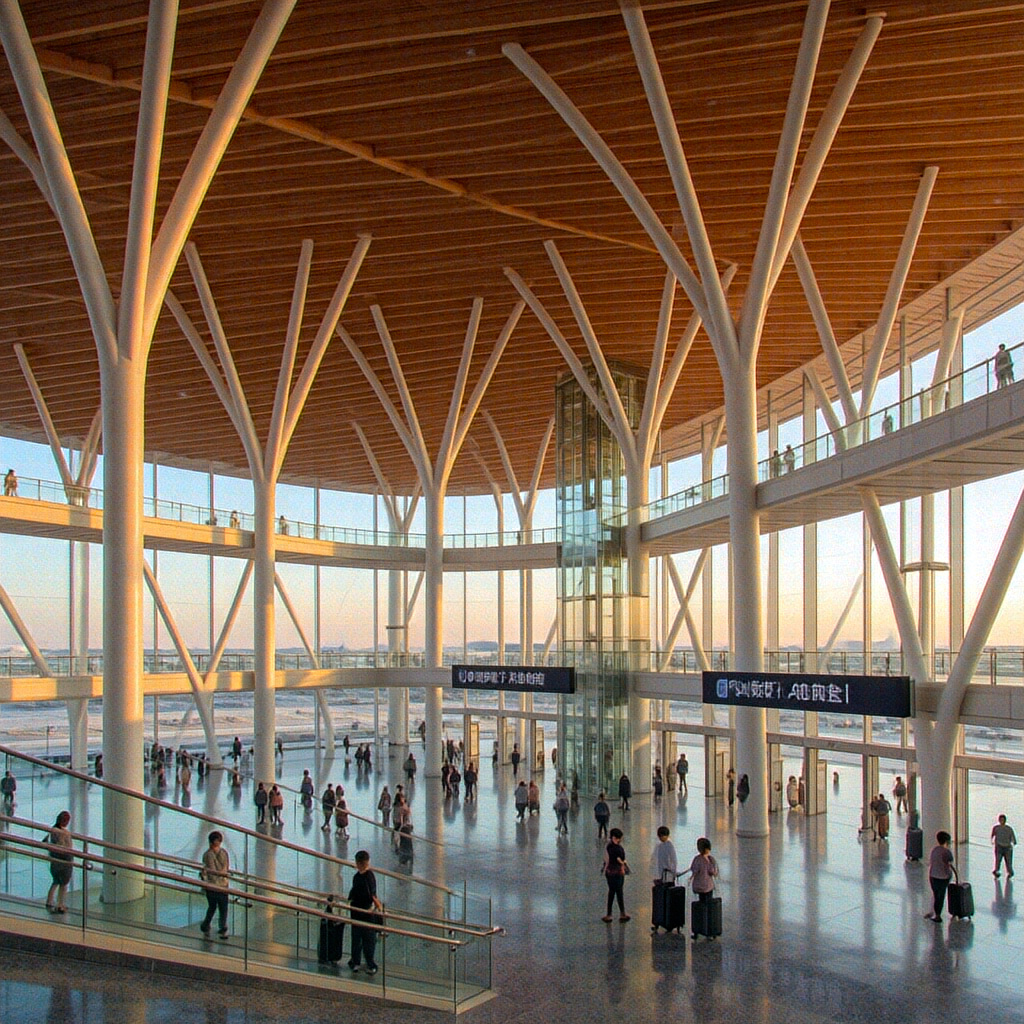}
{A futuristic international airport terminal with a sweeping timber roof, branching white columns, multiple elevated walkways, glass elevators, travelers pulling luggage, bilingual gate signs, and aircraft visible beyond the panoramic windows, rendered as a highly detailed architectural photograph at golden hour.}
\hfill
\ltwpsample
{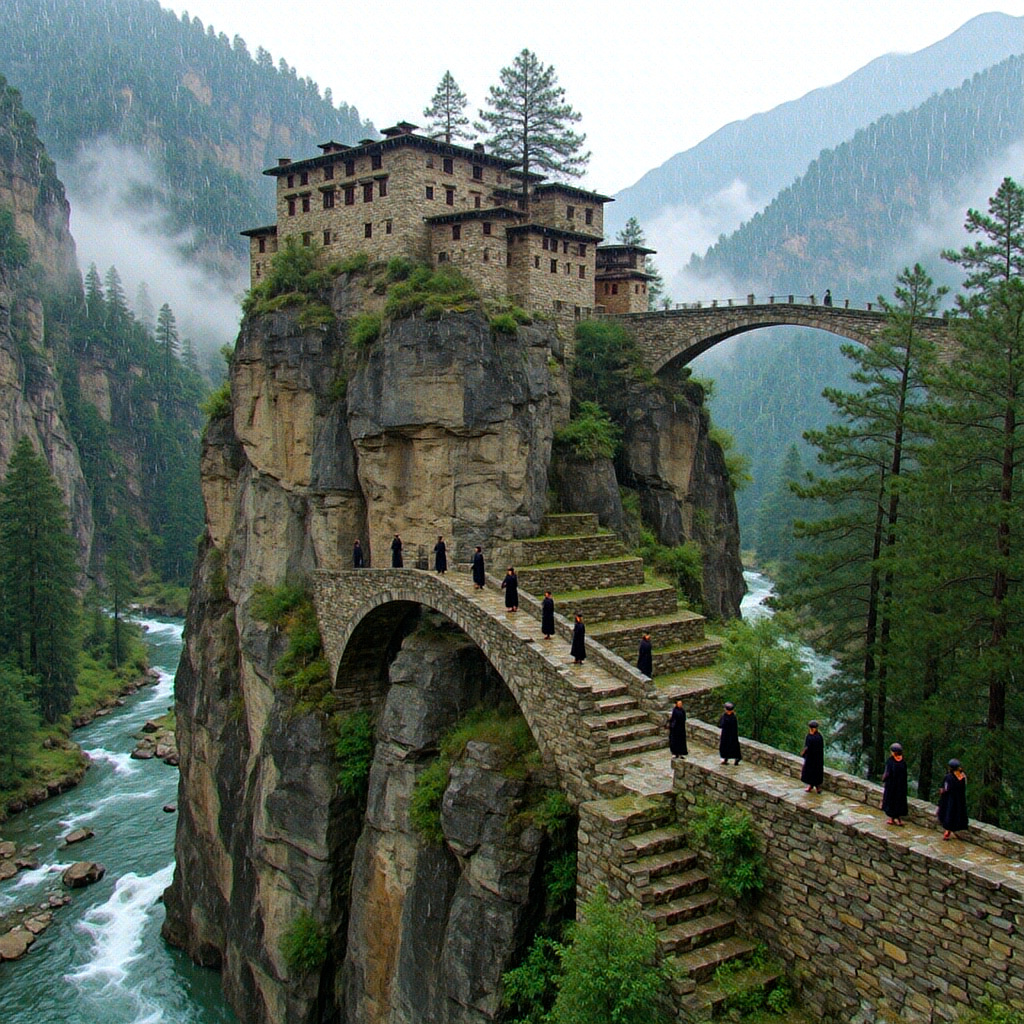}
{A remote stone monastery clings to a dramatic cliff above a winding river valley, connected by narrow arched bridges and steep stairways; tiny monks in dark robes cross the terraces as mist curls around pine-covered mountains in the background after rain.}
\par\vspace{0.35em}

\ltwpsample
{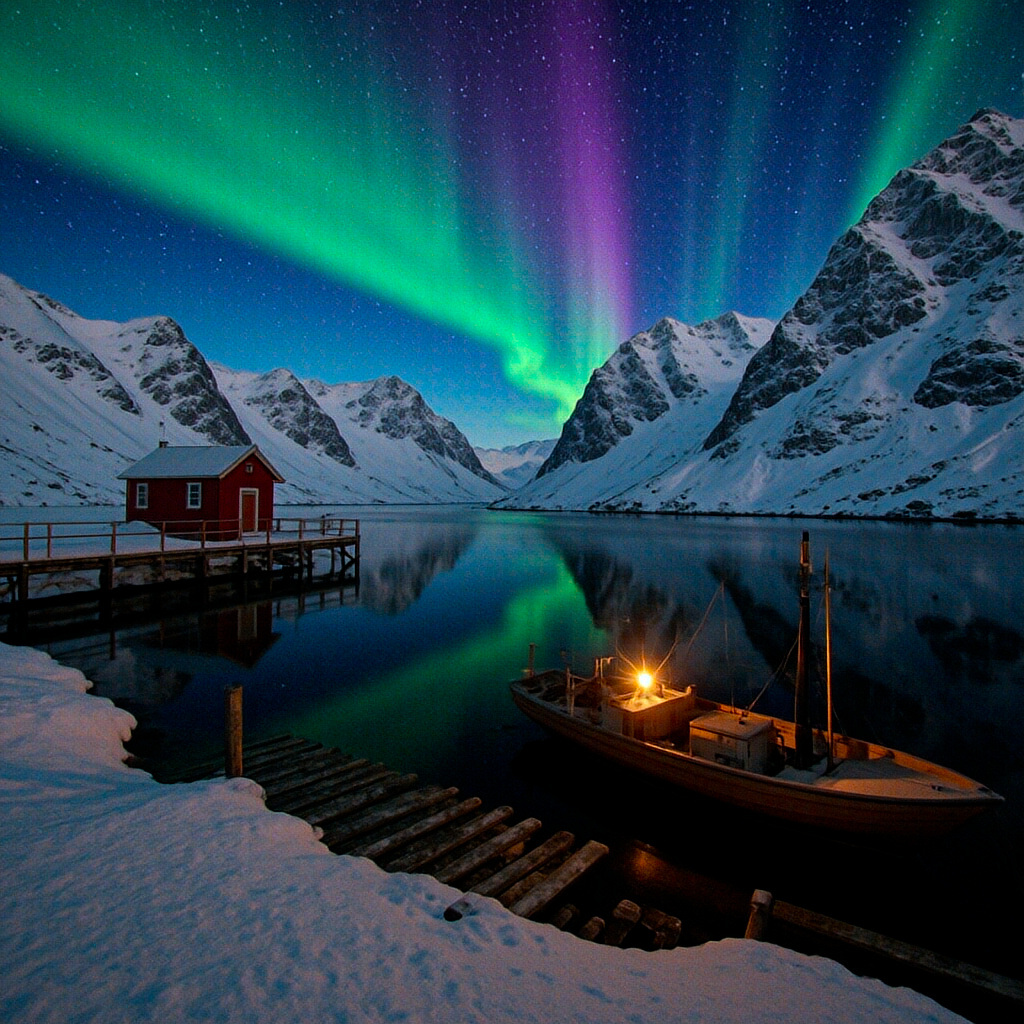}
{An Arctic fjord beneath the aurora borealis, with green and violet light reflected in dark water, jagged snow-covered mountains, a small red fishing cabin, a wooden pier, and a moored boat illuminated by one warm lamp under a star-filled sky.}
\hfill
\ltwpsample
{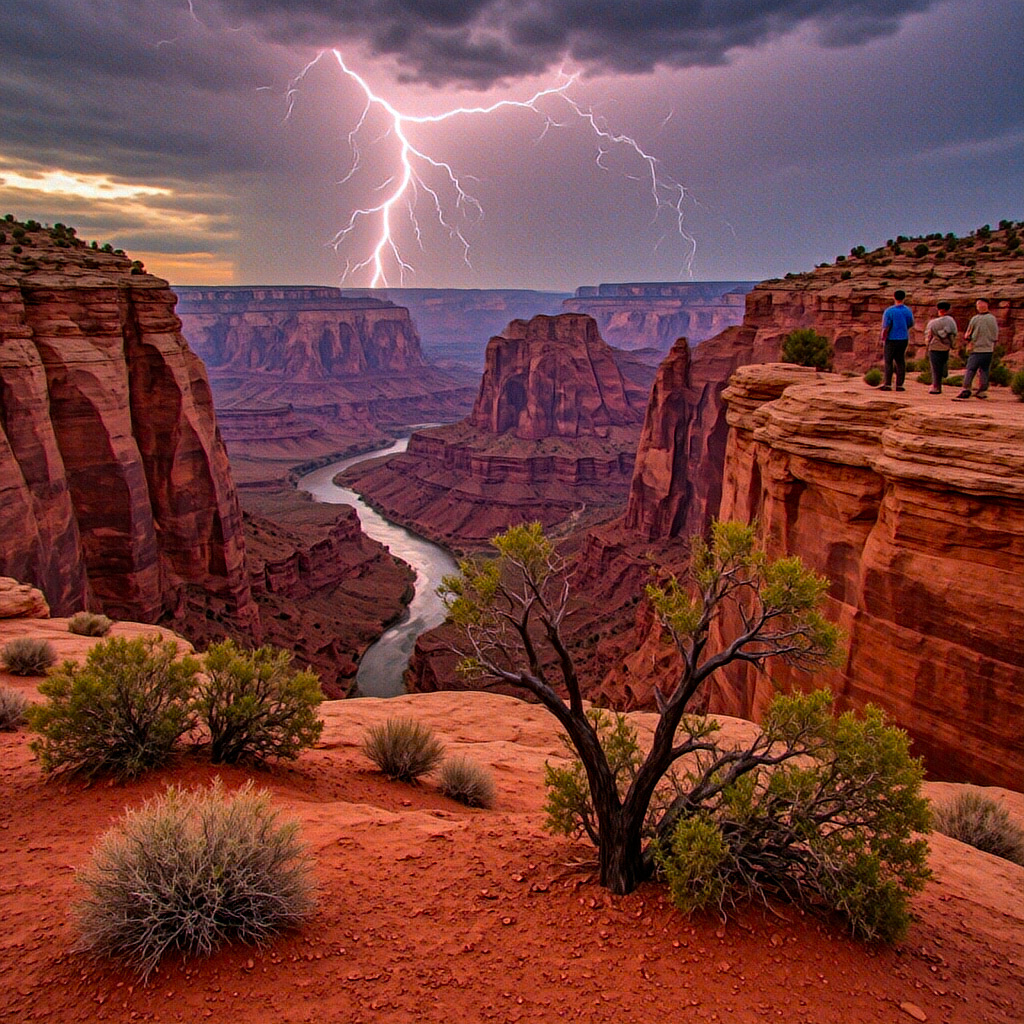}
{A vast red-rock canyon during an approaching thunderstorm, with branching lightning over distant mesas, a winding river far below, wind-bent desert shrubs in the foreground, and two hikers watching from a layered sandstone overlook in dramatic cinematic light.}
\hfill
\ltwpsample
{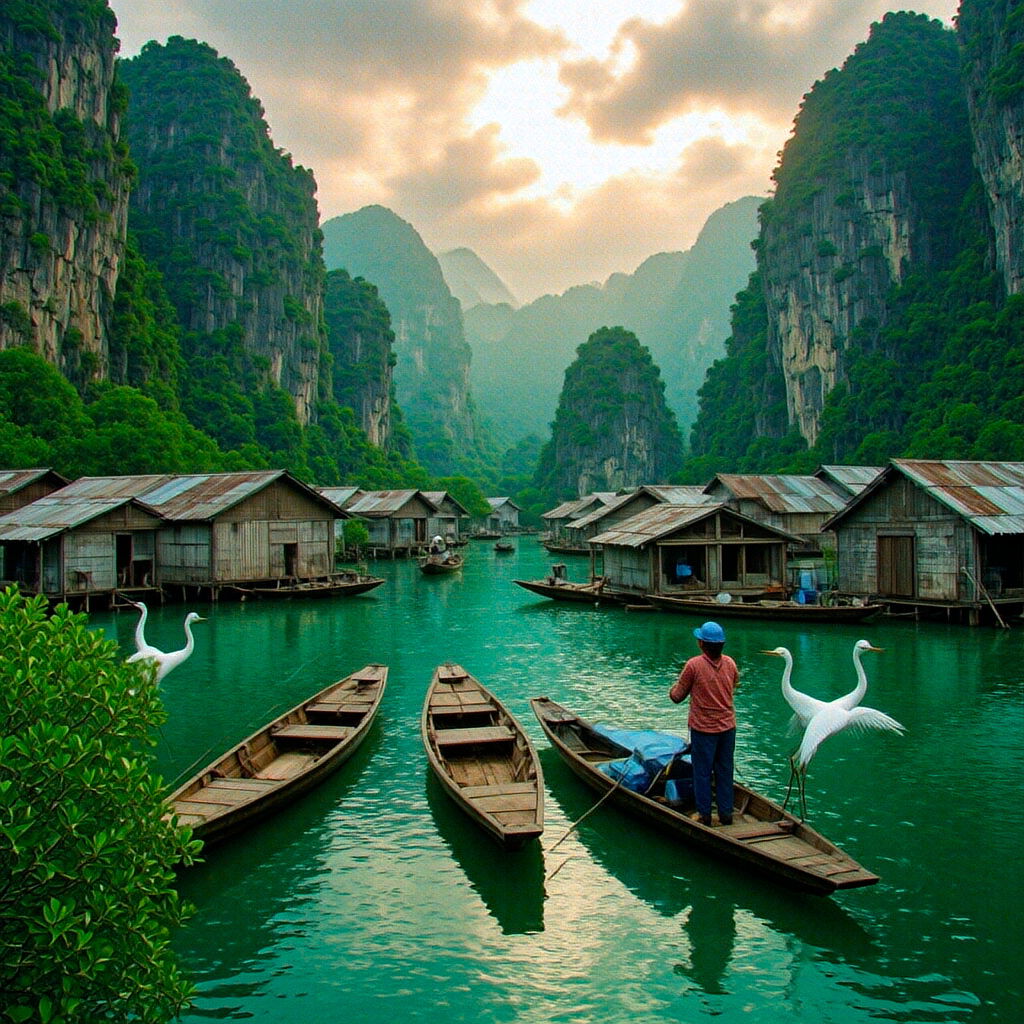}
{A tropical karst landscape where emerald limestone towers rise from a mangrove lagoon, fishermen guide narrow wooden boats between floating houses, white herons lift from the water, and late-afternoon sun breaks through humid clouds to illuminate the cliffs.}
\hfill
\ltwpsample
{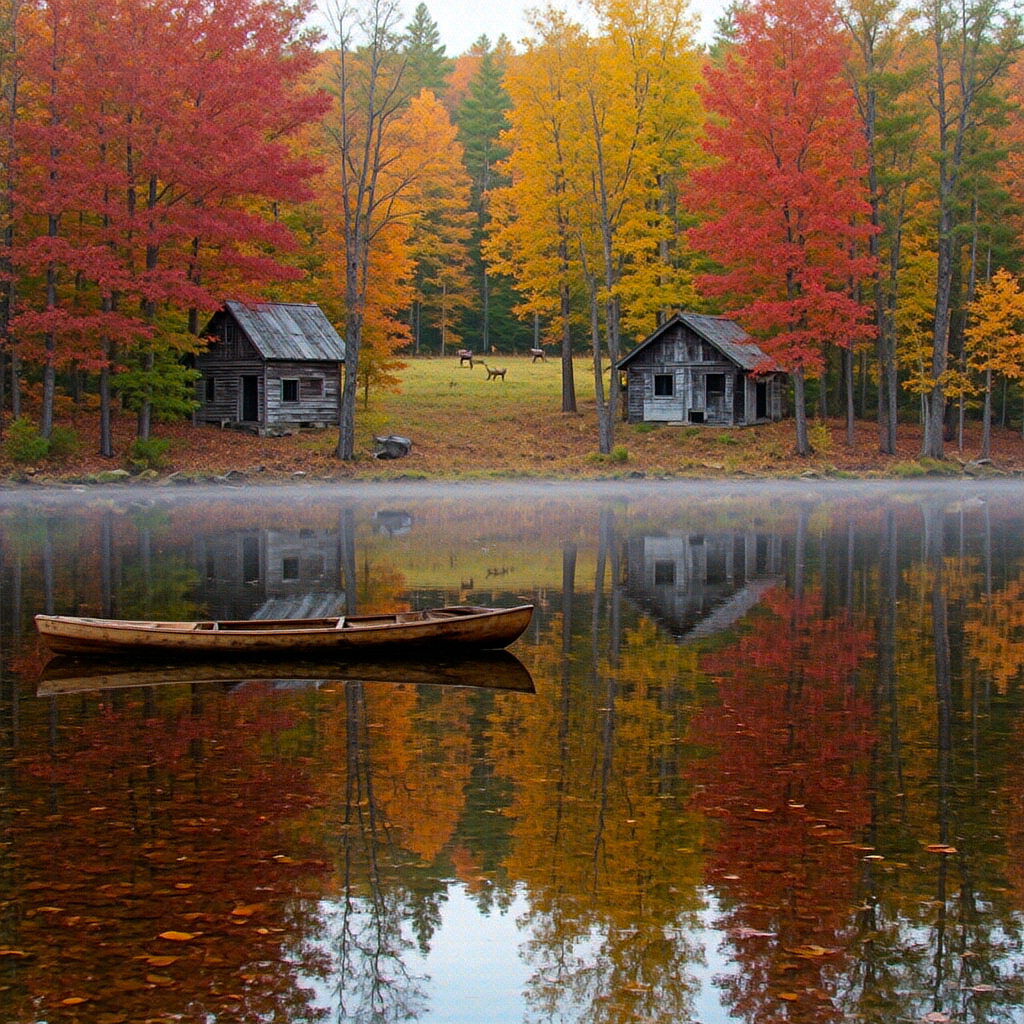}
{An autumn forest surrounding a glassy lake, with red and gold trees reflected in the water, a weathered cabin and canoe near the shore, deer crossing a distant clearing, and a thin layer of morning fog drifting between the trunks.}
\par\vspace{0.35em}

\ltwpsample
{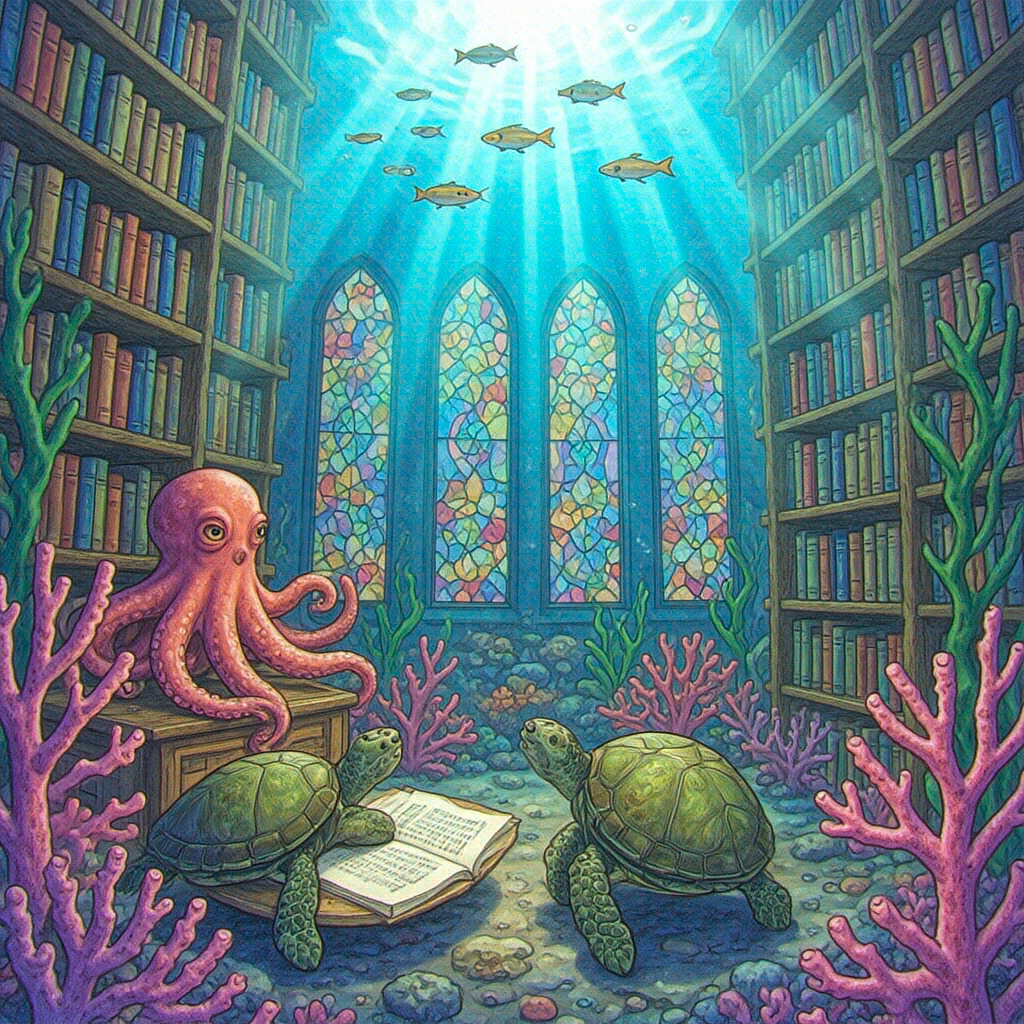}
{A luminous storybook illustration of an underwater library built inside a coral reef, where an octopus librarian arranges books, sea turtles read at shell-shaped tables, schools of fish pass stained-glass windows, and shafts of sunlight illuminate floating pages.}
\hfill
\ltwpsample
{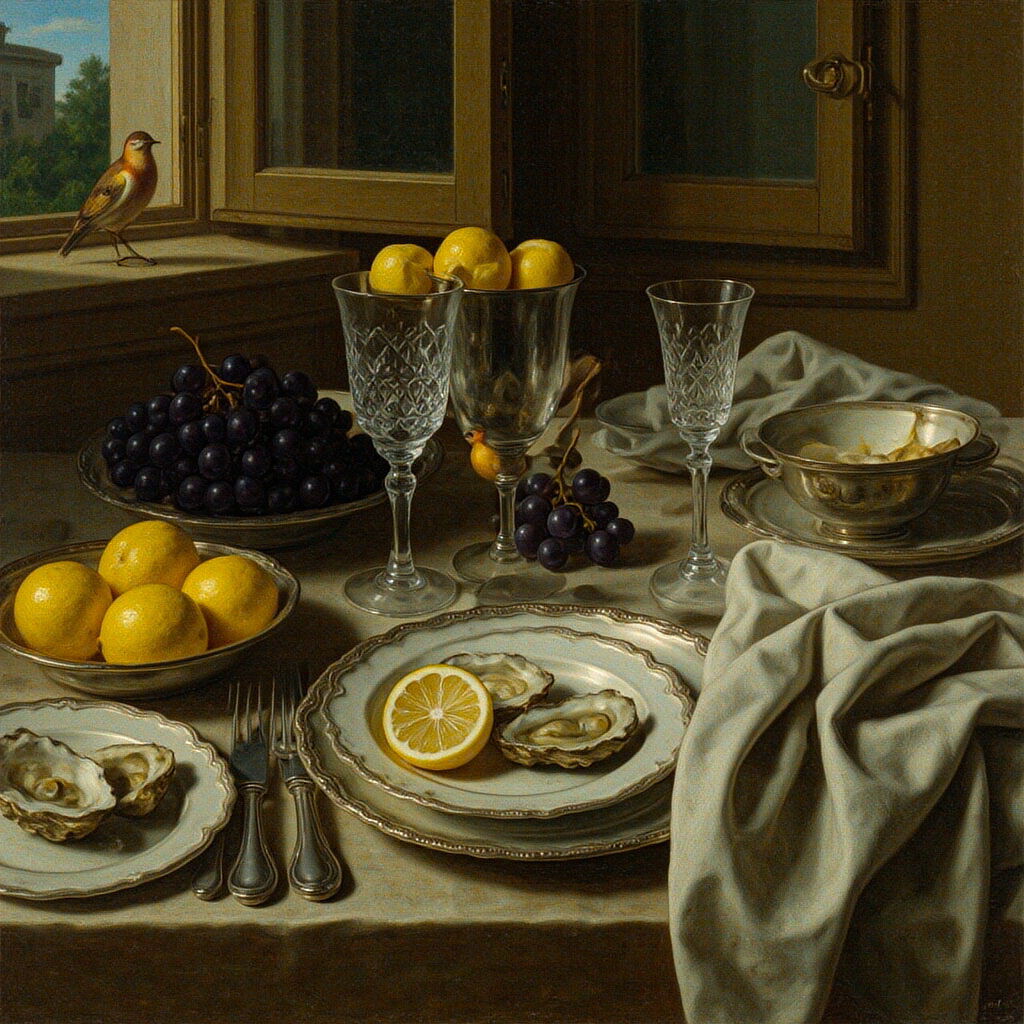}
{A Dutch Golden Age oil painting of an elaborate breakfast table with silverware, peeled lemons, oysters, dark grapes, a crystal goblet, folded linen, and a small songbird near an open window; dramatic chiaroscuro, deep shadows, and meticulous reflective surfaces.}
\hfill
\ltwpsample
{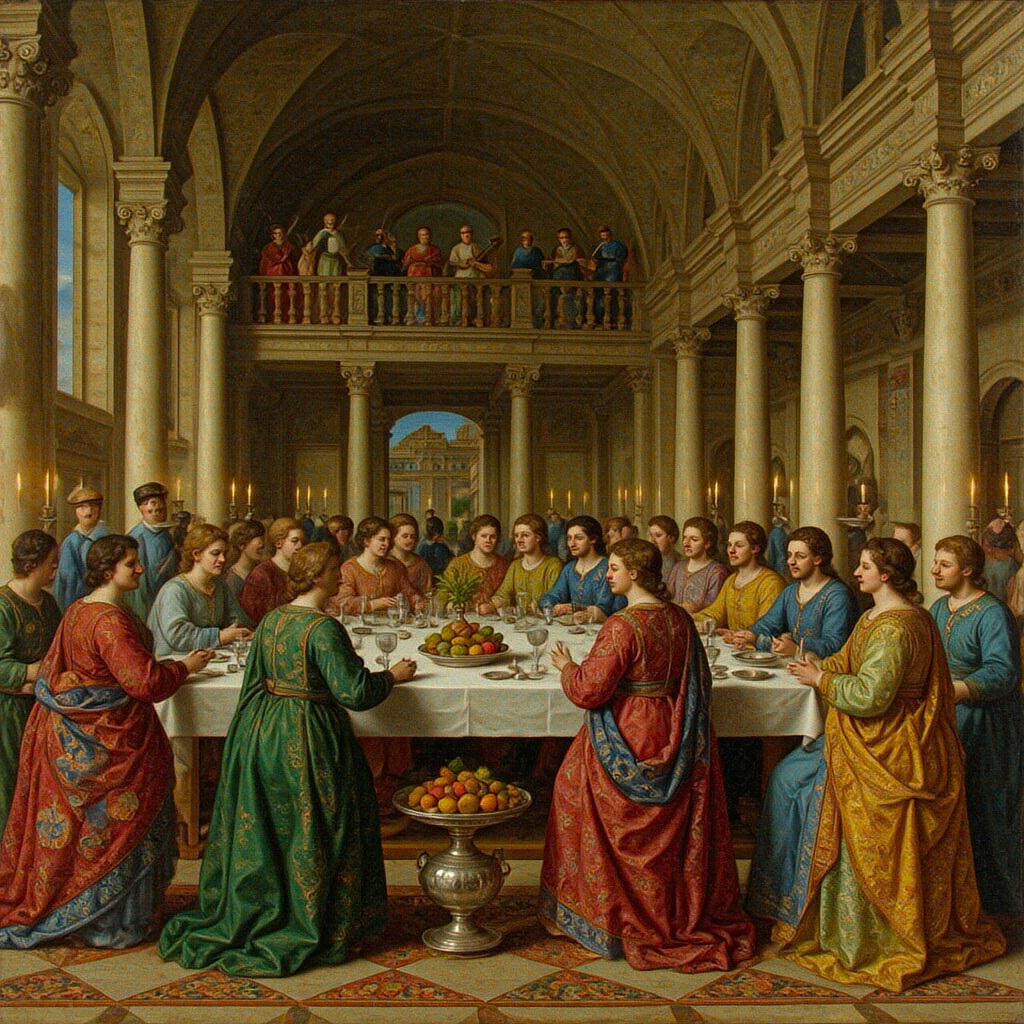}
{A large Renaissance-style oil painting of a multicultural banquet in a vaulted palace hall, with musicians on a balcony, servants carrying fruit and silver dishes, richly dressed guests conversing around a long table, intricate textiles, classical columns, and warm candlelight.}
\hfill
\ltwpsample
{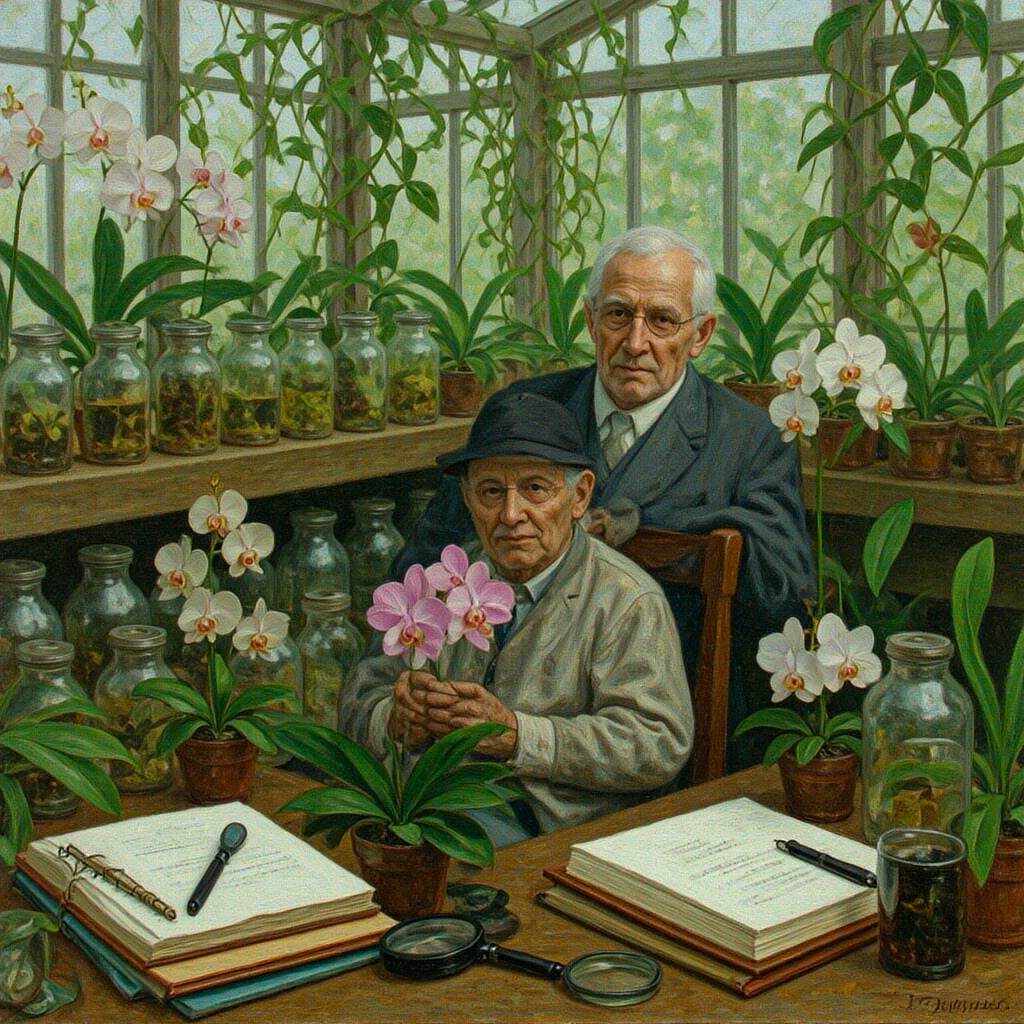}
{A formal oil portrait of an elderly botanist seated in a greenhouse, surrounded by specimen jars, orchids, notebooks, magnifying lenses, and climbing vines; the sitter holds a rare flower while diffuse window light reveals layered fabric and weathered hands.}
\par\vspace{0.35em}

\ltwpsample
{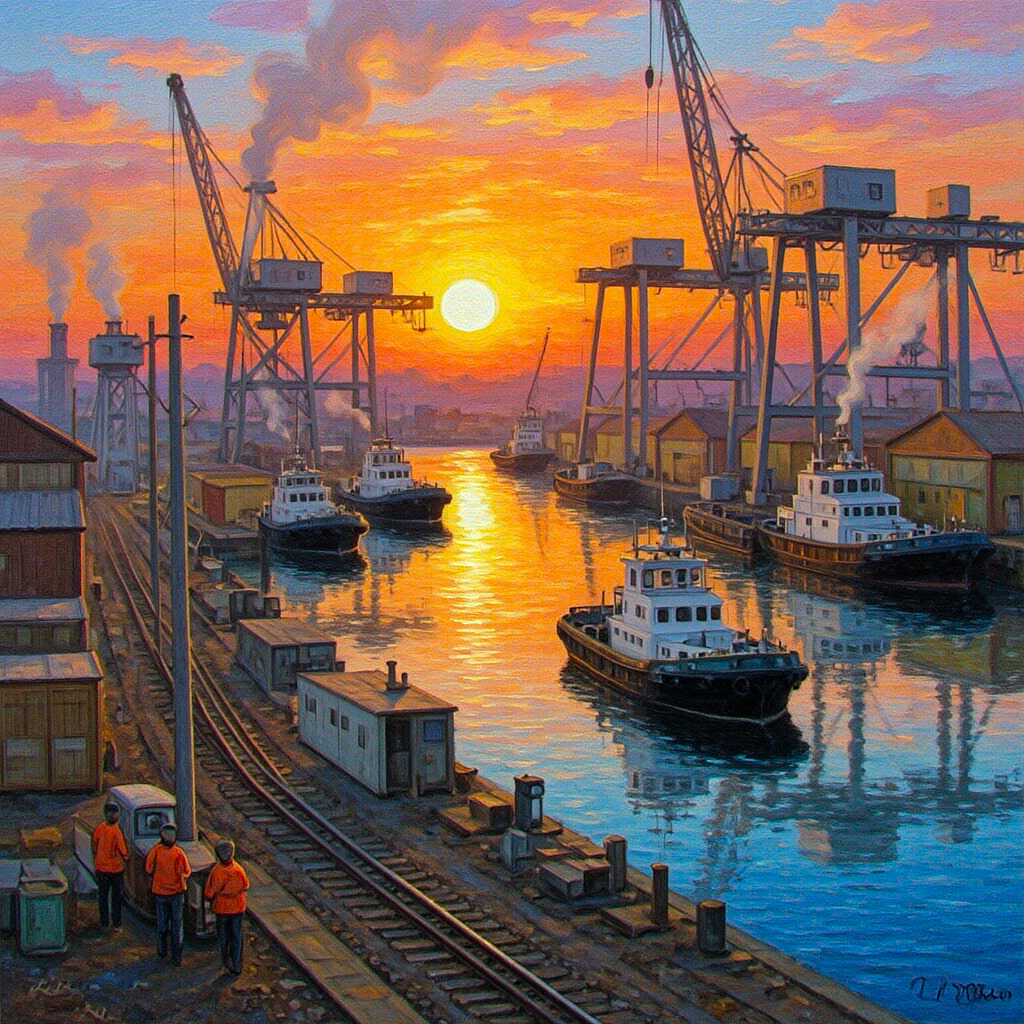}
{A panoramic oil painting of an industrial harbor at sunset, with cranes, warehouses, tugboats, rail lines, workers in orange jackets, smoke drifting above steel structures, and brilliant reflections across the water, combining precise architecture with luminous painterly color.}
\hfill
\ltwpsample
{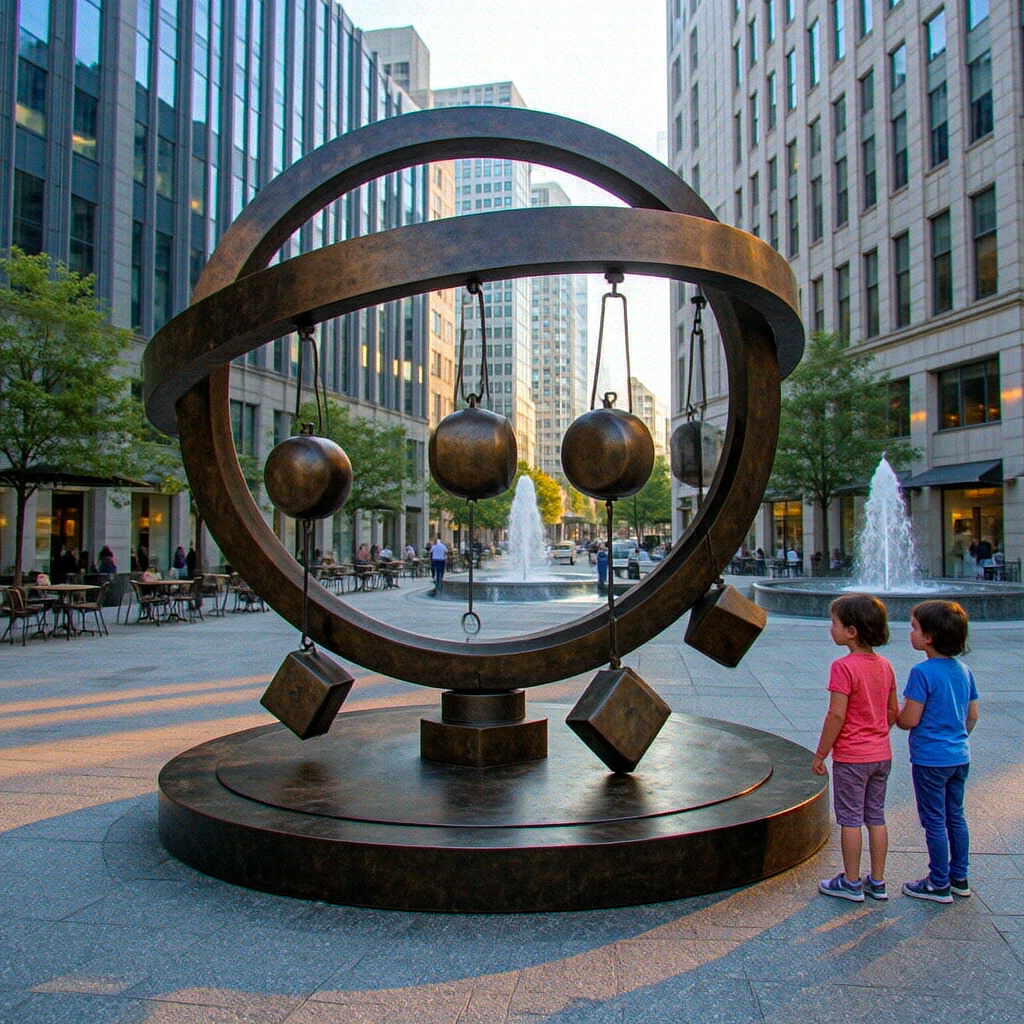}
{A large kinetic bronze sculpture in a contemporary city plaza, composed of rotating rings, counterweights, and suspended geometric forms; children watch it move while office towers, café tables, fountains, and long evening shadows fill the background.}
\hfill
\ltwpsample
{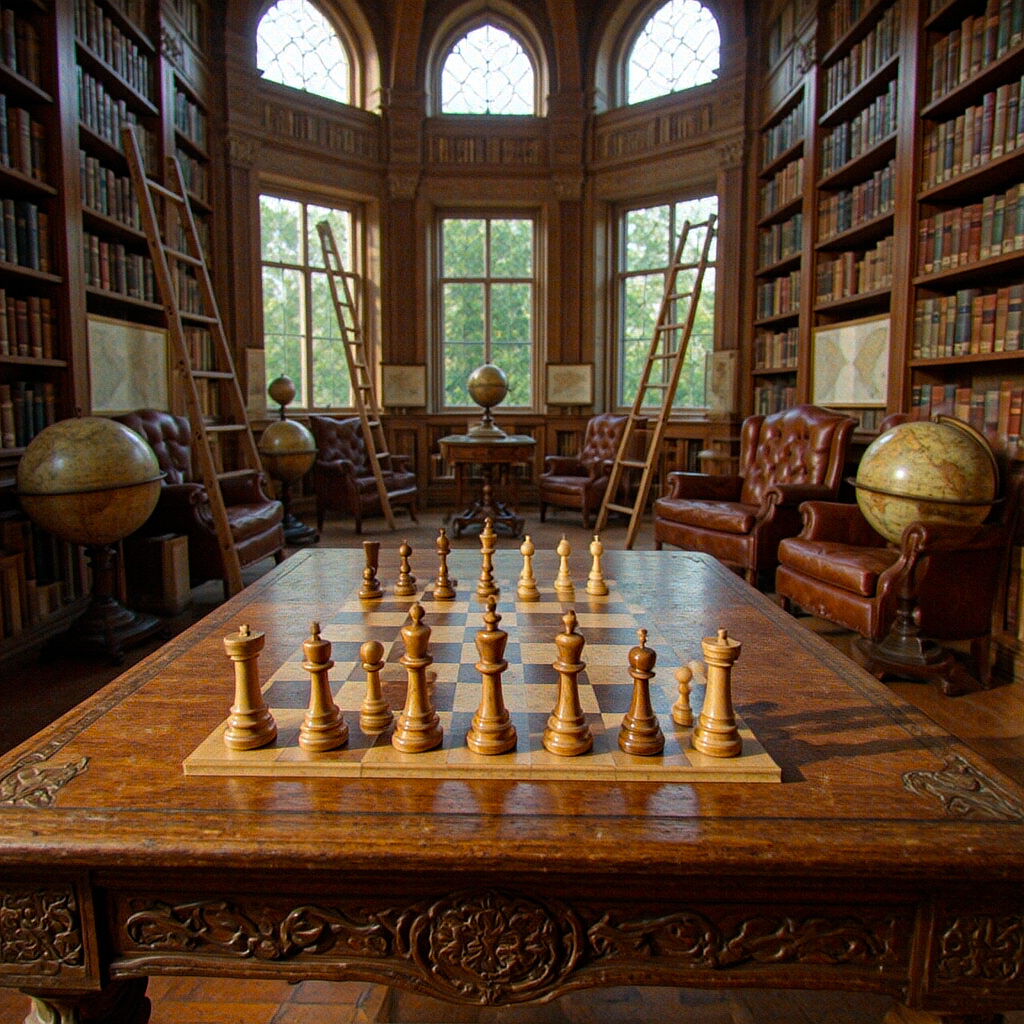}
{An antique ivory-and-ebony chess set sits midgame on a carved table in a grand library, surrounded by leather chairs, ladders, globes, maps, and floor-to-ceiling bookshelves; late-afternoon sunlight crosses the pieces through tall arched windows.}
\hfill
\ltwpsample
{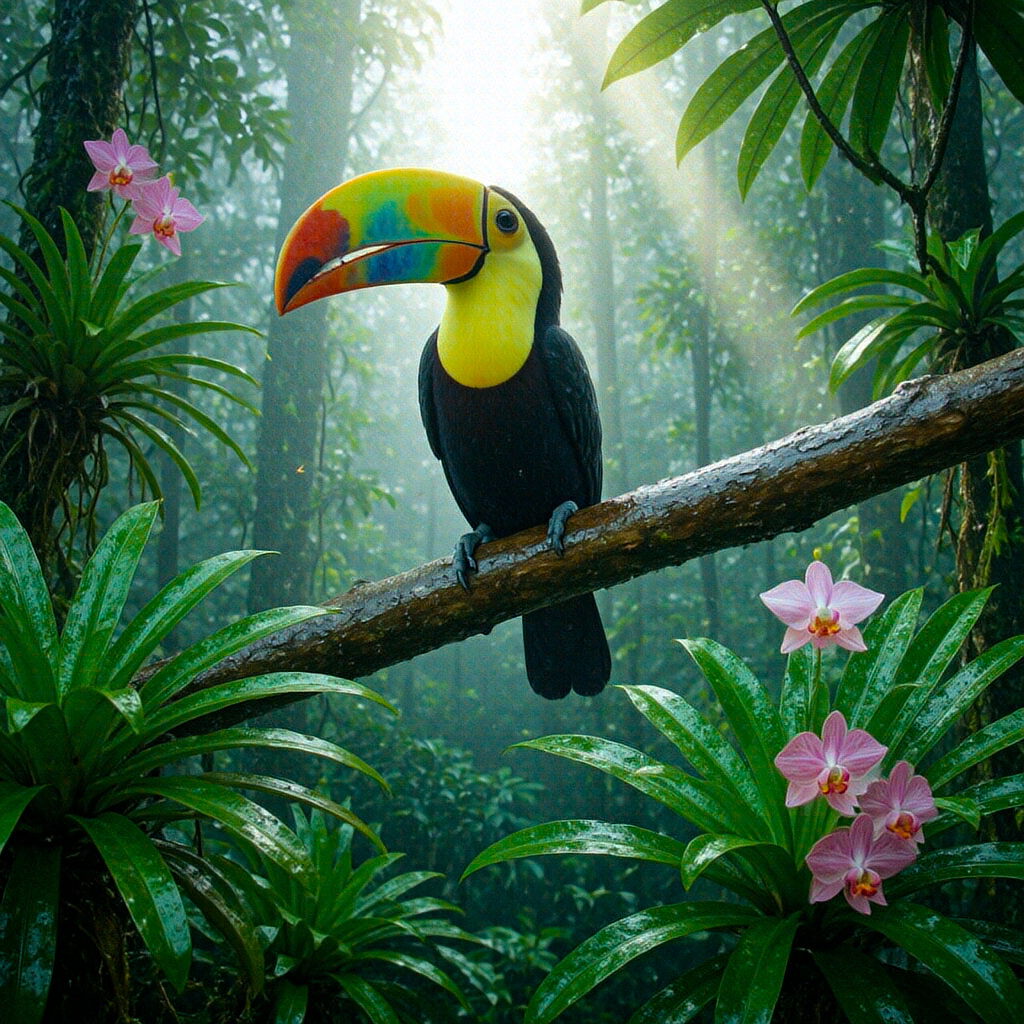}
{A keel-billed toucan with a large rainbow-colored beak perches on a wet branch in a misty cloud forest, surrounded by bromeliads, orchids, glossy leaves, drifting fog, and tiny insects illuminated by a narrow shaft of sunlight.}
\par\vspace{0.35em}

\ltwpsample
{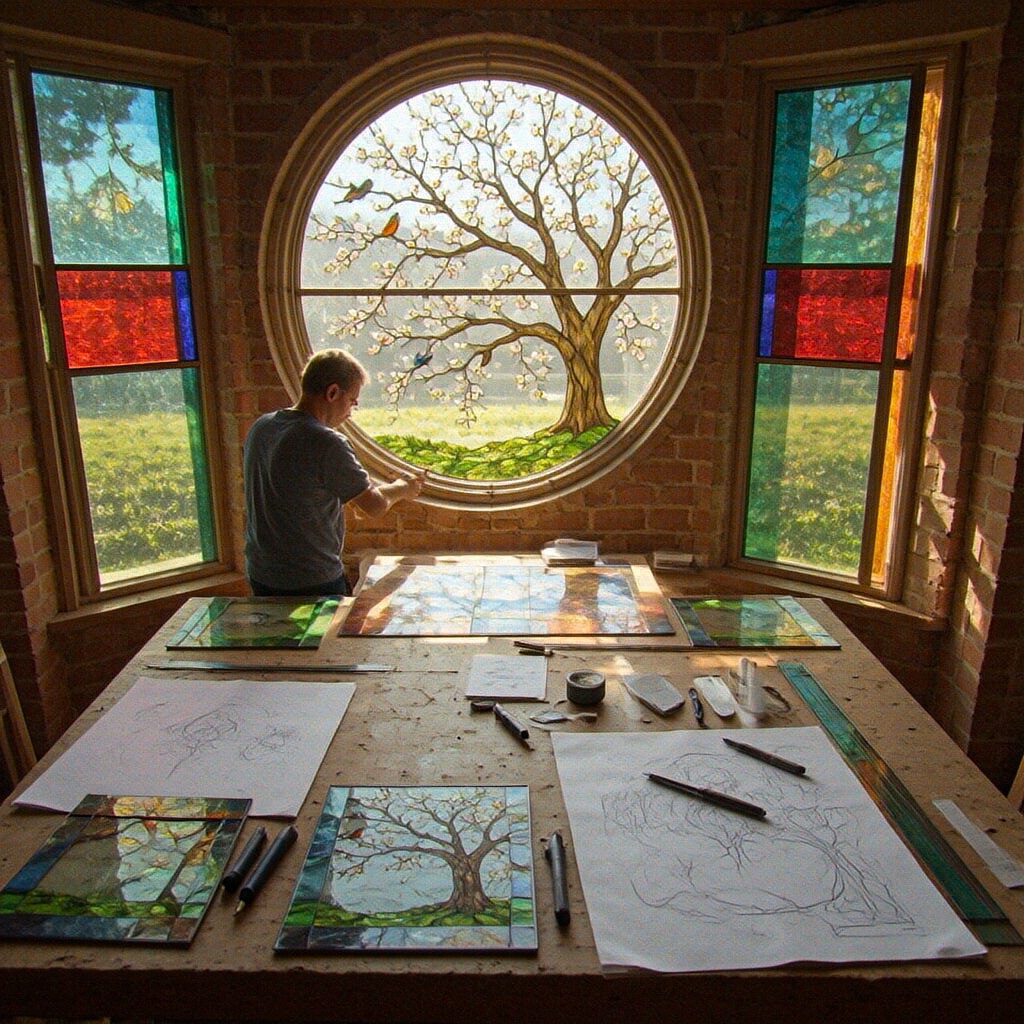}
{Inside a stained-glass workshop, two artisans assemble a large circular window showing birds and flowering trees, surrounded by colored glass sheets, lead strips, sketches, tools, and smaller finished panels glowing against the sunlit brick walls.}
\hfill
\ltwpsample
{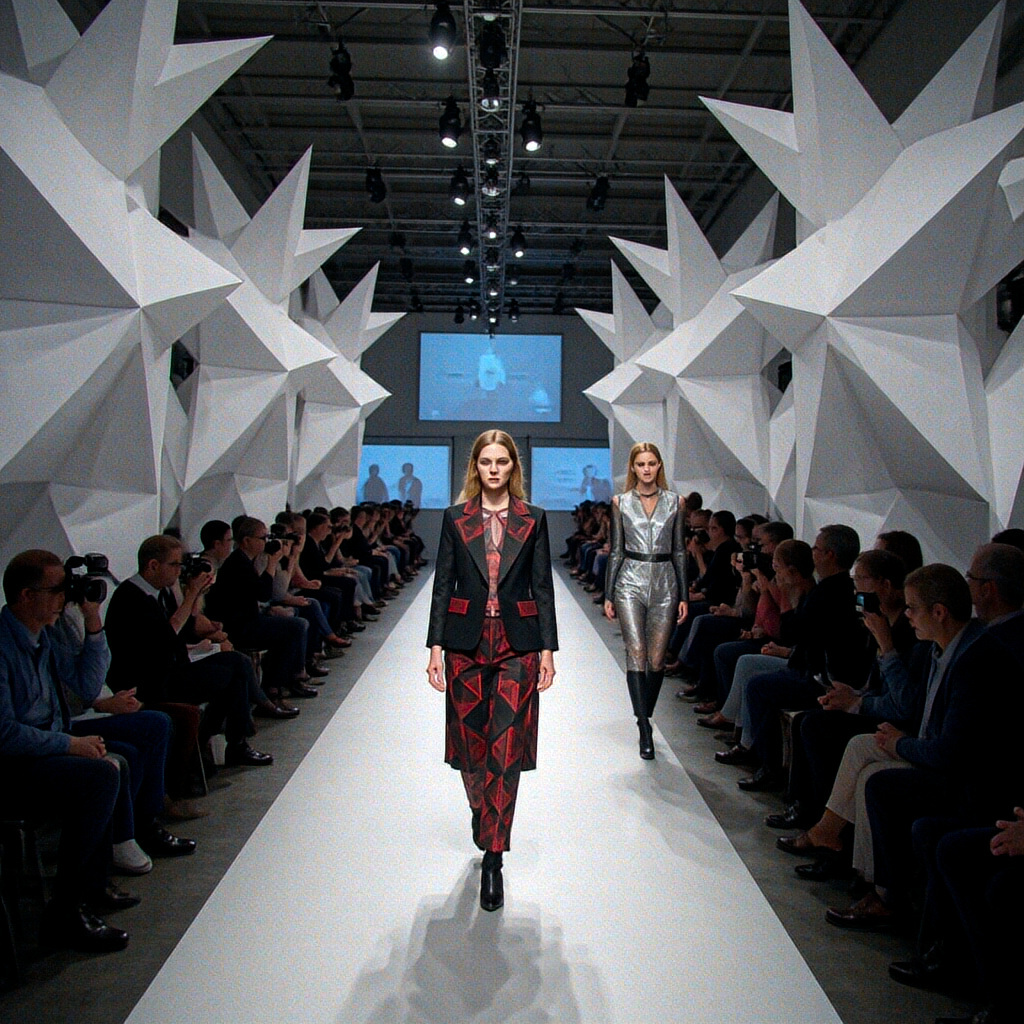}
{A fashion runway winds through a contemporary museum around towering white origami sculptures, while models wear geometric black, red, and silver clothing; photographers, seated guests, projection screens, and dramatic spotlights create a complex high-contrast scene.}
\hfill
\begin{minipage}[t]{0.22\textwidth}\strut\end{minipage}
\hfill
\begin{minipage}[t]{0.22\textwidth}\strut\end{minipage}
\par\vspace{0.35em}

\endgroup
\linenumbers

\FloatBarrier
\section{L2P-DMD Training Algorithm}
\label{sec:supp-algorithm}

\Cref{alg:l2p-dmd} gives our modified DMD update~\cite{yin2024dmd,yin2024dmd2}. $G_\theta$ is the complete pixel
sampler, $E$ includes all VAE preprocessing, and $\mathrm{sg}$ denotes
stop-gradient. Teacher CFG is applied only in $\operatorname{CleanPredict}(T_\psi)$.

\begingroup
\newcommand{\lTwoPHighlight}[1]{\colorbox{latorange!15}{\strut #1}}
\begin{algorithm}[H]
    \caption{L2P-DMD training core. Shaded operations highlight the encoder
    interface added to same-space DMD.}
    \label{alg:l2p-dmd}
    \small
    \begin{algorithmic}[1]
        \Require Initialized pixel sampler $G_\theta$, frozen latent teacher
        $T_\psi$, frozen VAE encoder $E$; initialize critic $C_\phi$ from $T_\psi$.
        \Require Critic updates per student update $r$, teacher CFG $\gamma$,
        time samplers $p_C,p_D$, corruption coefficients $\alpha_t,\sigma_t$,
        normalizer floor $\delta=10^{-5}$.
        \For{$k=1,2,\ldots$}
            \State Sample condition $c$ and student noise $\xi$.
            \State $x_C \gets \mathrm{sg}(G_\theta(\xi,c))$.
            \State \lTwoPHighlight{$\bar z_C \gets \mathrm{sg}(E(x_C))$}
                \Comment{Encode the detached critic-training sample.}
            \State Sample $t_C\sim p_C$ and $\epsilon_C\sim\mathcal{N}(0,I)$;
                $z^C_{t_C}\gets\alpha_{t_C}\bar z_C+\sigma_{t_C}\epsilon_C$.
            \State $\mathcal{L}_C \gets
                \ell_{\mathrm{native}}(C_\phi;z^C_{t_C},t_C,c,\bar z_C,\epsilon_C)$.
            \State $\phi \gets \operatorname{OptimizerStep}
                (\phi,\nabla_\phi\mathcal{L}_C)$.
            \If{$k\bmod r=0$}
                \State Sample fresh student noise $\xi$;
                    $x_0\gets G_\theta(\xi,c)$.
                \State \lTwoPHighlight{$z\gets E(x_0)$}
                    \Comment{Keep the input-gradient graph; freeze encoder weights only.}
                \State $\bar z\gets\mathrm{sg}(z)$;
                    sample $t\sim p_D$ and $\epsilon\sim\mathcal{N}(0,I)$.
                \State $z_t\gets\alpha_t\bar z+\sigma_t\epsilon$.
                \Statex \hspace{\algorithmicindent}\hspace{\algorithmicindent}
                    \textit{Compute the following predictions and correction without autograd:}
                \State $\hat z_T\gets
                    \operatorname{CleanPredict}(T_\psi,z_t,t,c;\gamma)$.
                \State $\hat z_C\gets
                    \operatorname{CleanPredict}(C_\phi,z_t,t,c;1)$.
                \State $w_z\gets
                    \max\!\left(\operatorname{mean}|\bar z-\hat z_T|,\delta\right)$;
                    $g_z\gets(\hat z_C-\hat z_T)/w_z$.
                \State $\mathcal{L}_{\mathrm{DMD}}\gets
                    \tfrac12\left\|z-\mathrm{sg}(z-g_z)\right\|_2^2$.
                \State \lTwoPHighlight{Backpropagate
                    $\mathcal{L}_{\mathrm{DMD}}$ through $E$ to $G_\theta$.}
                \State $\theta\gets\operatorname{OptimizerStep}
                    (\theta,\nabla_\theta\mathcal{L}_{\mathrm{DMD}})$;
                    keep $\psi$ and $E$ fixed.
            \EndIf
        \EndFor
        \State \Return $G_\theta$ \Comment{Direct-pixel inference; no VAE required.}
    \end{algorithmic}
\end{algorithm}
\endgroup

The critic uses its native denoising loss on detached encodings. For linear
flow, $\alpha_t=1-t$, $\sigma_t=t$, and the velocity target is
$\epsilon_C-\bar z_C$. A stochastic encoder reuses one reparameterized latent
sample throughout each student update. Replacing $E$ by the identity recovers
same-space DMD; latent matching instead gives
\begin{equation}
    \nabla_\theta\mathcal{L}_{\mathrm{DMD}}
    = J_{G_\theta}(\xi,c)^\top
      \underbrace{J_E(x_0)^\top}_{\text{through the frozen VAE encoder}}
      g_z .
    \label{eq:supp-encoder-dmd-gradient}
\end{equation}
Freezing $E$ preserves this input Jacobian; detaching $z$ would remove the
student gradient. \Cref{sec:supp-t2i-training} adds the training-only
text-to-image adversarial term.

\FloatBarrier
\section{ImageNet Experimental Setup}
\label{sec:supp-imagenet-setup}

\subsection{Pixel SiT Fine-Tuning Settings}
\label{sec:supp-pixelsit-finetuning}

\cref{tab:supp-pixelsit-finetuning} lists the pixel-space adaptation settings
used to obtain the VR242.5K initialization from pretrained latent SiT-XL/2.
This adaptation trains the full backbone and pixel heads together with
rank-256 LoRA adapters; the source teacher and SD-VAE remain frozen.
The ImageNet-256 distillation experiments using VR242.5K start from the EMA
checkpoint at 242,500 adaptation iterations.
The rank-16 interface aligns each normalized Oklab $16\times16$ pixel
patch with a $4\times2\times2$ SD-VAE latent patch through the
scale-calibrated Procrustes map used by AsymFlow. The adaptation retains
the latent SiT initialization of the intermediate Transformer blocks.
The projector and its scale buffers are retained in the DMD initializer.

\begin{center}
    \centering
    \small
    \setlength{\tabcolsep}{5.0pt}
    \captionsetup{hypcap=false}
    \captionof{table}{\textbf{Pixel SiT-XL fine-tuning settings.}
    These settings describe pixel-space adaptation before DMD distillation.
    The configured iteration budget and the checkpoint used downstream are
    reported separately.}
    \label{tab:supp-pixelsit-finetuning}
    \begin{tabular}{@{}p{0.34\textwidth}p{0.62\textwidth}@{}}
        \toprule
        Hyperparameter & Value \\
        \midrule
        Data / model & ImageNet-1K at $256^2$ / Pixel SiT-XL/16, subspace rank 16 \\
        Initialization / frozen modules & Latent SiT-XL/2 with Oklab Procrustes projector / source teacher and SD-VAE \\
        Trainable parameters & Full backbone and pixel heads plus rank-256 LoRA (dropout $0.05$); 776,590,336 parameters \\
        Pixel normalization & Oklab mean $(0.56,0.0,0.01)$, standard deviation $0.16$ \\
        Hardware / batch & $8\times$ H100 80GB; 32 per GPU, global 256 \\
        Optimizer / learning rates & AdamW8bit; base $10^{-4}$, \texttt{proj\_out} $10^{-3}$ \\
        Betas / weight decay / schedule & $(0.9,0.95)$ / $0$ / constant after 100-step warmup (initial factor $10^{-3}$) \\
        Precision / clipping & FP32 weights with BF16 autocast / norm 200 from step 100 \\
        Objective / weights & AsymFlowVR; MSE $10.0$, LPIPS $1.0$, VR shift $0.3$ \\
        Sigma floors / time sampling & Loss $0.05$, calibration $0.04$ / logit-normal $(0,1)$ with shift 17 \\
        Conditioning / guidance & Retention $0.9$ (dropout $0.1$) / source CFG $1.0$ \\
        EMA & Karras $\gamma=7$, every step from step 100 \\
        Budget / selected checkpoint & 400,000 / EMA at 242,500 steps (VR242.5K) \\
        Evaluation / checkpoints & FID-50K every 2,500 steps; checkpoints every 250 (mandatory every 1,000) \\
        \bottomrule
    \end{tabular}
\end{center}

\FloatBarrier
\subsection{Recovered Latent SiT Fine-Tuning}
\label{sec:supp-recovered-latent}

The recovered latent control starts from the released latent SiT-XL/2
checkpoint. We first perturb the floating-point weight tensors with at
least two dimensions,
\begin{equation}
    W'_l=W_l+0.05\,\operatorname{RMS}(W_l)\epsilon_l,
    \qquad \epsilon_l\sim\mathcal{N}(0,I),
    \label{eq:supp-latent-perturbation}
\end{equation}
using seed 20261028; one-dimensional biases and normalization parameters
are unchanged. We then fine-tune the full denoiser with its latent
flow-matching objective while freezing the SD-VAE. The selected model is
the EMA at 1,600 total recovery updates, not a directly noise-corrupted
checkpoint. Its formal no-CFG Heun-50 FID-50K is 11.4691.
The perturbation changes 202 of 403 tensors, containing 674,326,656
parameters, with aggregate relative $\ell_2$ change $0.0499962$.
We select an amplitude of 0.05 based on paired 2,048-sample calibration probes,
then verify degradation before recovery with formal FID-50K: the
perturbed teacher has unguided FID 14.3438, versus 9.3535 for the clean
latent reference.

\begin{center}
    \begin{minipage}{\textwidth}
    \centering
    \small
    \setlength{\tabcolsep}{5.0pt}
    \captionsetup{hypcap=false}
    \captionof{table}{\textbf{Recovered latent SiT training settings.}
    The recovery budget is separate from the 242.5K pixel-adaptation
    checkpoint and from downstream DMD training.}
    \label{tab:supp-recovered-latent}
    \begin{tabular}{@{}p{0.34\textwidth}p{0.62\textwidth}@{}}
        \toprule
        Hyperparameter & Value \\
        \midrule
        Initialization & Latent SiT-XL/2 with \cref{eq:supp-latent-perturbation} \\
        Model / trainability & 28 blocks, 16 heads, head dim. 72; full denoiser trainable, SD-VAE frozen \\
        Data / input & ImageNet-256 latents with random flip / $4\times32\times32$ \\
        Hardware / batch & $8\times$ H100 80GB, DDP / 64 per GPU, global 512 \\
        Optimizer / LR / schedule & Fused AdamW / $10^{-5}$ / constant without warmup \\
        Betas / epsilon / weight decay & $(0.9,0.95)$ / $10^{-8}$ / $0$ \\
        Objective / time sampling & Latent velocity MSE / uniform continuous time, shift 1, sigma floor $10^{-4}$ \\
        Dropout / clipping & Class $0.1$ / norm $1.0$ from update 0 \\
        Precision & FP32 weights with BF16 autocast; FP16 checkpoint weights and BF16 optimizer state \\
        EMA / seed & Momentum $0.999$ every step / base seed 2021 with rank shift \\
        Budget / selected EMA & 2,000 / 1,600 updates (819,200 sample presentations) \\
        Probe / formal evaluation & 2,048 / 50,000 samples; Heun-50, CFG 1.0; ADM reference, seed 2021 \\
        Checkpoint / probe cadence & Every 250 updates initially, then every 100 during refinement \\
        \bottomrule
    \end{tabular}
    \end{minipage}
\end{center}

The resumed trajectory evaluates EMA updates 1,000, 1,200, and 1,600 at
FID-50K $11.8806$, $11.7550$, and $11.4691$, respectively; update 1,600
initializes both the frozen teacher and online critic. Paired 2K probes only
monitor this trajectory and do not replace the formal selection measurements.

\begin{samepage}
\noindent\textbf{Offline unguided-FID evaluation.}
With Heun-50, 50,000 class-balanced samples, CFG 1.0, and seed 2021, the
Pixel SiT and recovered latent teachers obtain FID $11.9941$ and $11.4691$
(gap $0.5250$), respectively.
\par
\end{samepage}

\FloatBarrier
\subsection{ImageNet DMD Training Settings}
\label{sec:supp-imagenet-training}

\Cref{tab:supp-imagenet-dmd-shared} lists the shared DMD recipe;
\cref{tab:supp-imagenet-dmd-routes} gives route-specific initialization,
matching space, guidance, and budget.

\begin{center}
    \small
    \setlength{\tabcolsep}{5pt}
    \captionsetup{hypcap=false}
    \captionof{table}{\textbf{Shared ImageNet-256 DMD optimization settings.}
    These settings are distinct from the pixel-adaptation and recovered-teacher
    training recipes. Route-specific guidance, samplers, and budgets are listed
    separately.}
    \label{tab:supp-imagenet-dmd-shared}
    \begin{tabular}{@{}p{0.33\textwidth}p{0.63\textwidth}@{}}
        \toprule
        Setting & Value \\
        \midrule
        Task / student sampling & ImageNet-1K labels at $256^2$ / one endpoint from $t=1$ Gaussian noise; no inference CFG \\
        Pixel / latent noise shape & $3\times256\times256$ / $4\times32\times32$ \\
        Optimizers / learning rates & AdamW / $2\times10^{-6}$ for both student and critic \\
        Betas / weight decay / schedule & $(0.9,0.999)$ / $0$ / constant after 500-step warmup (initial factor $10^{-3}$) \\
        Clipping / update ratio / batch & Norm $10$ for both / $1{:}1$ / 8 per GPU, global 64 \\
        Precision / latent encoder & FP32 trainable weights, BF16 autocast and teacher / frozen FP32 SD-VAE encoder \\
        Student EMA & Momentum $0.9999$ from iteration zero \\
        Query times / warp & $u\in[0.02,0.98]$; $s_q(u)=qu/[1+(q-1)u]$ with route-specific $q$ \\
        Guidance / normalizer & Route-specific teacher CFG, unguided critic / mean absolute teacher residual, floor $10^{-5}$ \\
        Auxiliary objectives & None in the primary ImageNet routes \\
        Evaluation / checkpoints & One-step EMA FID-50K every 10,000 updates; checkpoints every 2,000 (mandatory every 20,000) \\
        \bottomrule
    \end{tabular}
\end{center}

\noindent\textbf{Initializer identifiers.}
In \cref{tab:supp-imagenet-dmd-routes}, L denotes pretrained latent
SiT-XL/2, P the pixel SiT EMA after 242,500 adaptation updates, and R
the recovered latent EMA after 1,600 updates, constructed in
\cref{sec:supp-recovered-latent}. J8 denotes the released AsymJiT-H/16
model with a rank-8 interface. J16 is the rank-16 critic initializer,
constructed from the common AsymJiT weights; the native AsymJiT teacher
and student remain rank 8.
The critic starts from the indicated initializer and is subsequently
trained online, whereas the teacher remains frozen.

\begin{center}
    \small
    \setlength{\tabcolsep}{4pt}
    \captionsetup{hypcap=false}
    \captionof{table}{\textbf{Initializers, matching recipes, and recorded
    budgets for the ImageNet routes.} Both the DMD query and critic-training
    samplers use uniform raw times with the listed shift. ``Cap / recorded''
    reports the configured update budget and the latest logged student update.}
    \label{tab:supp-imagenet-dmd-routes}
    \begin{tabular}{@{}p{0.19\textwidth}p{0.09\textwidth}p{0.09\textwidth}p{0.19\textwidth}p{0.06\textwidth}p{0.28\textwidth}@{}}
        \toprule
        Student / init & Teacher & Critic init & Matching / shift & CFG & Cap / recorded updates \\
        \midrule
        Latent SiT / L & L & L & Latent / $1$ & $1.6$ & 100,000 / 100,000 \\
        Pixel SiT / P & P & P & Pixel Oklab / $10$ & $1.4$ & 200,000 / 200,000 \\
        Pixel SiT / P & L & L & SD-VAE latent / $1$ & $1.6$ & 200,000 / 200,000 \\
        Pixel SiT / P & R & R & SD-VAE latent / $1$ & $1.6$ & 200,000 / 200,000 \\
        \midrule
        AsymJiT / J8 & J8 & J16 & Pixel RGB / $1$ & $1.4$ & 200,000 / 200,000 \\
        AsymJiT / J8 & L & L & SD-VAE latent / $1$ & $1.2$ & 100,000 / 100,000 \\
        AsymJiT / J8 & R & R & SD-VAE latent / $1$ & $1.2$ & 100,000 / 100,000 \\
        AsymJiT / J8 & P & P & Pixel Oklab / $10$ & $1.4$ & 200,000 / 152,650; last FID at 150,000 \\
        \bottomrule
    \end{tabular}
\end{center}

\paragraph{Recorded CFG searches.}
\label{sec:supp-imagenet-cfg-search}
The evaluated runtime teacher CFG values were: Latent SiT/L, $1.6$;
Pixel SiT/P, $\{1.0,1.2,1.4,2.0\}$; Pixel SiT/L with shifts 1 and 2,
$\{1.0,1.6,2.0,2.5,3.0\}$ and $\{1.2,1.4\}$; Pixel SiT/R,
$\{1.0,1.4,1.6,1.8,2.0\}$; AsymJiT/J8--J16,
$\{1.0,1.2,1.4,1.6,1.8\}$; AsymJiT/L,
$\{1.2,1.4,1.6,1.8\}$, including two full-interval runs at $1.4$;
AsymJiT/R, $1.2$; and AsymJiT/P, $1.4$.
These runtime values are independent of legacy job-name suffixes; the main
tables report the best observed one-step EMA FID-50K.

\FloatBarrier
\section{Additional ImageNet Results and Diagnostics}
\label{sec:supp-imagenet-results}

\subsection{Alternative Rules for Backpropagation and VAE Controls}
\label{sec:supp-backprop}

This section compares backpropagation through the frozen VAE encoder
(encoder VJP) with two decoded update heuristics and controls for the cost of
a frozen VAE round trip. All update rules reuse one posterior sample per
update, train the critic on detached encodings, and normalize in latent space.

\subsubsection{Decoded alternatives to backpropagation}

\begin{figure}[H]
    \centering
    \includegraphics[width=0.86\textwidth]{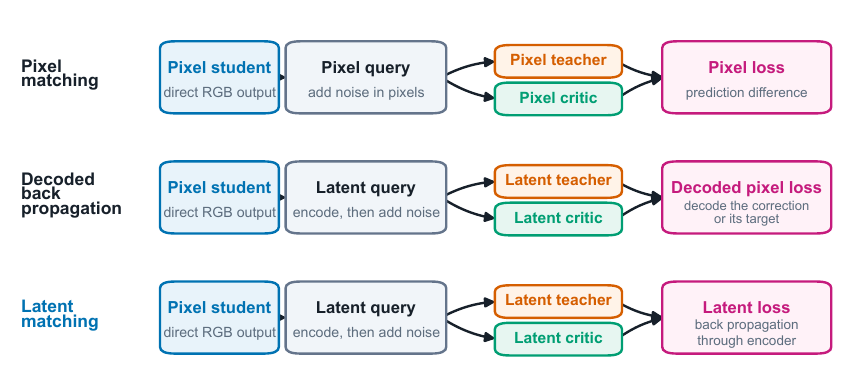}
    \caption{\textbf{Backpropagation across spaces is a design choice.}
    Pixel matching keeps every component in pixel space. A seemingly natural
    alternative encodes the sample, computes the correction in latent space,
    and decodes that correction or its target into a pixel loss. Our
    latent-matching system instead applies the objective to the encoded student
    output and uses backpropagation through the frozen encoder. The critic is fitted in a
    separate alternating update, and only the pixel student is retained for
    inference.}
    \label{fig:arm3-latent-to-pixel}
\end{figure}

For both decoded alternatives, $z=E(x_0)$, $z_0=\mathrm{sg}(z)$, and $g_z$
uses the same latent teacher, critic, corruption, and normalizer. Let $D$ be
the paired frozen decoder with its required preprocessing. The \textbf{decoded
step + restore} heuristic decodes the latent DMD step and adds a
$0.1$-weighted VAE reconstruction-restoration term:
\begin{equation}
    \widetilde g_x^{\mathrm{restore}}
    =D(z_0)-D(z_0-g_z)+0.1\bigl[x_0-D(z_0)\bigr].
    \label{eq:supp-decoder-restore}
\end{equation}
The first difference is the decoded DMD step; the bracketed term pulls the
student sample toward its VAE reconstruction and can compete with the DMD
update.

The \textbf{decoded endpoints} heuristic instead decodes the online critic
and frozen teacher clean estimates and treats their difference as a detached
pixel update:
\begin{equation}
    \begin{aligned}
        \widetilde g_x^{\mathrm{end}}
        &=\frac{D(z_{\mathrm{critic}})-D(z_{\mathrm{real}})}{w_z}, \\
        \mathcal{L}_{\mathrm{dec}}^{\mathrm{end}}
        &=\frac{1}{2}\left\|x_0-
          \mathrm{sg}(x_0-\widetilde g_x^{\mathrm{end}})\right\|_2^2.
    \end{aligned}
    \label{eq:supp-decoder-endpoints}
\end{equation}

The encoder-VJP baseline instead applies $J_E(x_0)^\top g_z$. The cycle
control retains this update and adds
\begin{equation}
    \mathcal{L}_{\mathrm{cycle}}
    =0.1\,\frac{1}{2}\left\|x_0-
      \mathrm{sg}\!\left[D(E(x_0))\right]\right\|_2^2.
    \label{eq:supp-latent-cycle}
\end{equation}

\begin{center}
    \centering
    \small
    \setlength{\tabcolsep}{5.0pt}
    \captionsetup{hypcap=false}
    \captionof{table}{\textbf{Current-recipe alternatives for transferring
    latent matching gradients to a pixel student.}
    All rows use the VR242.5K initialization, teacher CFG 1.6, and uniform
    (shift-1) sampling. The decoded variants are controls rather than co-equal
    proposed methods.}
    \label{tab:supp-backprop-ablation}
    \begin{tabular}{@{}llll@{}}
        \toprule
        Backpropagation & Pixel-space signal & Best FID $\downarrow$ & Outcome \\
        \midrule
        Decoded step + restore & $D(z_0)-D(z_0-g_z)+0.1[x_0-D(z_0)]$ & 47.36 @ 30k & 236.97 @ 200k \\
        Decoded endpoints & $D(z_{\mathrm{critic}})-D(z_{\mathrm{real}})$ & 382.72 @ 20k & stopped at 21.6k \\
        Encoder VJP & $J_E(x_0)^\top g_z$ & \textbf{7.15 @ 200k} & stable \\
        Encoder VJP + cycle 0.1 & $J_E(x_0)^\top g_z$ + cycle & \textbf{7.11 @ 200k} & neutral \\
        \bottomrule
    \end{tabular}
\end{center}

\begin{center}
    \centering
    \small
    \setlength{\tabcolsep}{4pt}
    \captionsetup{hypcap=false}
    \captionof{table}{\textbf{Current-recipe checkpoint trajectories.}
    Entries are one-step EMA FID-50K. All three runs use the VR242.5K
    initialization, teacher CFG 1.6, and uniform (shift-1) sampling.}
    \label{tab:supp-backprop-trajectories}
    \begin{tabular}{@{}lrrrrr@{}}
        \toprule
        Update rule & 30k & 50k & 100k & 130k & 200k \\
        \midrule
        Encoder VJP & 17.83 & 10.99 & 8.51 & 7.81 & 7.15 \\
        Encoder VJP + cycle 0.1 & 18.47 & 11.30 & 8.41 & 7.83 & 7.11 \\
        Decoded step + restore 0.1 & 47.36 & 56.03 & 118.10 & 186.50 & 236.97 \\
        \bottomrule
    \end{tabular}
\end{center}

The current endpoint rerun had already collapsed to 382.72 FID at 20k
updates and was stopped at 21.6k, so it does not provide a comparable
long-horizon trajectory.

\FloatBarrier
\subsubsection{Matched VAE round-trip control}

\begin{center}
    \centering
    \small
    \setlength{\tabcolsep}{5.0pt}
    \captionsetup{hypcap=false}
    \captionof{table}{\textbf{Matched cost of the frozen VAE round trip.}
    The round-trip route decodes and re-encodes the latent student output, so
    the gradient traverses the frozen decoder and encoder.  Entries are
    one-step EMA FID-50K; lower is better.}
    \label{tab:supp-vae-roundtrip}
    \begin{tabular}{@{}lccc@{}}
        \toprule
        Updates & Direct latent & Decode + encode & $\Delta$ \\
        \midrule
        30k & 6.60 & 7.65 & +1.05 \\
        50k & 4.72 & 4.98 & +0.27 \\
        70k & \textbf{3.98} & 4.59 & +0.61 \\
        100k & 4.26 & 4.55 & +0.30 \\
        \midrule
        Best & \textbf{3.98} @ 70k & 4.37 @ 90k & +0.40 \\
        \bottomrule
    \end{tabular}
\end{center}

\FloatBarrier
\subsection{ImageNet-256 Convergence Results}
\label{sec:supp-imagenet}

\begin{center}
    \centering
    \includegraphics[width=\textwidth]{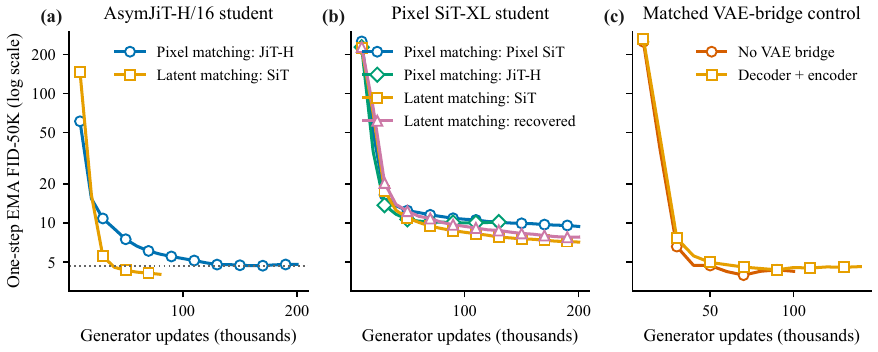}
    \captionsetup{hypcap=false}
    \captionof{figure}{\textbf{One-step ImageNet-256 convergence.} FID-50K is plotted on
    log-scale at 10k training step evaluation intervals. (a) Latent matching
    reaches the native AsymJiT-H pixel-matching result by 40k rather than 170k
    updates. (b) For the Pixel SiT student, the latent-matching runs start more
    slowly but overtake both pixel-matching runs; latent matching with native
    SiT reaches the pixel-matching result with Pixel SiT (9.43) by 80k rather than 200k
    updates. (c) Forcing the gradient to take a round trip through SD-VAE back to the same latent space has a relatively minor cost in quality and convergence speed.}
    \label{fig:imagenet-convergence}
\end{center}

\FloatBarrier

\clearpage

\section{Diagnosing the Pixel--Latent Supervision Gap}
\label{sec:supp-diagnostics}

In the main paper, latent matching outperforms pixel matching even though the
pixel teacher is larger and achieves better unguided sampling FID. This shows
that unguided sampling FID alone does not rank the evaluated configurations
by downstream distillation quality and motivates examining the supervision
signal.

One possibility is
that the pixel-space correction contains a larger fixed-point residual than
its latent-space counterpart. 
Below we test this hypothesis on ImageNet-256
and check several competing explanations.

Unless stated otherwise, every distillation run in this section trains the
same Pixel SiT-XL student initialized from the VR242.5K checkpoint using the
DMD loss alone, and reports one-step EMA FID-50K every 10k generator updates.
Pixel matching uses the Pixel SiT-XL teacher--critic pair with CFG~1.4 and
shift-10 noise sampling; latent matching uses Latent SiT-XL/2 with CFG~1.6
and uniform (shift-1) sampling.  Several exploratory runs were stopped once
their trend was clear and are compared at matched updates.

\subsection{How Large Is the Fixed-Point Residual?}
\label{sec:supp-wxb-disagreement}

To quantify such a residual, a natural thought experiment is to ask what
gradient a \emph{perfect} student would receive. Suppose the generator
already matches the real data distribution exactly.  Written in
clean-prediction form, the DMD update at guidance scale $w$ splits into two
parts:
\[
    y^{(w)}_{\mathrm{real}} - y_{\mathrm{critic}}
    = \underbrace{\bigl(y^{(w)}_{\mathrm{real}} - y^{(1)}_{\mathrm{real}}\bigr)}_{\text{intended CFG correction}}
    - \underbrace{\bigl(y_{\mathrm{critic}} - y^{(1)}_{\mathrm{real}}\bigr)}_{\text{spurious null residual}},
\]
where $y^{(1)}_{\mathrm{real}}$ is the frozen unguided teacher's clean
prediction from a noised sample and $y_{\mathrm{critic}}$ is the online
critic's prediction. When the student distribution matches the data
distribution, the critic's population optimum is also $f^\star$. If the
frozen teacher matched $f^\star$ and critic optimization reached that
optimum, the two would agree and the second term would vanish, leaving only
the intended CFG push. Any disagreement between the two injects a spurious
gradient even at the ideal fixed point.

This directly suggests a \emph{null test}: freeze the generator, feed real
ImageNet training images straight to a critic initialized from the frozen
teacher, and train it for 50k updates under each route's native denoising
objective.  To compare the resulting displacement across spaces on a common
dimensionless scale, we normalize it by the teacher's own denoising error on
a clean sample $y_0$:
\begin{equation}
    \rho(\sigma)=
    \frac{\bigl(\mathbb{E}\,\|y_{\mathrm{critic}}-
          y^{(1)}_{\mathrm{real}}\|_2^2\bigr)^{1/2}}
         {\bigl(\mathbb{E}\,\|y_0-
          y^{(1)}_{\mathrm{real}}\|_2^2\bigr)^{1/2}}.
    \label{eq:wxb-critic-disagreement}
\end{equation}
We evaluate \cref{eq:wxb-critic-disagreement} on 8,192 held-out ImageNet
validation images over ten noise levels $\sigma\in\{0.05,0.15,\dots,0.95\}$,
averaging squared errors within low ($\{0.05,0.15,0.25\}$), mid
($\{0.35,\dots,0.65\}$), and high ($\{0.75,0.85,0.95\}$) bands.  During the
null test, we also log an absolute-error version of the same ratio on
training images every 50 updates under each route's DMD noise sampler to
watch how the critic moves.

\begin{center}
    \centering
    \small
    \setlength{\tabcolsep}{6.0pt}
    \captionsetup{hypcap=false}
    \captionof{table}{\textbf{Null-test trajectory.}  Training-image
    disagreement (absolute errors normalized by the teacher's denoising
    error), averaged over 2.5k-update windows.  Both critics start at the
    teacher (zero initial disagreement), yet the pixel critic pulls away
    immediately.}
    \label{tab:wxb-null-trajectory}
    \begin{tabular}{@{}lcc@{}}
        \toprule
        Critic update window & Pixel critic & Latent critic \\
        \midrule
        0--2.5k      & 0.070 & 0.017 \\
        10k--12.5k   & 0.105 & 0.023 \\
        20k--22.5k   & 0.117 & 0.025 \\
        30k--32.5k   & 0.117 & 0.026 \\
        40k--42.5k   & 0.131 & 0.026 \\
        47.5k--50k   & 0.121 & 0.027 \\
        \bottomrule
    \end{tabular}
\end{center}

Even though both critics start at their teachers and maintain visually flat
training losses throughout (around 0.30 for pixel and 0.69 for latent),
their trajectories separate right away (\cref{tab:wxb-null-trajectory}).
The pixel critic drifts rapidly and saturates around 30k updates (averaging
0.124 over 40k--50k), whereas the latent critic's average disagreement over
40k--50k is roughly one-fifth as large (0.026 versus 0.124).

How big is this residual compared with the guidance signal we actually want?
Measuring $\|y_{\mathrm{critic}}-y^{(1)}_{\mathrm{real}}\|_1$ against
$\|y^{(w)}_{\mathrm{real}}-y^{(1)}_{\mathrm{real}}\|_1$ on training images
at each route's training CFG makes the gap concrete. In pixel space, the
spurious residual has 4.46$\times$, 3.20$\times$, and 0.81$\times$ the
magnitude of the intended CFG correction at low, mid, and high noise. Its
magnitude therefore exceeds the intended correction at low and mid noise;
this norm comparison alone does not determine their directional interaction.
In latent space, the same ratio is only 0.46$\times$, 0.25$\times$, and
0.17$\times$.

\subsection{Is This Residual Teacher Error or Just Critic Drift?}
\label{sec:supp-wxb-bound}

Seeing a large $\rho$ immediately raises a skeptical question: knowing that
the critic $C$ moved away from the teacher $T$ does not tell us \emph{why}.
In a high-dimensional space, $C$ could simply be drifting under SGD noise,
or $C$ and $T$ might be two equally good local minima with different
predictions.  And a flat curve in \cref{tab:wxb-null-trajectory} certainly
does not prove that $C$ has converged to the data-optimal denoiser
$f^\star(y_t,\sigma,c)=\mathbb{E}[y_0\mid y_t,\sigma,c]$.

It turns out we do not need the critic to converge in order to separate
genuine teacher error from symmetric drift.  For any denoiser $g$, its
held-out MSE risk decomposes orthogonally as
$L_\sigma(g)=\mathbb{E}\|g(y_t,\sigma,c)-y_0\|_2^2
=L_\sigma(f^\star)+\mathbb{E}\|g-f^\star\|_2^2$.  Taking the difference
between the teacher $T$ and the null-test critic $C$ gives
\begin{equation}
    \Delta L(\sigma)=L_\sigma(T)-L_\sigma(C)
    =\mathbb{E}\|T-f^\star\|_2^2-\mathbb{E}\|C-f^\star\|_2^2
    \le\mathbb{E}\|T-f^\star\|_2^2 .
    \label{eq:wxb-risk-bound}
\end{equation}
If $C$ and $T$ were merely wandering between equally accurate solutions,
$\Delta L$ would sit near zero.  Whenever the critic achieves strictly lower
held-out MSE than the teacher ($\Delta L(\sigma)>0$), $\sqrt{\Delta L(\sigma)}$
rigorously lower-bounds the teacher's distance to $f^\star$, regardless of
how far $C$ itself remains from optimal.  Normalizing by the same
denominator as \cref{eq:wxb-critic-disagreement} gives a convergence-free
lower bound on the teacher's normalized excess error:
\begin{equation}
    \rho_{\mathrm{LB}}(\sigma)=
    \sqrt{\frac{\max(\Delta L(\sigma),0)}{L_\sigma(T)}}.
    \label{eq:wxb-rho-lb}
\end{equation}
Because $\rho_{\mathrm{LB}}$ lower-bounds $\|T-f^\star\|$ whereas $\rho$
measures the step $\|C-T\|$, $\rho_{\mathrm{LB}}$ can legally exceed $\rho$
(for example, stepping halfway from $T$ toward $f^\star$ reduces squared
error by $\frac{3}{4}\|T-f^\star\|^2$ while moving only
$\frac{1}{4}\|T-f^\star\|^2$).  We estimate both quantities on the 8,192
ImageNet validation images at a common unguided scale, using paired draws
(identical class label, clean image, noise, and $\sigma$ for $T$ and $C$) so
that the risk difference has low variance.

\begin{center}
    \centering
    \small
    \setlength{\tabcolsep}{4.5pt}
    \captionsetup{hypcap=false}
    \captionof{table}{\textbf{Held-out critic disagreement $\rho$ and
    convergence-free lower bound $\rho_{\mathrm{LB}}$ on normalized teacher
    excess error} by noise band.  $\rho_{\mathrm{LB}}=0$ means this
    null-test critic gives no positive lower bound.  The last column reports
    the low-noise risk gap $\Delta L$ in units of its paired standard
    error.}
    \label{tab:wxb-rho-lb}
    \begin{tabular}{@{}llccccccc@{}}
        \toprule
        & & \multicolumn{3}{c}{$\rho$} & \multicolumn{3}{c}{$\rho_{\mathrm{LB}}$} & \\
        \cmidrule(lr){3-5}\cmidrule(lr){6-8}
        Space & Teacher & Low & Mid & High & Low & Mid & High & $\Delta L/\mathrm{s.e.}$ \\
        \midrule
        Pixel & AsymFlowVR 100k & 0.352 & 0.246 & 0.097 & 0.367 & 0.192 & 0.000 & 222 \\
              & AsymFlowVR 400k & 0.314 & 0.213 & 0.086 & 0.332 & 0.159 & 0.029 & 213 \\
              & Adapted SiT-XL, 242.5k & 0.314 & 0.202 & 0.085 & 0.329 & 0.147 & 0.000 & 191 \\
        \midrule
        Latent & Recovered, 1000 steps & 0.068 & 0.048 & 0.030 & 0.068 & 0.037 & 0.001 & 195 \\
               & Recovered, 1600 steps & 0.050 & 0.038 & 0.022 & 0.048 & 0.027 & 0.000 & 127 \\
               & Latent SiT-XL/2 & \textbf{0.029} & \textbf{0.033} & \textbf{0.023} & \textbf{0.005} & \textbf{0.000} & \textbf{0.000} & 2 \\
        \bottomrule
    \end{tabular}
\end{center}

The held-out test gives a clear answer (\cref{tab:wxb-rho-lb}).  For every
pixel teacher---even AsymFlowVR trained to 400k steps and the 242.5k-step
adapted SiT-XL---the further-fitted critic lowers validation MSE by around
200 standard errors at low noise ($\Delta L/\mathrm{s.e.}=191\text{--}222$),
certifying normalized teacher excess errors of at least
$\rho_{\mathrm{LB}}=0.33\text{--}0.37$ at low noise and $0.15\text{--}0.19$
at mid noise.  So the large pixel disagreement is not just random jitter
around an optimal denoiser: even well-trained pixel teachers still leave
substantial low- and mid-noise error headroom.  Latent SiT-XL/2 behaves
differently.  Its disagreement is an order of magnitude smaller
($\rho\approx 0.03$), and the null-test critic barely improves held-out risk
at all ($\rho_{\mathrm{LB}}=0.005$, 2~s.e., at low noise, and zero above
$\sigma\approx 0.25$).  The two recovered-latent controls sit cleanly in
between, stepping down from 1000 to 1600 recovery updates toward the full
latent teacher, while high-noise bounds stay near zero across all models.

\begin{figure}[t]
    \centering
    \includegraphics[width=0.56\linewidth]{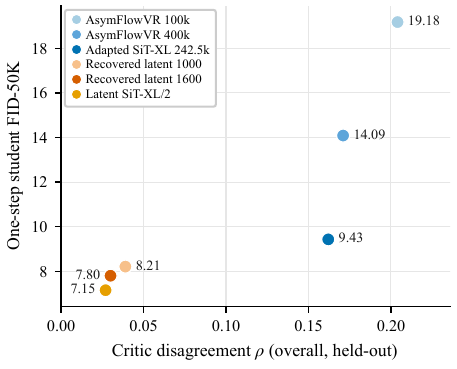}
    \caption{\textbf{Exploratory association between critic disagreement and
    student FID.}  Overall held-out $\rho$ versus one-step student FID-50K at
    200k updates.  The Pixel SiT-XL student and its VR242.5K initialization
    are fixed; blue and orange denote pixel- and latent-matching
    configurations, respectively.  Each configuration retains its native DMD
    recipe, so the plot is observational rather than a cross-space causal
    intervention.}
    \label{fig:wxb-teacher-strength}
\end{figure}

Does this diagnostic line up with actual distillation quality?  Holding the
Pixel SiT-XL student initialization fixed, one-step student FID at 200k
updates follows the exact same ordering as $\rho$ within each space
(\cref{fig:wxb-teacher-strength}): 19.18, 14.09, and 9.43 for the three
pixel teachers, and 8.21, 7.80, and 7.15 for the three latent teachers.  We
should still be careful not to overclaim from these numbers:
$\rho_{\mathrm{LB}}=0$ only means this particular critic found no positive
certificate (not that teacher error is literally zero), and because each
route keeps its native DMD recipe, \cref{fig:wxb-teacher-strength} documents
a consistent within-space association rather than a direct cross-space
causal proof.

\subsection{Could It Simply Be Encoder Smoothing or the Noise Schedule?}
\label{sec:supp-wxb-smoothing}

Even after seeing the null test, one might wonder whether latent matching
wins for a much more mundane reason.  Backpropagating through a frozen VAE
encoder $E$ could act as a spatial low-pass filter that simply smooths away
noisy high-frequency updates.  Alternatively, the gap might come from the
different noise samplers used in the two routes (uniform versus shift-10).

To check the smoothing idea first, we decompose the pixel-space update
received by the student into radial spatial-frequency bands and compare the
energy of the null-test update with that of a trained DMD run
(\cref{tab:wxb-spectrum}).  If encoder backpropagation helped mainly by
attenuating high frequencies, the two routes should look comparable at low
frequencies.  Instead, the gap spans the entire spectrum: even below $1/64$
cycles per pixel, the null-to-trained energy ratio is 0.04 for latent
matching versus 0.21 for pixel matching.  Moving up in frequency, pixel
matching's ratio climbs past one above $1/4$ cycles per pixel (1.27),
whereas latent matching drops to 0.02 in that same highest band even though
its trained update still puts most of its energy there.  Encoder
backpropagation is therefore not just erasing high-frequency gradients.
(Since these ratios compare two empirical updates without a ground-truth
gradient or orthogonal decomposition, they measure relative null energy
rather than an exact signal-to-noise split.)

\begin{center}
    \centering
    \small
    \setlength{\tabcolsep}{4.0pt}
    \captionsetup{hypcap=false}
    \captionof{table}{\textbf{Null-to-trained update energy ratio}
    by radial spatial-frequency band (cycles per pixel).  Lower means less
    null-test energy relative to the trained-run update.}
    \label{tab:wxb-spectrum}
    \begin{tabular}{@{}lccccccc@{}}
        \toprule
        Route & All & $<\!1/64$ & $1/64$--$1/32$ & $1/32$--$1/16$ & $1/16$--$1/8$ & $1/8$--$1/4$ & $>\!1/4$ \\
        \midrule
        Pixel $\to$ pixel, standard & 0.45 & 0.21 & 0.18 & 0.24 & 0.39 & 0.83 & 1.27 \\
        Latent $\to$ pixel, standard & \textbf{0.03} & \textbf{0.04} & \textbf{0.03} & \textbf{0.03} & \textbf{0.06} & \textbf{0.05} & \textbf{0.02} \\
        \bottomrule
    \end{tabular}
\end{center}

As a more direct check, if smoothing or the noise schedule were driving the
gain, we should be able to reproduce at least part of it by filtering pixel
gradients or swapping noise samplers (\cref{tab:wxb-isolation}).  Neither
intervention helps.  Applying the encoder's exact Gram-operator filter
$J_E^\top J_E\,g_x$ (via Jacobian--vector products) to the pixel-matching
update severely hurts training (37.8 vs.\ 13.5 FID at 40k updates), and a
Gaussian low-pass filter ($\sigma=1.5$ pixels) also degrades FID (13.6 vs.\
12.0 at 60k).  Swapping noise samplers between critics fails across every
tested pair: moving the pixel critic to uniform sampling stalls near 105 for
a pixel student and diverges (288 at 20k) for a latent student, while moving
the latent critic to shift-10 sampling slows convergence sharply for both
pixel (27.2 vs.\ 12.0 at 40k) and latent students (26.5 vs.\ 3.45 at 60k).
Simple gradient filtering and sampler choice therefore do not explain the
latent-matching advantage.

\begin{center}
    \centering
    \small
    \setlength{\tabcolsep}{5.0pt}
    \captionsetup{hypcap=false}
    \captionof{table}{\textbf{Gradient-filtering and noise-sampler
    interventions} (one-step EMA FID-50K at matched generator updates).
    Neither filtering pixel updates nor swapping noise schedules reproduces
    latent matching.}
    \label{tab:wxb-isolation}
    \begin{tabular}{@{}llcc@{}}
        \toprule
        Control & Setting (student / critic) & Intervened & Standard \\
        \midrule
        Gradient filtering
            & Encoder Gram $J_E^\top J_E\,g_x$ (pixel / pixel) & 37.8 @ 40k & 13.5 @ 40k \\
            & Gaussian low-pass, $\sigma{=}1.5$ px (pixel / pixel) & 13.6 @ 60k & 12.0 @ 60k \\
        \midrule
        Sampler swap
            & Pixel / pixel + uniform (CFG 1.6) & stalls ($\sim$105) & 9.43 @ 200k \\
            & Pixel / latent + shift-10 (CFG 1.4) & 27.2 @ 40k & 12.0 @ 40k \\
            & Latent / pixel + uniform & diverges (288 @ 20k) & --- \\
            & Latent / latent + shift-10 & 26.5 @ 60k & 3.45 @ 60k \\
        \bottomrule
    \end{tabular}
\end{center}

\FloatBarrier
\section{Text-to-Image Experimental Setup}
\label{sec:supp-t2i-training}

The adversarial component below is used only for the large-scale
text-to-image experiments. The ImageNet-256 experiments use DMD without an
adversarial objective.

\subsection{Model-Specific DMD Training Settings}
\label{sec:supp-t2i-dmd-settings}

These settings cover downstream DMD after the initialization in
\cref{sec:supp-t2i-initialization}. \Cref{tab:supp-t2i-dmd-common} reports
the shared recipe, \cref{tab:supp-t2i-dmd-models} gives the model-specific
choices, and \cref{tab:supp-t2i-dmd-checkpoints} records the training seeds,
completed budgets, and selected checkpoints. For each route, the frozen
teacher and online critic backbone are initialized from the same released
weights; the critic's STAC parameters are initialized separately, and only
the student loads the ODE-warmup initializer.

\begin{center}
    \begin{minipage}{\textwidth}
    \centering
    \small
    \setlength{\tabcolsep}{5pt}
    \captionsetup{hypcap=false}
    \captionof{table}{\textbf{Shared text-to-image DMD optimization settings.}
    These settings apply to every reported text-to-image DMD experiment;
    they are distinct from the ImageNet and ODE-warmup settings.}
    \label{tab:supp-t2i-dmd-common}
    \begin{tabular}{@{}p{0.32\textwidth}p{0.64\textwidth}@{}}
        \toprule
        Setting & Value \\
        \midrule
        Student / resolution & Four-step direct-pixel model / $1024\times1024$ \\
        Hardware / batch & Eight GPUs; 2 per GPU, global 16, no accumulation \\
        Random seed & Row-specific base seed in \cref{tab:supp-t2i-dmd-checkpoints}, plus GPU rank \\
        Optimizers / learning rates & AdamW; student $2\times10^{-6}$, online critic $5\times10^{-7}$ \\
        Adam / schedule / clipping & $(\beta_1,\beta_2)=(0.9,0.999)$, $\epsilon=10^{-8}$, WD $0.01$; constant LR; norm 10 \\
        Update and time sampling & Critic:student $5{:}1$; uniform rollout step; $u\sim U[0.02,0.98]$, $t=qu/[1+(q-1)u]$ \\
        STAC & Four context tokens, width 1024, separate LR $5\times10^{-5}$ \\
        Guidance / EMA & Critic CFG 1 (FLUX.2-dev uses embedded guidance); no inference CFG or student EMA \\
        DMD conditions / negative prompt & Synth100K (100,000 text-only prompts) / empty \\
        Pixel GAN / real pairs & Frozen Qwen3.5-0.8B backbone at $1024^2$ / LAION-1024 100K subset \\
        GAN optimization & AdamW head, LR $10^{-4}$, betas $(0.9,0.95)$, WD 0; 200-step warmup; final two rollout steps \\
        \bottomrule
    \end{tabular}
    \end{minipage}
\end{center}

\FloatBarrier
\begin{center}
    \begin{minipage}{\textwidth}
    \centering
    \small
    \setlength{\tabcolsep}{4pt}
    \captionsetup{hypcap=false}
    \captionof{table}{\textbf{Model-specific text-to-image DMD settings.}
    Row IDs map configurations to \cref{tab:supp-t2i-dmd-checkpoints}.
    ``Teacher / critic initialization'' gives the released checkpoint used
    for the frozen teacher and the online critic backbone's initial weights.
    ``Ext'' denotes external teacher CFG; ``Emb'' denotes embedded
    guidance supplied to both teacher and critic. $q$ is the score-time
    shift, not the student's rollout-grid shift. Settings shared across rows are in \cref{tab:supp-t2i-dmd-common}.}
    \label{tab:supp-t2i-dmd-models}
    \begin{tabular}{@{}p{0.16\textwidth}p{0.37\textwidth}p{0.13\textwidth}p{0.11\textwidth}p{0.13\textwidth}@{}}
        \toprule
        Student / row ID & Teacher / critic initialization & Guidance & DMD space & \shortstack{Score-time\\shift $q$} \\
        \midrule
        AsymFLUX / A-P & AsymFLUX.2 Klein-9B (pixel) & Ext 3.5 & Pixel & 17 \\
        AsymFLUX / A-L & FLUX.2 Klein Base 9B & Ext 3.5 & Latent & 6.0496 \\
        \midrule
        PixelDiT / P-P & PixelDiT-1.3B & Ext 2.75 & Pixel & 4 \\
        PixelDiT / P-L4 & FLUX.2 Klein Base 4B & Ext 3.5 & Latent & 6.0496 \\
        PixelDiT / P-L9 & FLUX.2 Klein Base 9B & Ext 3.5 & Latent & 6.0496 \\
        PixelDiT / P-L32 & FLUX.2-dev 32B & Emb 3.5 & Latent & 6.0496 \\
        PixelDiT / P-L3 & FLUX.3 Image Klein Base (5.6B) & Ext 3.5 & Latent & 6.0496 \\
        \midrule
        DreamLite / D1 & FLUX.2 Klein Base 4B & Ext 2.0 & Latent & 6.0496 \\
        DreamLite / D6 & FLUX.2 Klein Base 9B & Ext 2.0 & Latent & 6.0496 \\
        DreamLite / D4 & FLUX.2-dev 32B & Emb 2.0 & Latent & 6.0496 \\
        DreamLite / D24 & FLUX.3 Image Klein Base (5.6B) & Ext 2.0 & Latent & 6.0496 \\
        \bottomrule
    \end{tabular}
    \end{minipage}
\end{center}

For latent matching, $q=\exp(1.8)=6.0496474644$. FLUX.2-dev supplies its
listed embedded guidance to both latent networks; unlike external CFG, it
does not invoke a two-pass teacher. All latent rows use the encoder VJP.

\FloatBarrier
\begin{center}
    \begin{minipage}{\textwidth}
    \centering
    \small
    \setlength{\tabcolsep}{4pt}
    \captionsetup{hypcap=false}
    \captionof{table}{\textbf{Training seeds, completed iterations, and
    checkpoints used for the main text-to-image student results.}
    Row IDs refer to the configurations in \cref{tab:supp-t2i-dmd-models}.
    Completed iterations count critic updates; student updates occur
    every five iterations. Budgets exclude ODE warmup.}
    \label{tab:supp-t2i-dmd-checkpoints}
    \begin{tabular}{@{}lrrr@{}}
        \toprule
        Row ID & Base seed & \shortstack[r]{Completed\\iterations}
        & \shortstack[r]{Selected\\iteration} \\
        \midrule
        A-P & 10 & 10,000 & 3,000 \\
        A-L & 10 & 10,000 & 3,000 \\
        \midrule
        P-P & 10 & 20,000 & 20,000 \\
        P-L4 & 10 & 20,000 & 20,000 \\
        P-L9 & 10 & 20,000 & 20,000 \\
        P-L32 & 10 & 20,000 & 20,000 \\
        P-L3 & 10 & 20,000 & 20,000 \\
        \midrule
        D1 & 10 & 10,000 & 10,000 \\
        D6 & 11 & 10,000 & 10,000 \\
        D4 & 10 & 10,000 & 10,000 \\
        D24 & 10 & 10,000 & 10,000 \\
        \bottomrule
    \end{tabular}
    \end{minipage}
\end{center}

Selected checkpoints store student and online critic states together; no DMD
student EMA is substituted during evaluation. The full 10,000-iteration runs
contain 2,000 student updates, and the 20,000-iteration runs contain 4,000.
The selected AsymFLUX checkpoints correspond to 600 student updates.

\subsection{Training Data Construction and Benchmark Separation}
\label{sec:supp-t2i-data}

Every DMD student and critic uses Synth100K text-only conditions; the Qwen
GAN independently uses LAION-1024 100K image--caption pairs. LAION captions
and FLUX-Reason prompts are never DMD conditions.

\paragraph{Synth100K prompt construction.}
Synth100K splits evenly between dense and object-centric compositional
prompts.
\Cref{tab:supp-synthetic-prompts}
 breaks down the prompt distribution.

\begin{center}
    \begin{minipage}{\textwidth}
    \centering
    \small
    \setlength{\tabcolsep}{4.0pt}
    \captionsetup{hypcap=false}
    \captionof{table}{\textbf{Synth100K prompt families and template structure.}
    Shares are relative to all 100,000 prompts. Each 50K branch sums to
    50\%; the dense branch uses proportions 25/20/20/15/10/10 within its
    branch, and the object branch uses 17/17/17/17/16/16.}
    \label{tab:supp-synthetic-prompts}
    \begin{tabular}{@{}p{0.21\textwidth}rrp{0.58\textwidth}@{}}
        \toprule
        Family & Prompts & Share & Scene graph and rendered constraints \\
        \midrule
        Exact count & 12,500 & 12.5\% & Exactly 3--12 colored objects in a sampled layout; an exact marked subgroup; no extra objects. \\
        Attribute binding & 10,000 & 10\% & Three objects with distinct colors, sizes, and markings; bind each attribute to its named object. \\
        Spatial relations & 10,000 & 10\% & Four colored objects in horizontal, vertical, depth, or corner order; preserve order and visible gaps. \\
        Text rendering & 7,500 & 7.5\% & A sign, poster, package, or similar carrier displays a quoted uppercase phrase; no additional writing. \\
        Long composition & 5,000 & 5\% & Six colored entities distributed across foreground, middle ground, and background, with lighting and scene constraints. \\
        Actions and poses & 5,000 & 5\% & An actor performs a specified action with two colored objects while a nearby animal is present; the actor's hands are visible, and the pose has physically plausible support. \\
        \midrule
        Single object & 8,500 & 8.5\% & Exactly one colored foreground object in a sampled scene and framing; no duplicate. \\
        Two objects & 8,500 & 8.5\% & Exactly two distinct colored foreground objects, similar scale, fully visible and non-overlapping. \\
        Counting & 8,500 & 8.5\% & Exactly 2--8 colored objects in a row, balanced rows, an arc, or a grid; no hidden extras. \\
        Colors & 8,500 & 8.5\% & One object's main body has a named color under neutral lighting against a sampled background. \\
        Position & 8,000 & 8\% & Two objects with a specified left/right, above/below, or front/behind relation and its inverse. \\
        Color attribution & 8,000 & 8\% & Two objects with distinct named colors; prohibit swapping, blending, or repeating their colors. \\
        \bottomrule
    \end{tabular}
    \end{minipage}
\end{center}

A base construction seed of 20261104 drives prompt generation and the final
shuffle through deterministic offsets, producing 100,000 unique 40--121-word
captions. Duplicates are resampled, and the generator never reads benchmark
prompt files. The resulting prompt JSONL SHA-256 is
\nolinkurl{913f18ec5e761c2637a8c29f26f36eb8dd32e47014e2875be4be41410f88cf4f}.

\paragraph{LAION-1024 100K subset selection and pairing.}
From the first 1.5M LAION-3M metadata rows, we retain deduplicated 15--120-word
captions whose source image needs neither upsampling nor square cropping to
$1024^2$. TF--IDF features are reduced to 128 dimensions and clustered with
1,000-way MiniBatchKMeans. The SVD, clustering, and within-cluster
permutations use seed 0; deterministic round-robin selection retains 100,000
of 112,125 eligible pairs. The final Arrow-row shuffle also uses seed 0. The
preserved PNG bytes and metadata occupy 143 Arrow shards. During GAN training,
the same caption $c$ conditions real pair $(x_r,c)$ and independently generated
$x_g$; the DMD batch is drawn separately from Synth100K.

\paragraph{Text encoding and truncation.}
FLUX.2 Klein uses one shared 128-token Synth100K cache for P2P and L2P;
FLUX.2-dev uses 192 tokens, and FLUX.3 preserves official variable-length
contexts padded to multiples of 80 (62,052 length-80 and 37,948 length-160).
DreamLite uses its native Qwen3-VL \texttt{[Generate]:} template with 512
positions; PixelDiT uses official Gemma-2 CHI preprocessing with 300
positions; Qwen3.5 GAN captions use 96 caption tokens before the image--text
chat template, excluding visual and template tokens.

\paragraph{Raw-text benchmark overlap audit.}
After Unicode normalization and case folding, neither Synth100K nor the
LAION-1024 100K subset has an exact or near match to the 10,000 HPSv3 COCO
recaptions or 1,065 DPG-Bench prompts: all four stream--benchmark pairs have
0 exact and 0 near matches. ``Near'' includes exact matches and uses
word-trigram Jaccard similarity $\geq0.8$; this audit covers raw text rather
than semantic or image-level duplication.

\subsection{Pixel Adaptation and ODE Initialization}
\label{sec:supp-t2i-initialization}

\Cref{tab:supp-t2i-initialization} summarizes the three student initializers.
ODE targets come from each multi-step pixel base, not the later latent DMD
teacher; FLUX-Reason captions are used only at this stage. All downstream DMD
updates instead use Synth100K. Each warmup uses endpoint MSE with AdamW for
3,000 updates, global batch 8 on eight GPUs, uniform sampling over four student
times, and four-step re-noise inference without CFG.

\begin{center}
    \centering
    \small
    \setlength{\tabcolsep}{4.5pt}
    \captionsetup{hypcap=false}
    \captionof{table}{\textbf{Pixel adaptation and ODE initialization for
    $1024\times1024$ text-to-image students.} Target CFG is used only to
    generate warmup targets and is distinct from downstream DMD teacher
    guidance. Student grids list the four prediction times, excluding the
    terminal zero.}
    \label{tab:supp-t2i-initialization}
    \begin{tabular}{@{}p{0.18\textwidth}p{0.245\textwidth}p{0.245\textwidth}p{0.245\textwidth}@{}}
        \toprule
        Setting & AsymFLUX.2 Klein 9B & PixelDiT-1.3B & Pixel DreamLite \\
        \midrule
        Pixel base &
        Released AsymFLUX.2 Klein-9B adapter of FLUX.2 &
        Released PixelDiT-1.3B native-RGB weights &
        DreamLite Base adapted with AsymFlowVR \\
        Pixel adaptation before ODE &
        Released adapter; no new pixel adaptation &
        None; native RGB model &
        100K self-generated images; MSE weight 10, LPIPS weight 1;
        LR $2\times10^{-5}$ ($10\times$ for pixel branch);
        6,000-step budget, 3,000-step EMA selected \\
        \midrule
        ODE target solver / steps &
        Euler / 28 &
        Official flow DPM-Solver, order 2, multistep / 50 &
        Euler / 28 \\
        Target CFG / negative prompt &
        $4.0$ / empty &
        $3.5$ / empty &
        $3.5$ / empty \\
        Target / student time shift &
        $17$ &
        $4$ &
        $17$ \\
        Clean-prediction clamp during target generation &
        No &
        No additional clamp &
        Yes; sRGB-gamut clamp \\
        ODE target images &
        10,000 &
        10,000 &
        4,000 \\
        Warmup corruption noise &
        Fresh Gaussian; target's initial noise not reused &
        Exact paired FP32 initial noise from the target trajectory &
        Fresh Gaussian; target's initial noise not reused \\
        \midrule
        Warmup learning rate &
        $2\times10^{-6}$ &
        $2\times10^{-6}$ &
        $1\times10^{-5}$ \\
        Student prediction times &
        $[1.0000,0.9808,$\newline $0.9444,0.8500]$ &
        $[1.0000,0.9231,$\newline $0.8000,0.5714]$ &
        $[1.0000,0.9808,$\newline $0.9444,0.8500]$ \\
        \bottomrule
    \end{tabular}
\end{center}

\begin{samepage}
For all three warmups, let $x^\star$ denote the fixed pixel endpoint target
and $\epsilon$ the corruption noise specified in the table. We sample one
of the four student times uniformly and minimize
\begin{equation}
    x_t=(1-t)x^\star+t\epsilon,\qquad
    \widehat x_0=x_t-t\,v_\theta(x_t,t,c),\qquad
    \mathcal{L}_{\mathrm{warm}}=
    \operatorname{MSE}(\widehat x_0,x^\star).
    \label{eq:supp-t2i-warmup}
\end{equation}
\end{samepage}
Here $v_\theta$ is the native pixel velocity and
$t=s_q(u)$ for $u\in\{1,3/4,1/2,1/4\}$. Inference re-noises successive
endpoint predictions and returns pixels directly. Every teacher--critic
comparison within a family starts from the same initializer.

\subsection{Text-Conditioned Pixel GAN}
\label{sec:supp-t2i-gan}

Every row uses a training-only pixel GAN with a frozen Qwen3.5-0.8B backbone
and a trainable 256-wide binary MLP head. A LAION real pair $(x_r,c)$ and an
independent student sample $x_g$ share caption $c$; both images are $1024^2$
and captions are truncated to 96 tokens. The hinge losses are
\begin{equation}
    \begin{aligned}
        \mathcal{L}_{D}
        &= \tfrac{1}{2}\mathbb{E}\!\left[
          \left(1-D_\psi(x_r,c)\right)_{+}\right] \\
        &\quad+\tfrac{1}{2}\mathbb{E}\!\left[
          \left(1+D_\psi(x_g,c)\right)_{+}\right], \\
        \mathcal{L}_{\mathrm{adv}}
        &= -\mathbb{E}\!\left[D_\psi(x_g,c)\right].
    \end{aligned}
    \label{eq:qwen-gan-loss}
\end{equation}
We match the adversarial and DMD pixel-gradient norms. For an independently
drawn DMD sample $x_d$, set
$g_{\mathrm{DMD}}=\nabla_{x_d}\mathcal{L}_{\mathrm{DMD}}$ and
$g_{\mathrm{adv}}=\nabla_{x_g}\mathcal{L}_{\mathrm{adv}}$. We use
\begin{equation}
    \lambda
    = \min\!\left(
        \frac{r\lVert g_{\mathrm{DMD}}\rVert_2}
             {\lVert g_{\mathrm{adv}}\rVert_2+10^{-12}},
        10^4
      \right),
    \qquad r=1.
    \label{eq:qwen-gan-weight}
\end{equation}
The two gradients are accumulated. The discriminator head is updated every
iteration with learning rate $10^{-4}$. Adversarial student updates begin
after 200 discriminator steps and sample uniformly from the final two rollout
steps. Qwen and the head are discarded at inference.

\subsection{Matched-Hardware Training Cost}
\label{sec:supp-t2i-cost}

We time the two AsymFLUX-9B recipes on the same eight H200 141GB GPUs. Two fresh-process repeats use opposite
orders. Each repeat includes 20 warmup and 40 timed iterations per recipe,
yielding 80 timed iterations per recipe across both repeats (16 student-update
and 64 critic-only iterations). Rollout settings are balanced, and the GAN is
active from the start for this short probe.

\begin{center}
    \begin{minipage}{\textwidth}
    \centering
    \small
    \setlength{\tabcolsep}{4.5pt}
    \captionsetup{hypcap=false}
    \captionof{table}{\textbf{AsymFLUX-9B training cost on matched H200 hardware.}
    Latency is the mean synchronized rank-maximum iteration time across
    both repeats; memory is the measured peak per GPU. Absolute and
    relative changes are L2P minus P2P, with P2P as the denominator.}
    \label{tab:supp-asymflux-training-cost}
    \begin{tabular}{@{}lrrrr@{}}
        \toprule
        Metric & P2P & L2P & Absolute change & Relative change \\
        \midrule
        Mean training iteration (s) & 5.5414 & 5.6419 & $+0.1005$ & $+1.81\%$ \\
        Student-update iteration (s) & 9.7843 & 10.1035 & $+0.3192$ & $+3.26\%$ \\
        Critic-only iteration (s) & 4.4807 & 4.5265 & $+0.0459$ & $+1.02\%$ \\
        \midrule
        Peak allocated VRAM/GPU (GiB) & 53.66 & 69.74 & $+16.08$ & $+29.98\%$ \\
        Peak reserved VRAM/GPU (GiB) & 61.75 & 84.14 & $+22.39$ & $+36.25\%$ \\
        NVML sampled peak/GPU (GiB) & 63.46 & 85.85 & $+22.39$ & $+35.28\%$ \\
        \midrule
        GPU-hours, 80 timed iterations & 0.9851 & 1.0030 & $+0.0179$ & $+1.81\%$ \\
        \bottomrule
    \end{tabular}
    \end{minipage}
\end{center}

Timing includes cached inputs and all optimization/GAN work but excludes
setup, reporting, checkpoints, previews, and evaluation. GPU-hours cover
only the timed windows; reserved/NVML memory includes allocator caching.
In summary L2P does not significantly slow down training.

\subsection{HPSv3 and DPG-Bench Evaluation Protocol}
\label{sec:supp-t2i-evaluation}

All models generate at $1024^2$. Students use four-step unguided
sampling; references use their released inference procedure.

\begin{center}
    \centering
    \small
    \setlength{\tabcolsep}{4.5pt}
    \captionsetup{hypcap=false}
    \captionof{table}{\textbf{Text-to-image evaluation settings.}
    The HPSv3 prompt set is the paper's COCO recaption set, not a
    separate HPSv3-provided prompt benchmark. Higher is better for
    both metrics.}
    \label{tab:supp-t2i-evaluation}
    \begin{tabular}{@{}p{0.22\textwidth}p{0.36\textwidth}p{0.36\textwidth}@{}}
        \toprule
        Setting & HPSv3 & DPG-Bench \\
        \midrule
        Prompt set & 10,000 COCO recaptions in fixed JSON order & All 1,065 official ELLA prompts, sorted by filename stem \\
        Samples per prompt & 1 & 4; 4,260 images in total \\
        Generation seed & $i$ for zero-based prompt index $i$ & $4{,}000{,}000+4i+s$, $s\in\{0,1,2,3\}$ \\
        Scorer & HPSv3 reward model; scoring batch size 4 & Official ELLA mPLUG VQA scorer; four $1024$-pixel panels per grid \\
        Aggregation & Mean predicted preference reward over 10,000 pairs & Dependency-aware question score per image, then mean over four samples and all prompts, multiplied by 100 \\
        \bottomrule
    \end{tabular}
\end{center}

DPG uses all four samples with no best-of-four selection. HPSv3 standard
deviations are across image--prompt scores. Prompt lists
and numeric seeds are shared across models, although heterogeneous noise
spaces yield different noise tensors.

\end{document}